\documentclass[10pt,twocolumn,letterpaper]{article}

\usepackage[pagenumbers]{wacv} 

\usepackage{multirow}
\usepackage{xcolor}
\newcommand{\ns}[1]{\textcolor{black}{#1}}

\usepackage{graphicx}  
\usepackage{pgfplots}
\pgfplotsset{compat=1.18}
\usepackage{array}
\usepackage{subcaption}

\usepackage{multirow}
\usepackage{import}
\usepackage{xifthen}
\usepackage{transparent}
\graphicspath{{figures/}}
\usepackage{booktabs}
\usepackage{placeins}
\usepackage{tikz}
\usetikzlibrary{spy}
\usepackage{listings}
\usepackage{xcolor}

\newcolumntype{C}[1]{>{\centering\arraybackslash}m{#1}}
\usepackage{pifont}
\newcommand{\cmark}{\ding{51}}  
\newcommand{\xmark}{\ding{55}}  

\usepackage{longtable}
\usepackage{booktabs}
\usepackage{array}
\usetikzlibrary{shadows.blur}
\usepackage{fvextra}

\makeatletter
\renewcommand{\paragraph}{%
  \@startsection{paragraph}{4}{\z@}%
  {6pt plus 2pt minus 3pt}
  {-0.5em}
  {\normalfont\normalsize\bfseries}%
}
\makeatother
\usepackage{comment}

\definecolor{wacvblue}{rgb}{0.21,0.49,0.74}
\usepackage[pagebackref,breaklinks,colorlinks,allcolors=wacvblue]{hyperref}

\def\wacvPaperID{552} 
\def\confName{WACV}
\def\confYear{2027}

\title{Rain Beyond Streaks: \\Cascaded Realistic Rain and Wet Surface Synthesis via Scene Priors}

\author{
Melissa Tijink$^{1}$,
Kirill Tolstobrov$^{1}$,
Shunxin Wang$^{1}$,
Estefania Talavera$^{1}$,\\
Duc V. Le$^{1}$,
Luuk Spreeuwers$^{1}$,
Shaodi You$^{2}$,
Nicola Strisciuglio$^{1}$\\[0.8em]
\begin{minipage}[t]{0.45\textwidth}
\centering
$^{1}$University of Twente\\
Enschede, The Netherlands\\
\end{minipage}
\hfill
\begin{minipage}[t]{0.45\textwidth}
\centering
$^{2}$University of Amsterdam\\
Amsterdam, The Netherlands
\end{minipage}
}

\begin{document}
\maketitle
\begin{abstract}
Realistic rain synthesis is a persistent challenge for vision systems operating in adverse weather. Existing methods primarily focus on generating rain streaks, often neglecting consequent surface effects such as reflections and puddles, which are critical for perceptual realism and robustness of downstream vision tasks. We introduce CascadeRS, a rain synthesis framework that leverages depth and segmentation priors to render scene-consistent rain and surface effects. In addition to rain streaks, CascadeRS produces reflections and puddles aligned with scene semantics through a physics-inspired zero-shot, controllable generative pipeline. It produces photorealistic rainy urban scenes, with high perceptual realism, which we validate qualitatively with an extensive user study and quantitatively. We produce the Cityscapes-CascadeRS dataset, to study downstream task performance under real rainy conditions. Our results establish CascadeRS as a scalable and realistic rain-generation framework for evaluating and enhancing the robustness of vision in adverse rainy weather. Code and dataset are available via \url{https://github.com/nis-research/CascadeRainSynthesis}.
\end{abstract}

\section{Introduction}
\label{sec:introduction}

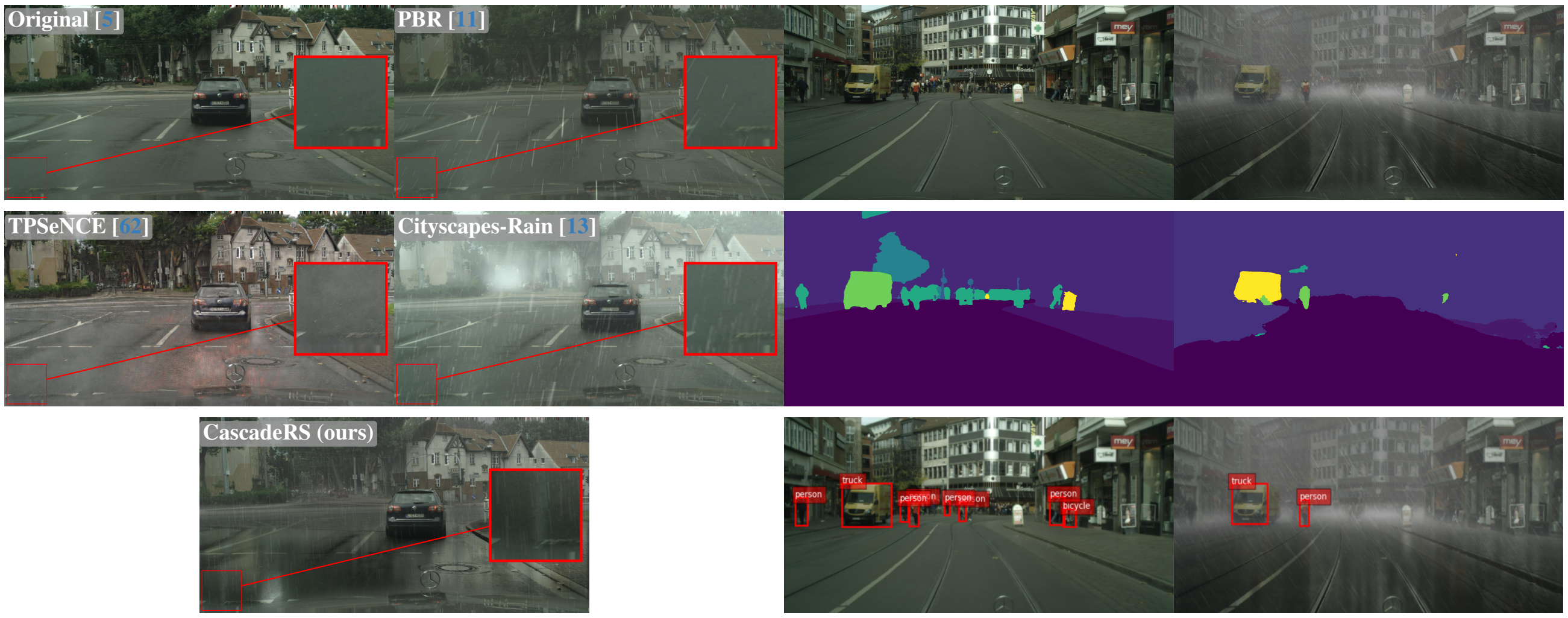
\begin{figure*}[t]
\centering
\resizebox{\linewidth}{!}{
\setlength{\tabcolsep}{1pt}
\renewcommand{\arraystretch}{1.0}

\begin{tabular}{@{}c@{}c@{}c@{}c@{}@{}}

\begin{tikzpicture}[spy using outlines={
  rectangle, red, line width=0.4pt, magnification=2.3, connect spies, size=1.0cm}]
  \node[anchor=center, inner sep=0] (img)
    {\includegraphics[width=0.245\textwidth]{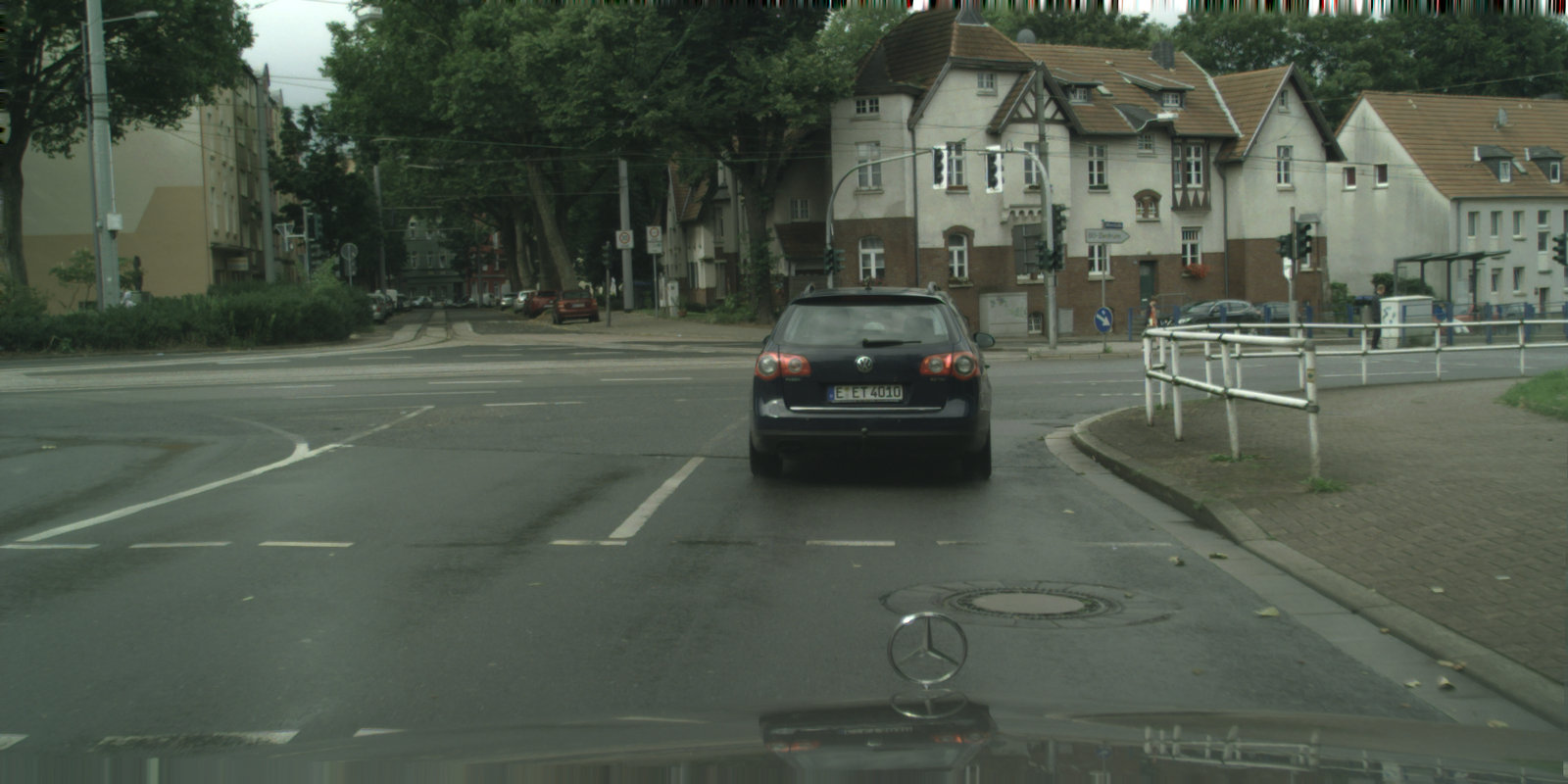}};
  \node[anchor=north west, text=white, font=\scriptsize\bfseries,
        fill=black!40!white, fill opacity=0.8, text opacity=1,
        inner sep=1pt, rounded corners=1pt]
        at ([yshift=-1pt]img.north west)
        {\textbf{Original}~\cite{Cityscapes}};
\spy on ($(img.south west)+(0.25,0.25)$)
     in node [right] at (1.05,0);
\end{tikzpicture} &
\begin{tikzpicture}[spy using outlines={
  rectangle, red, line width=0.4pt, magnification=2.3, connect spies, size=1.0cm}]
  \node[anchor=center, inner sep=0] (img)
    {\includegraphics[width=0.245\textwidth]{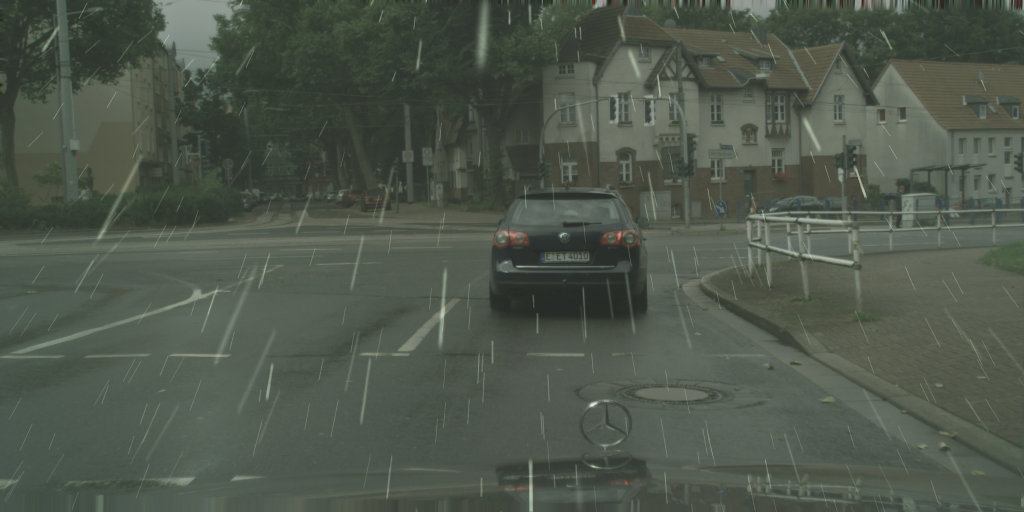}};
  \node[anchor=north west, text=white, font=\scriptsize\bfseries,
        fill=black!40!white, fill opacity=0.8, text opacity=1,
        inner sep=1pt, rounded corners=1pt]
        at ([yshift=-1pt]img.north west)
        {\textbf{PBR}~\cite{halder2019physicsbased}};
  \spy on ($(img.south west)+(0.25,0.25)$) in node [right] at (1.05,0);
\end{tikzpicture} &
\includegraphics[width=0.245\textwidth]{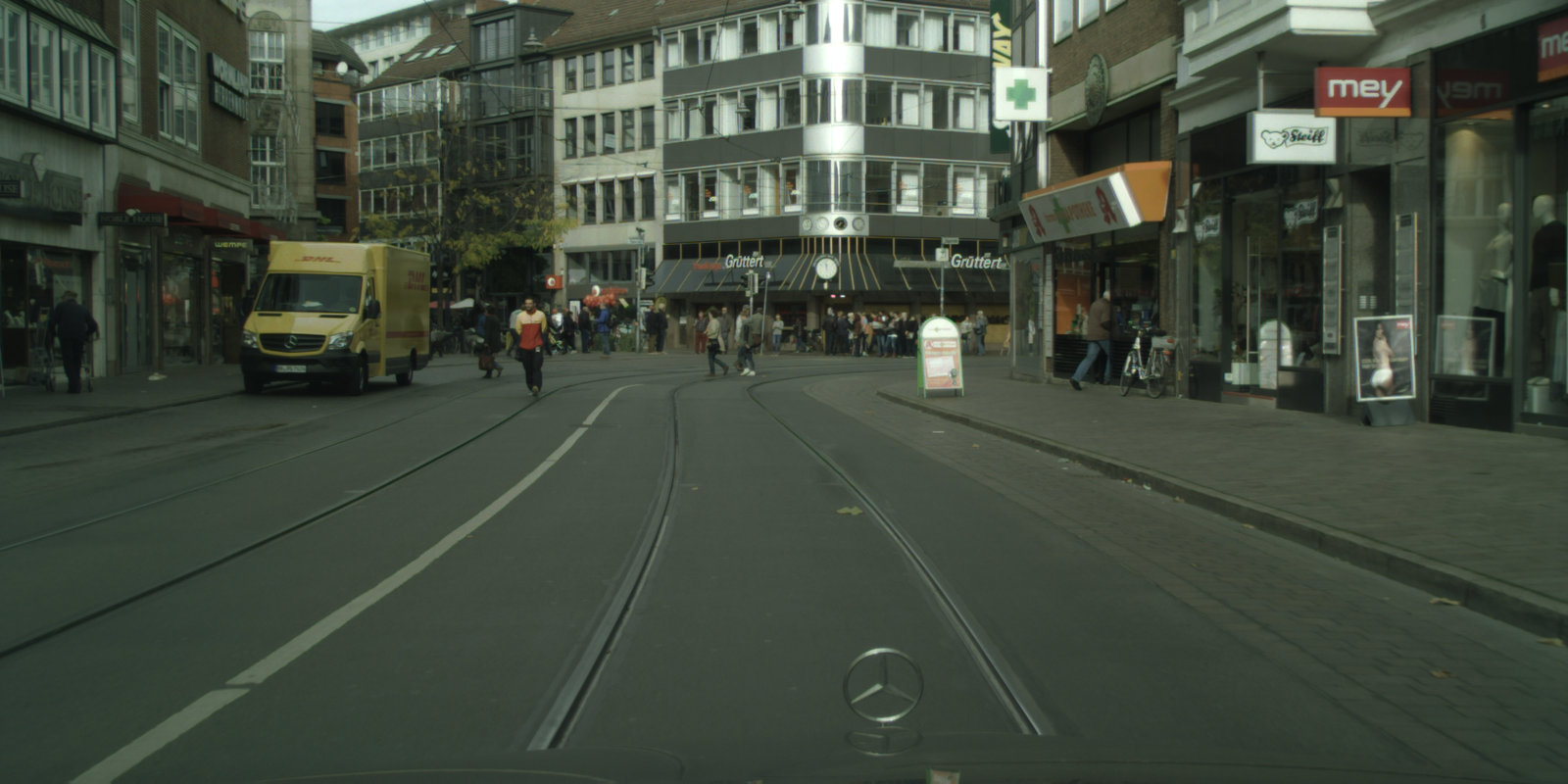} &
\includegraphics[width=0.245\textwidth]{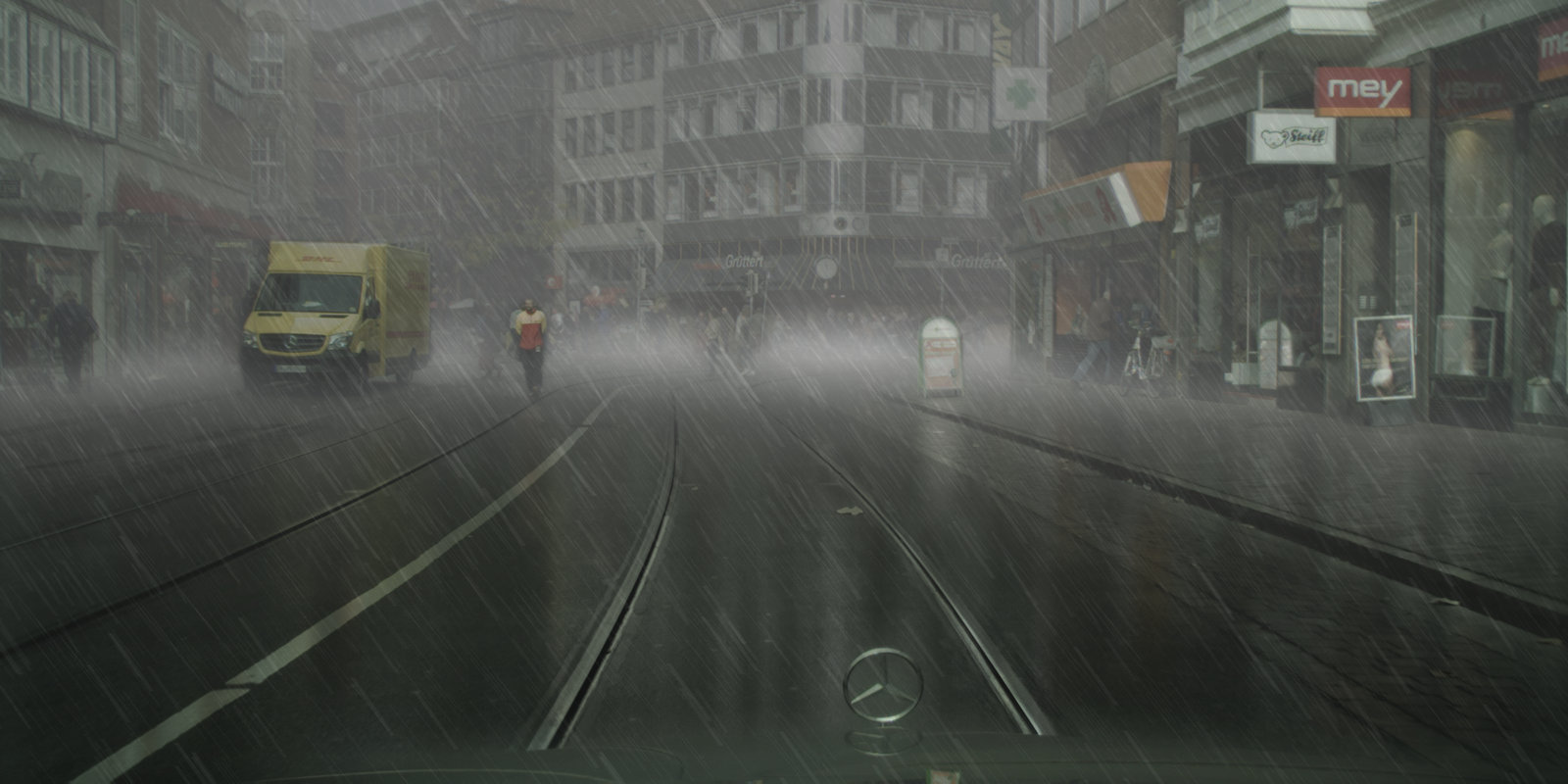} \\[-0pt]

\begin{tikzpicture}[spy using outlines={
  rectangle, red, line width=0.4pt, magnification=2.3, connect spies, size=1.0cm}]
  \node[anchor=center, inner sep=0] (img)
    {\includegraphics[width=0.245\textwidth]{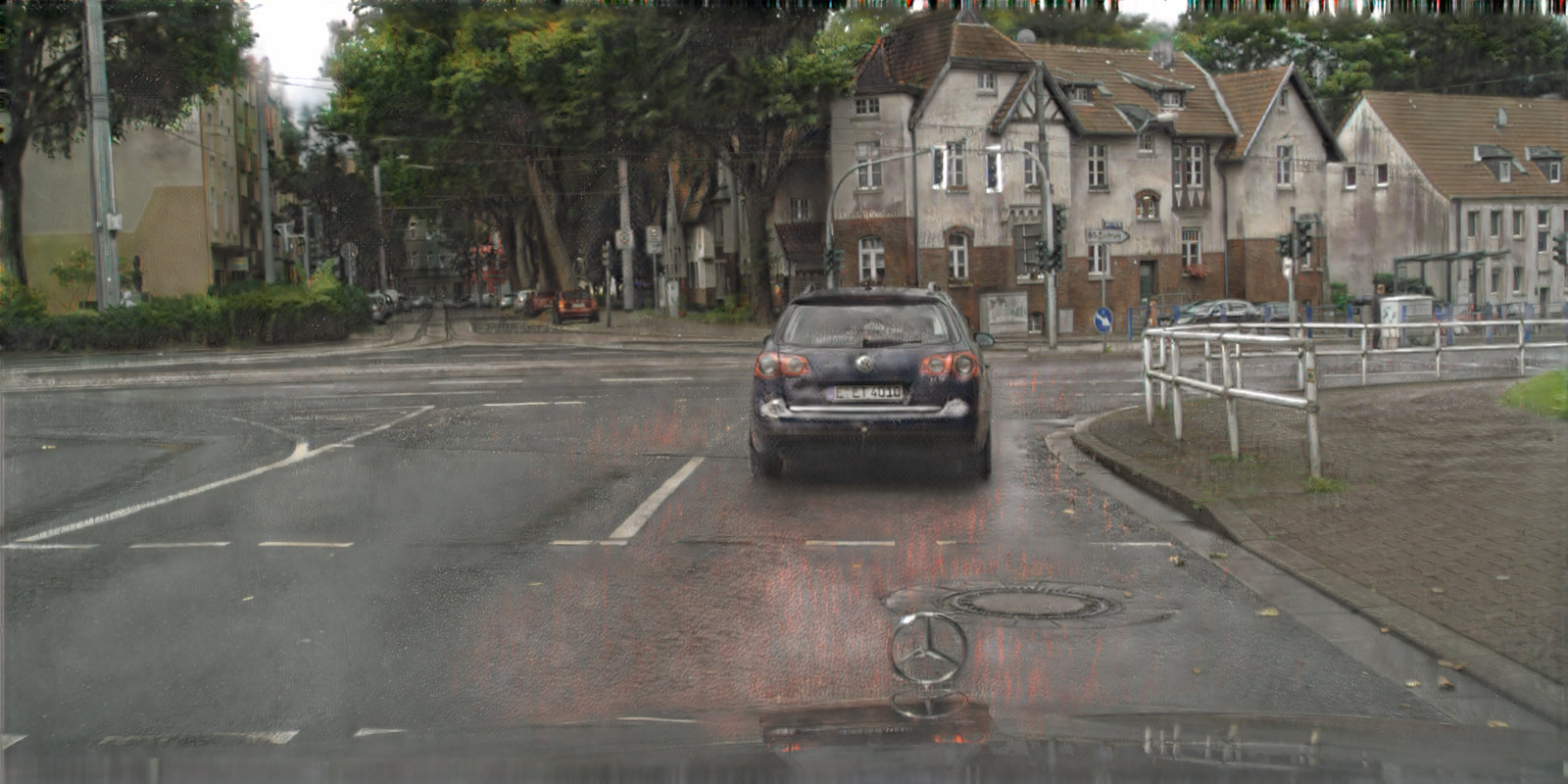}};
  \node[anchor=north west, text=white, font=\scriptsize\bfseries,
        fill=black!40!white, fill opacity=0.8, text opacity=1,
        inner sep=1pt, rounded corners=1pt]
        at ([yshift=-1pt]img.north west)
        {\textbf{TPSeNCE}~\cite{zheng2023}};
  \spy on ($(img.south west)+(0.25,0.25)$) in node [right] at (1.05,0);
\end{tikzpicture} &
\begin{tikzpicture}[spy using outlines={
  rectangle, red, line width=0.4pt, magnification=2.3, connect spies, size=1.0cm}]
  \node[anchor=center, inner sep=0] (img)
    {\includegraphics[width=0.245\textwidth]{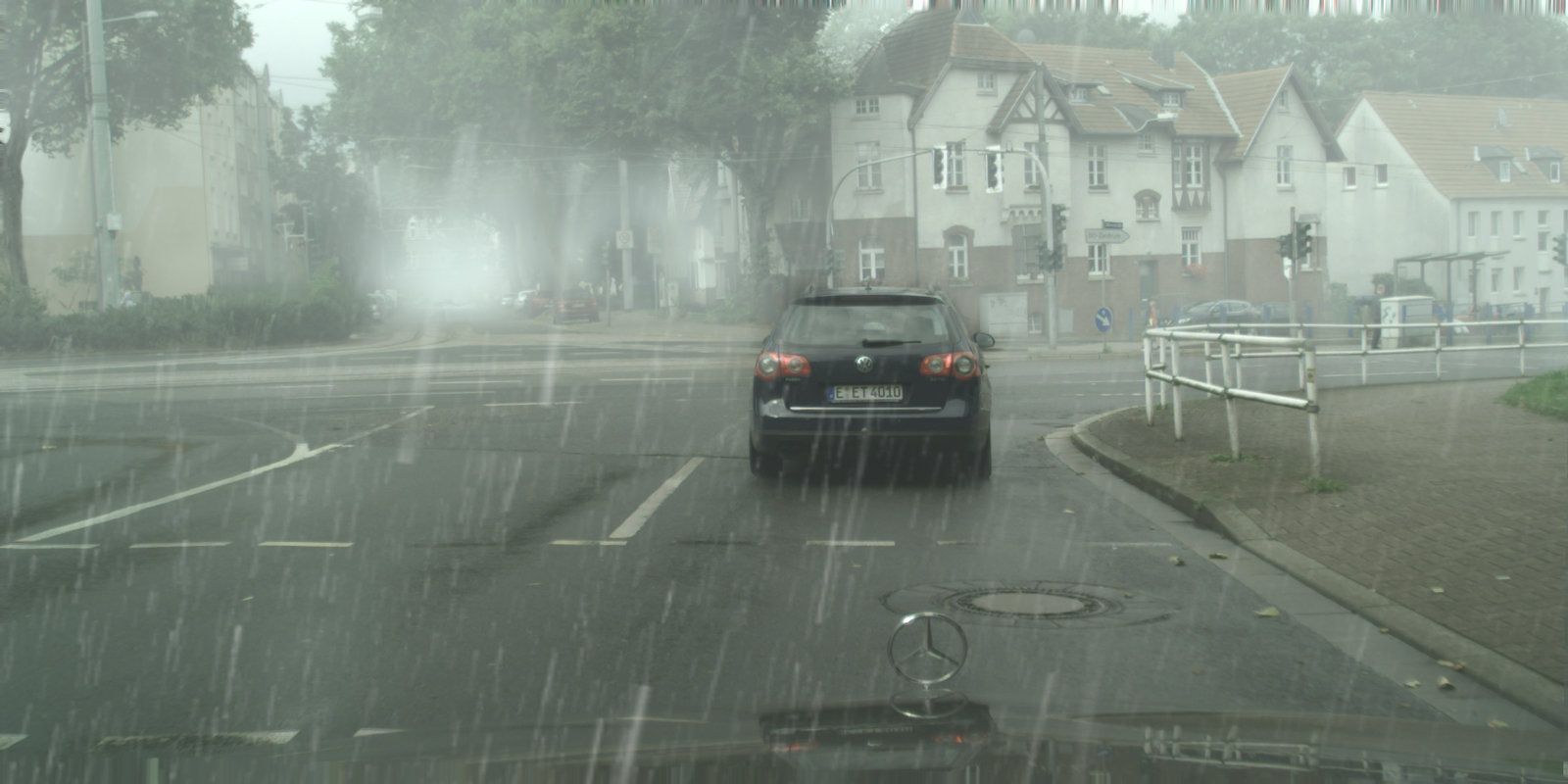}};
  \node[anchor=north west, text=white, font=\scriptsize\bfseries,
        fill=black!40!white, fill opacity=0.8, text opacity=1,
        inner sep=1pt, rounded corners=1pt]
        at ([yshift=-1pt]img.north west)
        {\textbf{Cityscapes-Rain}~\cite{DepthRainRemoval}};
  \spy on ($(img.south west)+(0.25,0.25)$) in node [right] at (1.05,0);
\end{tikzpicture} &
\includegraphics[width=0.245\textwidth]{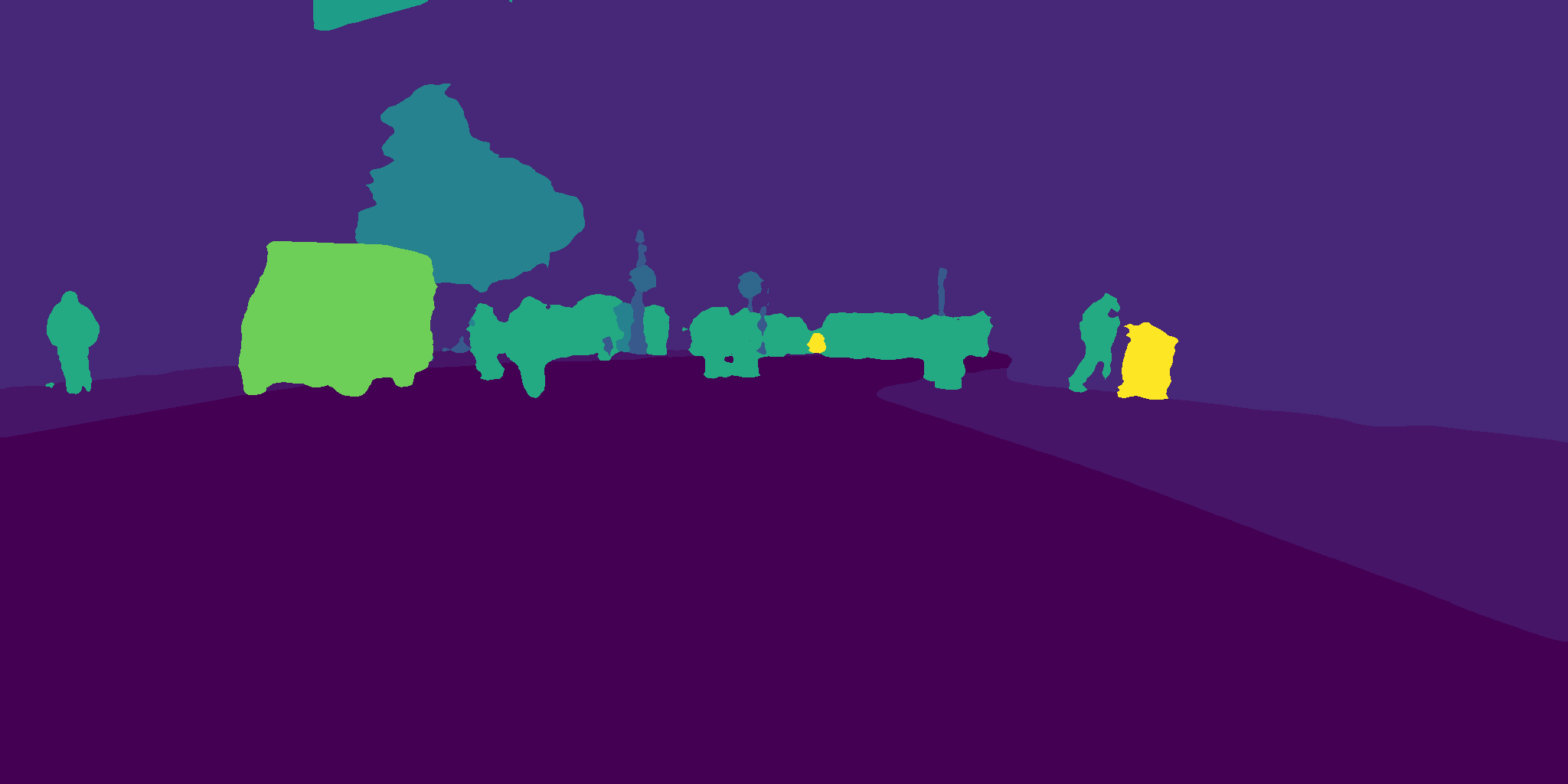} &
\includegraphics[width=0.245\textwidth]{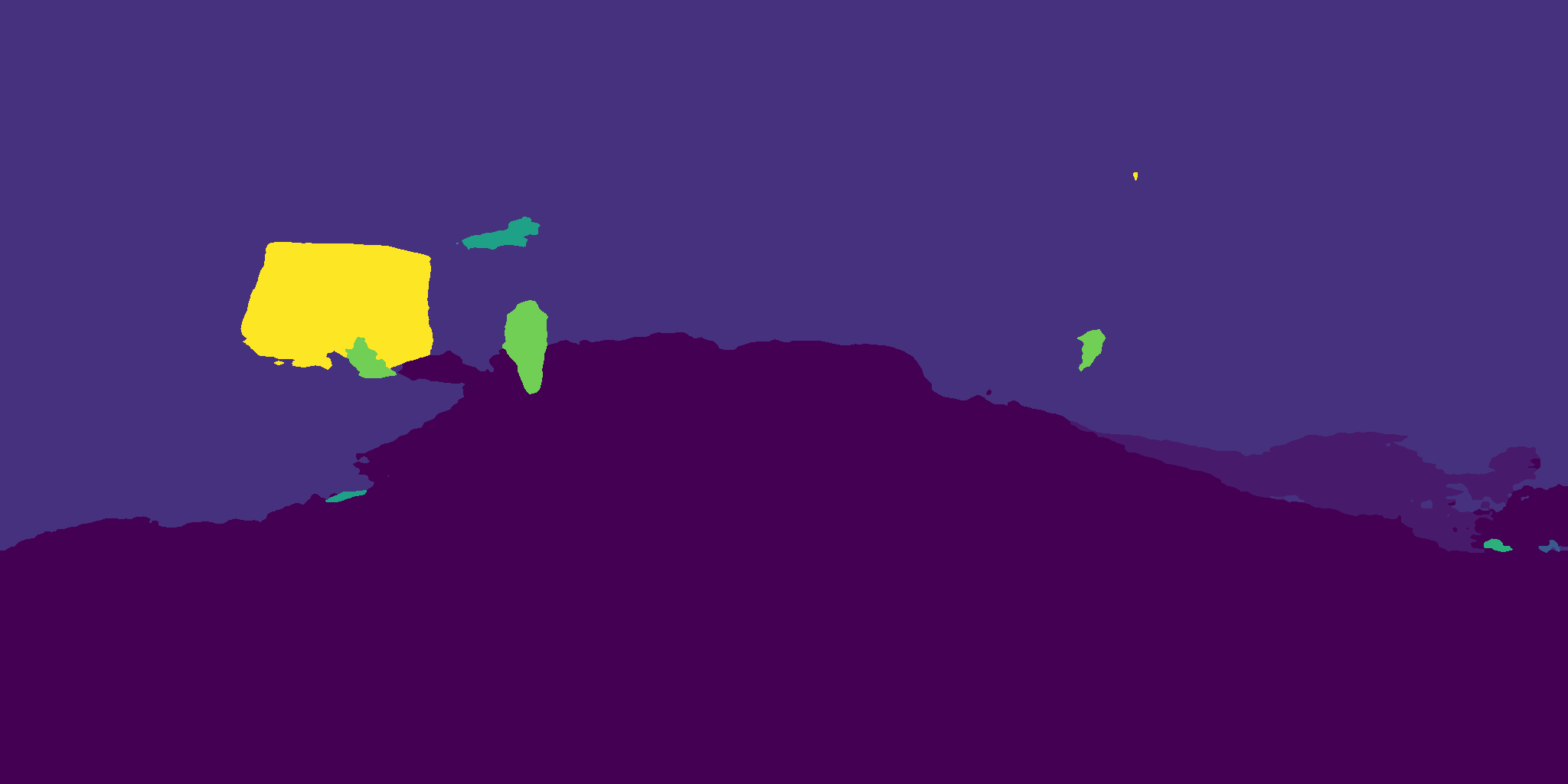} \\[-0pt]

\multicolumn{2}{c}{
  \makebox[0pt]{
  \begin{tikzpicture}[spy using outlines={
    rectangle, red, line width=0.4pt, magnification=2.3,
    connect spies, size=1.0cm}]
    \node[anchor=center, inner sep=0] (img)
      {\includegraphics[width=0.245\textwidth]{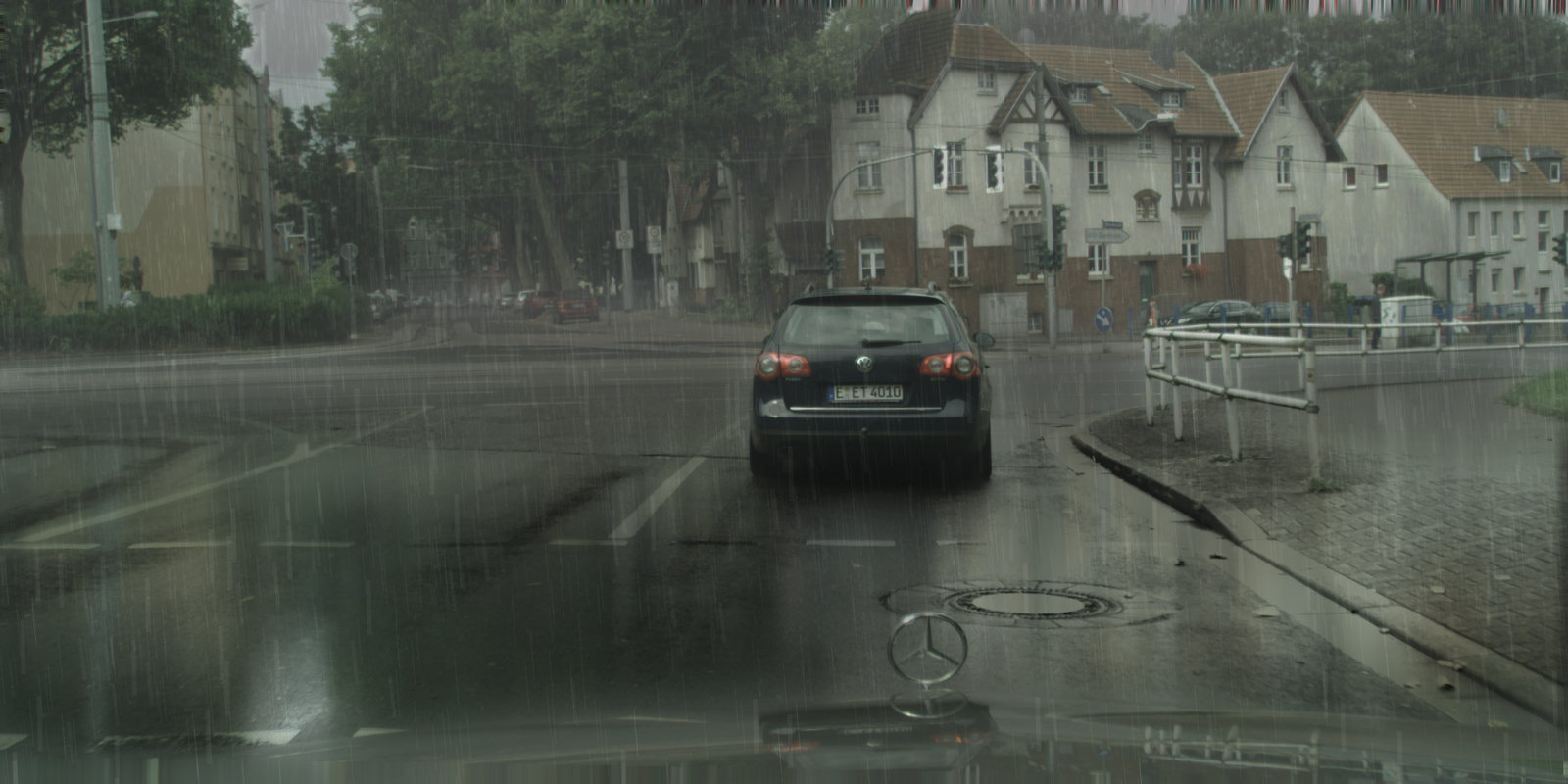}};
    \node[anchor=north west, text=white, font=\scriptsize\bfseries,
          fill=black!40!white, fill opacity=0.8, text opacity=1,
          inner sep=1pt, rounded corners=1pt]
          at ([yshift=-1pt]img.north west)
          {\textbf{CascadeRS (ours)}};
    \spy on ($(img.south west)+(0.25,0.25)$) in node [right] at (1.05,0);
  \end{tikzpicture}}
} &
\includegraphics[width=0.245\textwidth]{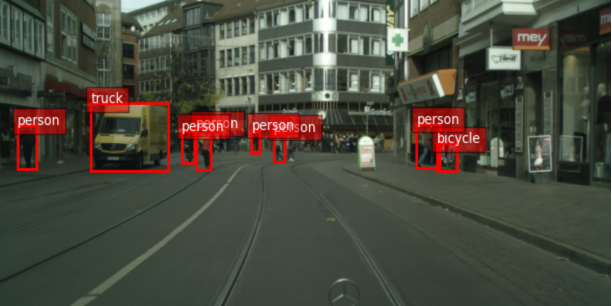} &
\includegraphics[width=0.245\textwidth]{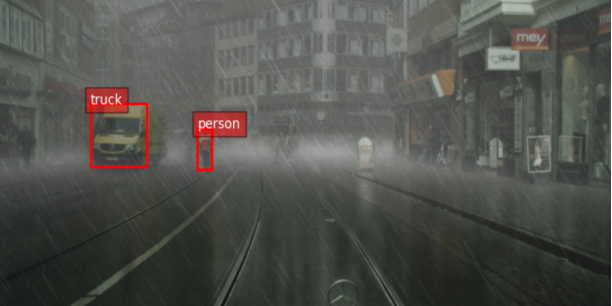} \\
\\[-5mm]
\multicolumn{2}{c}{\scriptsize\textbf{Comparison of rain rendering methods}} &
\multicolumn{2}{c}{\scriptsize\textbf{Effects of rain - clean (left) vs rainy (right)}} \\

\end{tabular}
}

\vspace{-4pt}
\caption{\textbf{Rain rendering and its impact on perception.}
In the first two columns we show the output of different rain rendering methods. 
Our CascadeRS approach produces the most realistic puddles and reflections, which are absent in prior synthetic datasets. The third and fourth columns illustrate how realistically rendered rain affects downstream vision tasks, such as semantic segmentation and object detection, where rain causes visible performance degradation. Background images are from \cite{Cityscapes}.}
\label{fig:combined_rain_impact}
\vspace{-4mm}
\end{figure*}


Rain is a common environmental factor that degrades image quality and challenges outdoor vision tasks such as autonomous driving and traffic surveillance~\cite{Wang2025,Jiang2024,Sang2024,Su2024}. 
It introduces visual distortions (streaks, blur, reduced contrast) and surface reflections that obscure details, reduce visibility, and alter object appearance (see Figure~\ref{fig:combined_rain_impact}).



Common strategies to address model prediction robustness under rain include de-raining~\cite{ZHANG2023Datadriven,Chen2025DerainingSurvey} and rain synthesis~\cite{Hoang2020DataAugmentation}. 
On one side, de-raining methods aim to recover a clean image from a rainy input. However, as supervised approaches, they require paired rainy–clean data, which are not available under real-world conditions. Despite good results~\cite{JORDER,DepthRainRemoval,SGI,IDT,zheng2023}, they often overfit to specific rain types due to the use of non-realistic synthetic training data with limited rain patterns, and may overcorrect by removing fine background details~\cite{deng2019drdnetdetailrecoveryimagederaining,chen2021robustrepresentationlearningfeedback,Wen_2024}. 
On the other side, synthesizing rain effects enables the training of vision models with rain-affected images and systematic evaluation of  performance under adverse rain conditions. It is a less costly approach compared to collecting real-world rain image data. Existing methods~\cite{Pang2025, Rain1400, DID-MDN, Zhang2020BeyondMonocular} oversimplify the modeling of rain using uniform streak patterns. Because they lack depth-based variability, realistic fog, proper interaction with lighting, and occlusion by scene objects, the generated rain exhibits low perceptual realism. Wet reflective surfaces and puddles also affect image perception, but are often not considered and modeled in existing methods \cite{SGI,IDT,DepthRainRemoval}. Examples are shown in Figure~\ref{fig:combined_rain_impact}, where methods like PBR \cite{halder2019physicsbased} and Cityscapes-Rain~\cite{DepthRainRemoval} do not account for rain-related surface effects. Generative AI models for rain synthesis require large amounts of training data (often created with rendering engines \cite{Su2024,CHOI2022,wang2020raingenerationrainremoval}) and might overfit to certain rain types. They lack controllable and principled  generation\ns{, resulting in issues with retaining fine-grained details or misalignment between prompts and generated images. We show examples in Appendix~\ref{supmat: rain synth with genai}}.
Moreover, real rain datasets are scarce \cite{Zhang2023Weatherstream,Zhong2022RainyWCity} and  evaluations mostly rely on synthetic data. 


In this paper, we address the generation of perceptually realistic rain images with a principled approach that is grounded in scene depth and semantics priors, and further uses physics-inspired priors. We propose \textbf{Cascade Rain Synthesis (CascadeRS)}, a framework with a learning-free rain synthesis module 
that produces realistic and physics-\ns{inspired} puddles, wet-surfaces, and reflections guided by the depth and semantics scene priors to ensure perspective coherence. 
CascadeRS only receives inputs from pre-trained depth-estimation and semantic segmentation models, while not requiring training data and learning itself, decomposes the scene into depth-aware layers, and generates rain and fog consistently with the scene geometry. We design it as a modular pipeline, 
enabling a controllable generation process. 
We generate the CityScape-CascadeRS dataset of rain-affected urban scenes integrating depth-guided atmospheric and surface rain synthesis, which we use to benchmark perceptual realism (quantitatively, with FID scores and qualitatively, with a user study) and impact on downstream segmentation and object detection robustness, in comparison to existing generative methods and datasets. 
\section{Related Works}
\label{sec:related_works}

\paragraph{Rain synthesis.}
Rain synthesis has been widely explored, with traditional physics-based rendering methods primarily focusing on modeling the appearance of rain streaks based on real-world optical and motion properties \cite{Garg2006Photorealistic,halder2019physicsbased}. While these approaches capture physically-plausible streaks, they do not model effects such as surface wetness, reflections, or occlusions, which are critical for perceptual realism. Fog is often included as an atmospheric complement to rain to enhance depth perception, and is a well-studied component~\cite{DepthRainRemoval,DGNL-Net,Nikolov2024Digiweather,SakaridisFog2018}. 
Several works focus on scene geometry or atmospheric scattering to improve rain placement and visibility \cite{DepthRainRemoval,halder2019physicsbased}. 
Recent deep-learning methods explored the possibility of controlling rain streak density, direction, or illumination \cite{Pang2025,zhou2025learningrendering}, and interaction with transmission maps~\cite{WangDerain2020}. 
 Generative and diffusion-based weather frameworks integrate richer volumetric or wet-surface cues as they are included in training data \cite{Dai2025RainyGS,qian2025weatheredit}, but often rely on paired training data, limiting their scalability and applicability to diverse real-world conditions. Furthermore, being end-to-end learning-based black-box models, the generative process is less controllable, thus making the quantitative evaluation of perceptual realism against other methods difficult. 
Image-to-image translation and unpaired adversarial learning approaches \cite{zheng2023,Liu_2021_ICCV} enhance visual quality of weather effects but require large domain-specific training datasets and involve heavy optimization pipelines, limiting adaptability to new environments. Similarly, de-raining methods exploit streak-based priors to simulate rain to be subsequently removed~\cite{SGI,IDT,DepthRainRemoval,Zhang2023RainNighttime}. They demonstrate the importance of modeling rain streaks even at nighttime, yet do not consider wet surface or reflective cues~\cite{Yan2025Derain}.
Meanwhile, geometry-aware rendering leverages depth and occlusion cues for spatial consistency~\cite{zhan2025larender,zhou2024hugsim}, underscoring the importance of considering scene geometry. However, their primary goal is not weather synthesis, and they showcase rain effects as a by-product. 
Existing methods mainly focus on rain streak synthesis, not considering consequent effects like puddles and surface reflections, which limits the realism of synthesized images and their feasibility to simulate real-world conditions in training or benchmarking settings. Our method advances rain synthesis beyond streak-level generation and improves perceptual realism by integrating scene semantics and depth-guided rendering of atmospheric effects with physics-\ns{inspired} surface effects, including reflections, puddles and fog/vapor.

\paragraph{Surface interactions.}
 Perceptual realism in rainy scenes depends not only on atmospheric effects but also on surface interactions, such as formation of wet-road reflections and puddles, which alter light transport, visibility, and scene appearance in ways that significantly impact both human observers and vision models~\cite{Paulin2023,Zhang2025Robustroad}. Surface effects such as cloud cover and puddles degrade segmentation performance more than rain streaks alone, underscoring their impact on performance of vision tasks~\cite{Khan_2019_ProcSy}. However, puddles remain underrepresented in autonomous driving and outdoor vision datasets~\cite{Kumar2023Waterpuddle,Zhang2025Robustroad}, despite their relevance to perception challenges and road safety. 
 While recent generative models for weather synthesis attempt to simulate puddles, they rely on generating large numbers of images and retaining only the top 4\% with the highest CLIP score~\cite{lin2025controllableweathersynthesisremoval}. This highlights limited reliability of controlling and scaling current generative approaches. 
In the Appendix~\ref{supmat: rain synth with genai}, we show how prompt-based generative models lack control and leave behind details of the original images. 

Existing works have laid the foundations to modeling rain streaks and learning-based weather synthesis, yet remain limited in capturing wet surface effects, considered key contributors to the synthetic-to-real gap \cite{Zhang_2023_survey}, and explicitly addressing the perceptual realism of generated effects. Among them, in Table~\ref{tab:related_comparison} we report single-image rain synthesis methods with public code, that address both physics- (PBR~\cite{halder2019physicsbased}, Cityscapes-Rain~\cite{DepthRainRemoval}) and learned-based paradigms (TPSeNCE~\cite{zheng2023}). PBR and Cityscapes-Rain render rain streaks and fog but omit puddles and reflections; TPSeNCE is GAN-based and able to synthesize surface effects, although without explicit control w.r.t. rain intensity. Furthermore, the characteristics of generated wet effects are tighten to the training data, making TPSeNCE suffer from displacement artifacts, which is also the case of~\cite{Tremblay2021} that extends PBR with a GAN model, and low-resolution.
With our CascadeRS, we generate realistic reflections and puddles alongside rain and fog in a procedural way, addressing the limitations of existing rain synthesis methods that do not account for wet surface effects. The principled design of CascadeRS enables a step-by-step controllable image synthesis, differently from the non-transparent end-to-end generative AI methods.

\begin{figure*}[ht]
    \centering
    \includegraphics[width=\linewidth]{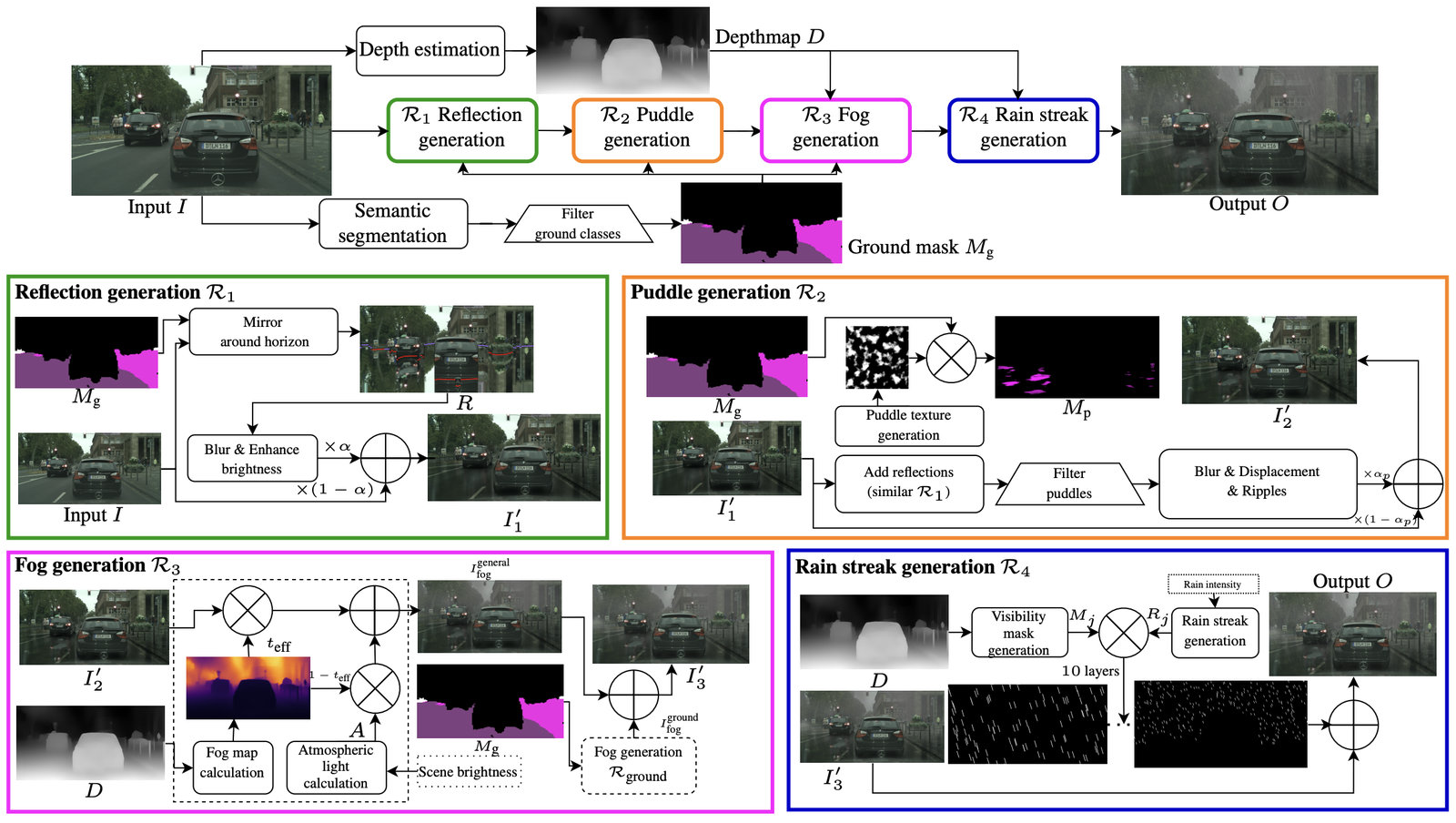}
    \vspace{-7mm} 
    \caption{\textbf{CascadeRS overview.} Overall pipeline (top row): depth and semantic cues guide sequential surface effects and rain synthesis. The colored boxes below illustrate the detail of each stage: reflection, puddle, fog, and  streak generation.}
    \label{fig:method-pipeline}
    \vspace{-5mm}
\end{figure*}

\section{Method: CascadeRS}
\label{sec:Method}

We propose CascadeRS,
that
leverages scene priors as depth and semantic maps, to procedurally generate rain and wet surface effects consistently with the scene geometry. 
Puddle shapes are generated using a physics-inspired approach~\cite{Perlin1985}, and placed on the image following local scene depth and surface semantics. Reflections are rendered using view-dependent image warping for spatial consistency. Rain and fog synthesis further couples scene-depth with surface accumulation effects, achieving full-scene weather rendering without dataset-specific training.

In Figure~\ref{fig:method-pipeline}, we show an overview of the proposed pipeline. Given an image $I \in \mathbb{R}^{H \times W \times 3}$ of a scene, we generate a rainy version $\hat{I}_{\text{rain}}$ 
as a parameterized transformation:
\vspace{-2mm}
\begin{equation}
    \hat{I}_{\text{rain}} = \mathcal{F}(I, D, S; \theta),
    \vspace{-1.5mm}
\end{equation}

\noindent where $D$ is the normalized depth map, $S$ the semantic segmentation map, and $\theta$ a set of user-controllable parameters such as rain intensity, reflectivity, fog density, and streak angle. Depth is estimated using the pretrained \mbox{MiDaS}~\cite{midas}, while segmentation of ground regions (\ie road, sidewalk, driveway) is obtained using the pretrained SegFormer~\cite{xie2021segformersimpleefficientdesign}. In Section~\ref{sec:ablations}, we show that the scene prior models can be plug-n-play replaced. 
Instead of learning $\mathcal{F}$ through data-driven optimization, we defined it as a composition $\mathcal{F} = \mathcal{R}_4 \circ \mathcal{R}_3 \circ \mathcal{R}_2 \circ \mathcal{R}_1,$ of four interpretable, scene-geometry-aware sequential operators for \textit{reflections} $\mathcal{R}_1$, \textit{puddles} $\mathcal{R}_2$, \textit{fog} $\mathcal{R}_3$, and \textit{rain streaks} $\mathcal{R}_4$. 
%
This cascaded formulation reflects the perceptual coherence and overlay order of these visual effects: wet surfaces are on the ground, puddles accumulate on them, fog forms an atmospheric layer perspectively above ground, and rain streaks overlay the scene. 

\paragraph{Reflection generation.} 
We first synthesize diffuse reflections characteristic of wet asphalt or soil.
The semantic priors of the scene, namely a segmentation map $M_{\text{g}}(x, y)$ of the ground regions (\textit{road}, \textit{sidewalk}, and \textit{driveway}),
define the spatial support for reflections. 
In particular, the highest line of the semantic map containing ground pixels 
yields a horizon reference $y_i$ from which we  compute a mirrored reflection map $R(x, y) = I(x, 2y_i - y)$ (with $y < y_i$). 
This allows adaptation to varying perspectives and terrain (Figure~\ref{fig:method-pipeline}, $\mathcal{R}_1$). 
%
%
%
We account for the scattering and defocus of reflection by applying a vertical motion blur $\mathcal{V}(\cdot)$ and a Gaussian blur $\mathcal{G}(\cdot)$. As image brightness influences the appearance of reflections, we modulate them to enhance reflective appearance near bright regions such as sky or headlights. Specifically, we enhance the brightness of the original grayscale image with operator $\mathcal{B}(\ell)$, resulting in the final luminance-weighted reflection: 
{\setlength{\abovedisplayskip}{4pt}%
 \setlength{\belowdisplayskip}{4pt}%
 \setlength{\abovedisplayshortskip}{0pt}%
 \setlength{\belowdisplayshortskip}{0pt}
\begin{equation}
    R'(x, y) = \mathcal{G}\!\big(\mathcal{V}(R(x, y))\big)\, \mathcal{B}(\ell(x, y)).   
\end{equation}

\noindent Finally, we blend the reflection with the original image using the semantic mask of the ground $M_{\text{g}}$ as a spatial guide to compute a blending coefficient ${\alpha(x, y) = \lambda_\alpha \mathcal{G}\!\big(M_{\text{g}}(x, y)\big)}$, where $\lambda_\alpha{=}0.75$ controls global reflection strength and $\mathcal{G}$ (kernel size $201\times201$) 
ensures smooth transitions at mask boundaries. The image with added reflections is:
\begin{equation}
    I'_1(x, y) = (1 - \alpha(x, y))\, I(x, y) + \alpha(x, y)\, R'(x, y).
\end{equation}


\paragraph{Puddle generation.}
We synthesize puddles using multi-octave Perlin noise~\cite{Perlin1985}, 
which provides a spatially coherent, multi-scale stochastic field commonly used in graphics and simulation pipelines to emulate natural heterogeneity (e.g., terrain and water-like patterns)~\cite{perlinwater,perlinewater2}.
This step is not a hydrodynamic simulation; instead, Perlin noise acts as a controllable morphological prior that approximates irregular pooling boundaries induced by  surface micro-geometry.
We generate a water puddle mask $M_p(x,y)$ by thresholding a Perlin noise texture $\mathcal{T}(x, y) = \sum_{i=0}^{n} a^{i} \, \eta(f_i x, f_i y)$,
where $\eta$ denotes Perlin noise at spatial coordinates $(x, y)$ with frequency $f_i$, and amplitude scaled by a factor $a$. We ensure that puddles are applied on ground regions by multiplying the puddle mask by the ground mask $M_{\text{g}}$. 
We explicitly account for reflections within puddles and synthesize them using the approach in $\mathcal{R}_1$, resulting in $R_p(x,y)$. 
We further generate surface diffusion and motions with a sinusoidal wave map $\mathcal{D}_w(\cdot)$ that introduces low-frequency undulations, and a stochastic displacement $\mathcal{D}_c(\cdot)$ that adds high-frequency irregularities. Their composition produces a distorted reflection that captures the dynamic appearance of shallow water under rainfall. The result is further modulated by scene brightness to attenuate strong light sources and enriched by ripple perturbations via $\mathcal{R}_{\text{r}}(\cdot)$, which adds micro-splashes from elliptical noise textures to mimic raindrop impacts:
\begin{equation}
    R_{p,\text{final}}(x, y)
    = \mathcal{R}_{\text{r}}\!\big(\mathcal{D}_c(\mathcal{D}_w(R_p(x, y)))\, (1 - \ell(x, y))\big).
\end{equation}
The ripple-enhanced reflection is then blended with the previous stage output $I'_1$ using a spatially varying reflectivity coefficient $\alpha_p(x, y)=\lambda_p M_p(x, y)$ that takes into account the luminance map $L(x, y)$ derived from the RGB channels
~\cite{Luminance}. 
The image with added puddles is thus:
\begin{equation}
    I'_2(x, y) = (1 - \alpha_p)\, I'_1(x, y)
    + \alpha_p\, R_{p,\text{final}}(x, y).
\end{equation}

\paragraph{Fog generation.}
We add fog-like effects, 
using an atmospheric scattering model~\cite{1211417}, commonly employed in rain synthesis~\cite{DepthRainRemoval,Li2019HeavyRain}, and produce image $I'_3$ as: 
\begin{equation}
    I'_3(x, y) = I'_2(x, y)\, t_{\text{eff}}(x, y) + A\, (1 - t_{\text{eff}}(x, y)),
\end{equation}
where both the transmittance $t_{\text{eff}}$ and atmospheric light $A$ are determined by global scene properties, including depth, average brightness, and rain intensity (details are in Appendix~\ref{supmat: method implementation details}).
A near-ground fog layer is also synthesized to reproduce thin mist accumulating near wet surfaces during heavy rain. It is generated by blurring the ground mask $M_{\text{g}}$ and uses depth proximity to create a transmittance map that concentrates fog near the road plane. 
\begin{table}[!t]
\centering
\caption{\textbf{Comparison of rain synthesis.}
CascadeRS combines multiple weather effects with a training-free (TF) approach.}
\vspace{-2mm}
\label{tab:related_comparison}
\scriptsize
\setlength{\tabcolsep}{2.5pt}
\renewcommand{\arraystretch}{0.9}
\begin{tabular}{lcccccccc}
\toprule
\textbf{Method} & Rain & Fog & Pudd. & Refl. & TF & Control & Dataset agnostic & Synth. \\
\midrule
PBR~\cite{halder2019physicsbased} & \cmark & \xmark & \xmark & \xmark & \cmark & \cmark & \cmark & Physics \\
CS-Rain~\cite{DepthRainRemoval} & \cmark & \cmark & \xmark & \xmark & \cmark & \cmark & \xmark & Physics \\
TPSeNCE~\cite{zheng2023} & \cmark & \cmark & \cmark & \cmark & \xmark & \xmark & \cmark & GAN \\

\midrule
\textbf{CascadeRS} & \cmark & \cmark & \cmark & \cmark & \cmark & \cmark & \cmark & Physics \\
\bottomrule
\vspace{-8mm}
\end{tabular}
\end{table}

\paragraph{Rain streak generation.}
We synthesize rain streaks, 
as linear, semi-transparent distortions produced by falling droplets. Their appearance varies with depth: nearby streaks are sharper, brighter, and longer, while distant ones appear shorter and fainter due to occlusions and atmospheric scattering. To reproduce this behavior, we generate a stack of $N_L=10$ depth-guided rain layers, each representing a different visibility range and streak profile, and composite them over the fog-enhanced image~$I'_3$.
The layers differ in density, thickness, blur, and visibility range, enabling a smooth transition from prominent foreground streaks to subtle background ones. Several characteristics, including motion blur and streak visibility, adapt dynamically to the global rain intensity, producing longer and denser streaks under heavier rain. Streak orientation is stochastically perturbed around a near-vertical base direction to avoid uniform patterns and improve realism.
Each layer $j$ is generated by sampling sparse binary noise $\Psi_j(x, y)$, transforming it into elongated streaks using  motion-blur ~$\mathcal{V}$, and smoothing it with a Gaussian~$\mathcal{G}$:
\begin{equation}
    R_j(x, y) = \mathcal{G}\!\big(\mathcal{V}(\Psi_j(x, y))\big).
\end{equation}

Depth-aware composition is achieved by filtering each layer with a visibility mask $M_j(x, y)$ derived from the normalized depth map $D(x,y)$. Each layer is assigned a depth interval $(d_j^{\min}, d_j^{\max}]$, and the mask activates only where the scene depth falls within this range. This ensures that foreground layers produce prominent streaks, while background layers contribute with subtler, distant streaks.
The final rain-enhanced image is obtained by accumulating the layers of visible streaks over the fog-rendered input:
\begin{equation}
    O = I'_4(x, y) = I'_3(x, y) + \sum_{j=1}^{N_L} M_j(x, y)\, R_j(x, y).
\end{equation}


\section{Validation}
\label{sec:experimental_setup}
We compare CascadeRS to existing rain synthesis methods (with available code) and real rain scenes. We assess the realism and perceptual quality of the generated rain effects with an extensive user study, and a llm-as-judge evaluation. As our objective is to evaluate robustness of downstream vision tasks (segmentation and object detection) to real adverse weather rather than synthetic image restoration, we do not include image de-raining benchmarks. 
We also perform an ablation study to analyze the contribution of puddles and secondary rain effects, scene semantic and depth priors, and the impact of hyperparameters.
\paragraph{Datasets.}
We use the Cityscapes dataset~\cite{Cityscapes} 
as the baseline on which we apply rain synthesis. We generate CityScapes-CascadeRS starting from the validation set of Cityscapes, and produce images with ten rain intensity levels (see Appendix~\ref{supmat: effect of rain intensity}) to systematically analyze the effect of rain severity on perceptual realism and model robustness. 
We also use the RID~\cite{Li2019_realrain} and ACDC~\cite{sakaridis2024acdcadverseconditionsdataset} datasets, to quantify and compare the perceptual realism of rain synthesis models, and evaluate segmentation performance under real rain conditions.  RID contains outdoor scenes captured under varied rain intensities and lighting, while  the ACDC dataset provides real images captured under diverse adverse weather conditions such as rain, fog, snow, and nighttime. 
We further show the generalizability of our method \ns{(without any parameter tuning)} to other car-view datasets (KITTI~\cite{Geiger2013IJRR}, BDD100K~\cite{yu2020bdd100kdiversedrivingdataset} and Roadmarks~\cite{Wu2012Roadmarks}) and to side-view scenes (nuScenes dataset~\cite{nuscenes}).

\paragraph{Baselines.}
We compare CascadeRS against three publicly available synthetic rain generation methods and two real-world rainy datasets. The baselines are 
PBR~\cite{halder2019physicsbased},  TPSeNCE~\cite{zheng2023}, and Cityscapes-Rain~\cite{DepthRainRemoval} (see Table~\ref{tab:related_comparison})
\footnote{Since the code for RainyGS and WeatherEdit~\cite{Dai2025RainyGS,qian2025weatheredit} is not publicly released, we do not include them in our baselines.}. A qualitative comparison of the rain synthesis methods is included in Appendix~\ref{supmat: qual. comparison of rain synth methods}.

\subsection{Perceptual Realism}
\label{sec:realism}
We conducted a user-study\footnote{Approved by the Ethics committee of the main authors' institutions} via an online questionnaire 
and distributed through Amazon Mechanical Turk. 
We collected responses from 77 participants, and compared CascadeRS and the baselines. Furthermore, we studied the alignment of perceived realism between human and vision models, by replicating the evaluation using Gemini 2.5 Flash in an \textit{LLM-as-judge} evaluation setting. 

\begin{figure}[!t]
    \centering
        \includegraphics[width=\linewidth]{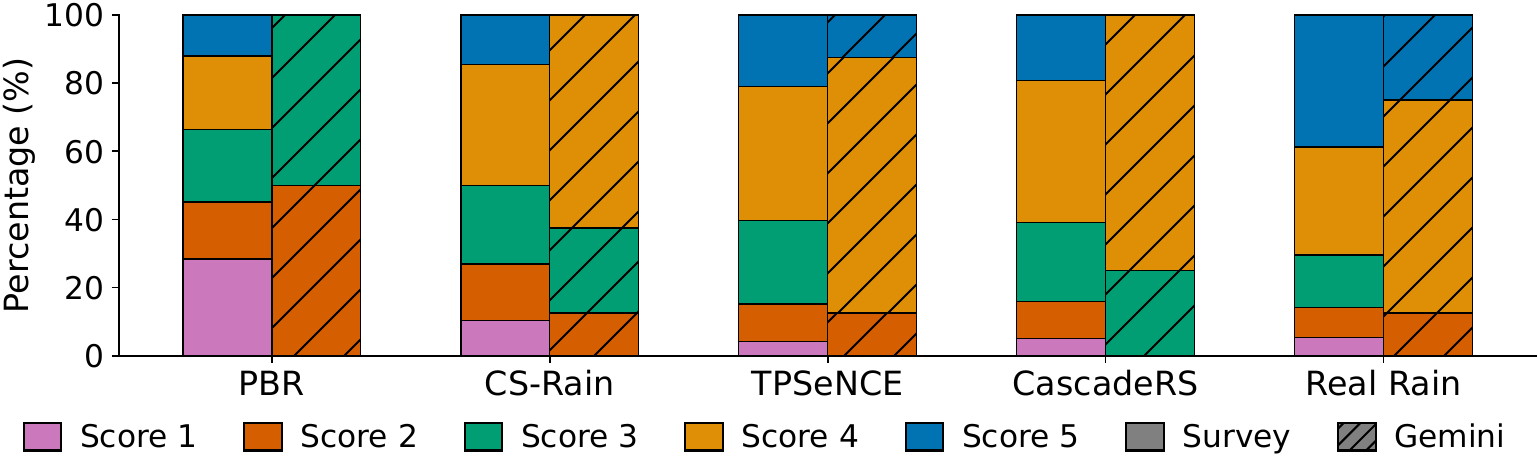}
        \captionof{figure}{\textbf{Realism ratings per method}, comparing user study (plain bars) and LLM-as-judge (striped bars). The stacked bars indicate score percentage (1: not realistic, to 5: very realistic). Detailed rating percentages are in Appendix~\ref{supmat: survey details}.}
        \label{Fig - experiments - single image scores}
        \vspace{-6mm}
    \end{figure}

\paragraph{Single-shot realism evaluation.}
We selected eight images and synthesized rainy versions with each method. 
For methods that allow control over rain intensity (PBR, Cityscapes-Rain and CascadeRS), we generate images that uniformly cover a range of weak, moderate, and heavy rainfall conditions (see Appendix~\ref{supmat: quant. eval of rain synth methods} and~\ref{supmat: survey details}). 

Participants were asked to rate the perceptual realism of each image individually using a five-point Likert scale (1 = Not realistic, 5 = Very realistic).
We report the distribution of the scores achieved by each method in Figure~\ref{Fig - experiments - single image scores}. Real rain images were rated as the most realistic, confirming the consistency of participant judgments.  CascadeRS achieved high realism ratings (19.3\% Score-5 and 41.6\% Score-4 responses), aligned with TPSeNCE (20.9\% Score-5 and 39.4\% Score-4 responses). However, in direct ranking comparisons (see next paragraph), CascadeRS clearly outperforms TPSeNCE.
Although TPSeNCE attempts to produce wetness effects and scores well in our study, it suffers from scene-geometry displacements (puddles that may result outside or not perspectively well-aligned with ground regions), and limited control of rain intensity. To not penalize TPSeNCE due to these inconsistencies, we generated several images and discarded those with improper effects. Thus, only higher-quality images generated by TPSeNCE were included in the study. In contrast, we did not specifically discard/select images generated with CascadeRS, as its principled approach ensures respecting scene geometry and allows for direct control of the intensity of rain effects. 
LLM-as-judge evaluations ranked similarly to the user survey: as these depend on the selected model and prompts, we consider the results complementary to, rather than a replacement for, the human study.

\paragraph{Ranking image realism.}
Participants were shown 14 sets of four rainy images, each set generated from the same base scene (see Figure~\ref{fig:combined_rain_impact}), and ranked the images from the most to the least realistic.  Rankings were aggregated using a Borda score, assigning 4 points for first place, 2 for second, 1 for third, and 0 for fourth. We report the scores in Table~\ref{tab:borda}, which clearly show that CascadeRS images were graded as the most realistic, with TPSeNCE distanced by a substantial margin.  
The LLM-as-judge assigned the highest score to Cityscapes-Rain and a lower score to TPSeNCE. While the reasons for this difference are unclear, it may reflect factors like image downscaling required for TPSeNCE processing or  how Gemini weighs visual cues. 
Overall, these findings highlight that human perception of CascadeRS as generating the most realistic and high-quality rain effects, considerably improving on physics- and learning-based approaches.

\paragraph{Quantitative realism analysis.}
We compute the Fr\'{e}chet Inception Distance (FID) between the distribution of synthetic images generated by our and other methods, and that of real rainy images in the RID~\cite{Li2019_realrain} and ACDC~\cite{sakaridis2024acdcadverseconditionsdataset} datasets, which we report in
Table~\ref{tab:FID}. 
CascadeRS produces achieves the lowest (best) FID score across the RID dataset, highlighting the visual realism of our generated images. The higher FID scores obtained on the ACDC dataset, are likely due to the fact that ACDC captures in-car viewpoints with frequent raindrop-on-lens artifacts, a phenomenon not currently modeled in CascadeRS and rarely synthesized by other methods. 
This also points out the similarity of TPSeNCE~\cite{zheng2023} to ACDC, as its training images include post-rain and raindrop effects. Additional KID, LPIPS, SSIM, and PSNR results are provided in Appendix~\ref{supmat: quant. eval of rain synth methods}.

\begin{figure}[!t]
\centering
\begin{minipage}{0.48\linewidth}
        \centering
        \setlength{\tabcolsep}{2pt}
        \renewcommand{\arraystretch}{1.1}
        \captionof{table}{\textbf{Borda ranking results.} Higher scores indicate greater perceived realism; CascadeRS ranks highest in the human study, while Gemini favors Cityscapes-Rain.}
        \label{tab:borda}
        \scriptsize
        \begin{tabular}{@{}lcc@{}}
        \toprule
         & \multicolumn{2}{c}{\textbf{Borda Score} $\uparrow$} \\
        \cmidrule(lr){2-3}
         
        \textbf{Method} & \textbf{User-study} & \textbf{LLM-as-judge} \\
        \midrule
         PBR & 1131 & 10 \\
         CS-Rain & 1604 & \textbf{34} \\
         TPSeNCE & 1718 & 10 \\
         CascadeRS & \textbf{2015} & 30\\
         \bottomrule
        \end{tabular}
\end{minipage}
\hfill
\begin{minipage}{0.48\linewidth}
        \captionof{table}{\textbf{FID scores across rain simulation methods} on ACDC and RID, averaged over different intensity levels (results per intensity are in Appendix~\ref{supmat: quant. eval of rain synth methods}).}
        \centering
        \label{tab:FID}
        \setlength{\tabcolsep}{5pt}
        \renewcommand{\arraystretch}{1.1}
        \scriptsize
        \begin{tabular}{lcc}
        \toprule
        ~ & \multicolumn{2}{c}{\textbf{FID} $\downarrow$} \\
        \cmidrule{2-3}
        \textbf{Method} & \textbf{ACDC} & \textbf{RID} \\
        \midrule
        PBR & 200.73 & 167.06 \\
        CS-Rain & 150.79 & 147.96 \\
        TPSeNCE & \textbf{111.87} & 113.64 \\
        CascadeRS & 123.33 & \textbf{98.32} \\
        \bottomrule
        \end{tabular}
\end{minipage}
\end{figure}

\subsection{Evaluation on downstream tasks}

\paragraph{Downstream task robustness in rainy scenes.}
We evaluate the robustness of three pretrained segmentation models, SegFormer~\cite{xie2021segformersimpleefficientdesign}, FastSeg~\cite{zhang2020fastseg}, and GCNet~\cite{yang2025goldencudgel}, on scene with increasing rain intensity generated with CascadeRS. Figure~\ref{fig:experiments-boxplots} shows the distribution of per-image mIoU 
with ACDC and the original Cityscapes validation set providing real-rain and clear-weather references, respectively. 
For SegFormer, CascadeRS substantially overlaps with the performance observed on ACDC, while FastSeg and GCNet show stronger degradation, particularly at higher rain intensities (see Appendix~\ref{supmat: cross-dataset segm analys.} and ~\ref{supmat: effect rain intensity on segmentation}). This indicates that the challenges induced by CascadeRS with controllable rain intensity make strong robustness stress tests for existing segmentation models. PBR causes the greatest degradation, with performance substantially below that on real rain due to its more extreme yet unrealistic streak effects. Cityscapes-Rain and TPSeNCE generally remain closer to clear-weather performance, thus making them less useful for investigating robustness issues in segmentation. 
We also evaluate the robustness of the DINO-teacher~\cite{lavoie2025large} object detector to rainy images and observe similar trends in results to the segmentation tests (see Appendix~\ref{supmat: object detection}).
CascadeRS spans a broad range of segmentation challenges while providing controllable rain intensity and high perceptual realism. These results support its use as a systematic benchmark for evaluating downstream vision robustness in adverse weather conditions.

\begin{figure}[!t]
    \centering

    \includegraphics[width=\columnwidth]{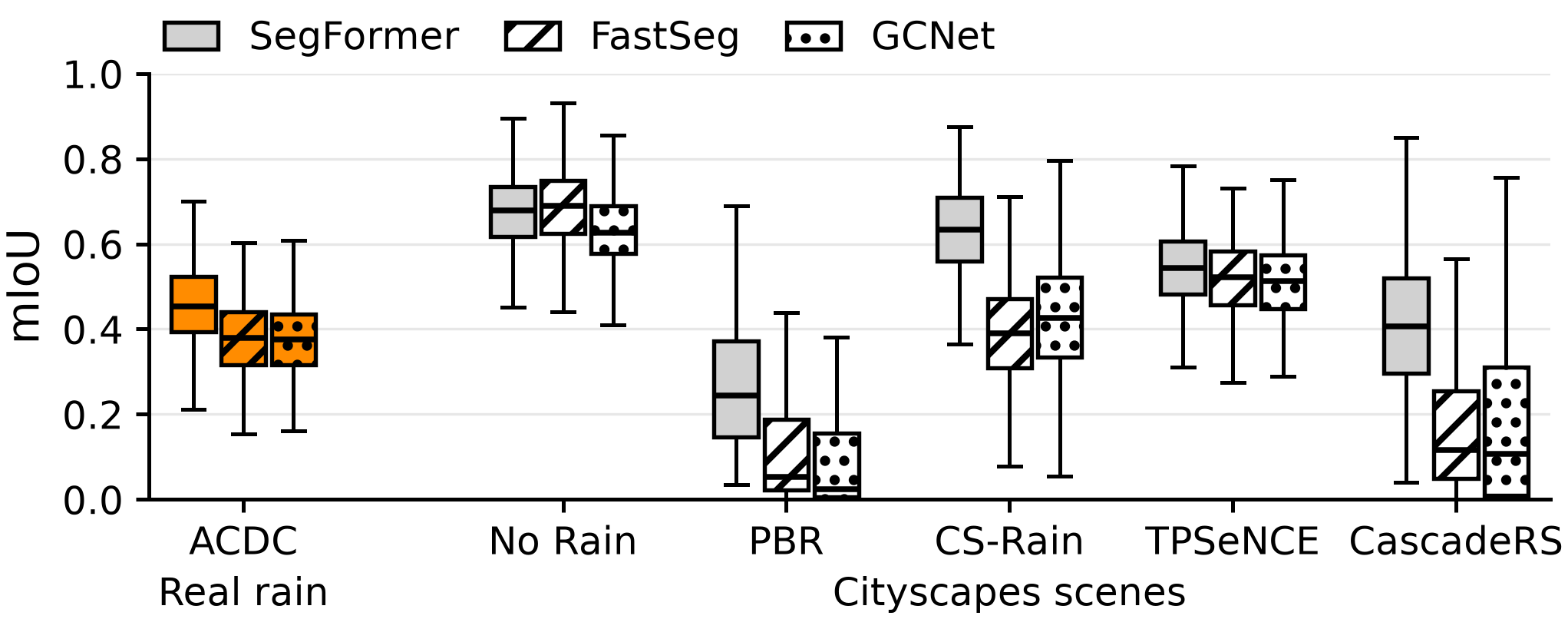}
    \caption{\textbf{mIoU across datasets.}
    Boxplots show per-image mIoU aggregated over all images and, where applicable, rain intensities. ACDC \cite{sakaridis2024acdcadverseconditionsdataset} contains real rainy images.}
    \label{fig:experiments-boxplots}
\vspace{-5mm}
\end{figure}

\paragraph{CascadeRS augmentations for segmentation.}


We investigate synthetic rain as data augmentation for SegFormer-B2 and GCNet: we compare CascadeRS (CRS) against Cityscapes-Rain (CS-R) using a 91/9\% Cityscape original/synthetic split for training, sampled across different rain intensity levels (limited by the availability of Cityscapes-Rain scenes).  Both models are evaluated on real rain from ACDC and clean Cityscapes validation (Table~\ref{tab:segmentation-augmentation}). CascadeRS improves FWIoU on ACDC for both SegFormer-B2 (0.7824~$\rightarrow$~0.8059) and GCNet (0.7241~$\rightarrow$~0.7958), surpassing Cityscapes-Rain; mIoU remains approximately unchanged for SegFormer-B2 and improves for GCNet. Performance on clean Cityscapes is largely preserved. These results indicate improved robustness to real rain with limited impact on clean-weather performance.

We also investigate training with 30\% rain-augmented data. For both SegFormer-B2 and GCNet, the largest and most consistent improvements are observed for \textit{road} and \textit{sidewalk}. For SegFormer-B2, IoU on these classes improves by +0.1048 and +0.1295, while for GCNet it improves by +0.1935 and +0.2344 IoU on ACDC. Large homogeneous classes such as \textit{sky} and \textit{vegetation} show only small changes, while reductions and higher variability occur mainly for rare classes such as \textit{bus} and \textit{bicycle}. The benefits of CascadeRS are effectively strong for road-surface classes. Per-class results (w/ multi-seed runs) are in Appendix~\ref{supmat: cross-dataset segm analys.}.

\begin{table}[!t]
    \centering
    \setlength{\tabcolsep}{2.5pt}
    \renewcommand{\arraystretch}{0.9}
    \caption{\textbf{Segmentation results} of SegFormer-B2 \cite{xie2021segformersimpleefficientdesign} and 
    GCNet \cite{yang2025goldencudgel} on ACDC and Cityscapes validation sets. Models are retrained on original Cityscapes (CS), Cityscapes mixed with Cityscapes-Rain (CS-R), and mixed with CascadeRS (CRS).}
    \vspace{-2mm}
    \scriptsize

    \resizebox{\columnwidth}{!}{%
    \begin{tabular}{@{}llcccccc@{}}
    \toprule
    & & \multicolumn{3}{c}{\textbf{ACDC}}
        & \multicolumn{3}{c}{\textbf{CS-val}} \\
    \cmidrule(lr){3-5}
    \cmidrule(lr){6-8}
    \textbf{Model} & \textbf{Metric}
        & \textbf{CS} & \textbf{CS-R} & \textbf{CRS}
        & \textbf{CS} & \textbf{CS-R} & \textbf{CRS} \\
    \midrule

    \multirow{2}{*}{SegFormer-B2}
        & mIoU $\uparrow$
        & 0.4259 & 0.3882 & \textbf{0.4265}
        & 0.6813 & \textbf{0.6886} & 0.6679 \\
        
        & FWIoU $\uparrow$
        & 0.7824 & 0.7753 & \textbf{0.8059}
        & \textbf{0.9034} & 0.9015 & 0.9016 \\

    \midrule

    \multirow{2}{*}{GCNet}
        & mIoU $\uparrow$
        & 0.3546 & \textbf{0.4047} & 0.4019
        & \textbf{0.7156} & 0.7142 & 0.7086 \\

        & FWIoU $\uparrow$
        & 0.7241 & 0.7760 & \textbf{0.7958}
        & \textbf{0.9106} & 0.9093 & 0.9077 \\

    \bottomrule
    \end{tabular}}

    \label{tab:segmentation-augmentation}
    \vspace{-4mm}
\end{table}

\paragraph{CascadeRS augmentations for object detection.}
We further evaluate whether synthetic rain improves robustness for object detection by retraining DINO Teacher \cite{lavoie2025large}, which uses a DINOv2 ViT-L backbone with detection heads trained  on eight Cityscapes instance classes (e.g., car, person, bus, and bicycle). We compare training on the original Cityscapes data (CS), a 91/9\% mixture with Cityscapes-Rain (CS-R)  or CascadeRS (CRS), and augmentation of the complete Cityscapes training set with an additional full CascadeRS set (CS+CRS). For the rainy training data, rain intensities are uniformly sampled from the available intensity levels. As shown in Table~\ref{tab:detection-augmentation}, none of the rainy training configurations consistently improves detection performance on ACDC without compromising Cityscapes performance, and the differences between the training settings are generally small. This result should be considered in the context of the Cityscapes detection task. Detection is restricted to eight instance classes, whereas the greatest improvements from rainy training in our semantic segmentation experiments occur predominantly for ground-related classes, such as road and sidewalk, which are not represented in the detection evaluation. Nevertheless, rain does affect detection performance: our supplementary experiments show that the pretrained DINO Teacher degrades with increasing rain intensity on Cityscapes validation images synthesized using the rain-generation methods. Thus, while rain constitutes a relevant domain shift for object detection, simply mixing synthetic rainy images into the training set does not provide a clear robustness benefit under the evaluated training schemes. Alternative strategies, such as emphasizing higher rain intensities that more closely reproduce the performance degradation observed on ACDC, may therefore be required.

\begin{table}[!t]
    \centering
    \setlength{\tabcolsep}{2.2pt}
    \renewcommand{\arraystretch}{0.9}
    \caption{\textbf{Object detection results} of
    DINO Teacher on ACDC and Cityscapes, trained on the original Cityscapes (CS), Cityscapes mixed -- by a 91/9\% ratio -- with Cityscapes-Rain (CS-R) and with CascadeRS (CRS), and  Cityscapes training set plus its complete augmented version with CascadeRS.}
    \vspace{-2mm}
    \scriptsize
    \begin{tabular}{@{}lcccccccc@{}}
    \toprule
    & \multicolumn{4}{c}{\textbf{ACDC}}
    & \multicolumn{4}{c}{\textbf{CS-val}} \\
    \cmidrule(lr){2-5}
    \cmidrule(lr){6-9}
    \textbf{Metric}
    & \textbf{CS} & \textbf{CS-R} & \textbf{CRS} & \textbf{CS+CRS}
    & \textbf{CS} & \textbf{CS-R} & \textbf{CRS} & \textbf{CS+CRS} \\
    \midrule
    AP $\uparrow$
    & \textbf{18.03} & 16.70 & 17.68 & \textbf{18.06}
    & 32.18 & \textbf{32.76} & 31.99 & 32.17 \\

    AP@50 $\uparrow$
    & \textbf{43.41} & 39.41 & 41.79 & 40.46
    & 61.16 & \textbf{62.11} & 60.64 & 61.68 \\
    \bottomrule
    \end{tabular}
    \label{tab:detection-augmentation}
\end{table}

\subsection{Ablations}
\label{sec:ablations}

We assess the contribution of the CascadeRS components: we incrementally generate rain streaks with fog, reflections, and puddles, and evaluate perceptual realism as part of the user study, and semantic segmentation performance. Examples images are in Appendix~\ref{supmat: ablation study - visual examples}.

\paragraph{Perceptual realism.}
We use base images from Cityscapes and produce versions with synthetic rain effects in three consecutive stages: (1) rain streaks with fog, (2) reflections, and (3) puddles, with randomly sampled rain intensities. User study participants ranked these images from most to least realistic. 
Each method component improves the perceptual realism of the synthesized rain effects. Borda score on streak-only images (1073) improves when adding reflections (1272), and further increases with puddles (1351). The configuration with reflections and puddles achieves the highest scores in the LLM-as-judge test (20 and 26, vs 2 with only streask), highlighting the relevance of the contributions of our method.  Reflections added on top of rain streaks contribute to the largest relative gain, showing the importance of surface interactions and depth-consistent illumination in perceived photorealism.

\begin{figure}
    \centering
    \includegraphics[width=\linewidth]{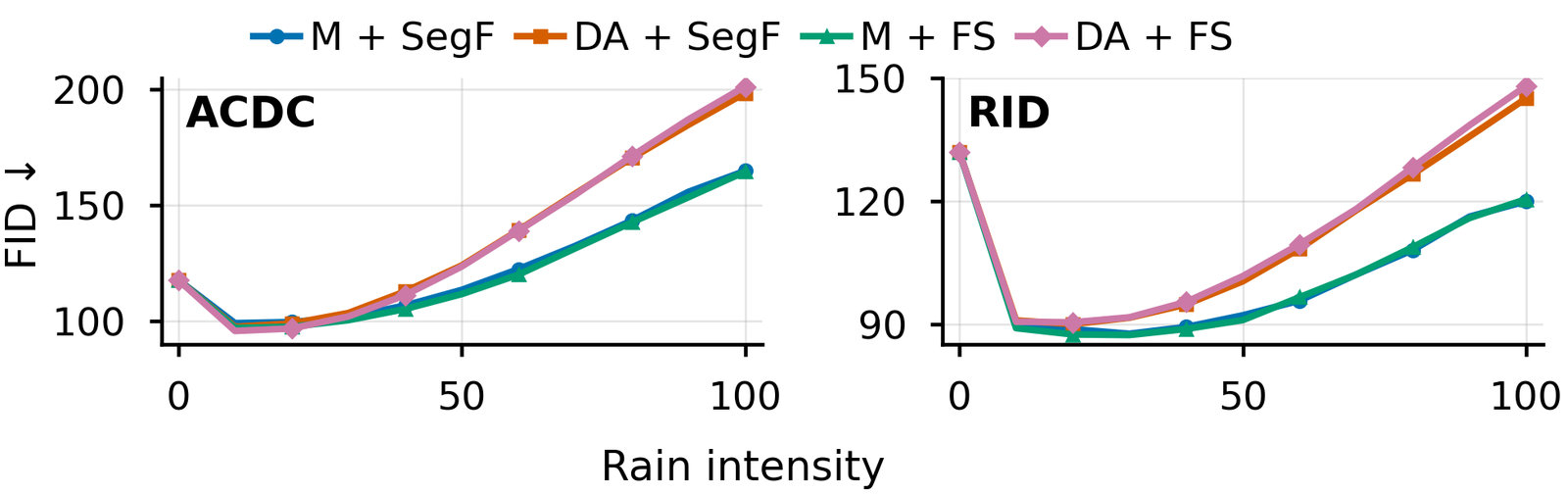}
    \vspace{-8mm}
    \caption{\textbf{FID scores of various depth and segmentation model compositions.} M = MiDaS~\cite{midas}, DA = DepthAnything~\cite{depthanything}, SegF = SegFormer~\cite{xie2021segformersimpleefficientdesign}, and FS = FastSeg~\cite{zhang2020fastseg}. Replacing the depth model has a negative effect on the FID score.}
    \label{fig:Experiments-modular-ablation}
    \vspace{-5mm}
\end{figure}

\paragraph{Depth and segmentation priors.}
We study the contribution of pretrained depth and semantic segmentation models as scene priors. We consider DepthAnything~\cite{depthanything} as an alternative to MiDaS and FastSeg~\cite{zhang2020fastseg} to SegFormer. We use ACDC as a reference, and report the FID scores achieved with different depth/segmentation priors in Figure \ref{fig:Experiments-modular-ablation}.  Replacing the depth model affects FID scores as it induces stronger fog effects at higher rain intensity. The semantic segmentation model does not determine perceptual changes. Supplementary qualitative examples and a DINOv3 UMAP analysis (Appendix~\ref{supmat: effect pretrained priors} further show that these changes primarily affect appearance, while preserving semantic image representation.

\paragraph{CascadeRS components vs segmentation robustness.}
We measure the impact of incrementally generating rain streaks, reflections and puddles on the performance of pretrained SegFormer-B2 and FastSeg. In Figure~\ref{Fig - experiments - ablation segmentation}, we report the results as the intensity of synthetic rain increases and for images with different synthetic effects. 
Among the CascadeRS components, reflections produce the largest performance drop compared to sole rain streaks, while puddles cause a slight further decline, highlighting the challenging effects that our method synthesizes. Out novel wet surface effects generation pipeline allows to simulate real rain scenes more effectively and under stronger domain shifts than those accounted for by pretrained models. 

\begin{figure}[!t]
    \centering
    \def\svgwidth{\linewidth} 
    \resizebox{0.9\linewidth}{!}{
    \input{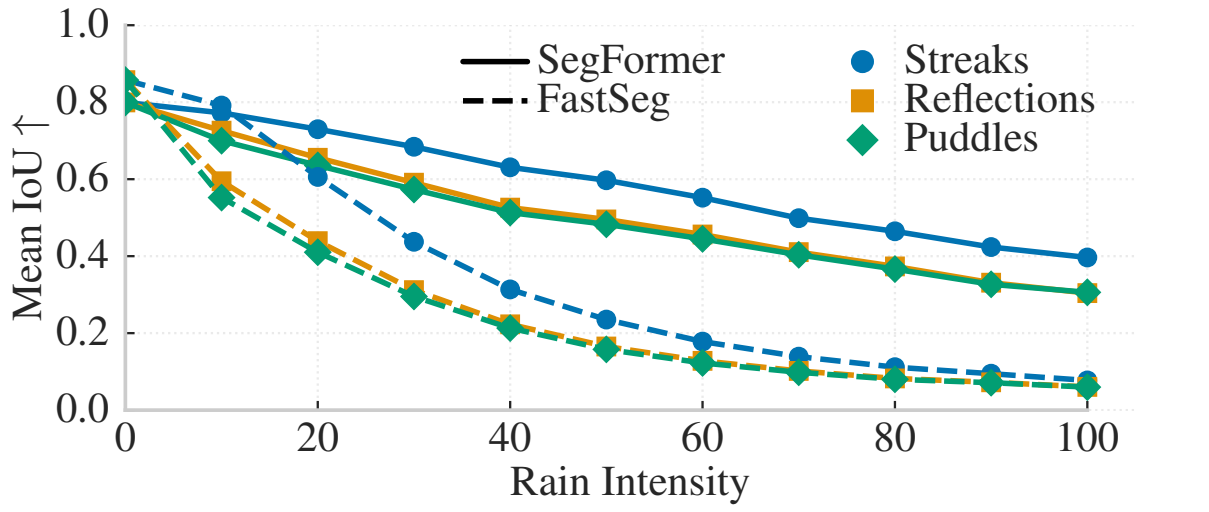}}
    \vspace{-2mm}
    \caption{\textbf{Segmentation results for the ablation of CascadeRS components} under varying rain intensities.}
    \label{Fig - experiments - ablation segmentation}
    \vspace{-6mm}
\end{figure}

\paragraph{Parameter sensitivity. }
We study the sensitivity to five principal parameters of CascadeRS (i.e. rain intensity, fog density, puddle resolution, puddle persistence, and puddle reflection) and show results of FID, LPIPS, mIoU and AP metrics in Figure~\ref{fig:paramter-sweep-compact} (for reporting purposes, parameter values are normalized in $[0,1]$).  Rain intensity is the primary user-controllable parameter that influences the formation of rain streaks, fog, and puddles; as expected, it affects visual appearance and downstream performance. Other parameters have a minor to negligible impact, indicating high stability of the output. Limited variation across puddle settings indicates robustness to their configuration. Further results are in Appendix~\ref{supmat: parameter robustness}.

\begin{figure}
    \centering
    \includegraphics[width=\linewidth]{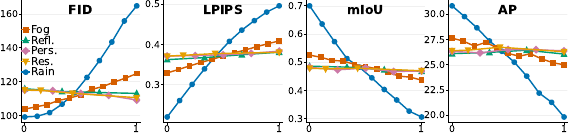}
    \caption{\textbf{Parameter sensitivity analysis.}
Effect of five CascadeRS parameters on FID, LPIPS, mIoU, and AP. Fog = fog density, Refl. = puddle reflection, Pers. = puddle persistence, Res. = puddle resolution, and Rain = rain intensity.}
    \label{fig:paramter-sweep-compact}
     \vspace{-4mm}
\end{figure}

\begin{figure}[t]
\centering
\captionsetup[subfigure]{labelformat=empty} 
\setlength{\tabcolsep}{1pt}
\renewcommand{\arraystretch}{1.0}


\begin{subfigure}[b]{0.24\columnwidth}
  \centering
  \includegraphics[width=\linewidth]{figures/dataset_render/cropped/b1c66a42-6f7d68ca.jpg}
  
\end{subfigure}
\hfill
\begin{subfigure}[b]{0.24\columnwidth}
  \centering
  \includegraphics[width=\linewidth]{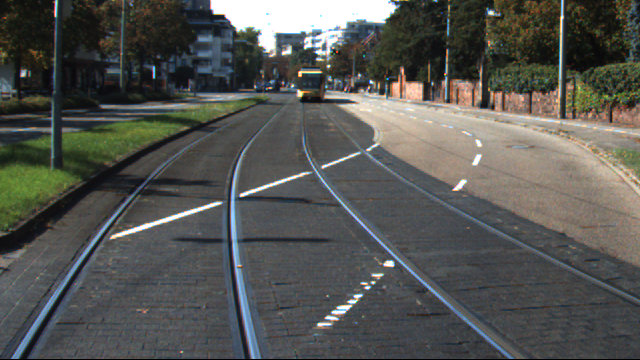}
  
\end{subfigure}
\hfill
\begin{subfigure}[b]{0.24\columnwidth}
  \centering
  \includegraphics[width=\linewidth]{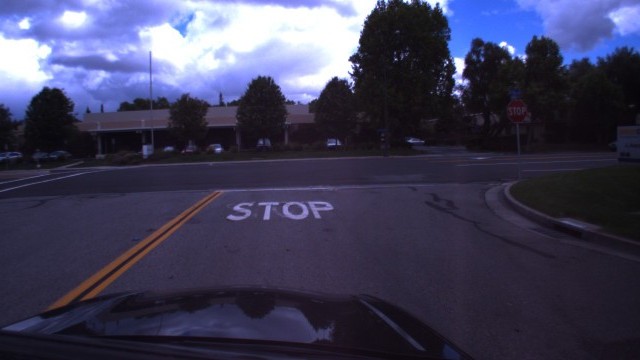}
 
\end{subfigure}
\hfill
\begin{subfigure}[b]{0.24\columnwidth}
  \centering
  \includegraphics[width=\linewidth]{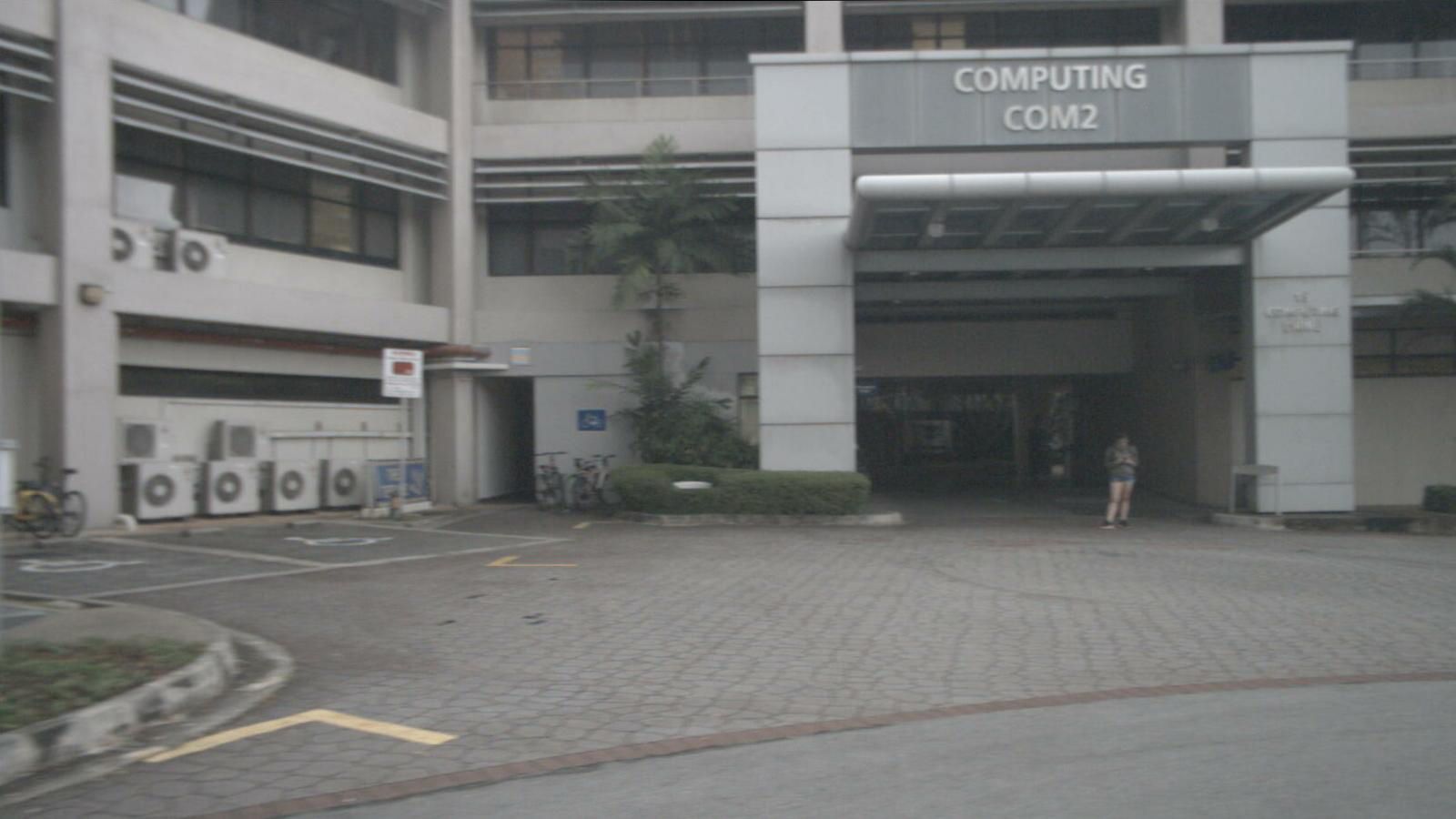}
 
\end{subfigure}

\begin{subfigure}[b]{0.24\columnwidth}
  \centering
  \includegraphics[width=\linewidth]{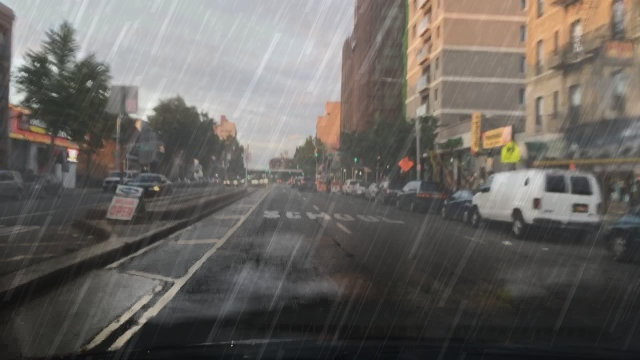}
  \caption{BDD100K~\cite{yu2020bdd100kdiversedrivingdataset}}
\end{subfigure}
\hfill
\begin{subfigure}[b]{0.24\columnwidth}
  \centering
  \includegraphics[width=\linewidth]{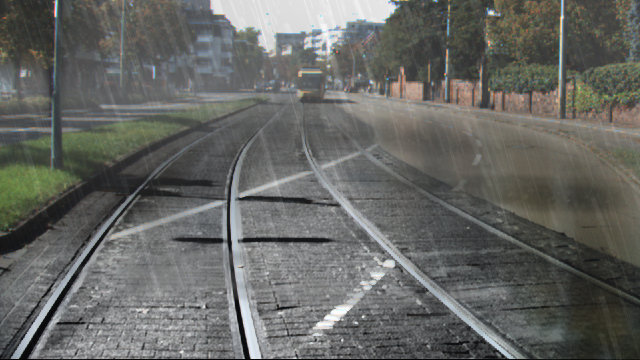}
  \caption{KITTI~\cite{Geiger2013IJRR}}
\end{subfigure}
\hfill
\begin{subfigure}[b]{0.24\columnwidth}
  \centering
  \includegraphics[width=\linewidth]{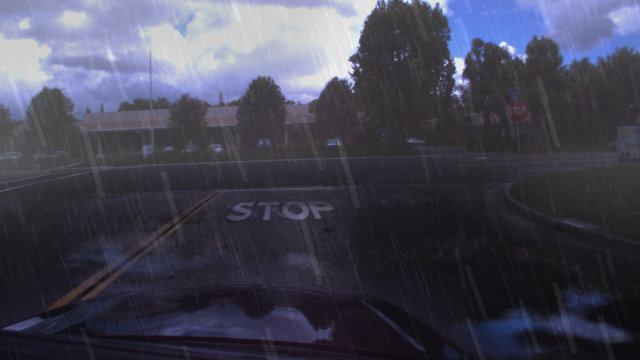}
   \caption{Roadmarks~\cite{Wu2012Roadmarks}}
\end{subfigure}
\hfill
\begin{subfigure}[b]{0.24\columnwidth}
  \centering
  \includegraphics[width=\linewidth]{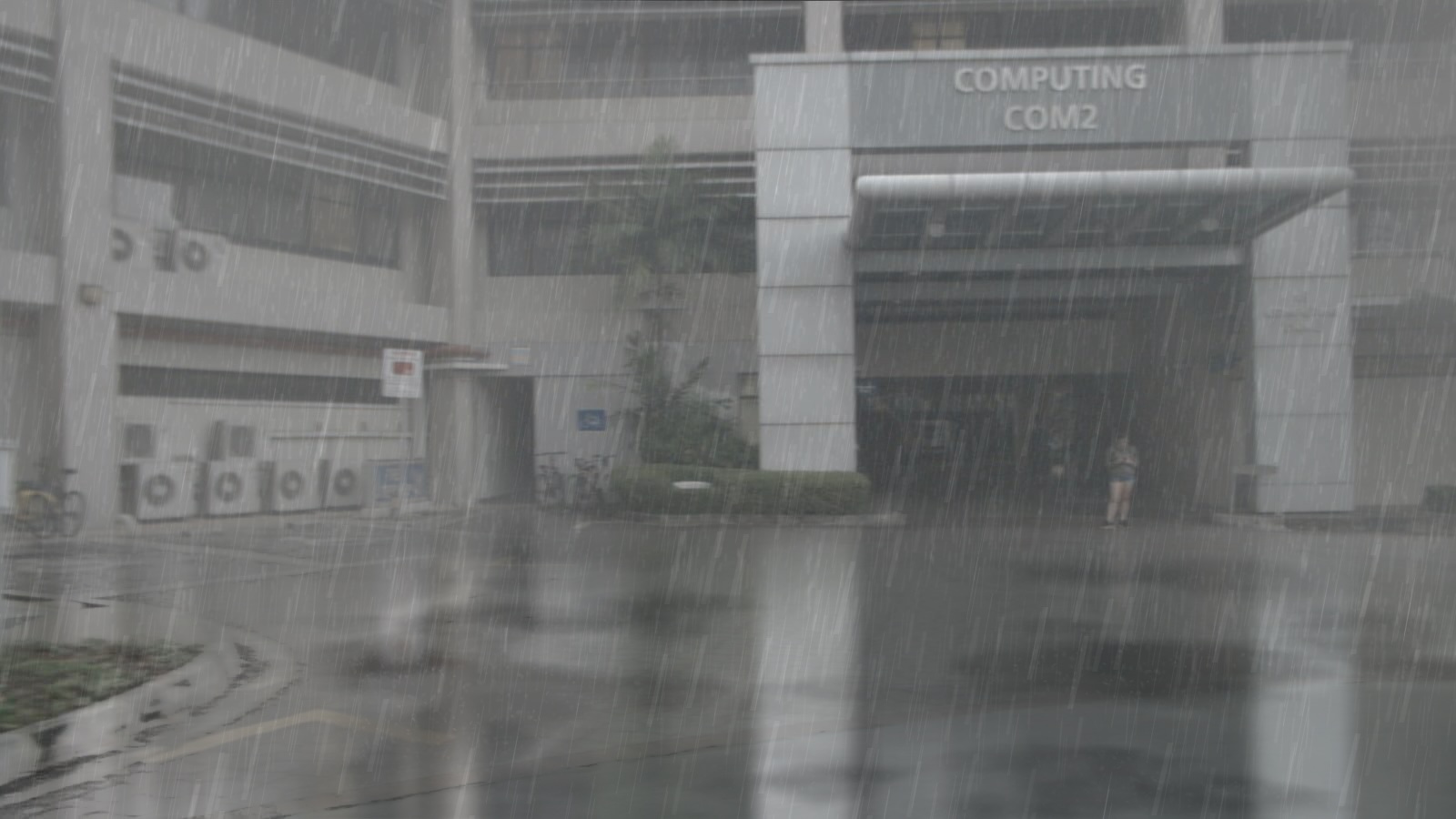}
    \caption{nuScenes~\cite{nuscenes}}
\end{subfigure}

\vspace{-0.5em}
\caption{\textbf{Generalizability of CascadeRS:} rain effects synthesized on images from datasets with different road-view perspective. Additional examples are in Appendix~\ref{supmat: qual. results across datasets}.}
\vspace{-6mm}
\label{fig:dataset_pairs_cropped}
\end{figure}

\subsection{Discussion}
The principled construction of CascadeRS makes it a generalizable rain synthesis pipeline. It can produce realistic and spatially coherent rain, reflection, and puddle effects across diverse datasets. Figure~\ref{fig:dataset_pairs_cropped} shows rain synthesized with CascadeRS on the KITTI~\cite{Geiger2013IJRR}, BDD100K~\cite{yu2020bdd100kdiversedrivingdataset}, Roadmarks~\cite{Wu2012Roadmarks}, and nuScenes~\cite{nuscenes} datasets, demonstrating its applicability to diverse scene resolutions, lighting conditions, and camera geometries without  parameter adjustments.
We focused on daylight images: extending CascadeRS to nighttime or dawn conditions~\cite{Zhang2023RainNighttime,zhang2025nighttimederaining} would involve adjusting physical parameters, such as the reflectivity coefficient, to better capture low-light behavior.
The modular pipeline of CascadeRS makes it flexible and adaptable by integrating different prior depth and semantic segmentation models. For instance, one can plug-in improved models that provide stronger priors or use depth- and semantic-prior models trained e.g. for night vision or to synthesize rain for tasks related to outdoor surveillance, or UAV navigation.
\section{Conclusion}
\label{sec:conclusion}



We presented CascadeRS, a generative pipeline that leverages scene depth and semantic priors to render physically consistent rain and wet-surface effects, including reflections and puddles. By exploiting pretrained depth and segmentation models as scene priors, its rain synthesis modules remain procedural and require no training. CascadeRS improves the perceptual realism of synthetic rain, as demonstrated through user studies and model-based evaluations. It enables benchmarking vision models under controllable adverse rain conditions using the CascadeRS-Cityscapes dataset, while also providing consistent data augmentation to improve robustness in real-world rain. Since CascadeRS does not rely on training images, it is not dataset-specific, and its modular design allows any depth or semantic segmentation model to be integrated into the generation process. We further demonstrated its generalizability to both front-view and side-view driving datasets. CascadeRS thus provides a flexible framework for advancing vision systems in challenging rain conditions.


\section*{Acknowledgements}
This work was carried out as part of the FP5-TRANS4M-R project,
funded by the European Union through Europe's Rail Joint
Undertaking and supported by its members. Views and opinions expressed are, however, those of the
author(s) only and do not necessarily reflect those of the
European Union or Europe's Rail Joint Undertaking.
Neither the European Union nor the granting authority
can be held responsible for them.

{
    \small
    \bibliographystyle{ieeenat_fullname}
    \bibliography{main}
}

\appendix
\clearpage
\setcounter{page}{1}
\maketitlesupplementary



We provide additional qualitative and quantitative results, implementation details, and study descriptions that complement the main paper. We first illustrate how the global rain-intensity parameter affects the rendering pipeline, followed by visual examples from our ablation study. We then present an extended qualitative comparison against baseline methods and provide a more extensive FID evaluation. 

Further materials include the detailed results of the semantic segmentation study and additional qualitative results across multiple datasets to demonstrate the robustness and generalizability of the proposed approach. The effect of using pretrained depth and segmentation models is evaluated with quantitative and qualitative results. We also include implementation details, focusing on how global quantities influence internal parameters. Finally, we describe the design, setup, and prompts used in the human and model-based survey evaluations.

\section{Effect of Rain Intensity \label{supmat: effect of rain intensity}}
Figure~\ref{fig:intensity_sweep} illustrates the influence of the global rain-intensity parameter used throughout CascadeRS. As intensity increases, all rain-dependent modules respond accordingly: the rain-streak generator increases layer density and motion-blur length, the fog module increases the scattering coefficient and atmospheric light strength, and the wet-surface and puddle components apply stronger reflectivity and more pronounced water accumulation. Importantly, the rain \emph{direction} is sampled randomly for each rendering and is therefore decoupled from the intensity parameter; intensity controls only the magnitude of rain effects, not their orientation. The resulting progression, from subtle streaks and light haze at low intensity to heavy streaks, dense fog, and stronger surface effects at high intensity, demonstrates how a single global parameter modulates the full rendering pipeline in a consistent and physically meaningful manner.

\begin{figure*}[t]
\centering
\setlength{\tabcolsep}{3pt}
\renewcommand{\arraystretch}{0}

\captionsetup[sub]{skip=2pt}

\newcommand{\intensityimg}[1]{
\includegraphics[
    width=\linewidth,
    trim=0pt 0pt 450pt 0pt,
    clip
]{figures/intensity/lindau_000058_000019_leftImg8bit_rain_wp_#1.jpg}
}

\begin{tabular}{ccc}
  \begin{subfigure}[b]{0.27\linewidth}\centering\intensityimg{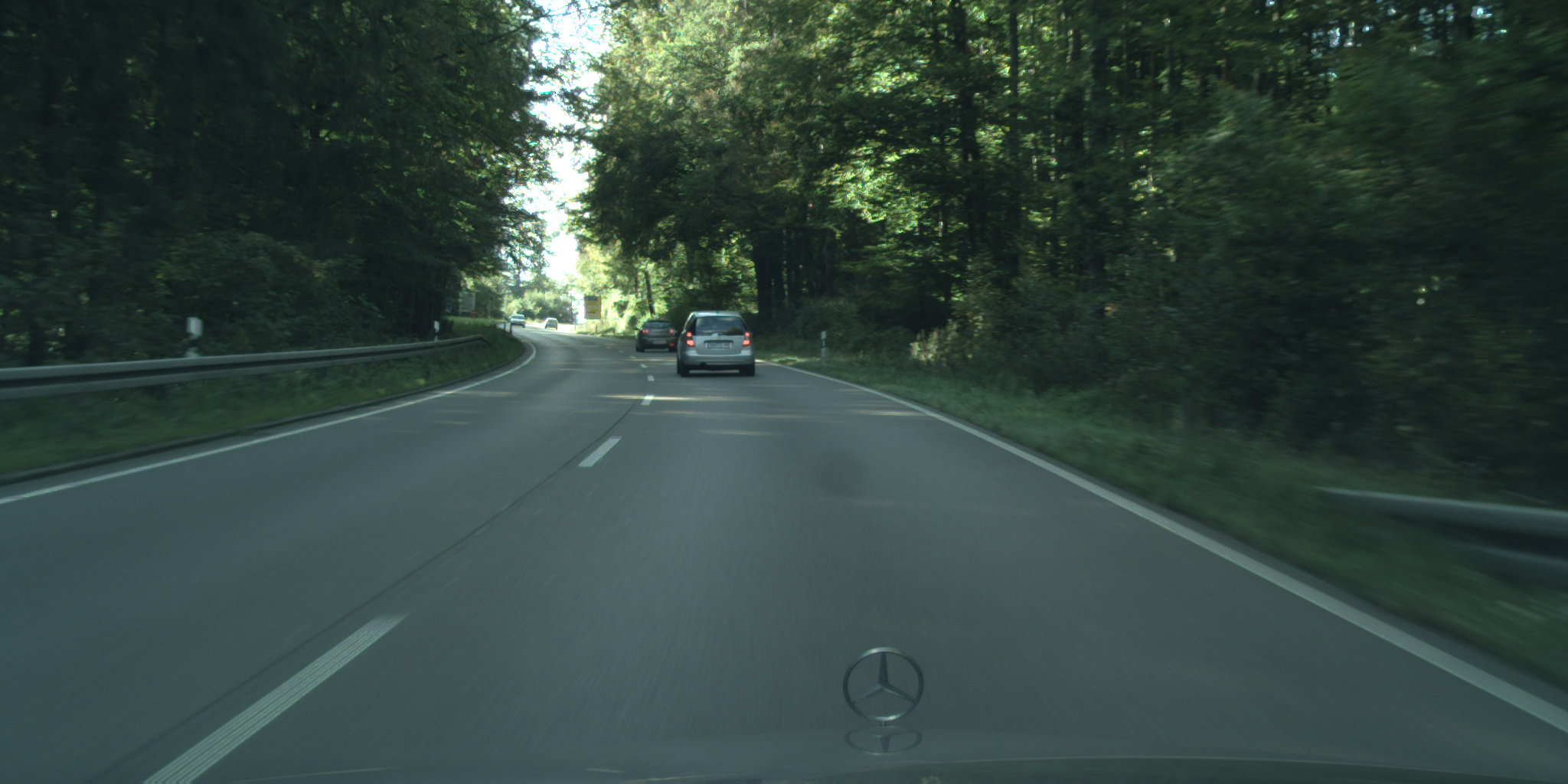}\caption{}\end{subfigure} &
  \begin{subfigure}[b]{0.27\linewidth}\centering\intensityimg{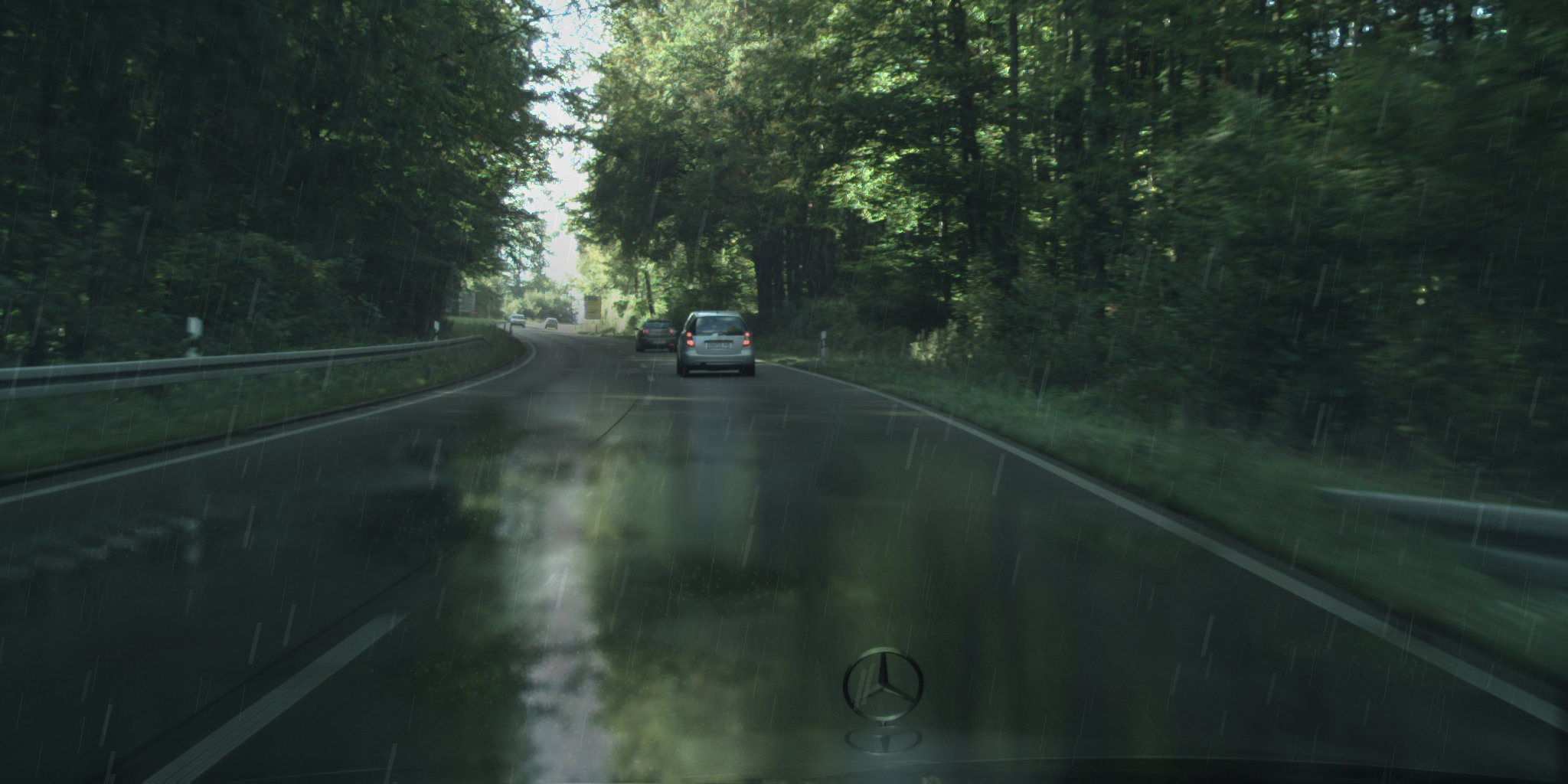}\caption{}\end{subfigure} &
  \begin{subfigure}[b]{0.27\linewidth}\centering\intensityimg{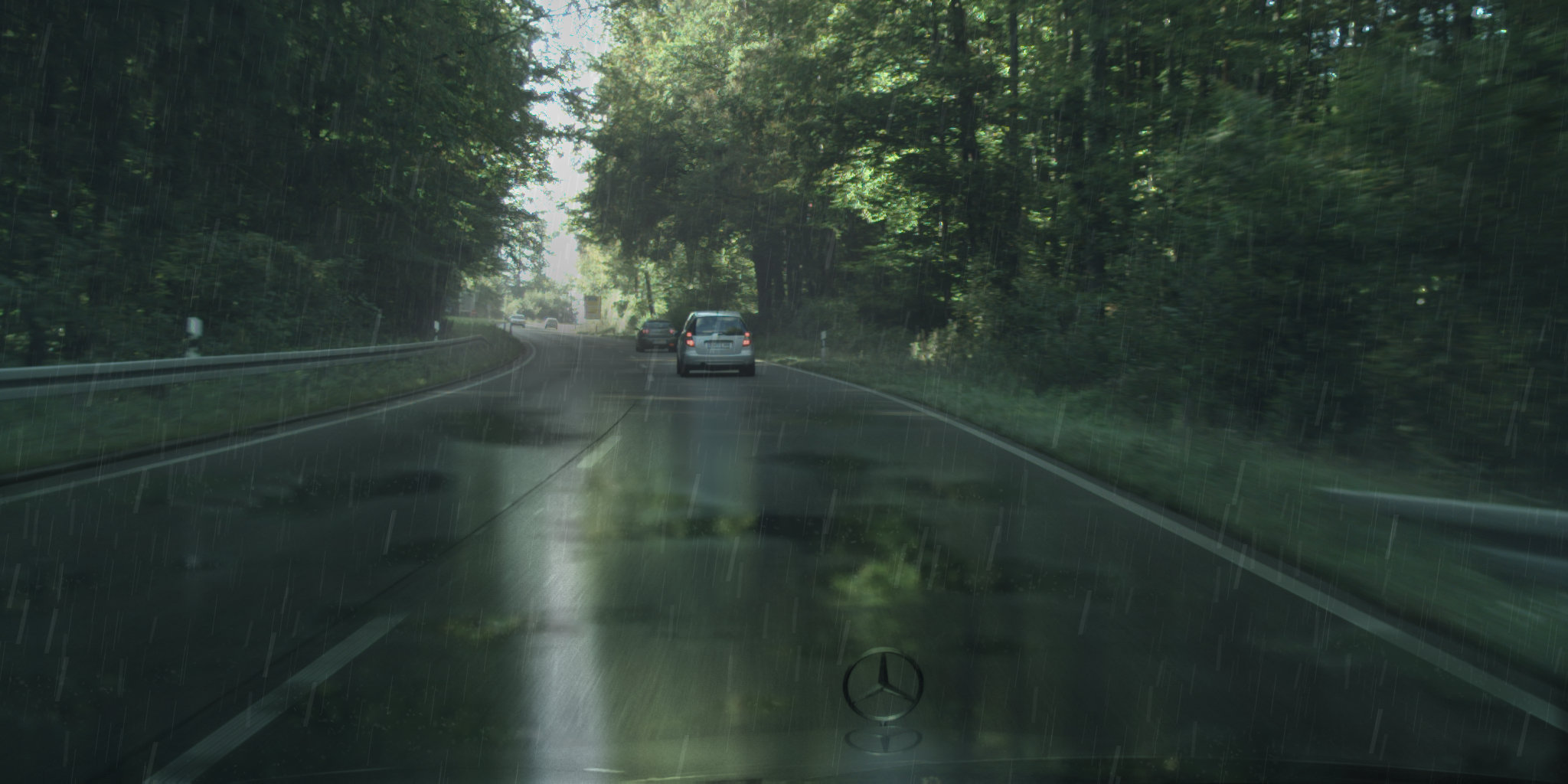}\caption{}\end{subfigure} \\[2pt]

  \begin{subfigure}[b]{0.27\linewidth}\centering\intensityimg{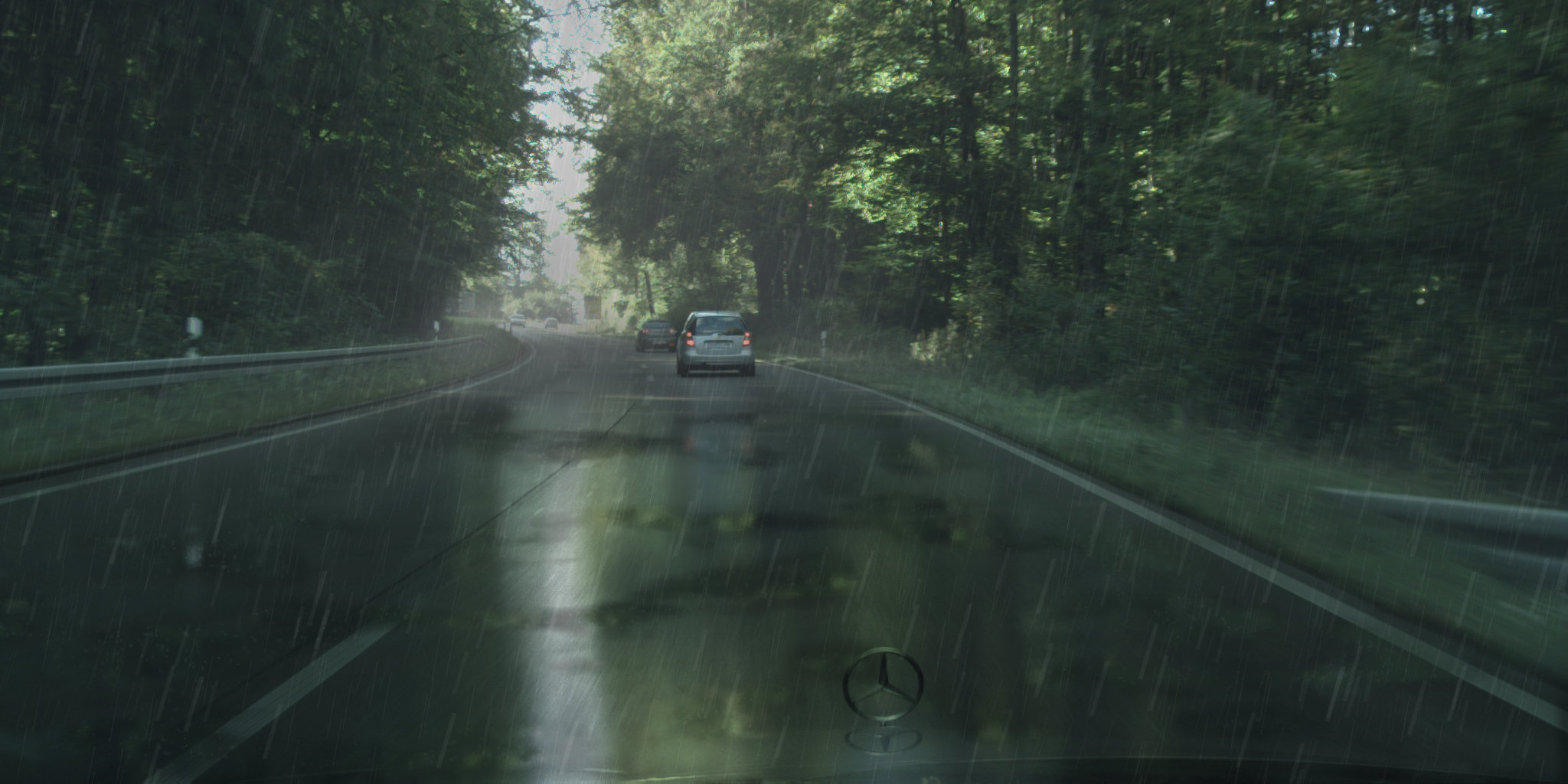}\caption{}\end{subfigure} &
  \begin{subfigure}[b]{0.27\linewidth}\centering\intensityimg{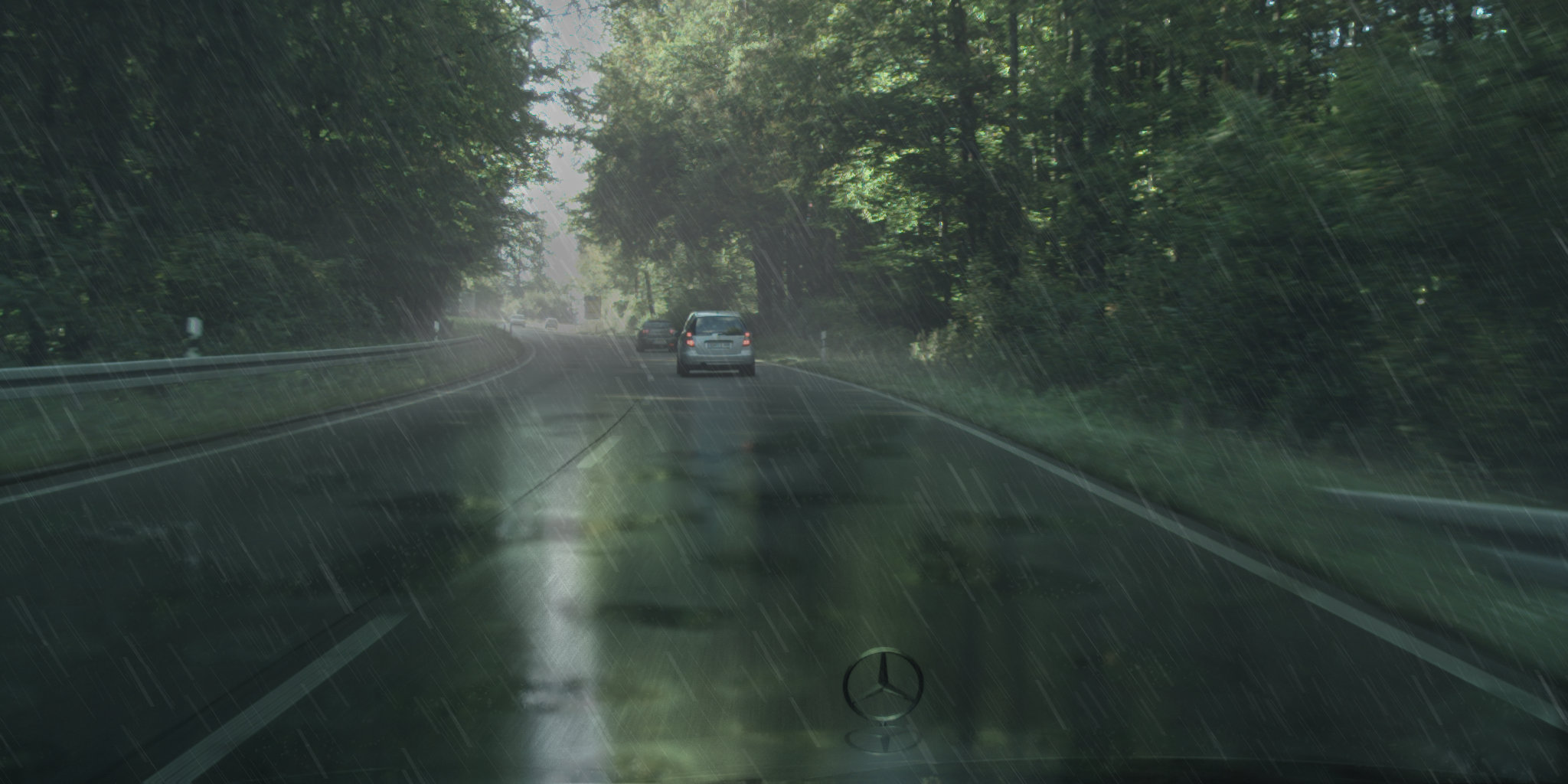}\caption{}\end{subfigure} &
  \begin{subfigure}[b]{0.27\linewidth}\centering\intensityimg{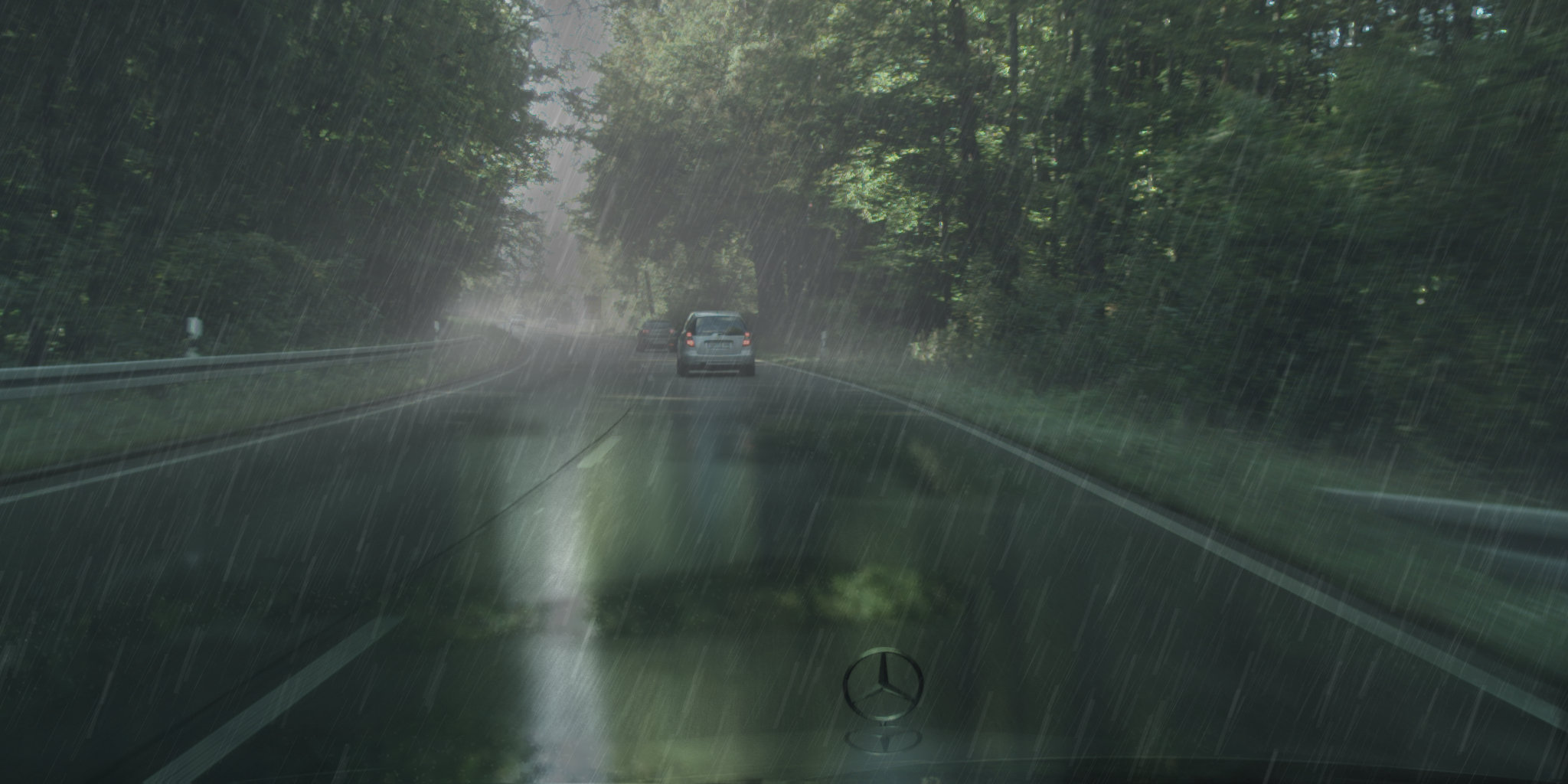}\caption{}\end{subfigure} \\[2pt]

  \begin{subfigure}[b]{0.27\linewidth}\centering\intensityimg{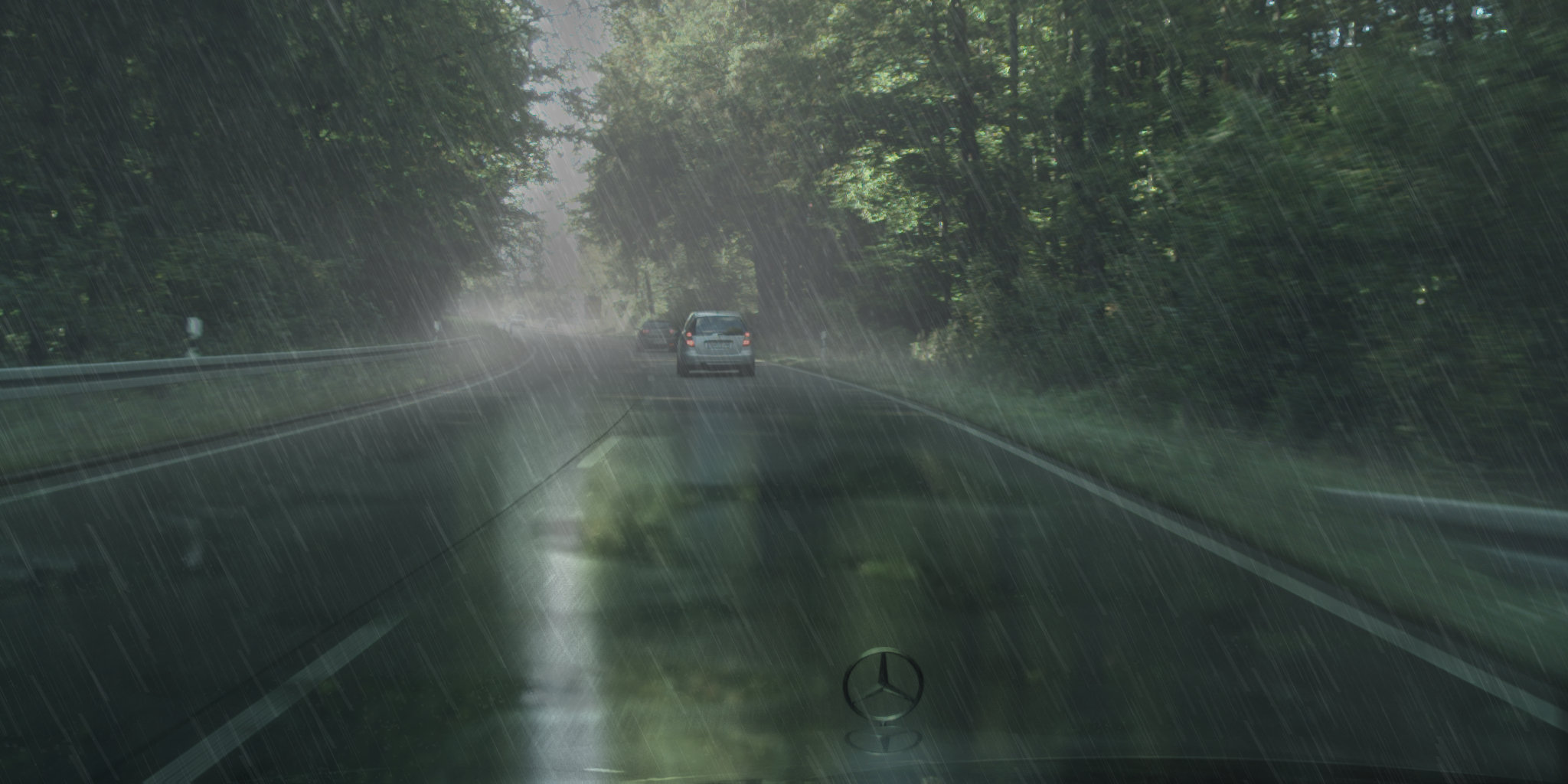}\caption{}\end{subfigure} &
  \begin{subfigure}[b]{0.27\linewidth}\centering\intensityimg{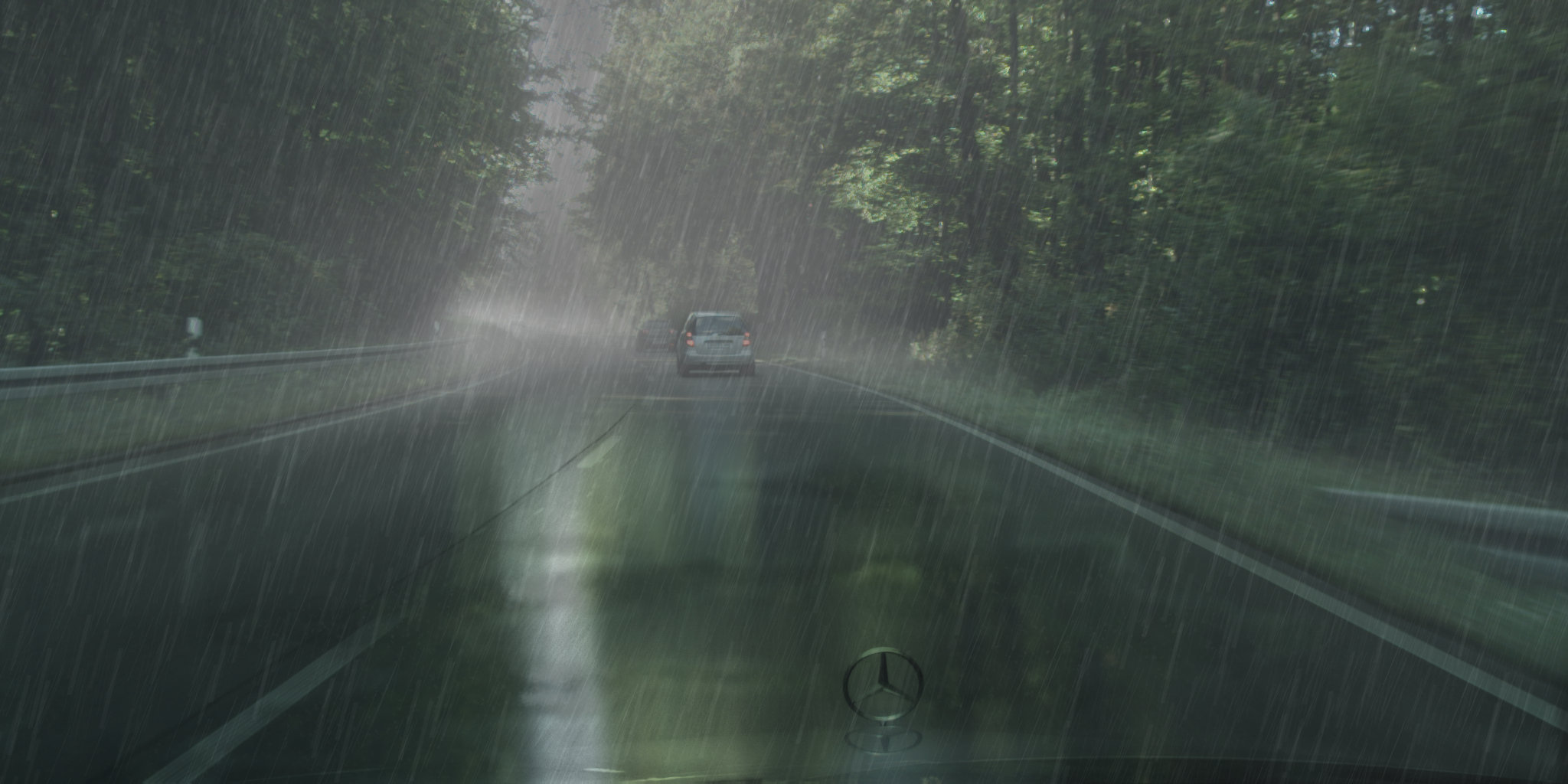}\caption{}\end{subfigure} &
  \begin{subfigure}[b]{0.27\linewidth}\centering\intensityimg{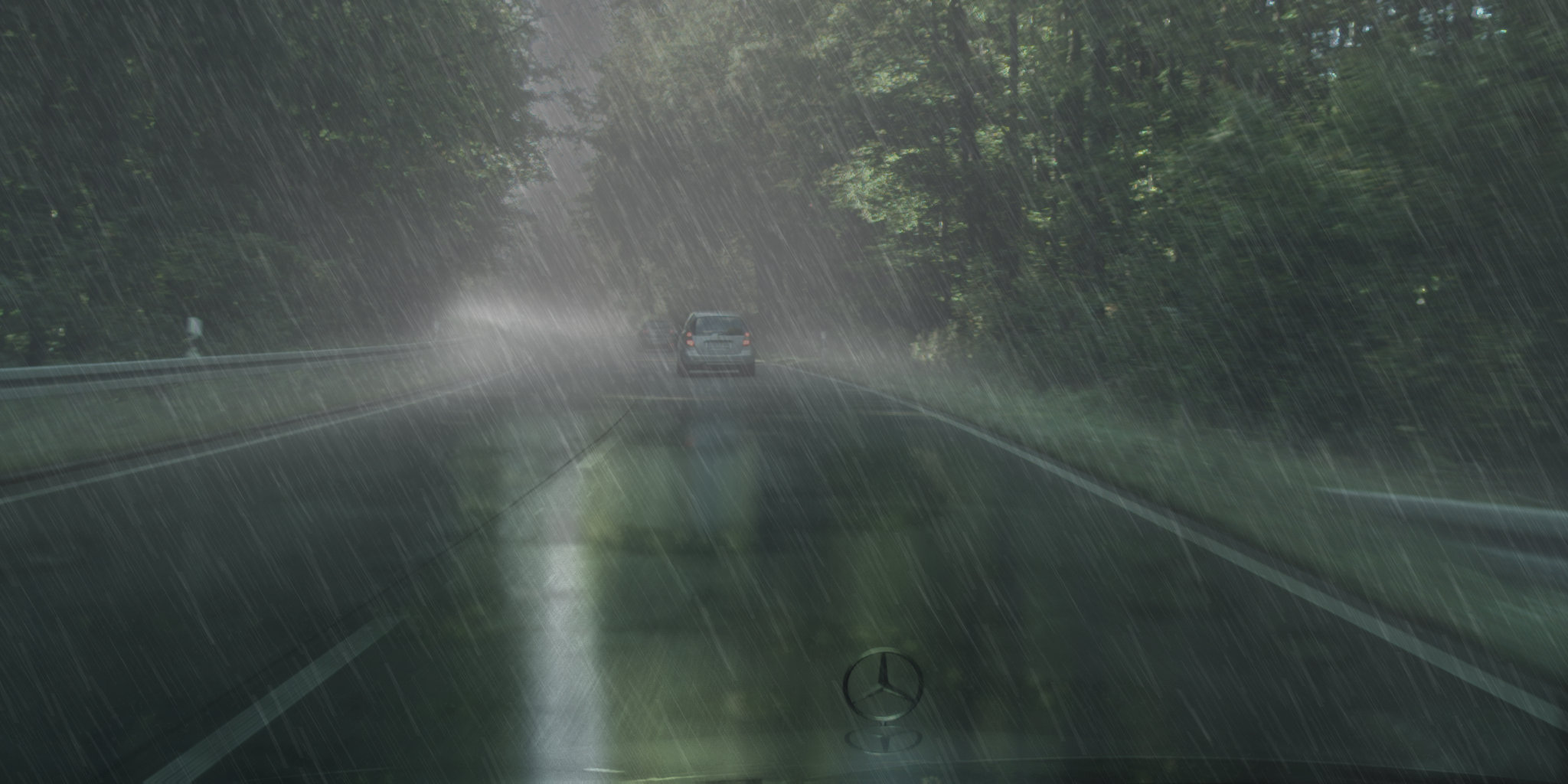}\caption{}\end{subfigure} \\[2pt]

  \begin{subfigure}[b]{0.27\linewidth}\centering\intensityimg{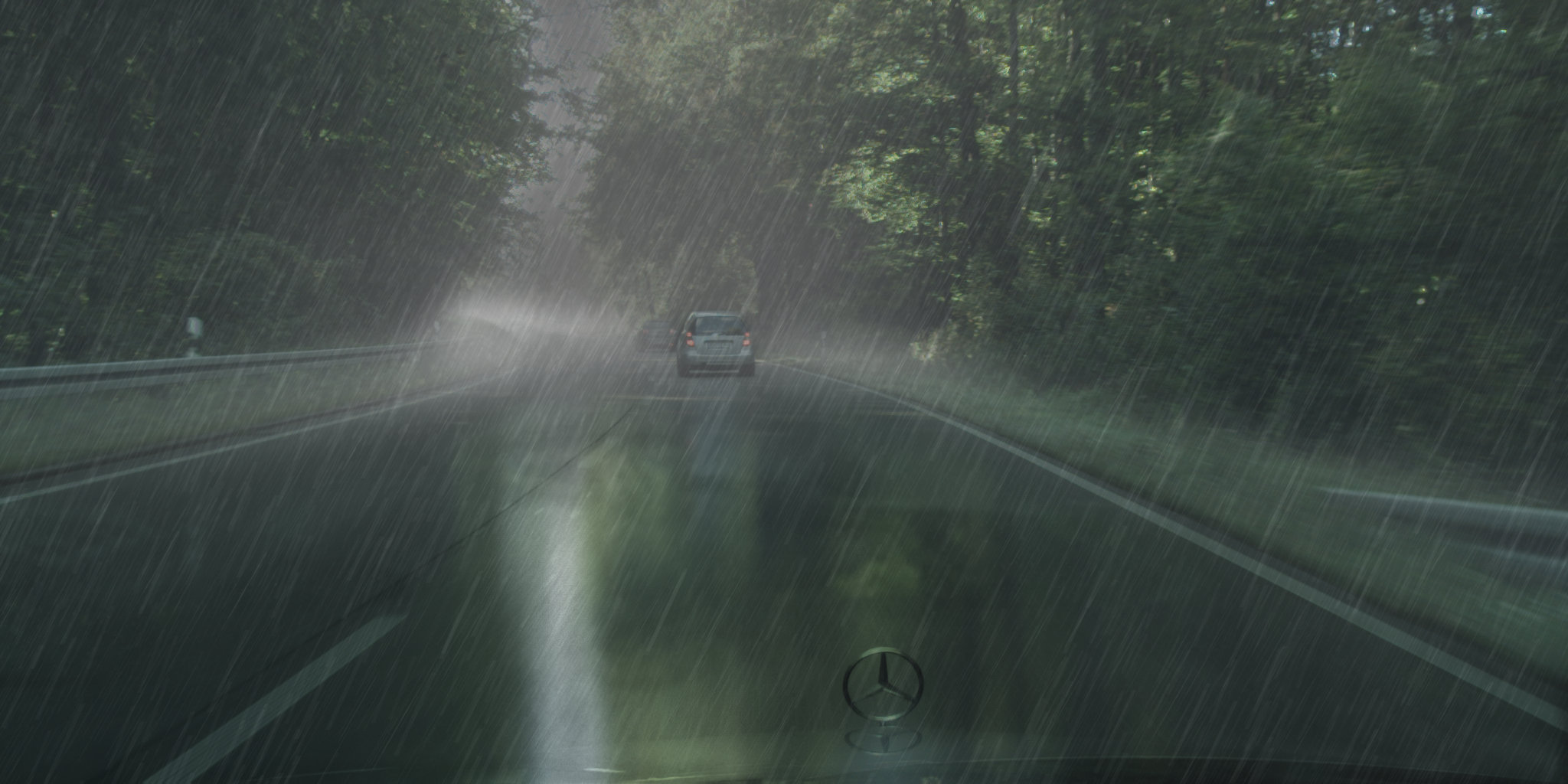}\caption{}\end{subfigure} &
  \begin{subfigure}[b]{0.27\linewidth}\centering\intensityimg{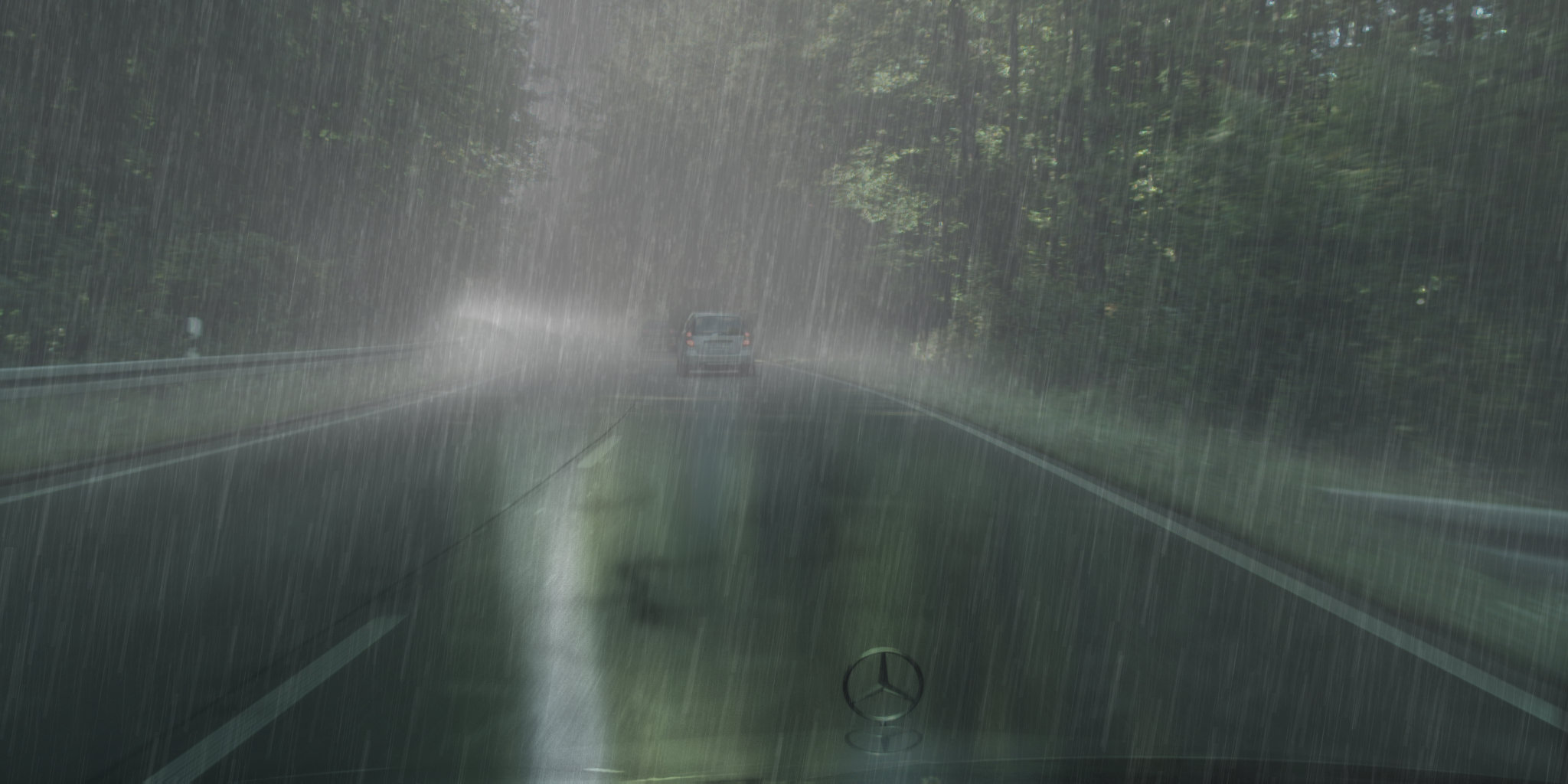}\caption{}\end{subfigure} &
  {}
\end{tabular}

\caption{\textbf{Rain-intensity sweep on a (cropped) Cityscapes sample~\cite{Cityscapes}.} 
Subfigures (a)–(k) correspond to intensities 0–100. Increasing intensity leads to denser streaks, thicker fog, and stronger atmospheric degradation in CascadeRS.}
\label{fig:intensity_sweep}
\end{figure*}


\section{Ablation Study: Visual Examples \label{supmat: ablation study - visual examples}}
Figures~\ref{tab:cascaders_multiscene} and~\ref{fig:combined_zoom_ablations} illustrate the contribution of the main components of CascadeRS (rain streaks, wet-surface reflections, and puddles) across multiple scenes. These examples complement the quantitative ablation results in the main paper and provide a detailed visual breakdown of how each rendering module contributes to the overall realism.

Figure~\ref{tab:cascaders_multiscene} shows several Cityscapes scenes processed with the sequential stages of the pipeline. The \emph{Rain streaks} version adds only depth-aware rain streaks and fog, the \emph{Reflections} stage introduces wet-road reflections driven by scene geometry and brightness cues, and the \emph{Puddles} stage adds physically plausible puddle shapes and distortions. The progression demonstrates how local surface interactions and water accumulation significantly enhance perceptual realism beyond streak-only rendering.

\begin{figure*}[t]
\centering
\renewcommand{\arraystretch}{1.1}
\setlength{\tabcolsep}{2pt}
\caption{\textbf{Progressive illustration of the CascadeRS components across multiple Cityscapes~\cite{Cityscapes} scenes.}
Each row corresponds to a different image, showing the sequential addition of (1) rain streaks, (2) wet-surface reflections, and (3) puddles guided by scene geometry.
Together, these effects yield realistic and coherent rainy appearances.}
\vspace{3pt}
\scriptsize
\begin{tabular}{C{0.22\linewidth} C{0.22\linewidth} C{0.22\linewidth} C{0.22\linewidth}}
\toprule
\textbf{Original} & \textbf{Rain Streaks} & \textbf{Reflections} & \textbf{Puddles} \\
\midrule

\includegraphics[width=\linewidth]{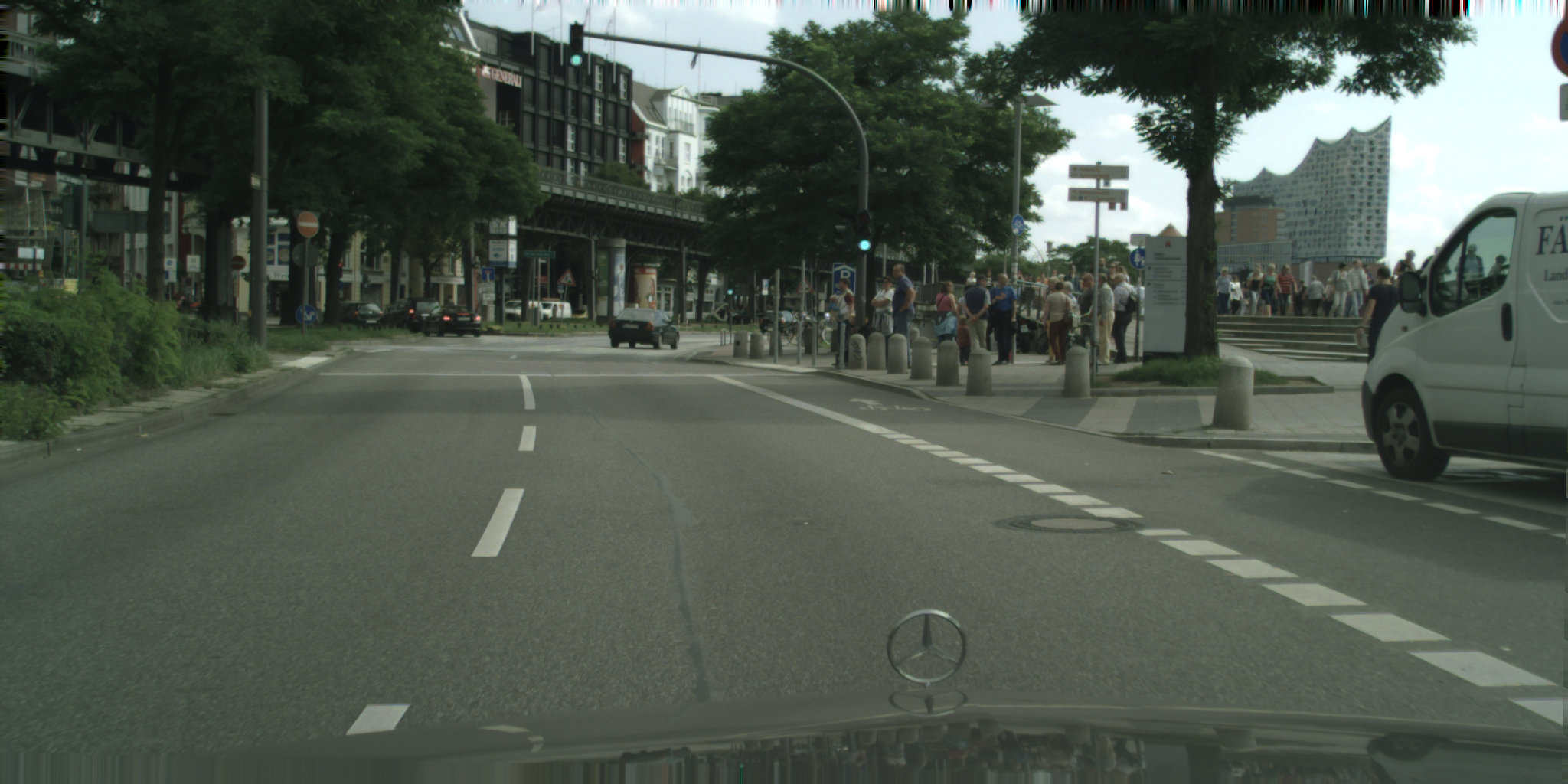} &
\includegraphics[width=\linewidth]{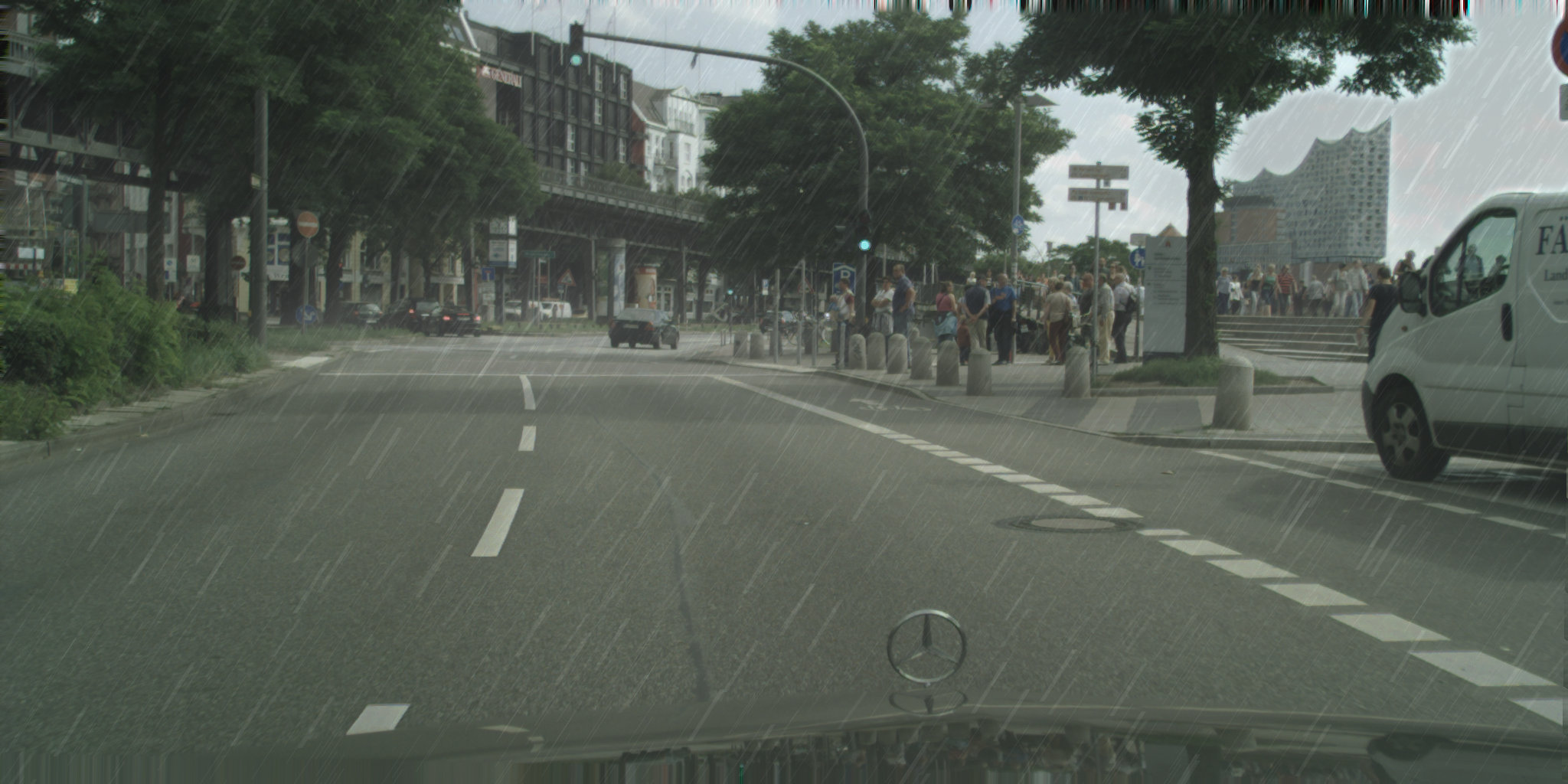} &
\includegraphics[width=\linewidth]{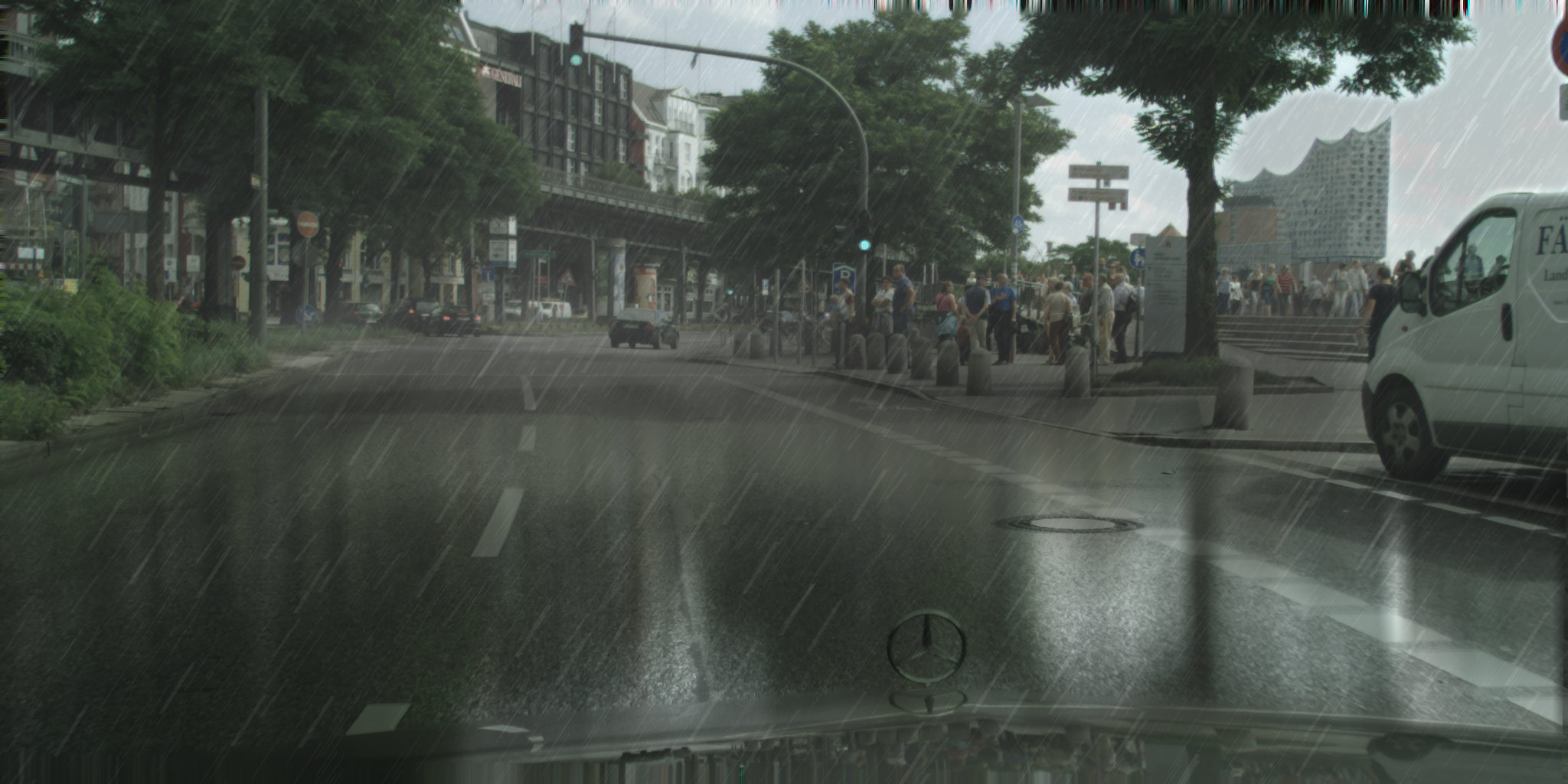} &
\includegraphics[width=\linewidth]{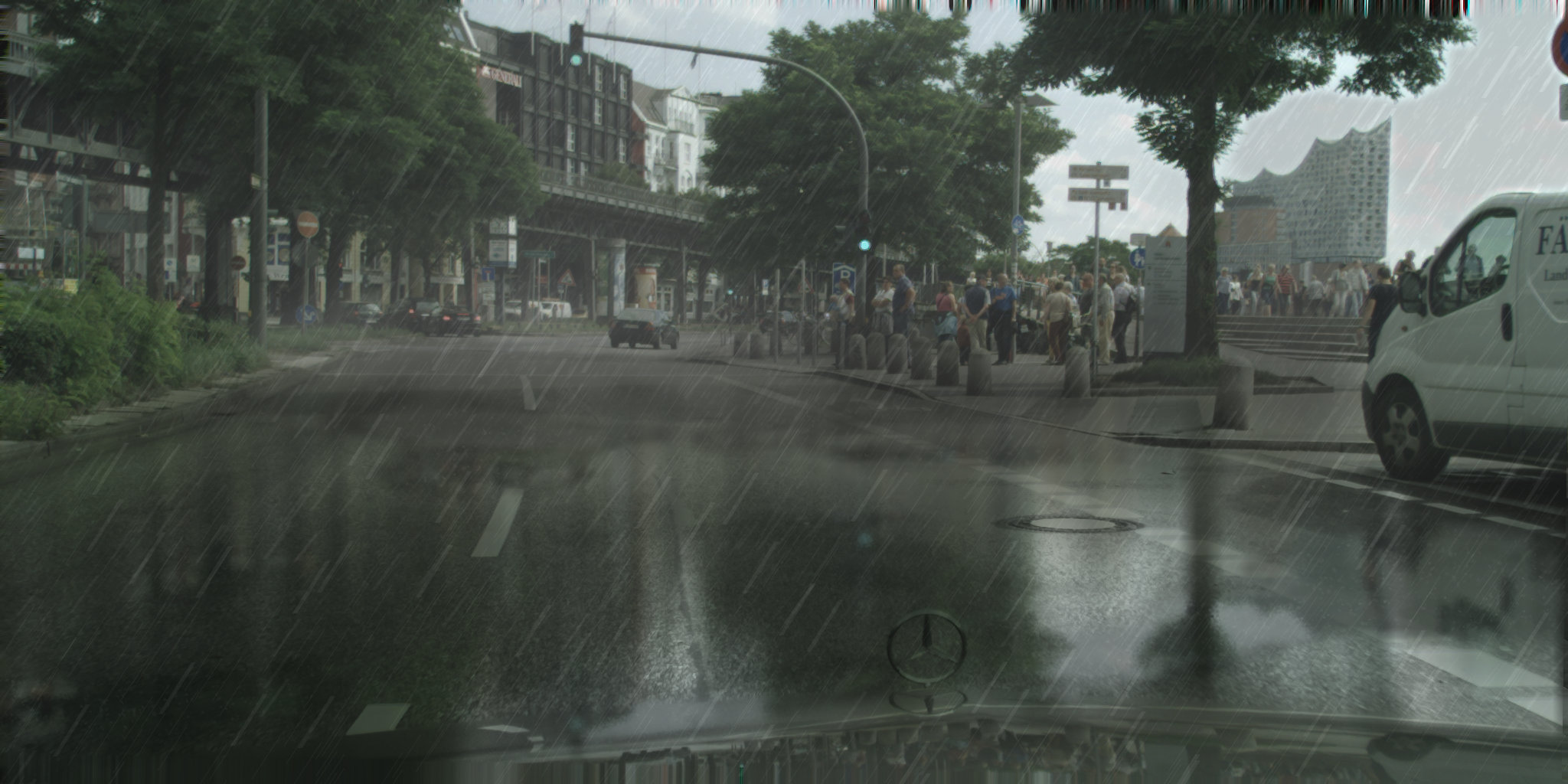} \\[-1mm]

\includegraphics[width=\linewidth]{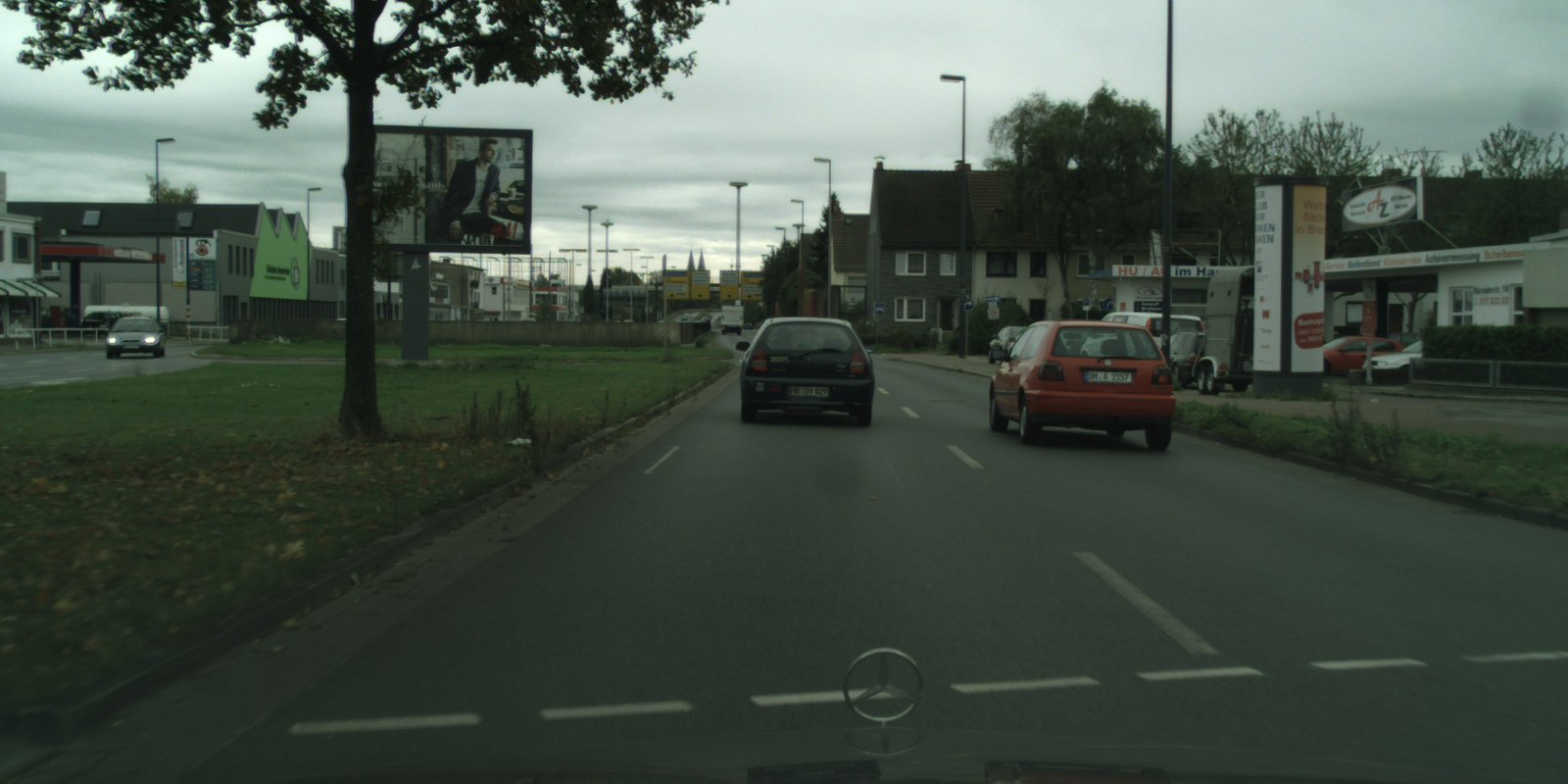} &
\includegraphics[width=\linewidth]{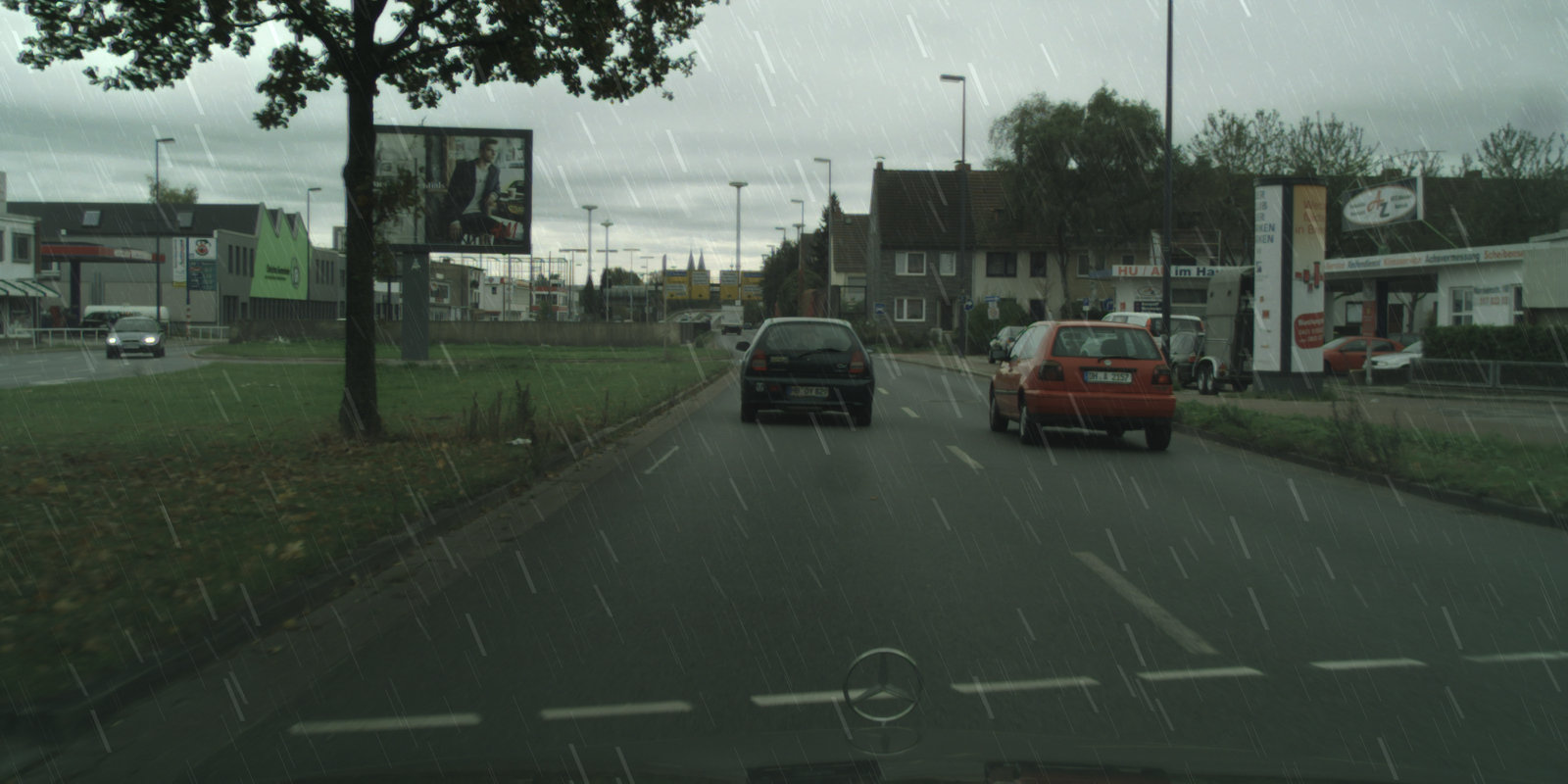} &
\includegraphics[width=\linewidth]{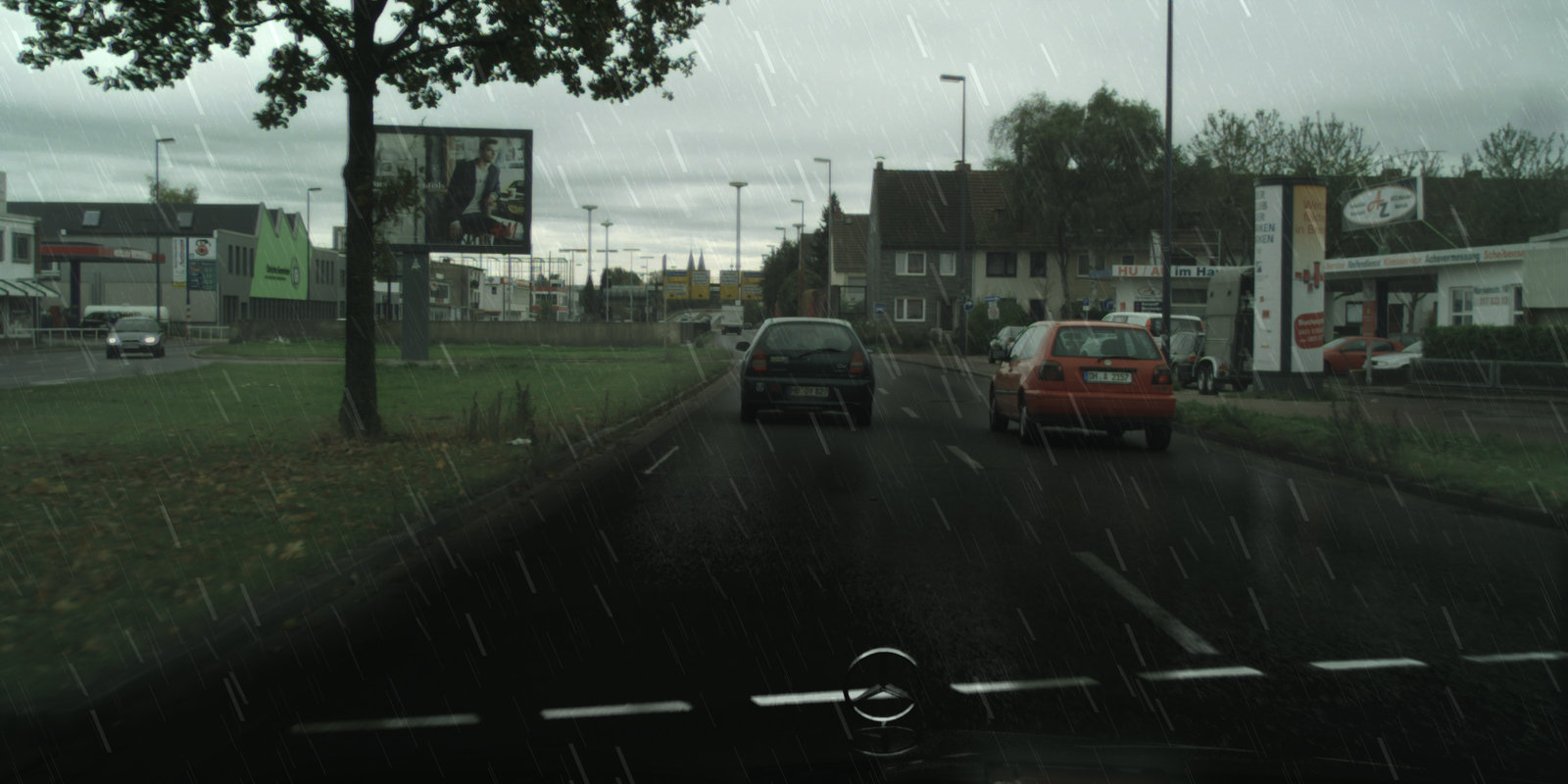} &
\includegraphics[width=\linewidth]{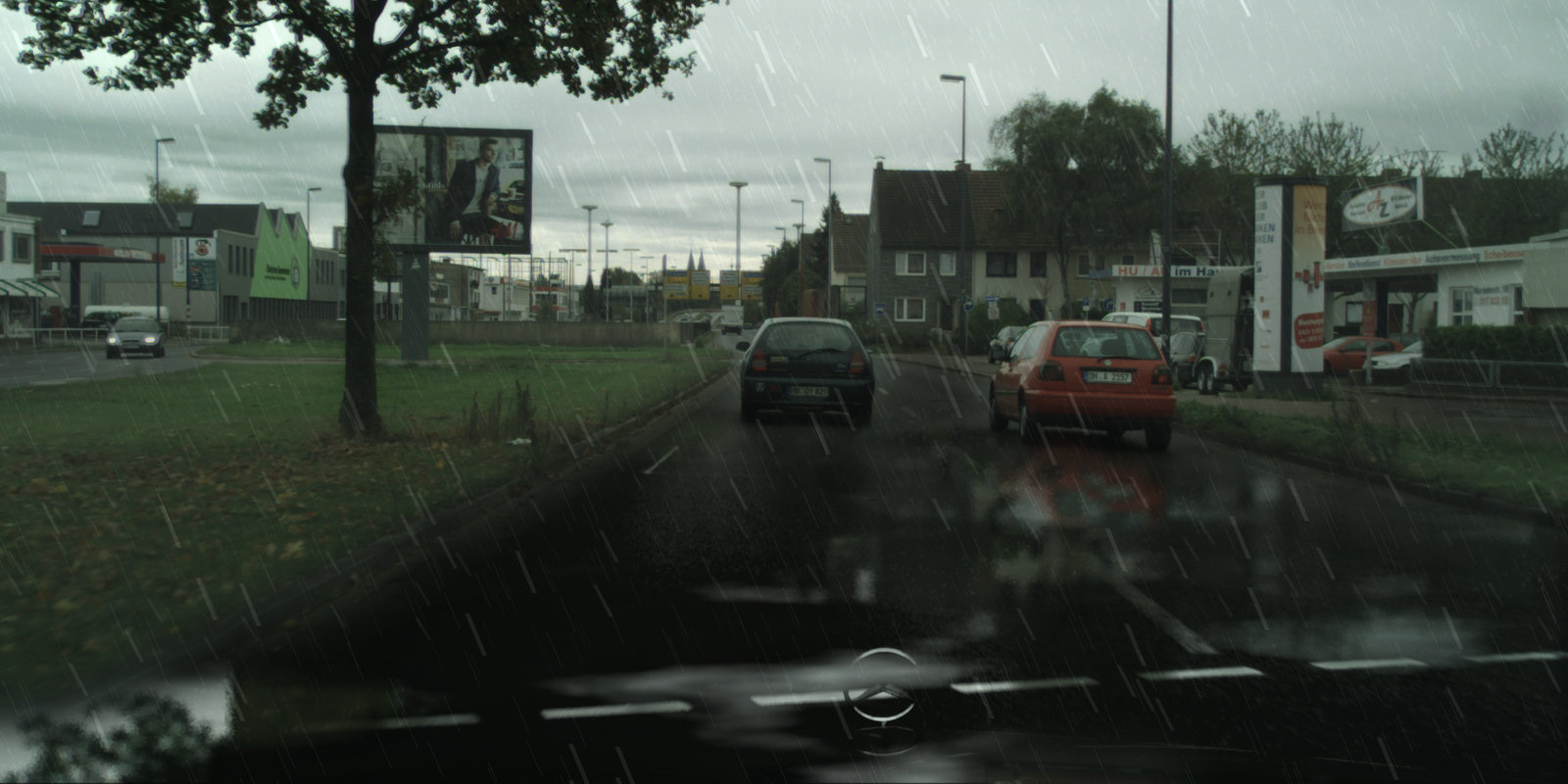} \\[-1mm]

\includegraphics[width=\linewidth]{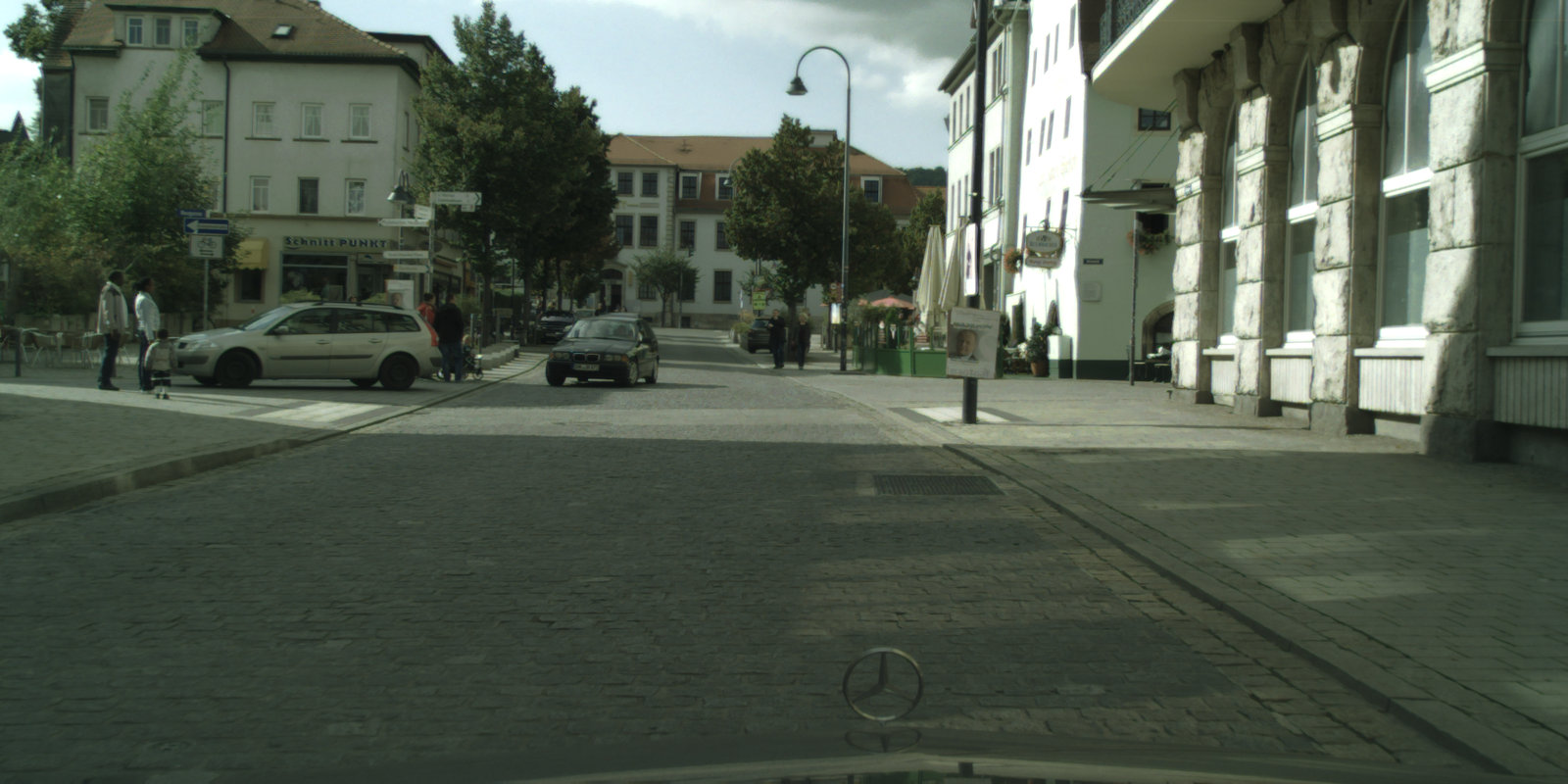} &
\includegraphics[width=\linewidth]{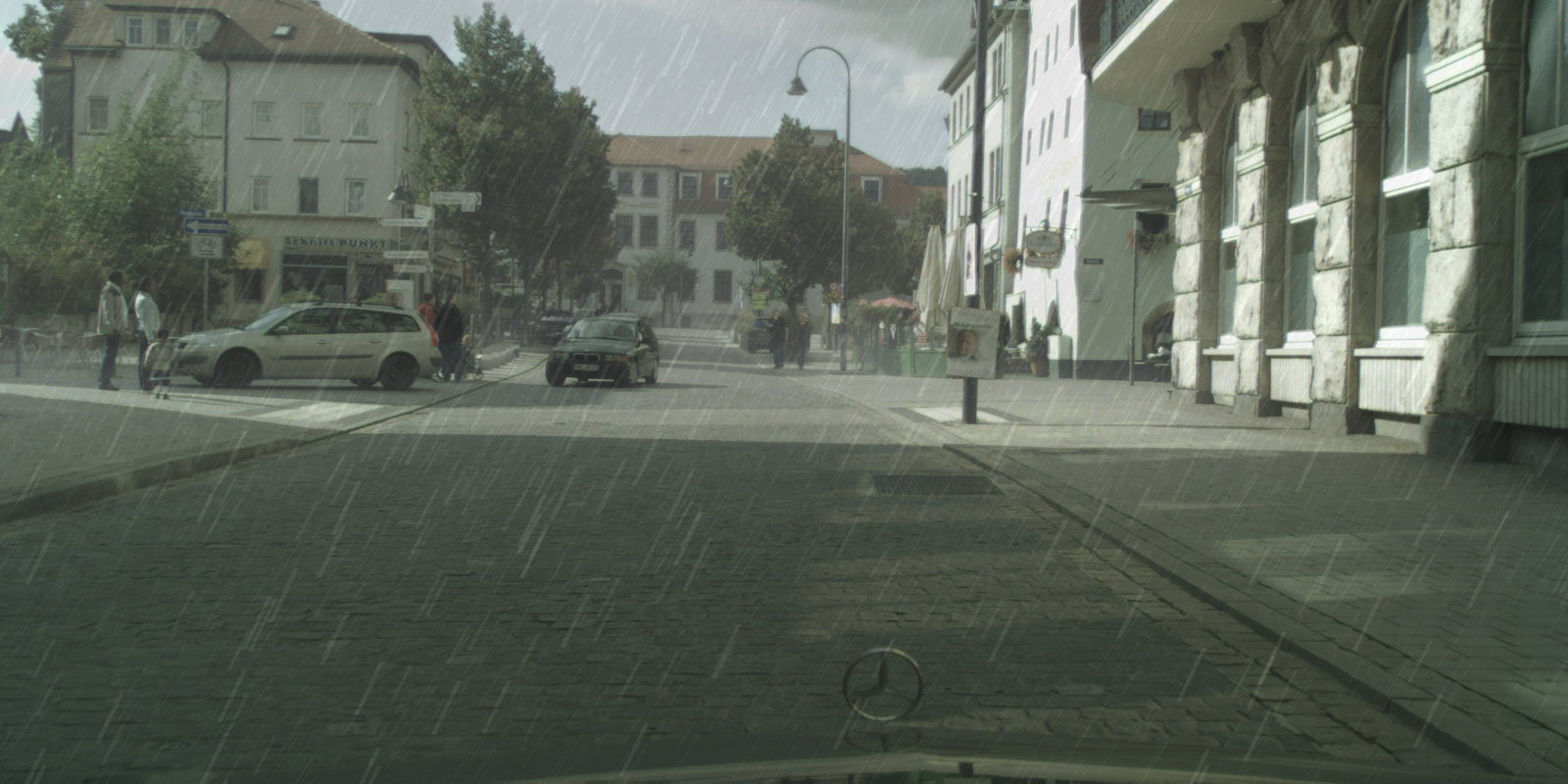} &
\includegraphics[width=\linewidth]{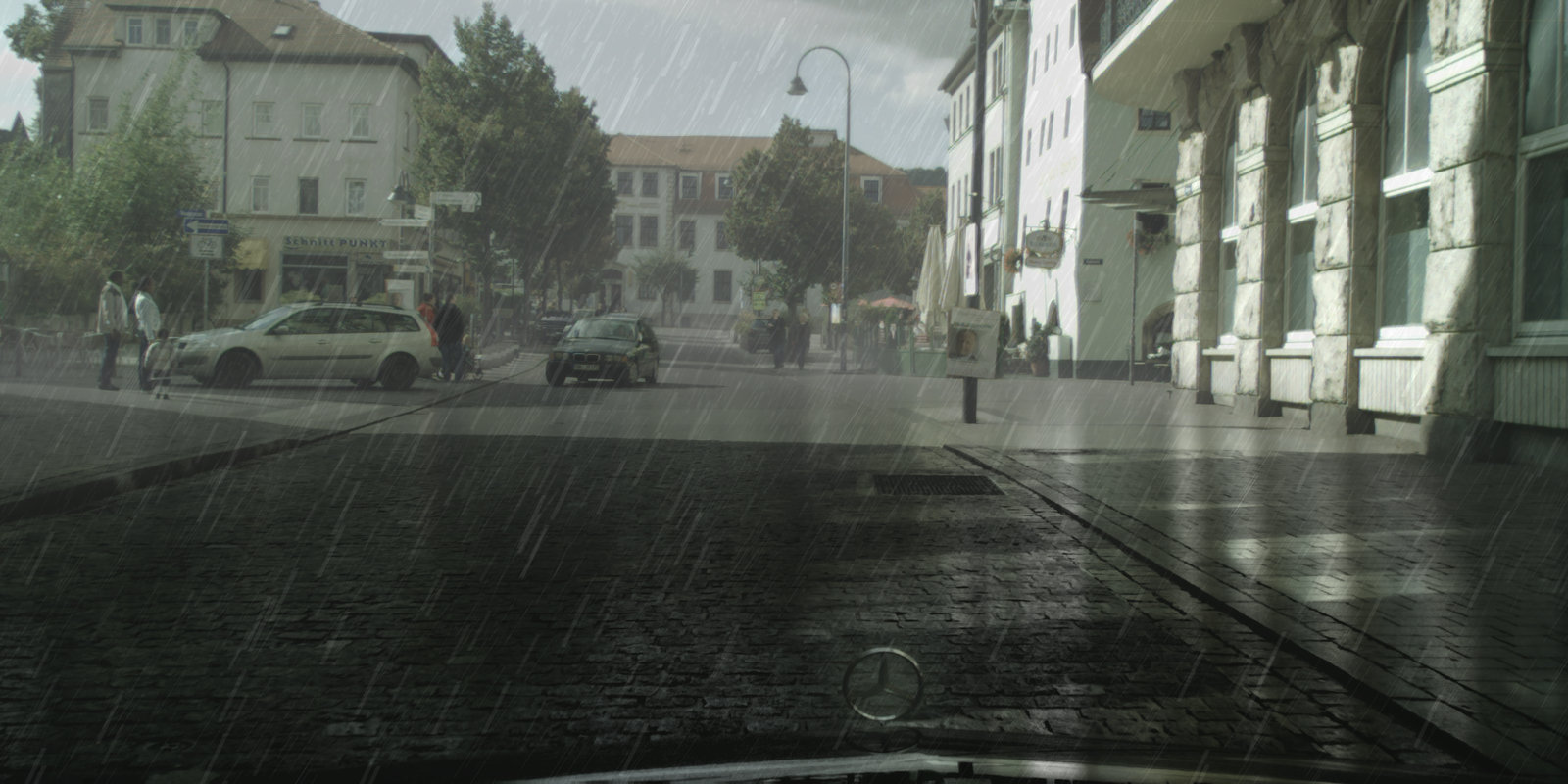} &
\includegraphics[width=\linewidth]{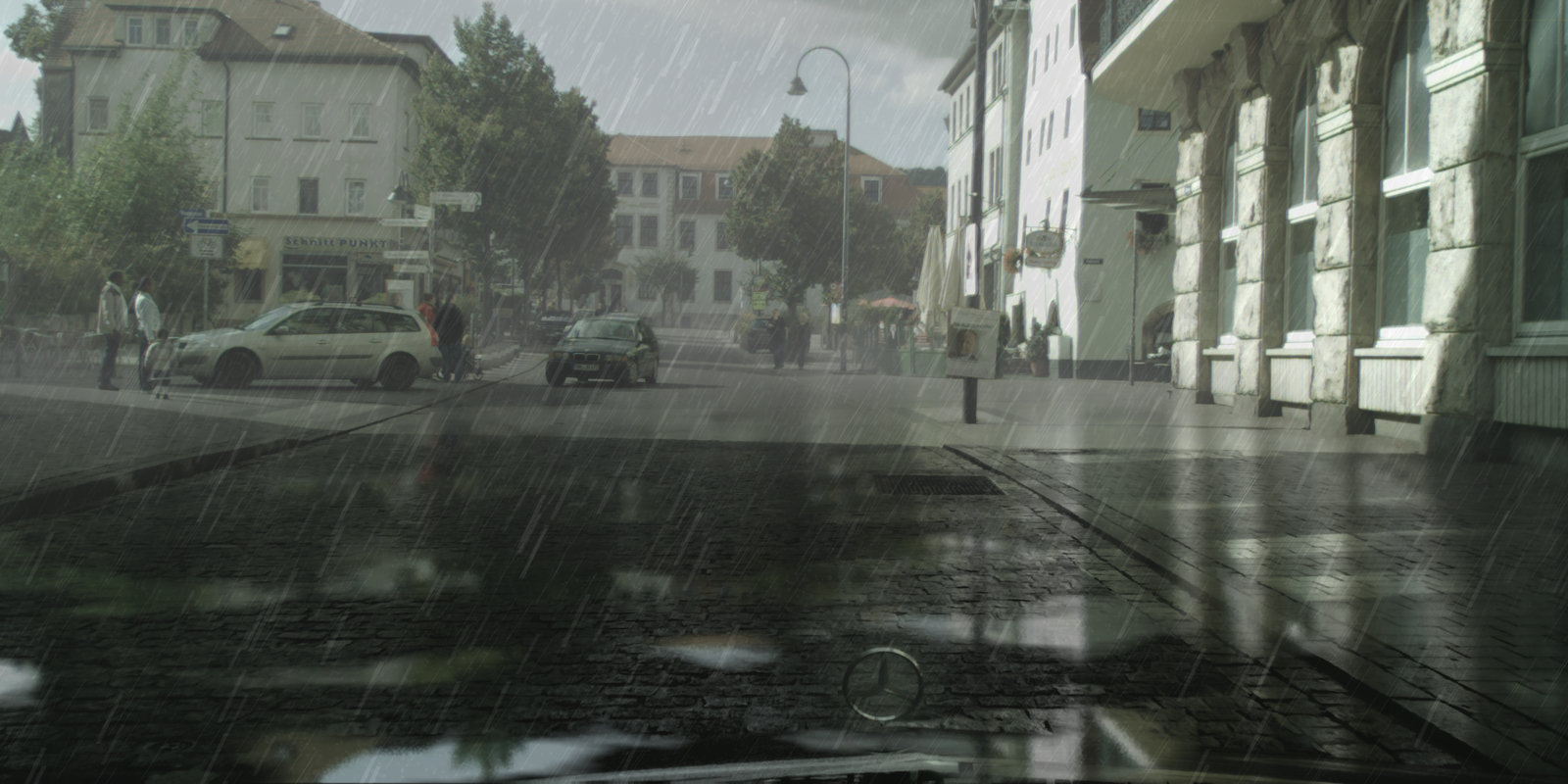} \\[-1mm]

\includegraphics[width=\linewidth]{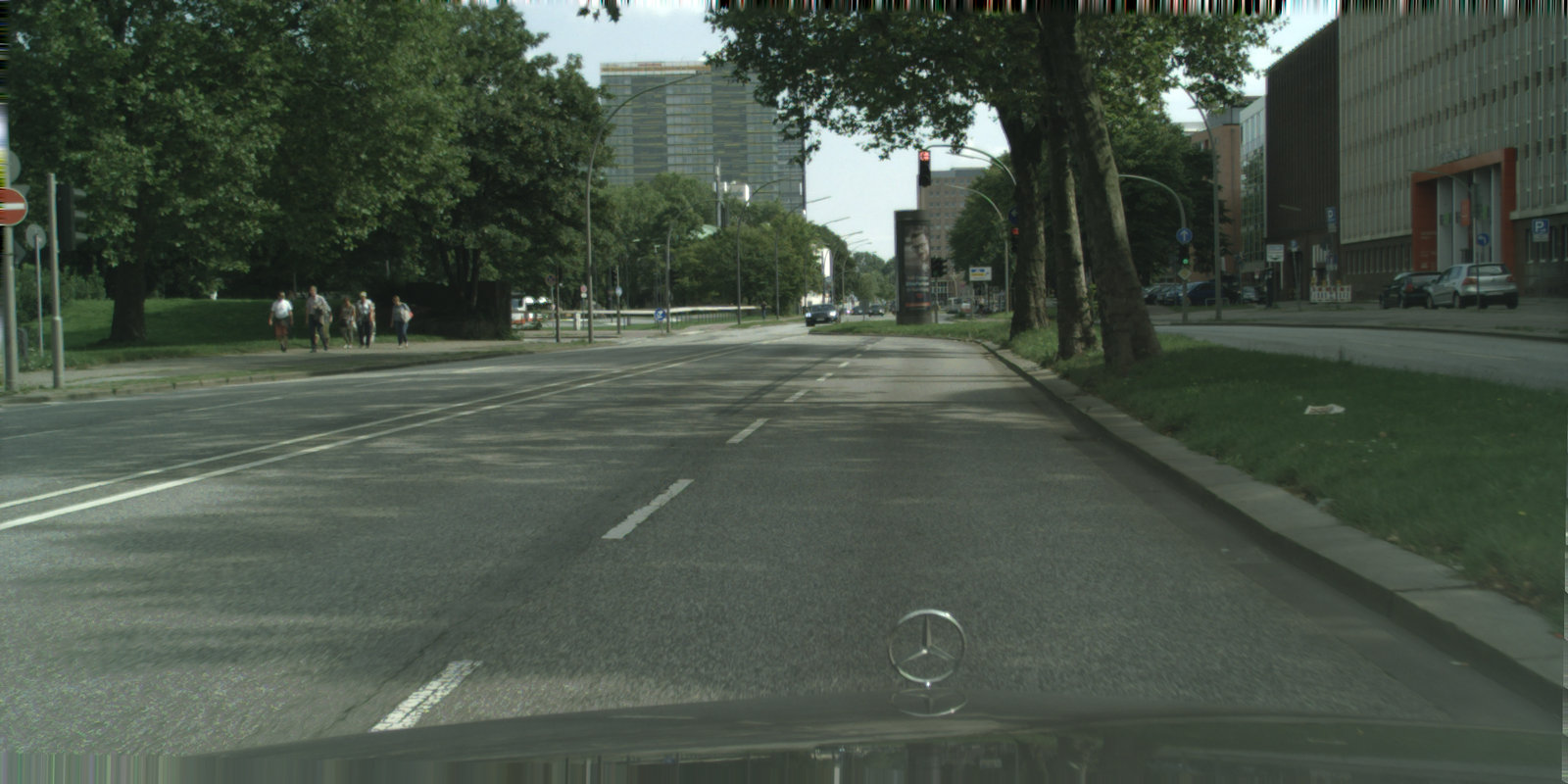} &
\includegraphics[width=\linewidth]{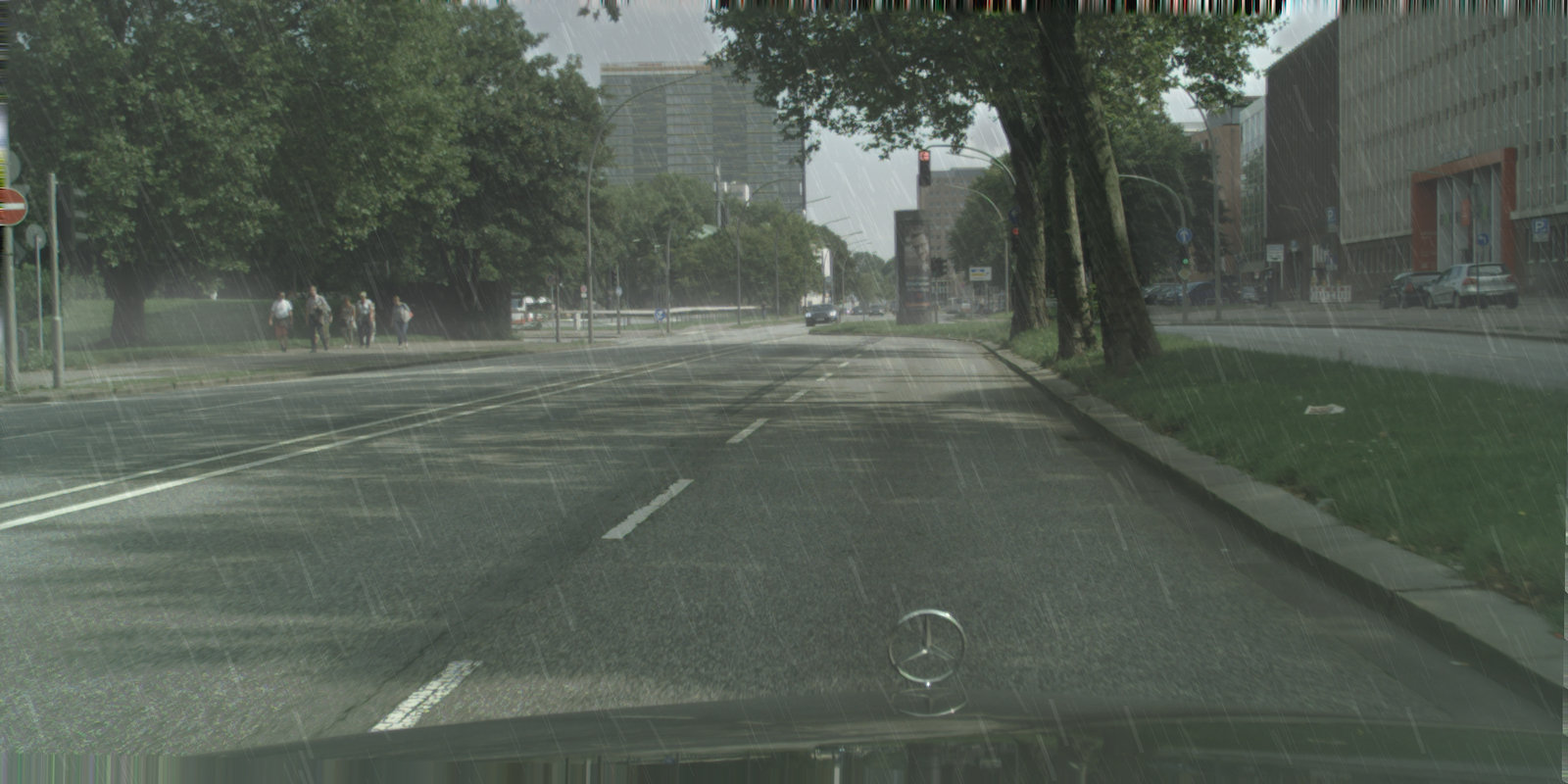} &
\includegraphics[width=\linewidth]{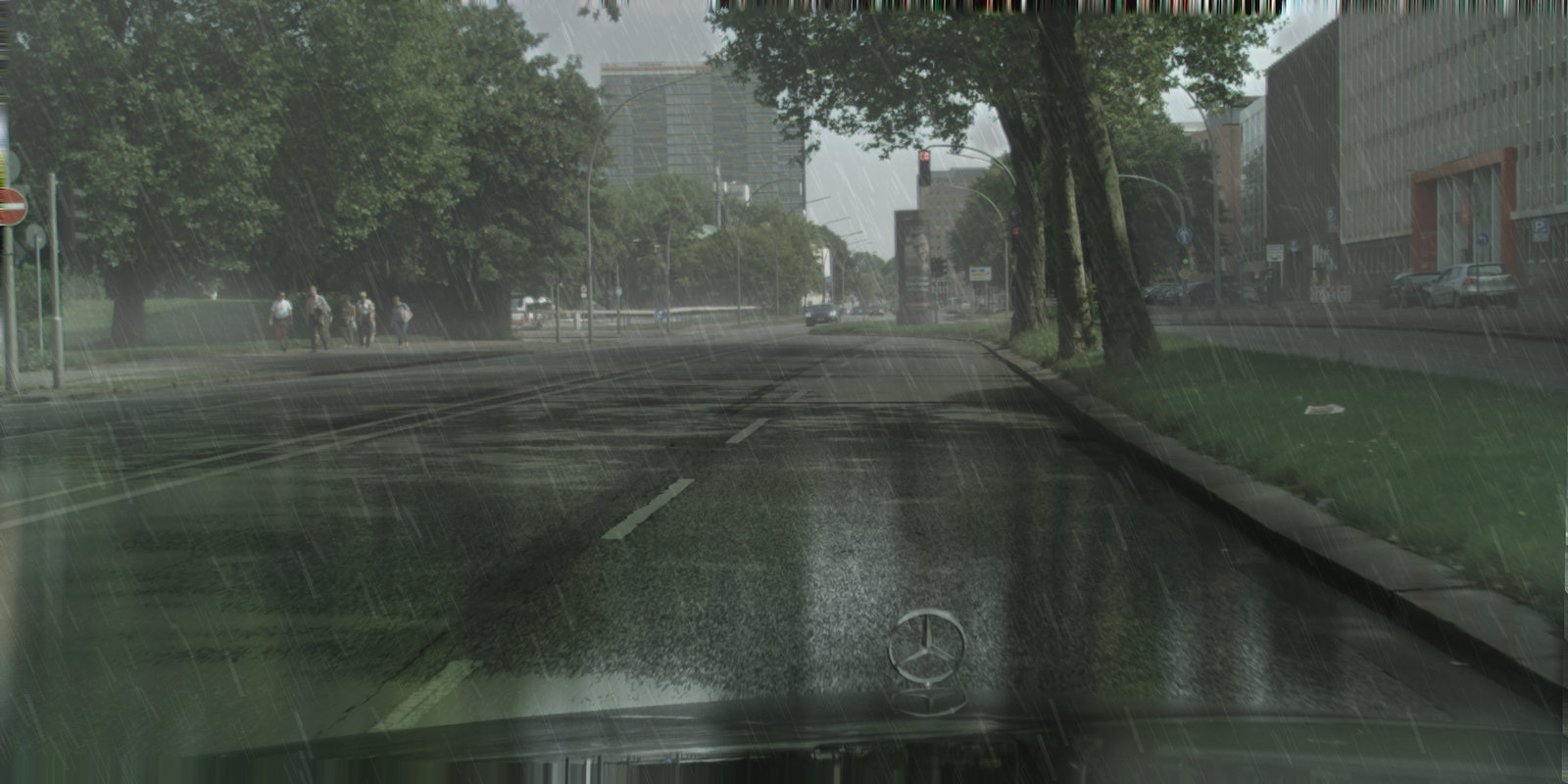} &
\includegraphics[width=\linewidth]{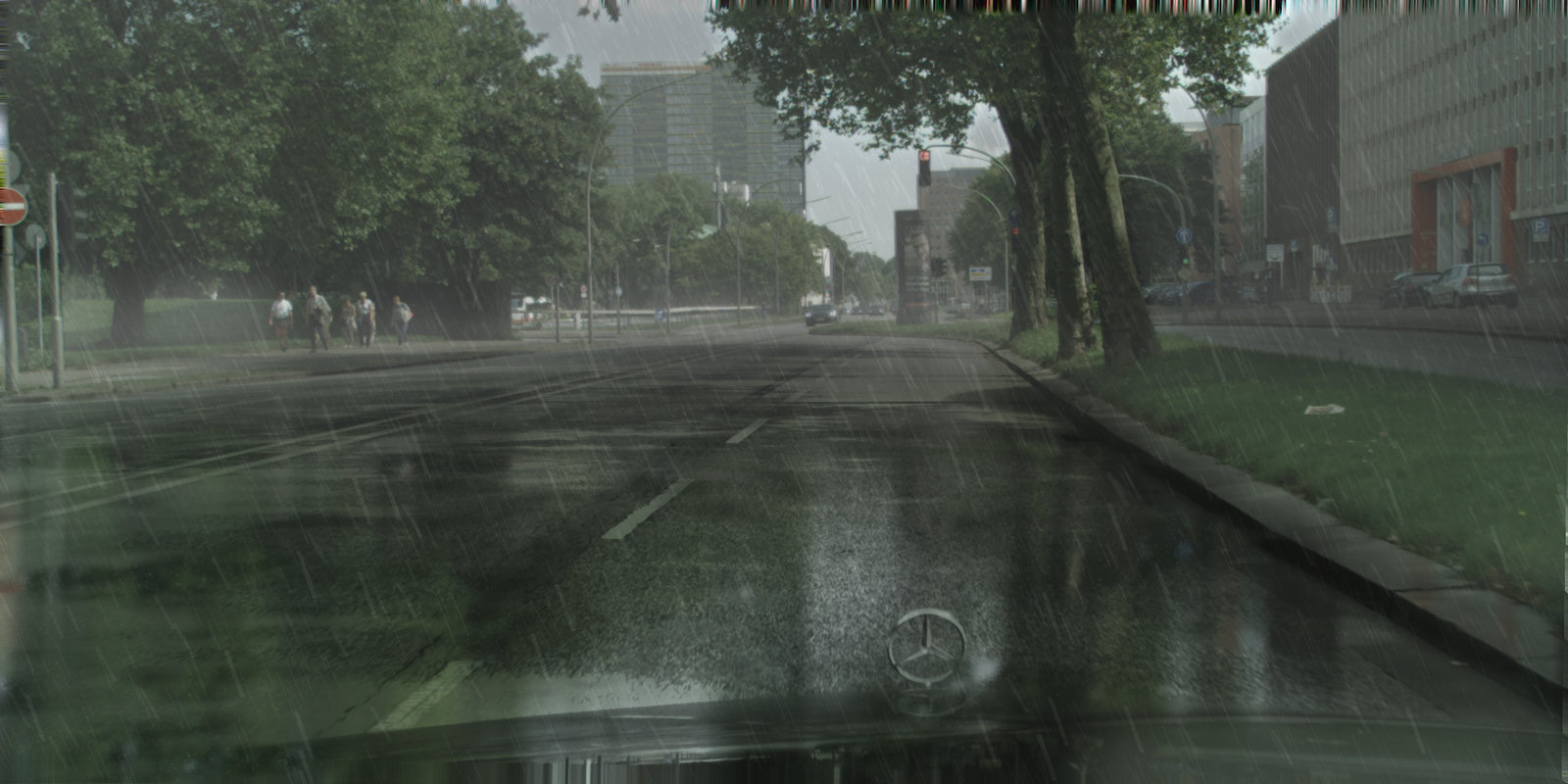} \\[-1mm]

\includegraphics[width=\linewidth]{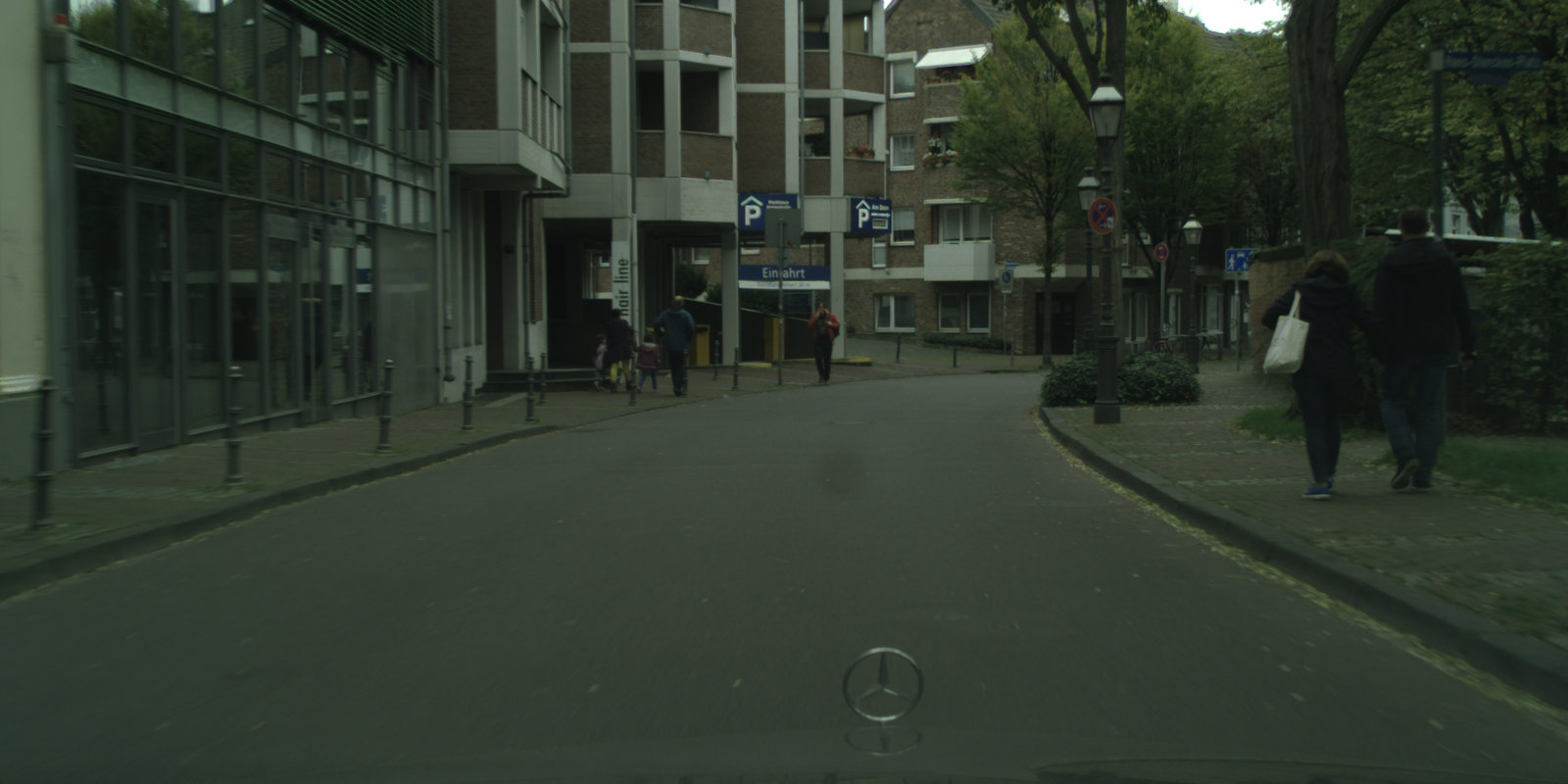} &
\includegraphics[width=\linewidth]{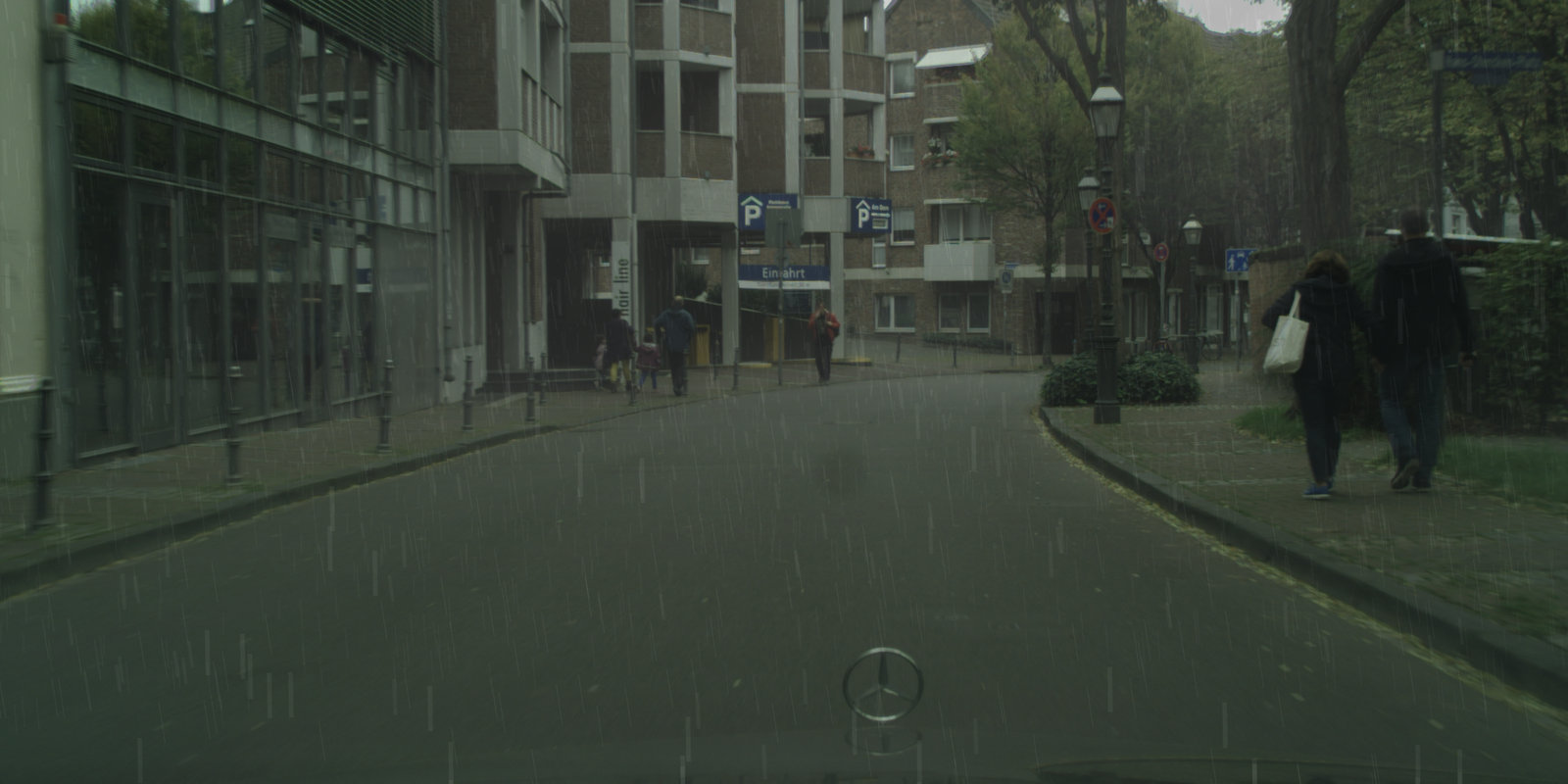} &
\includegraphics[width=\linewidth]{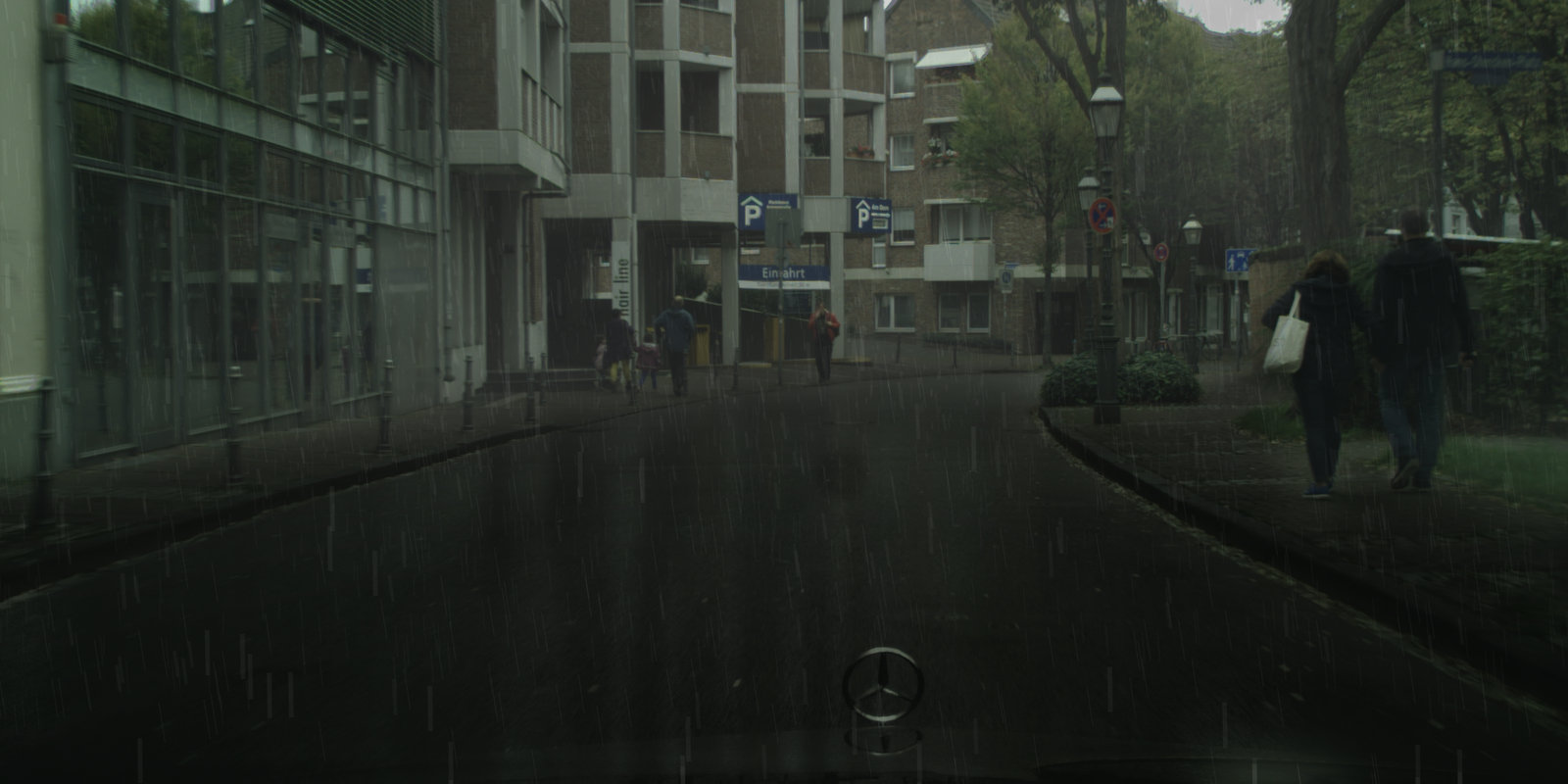} &
\includegraphics[width=\linewidth]{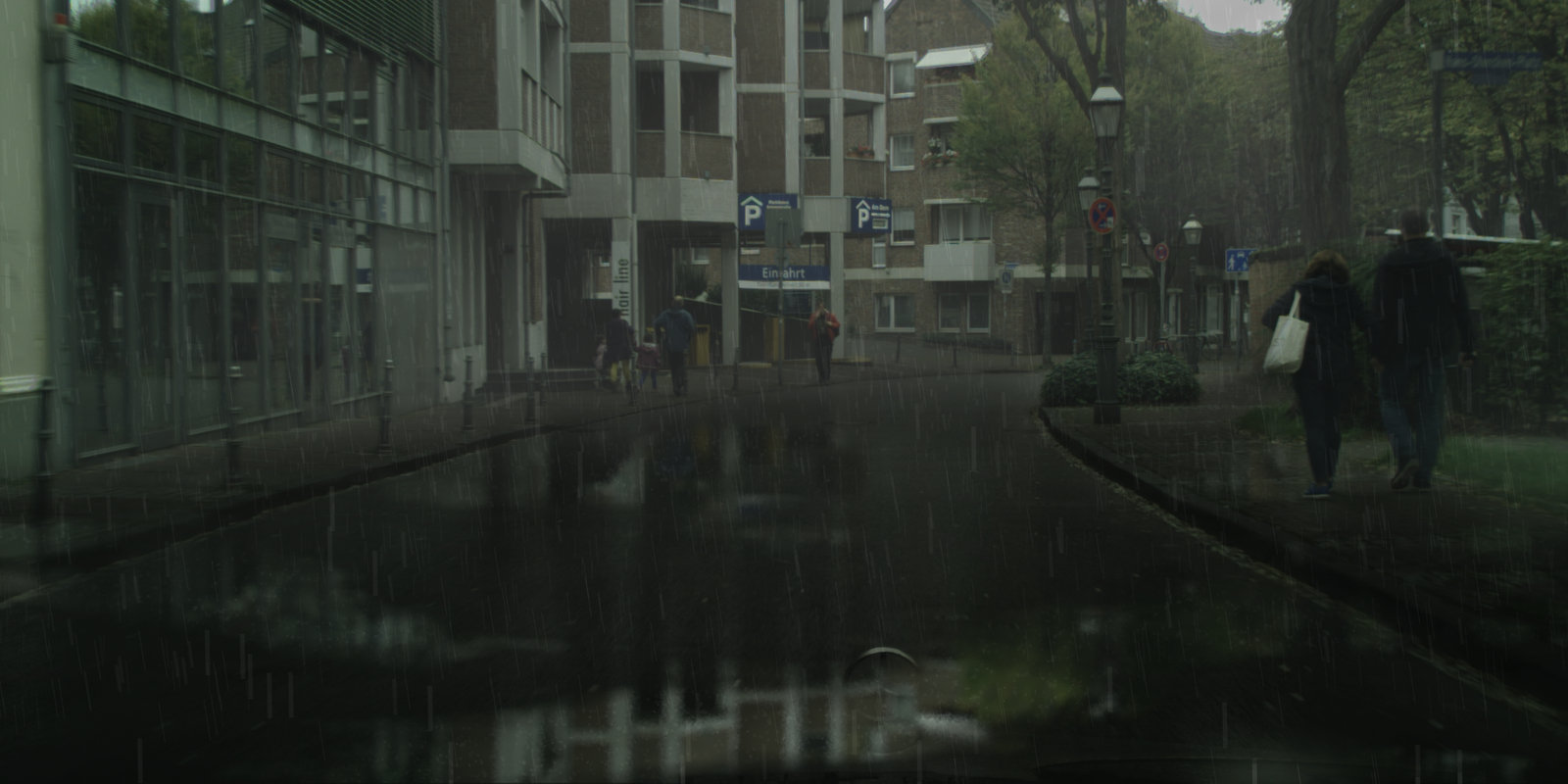} \\ [-1mm]

\includegraphics[width=\linewidth]{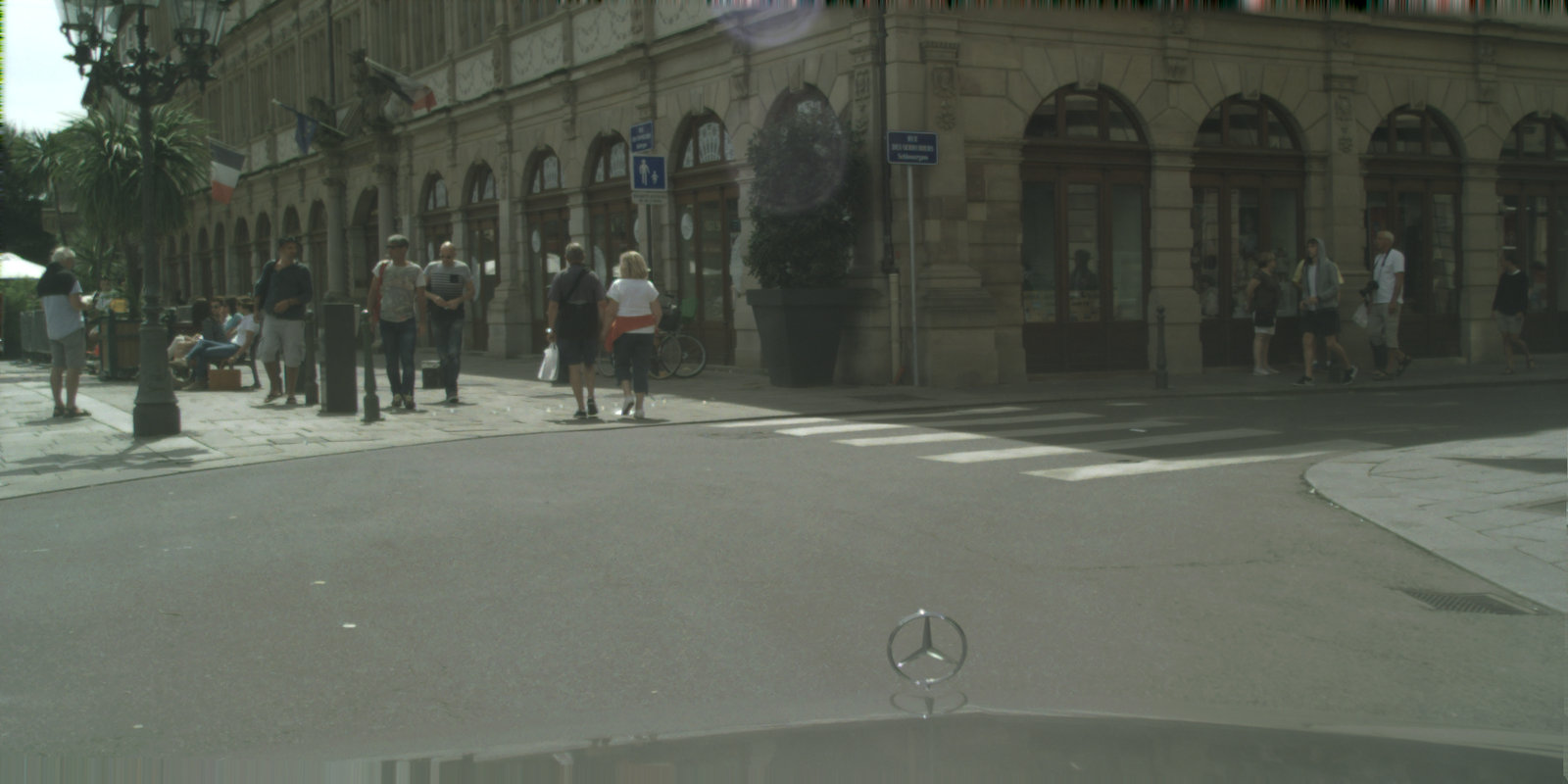} &
\includegraphics[width=\linewidth]{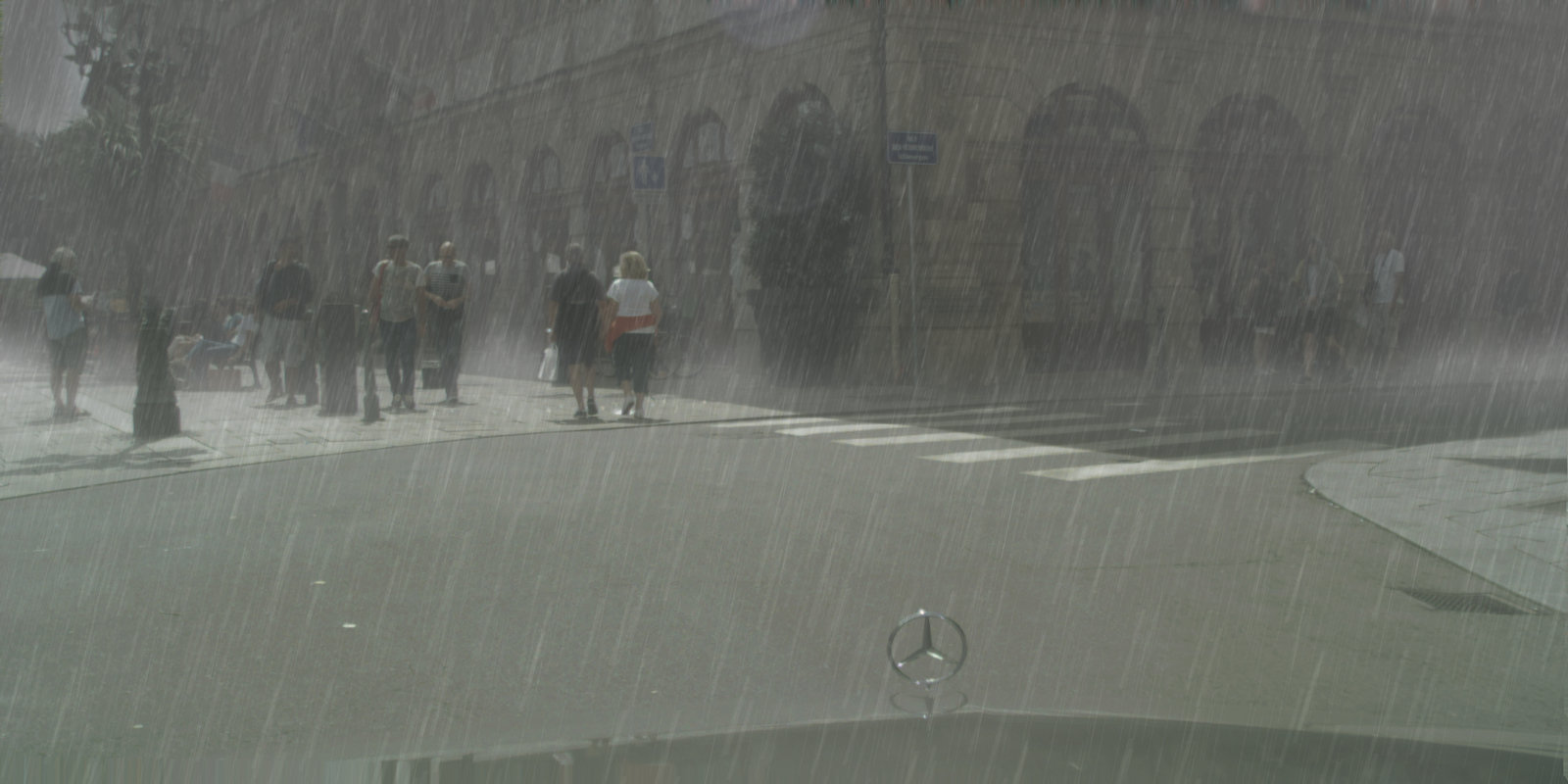} &
\includegraphics[width=\linewidth]{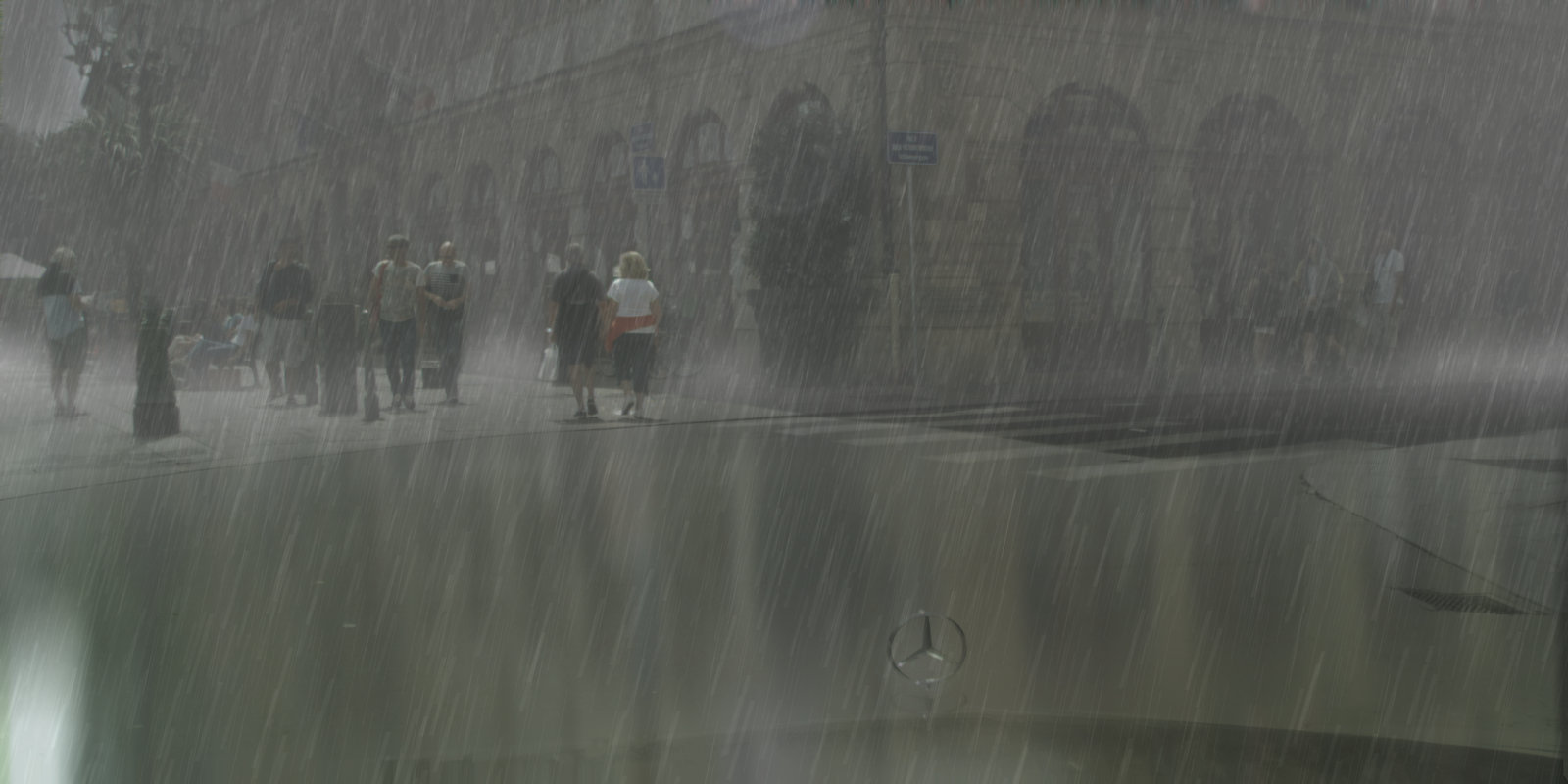} &
\includegraphics[width=\linewidth]{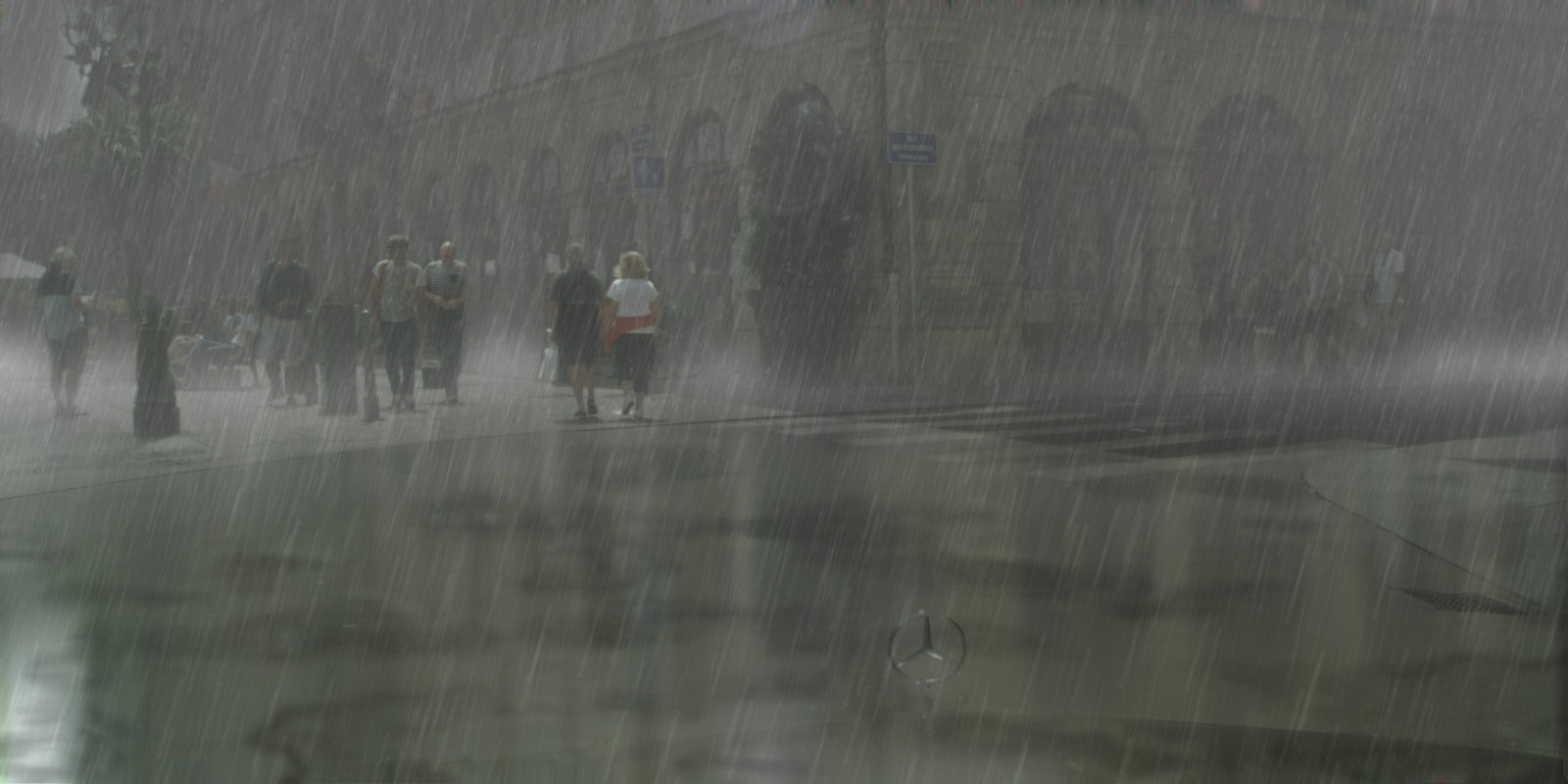} \\[-1mm]

\includegraphics[width=\linewidth]{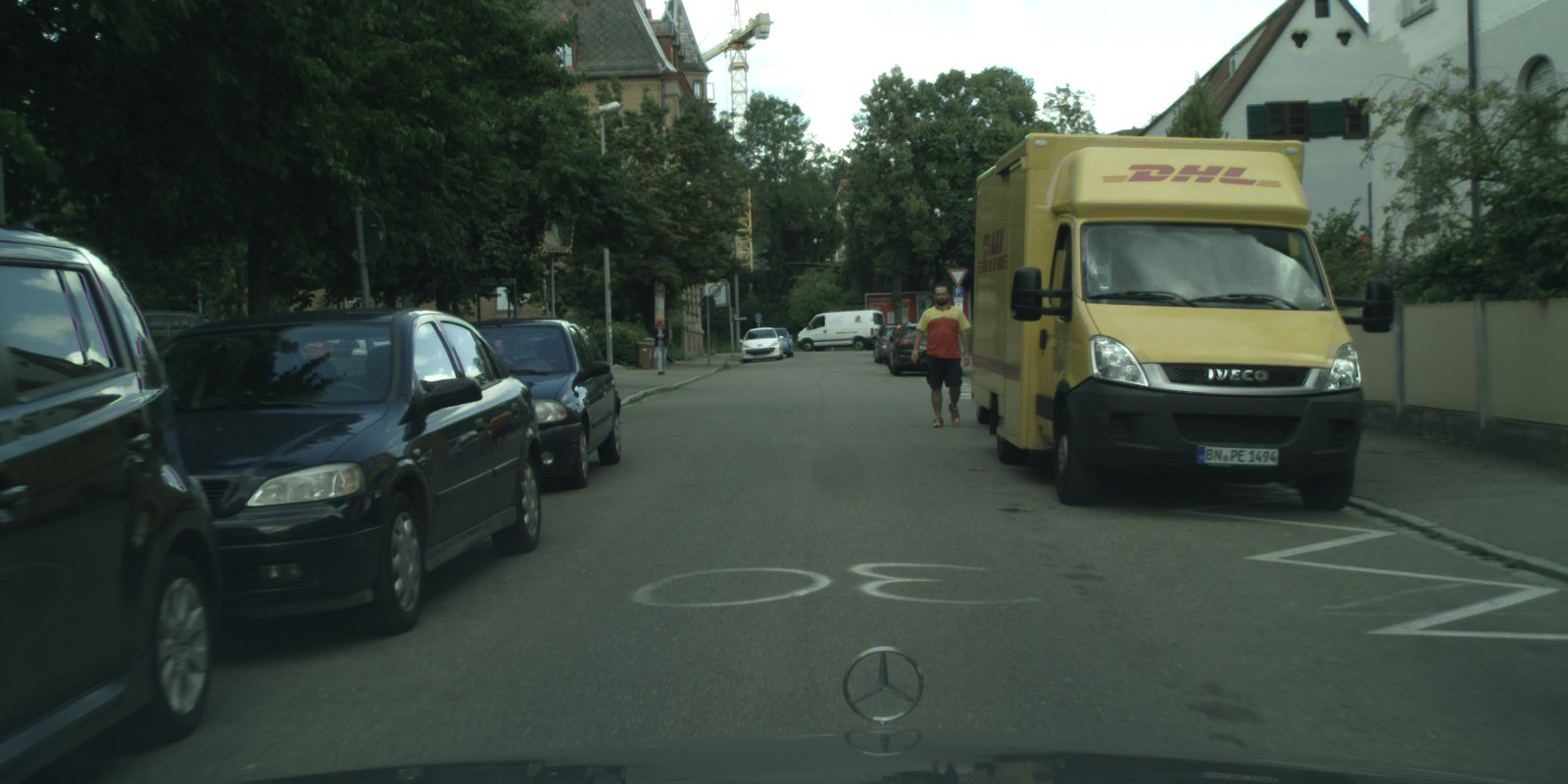} &
\includegraphics[width=\linewidth]{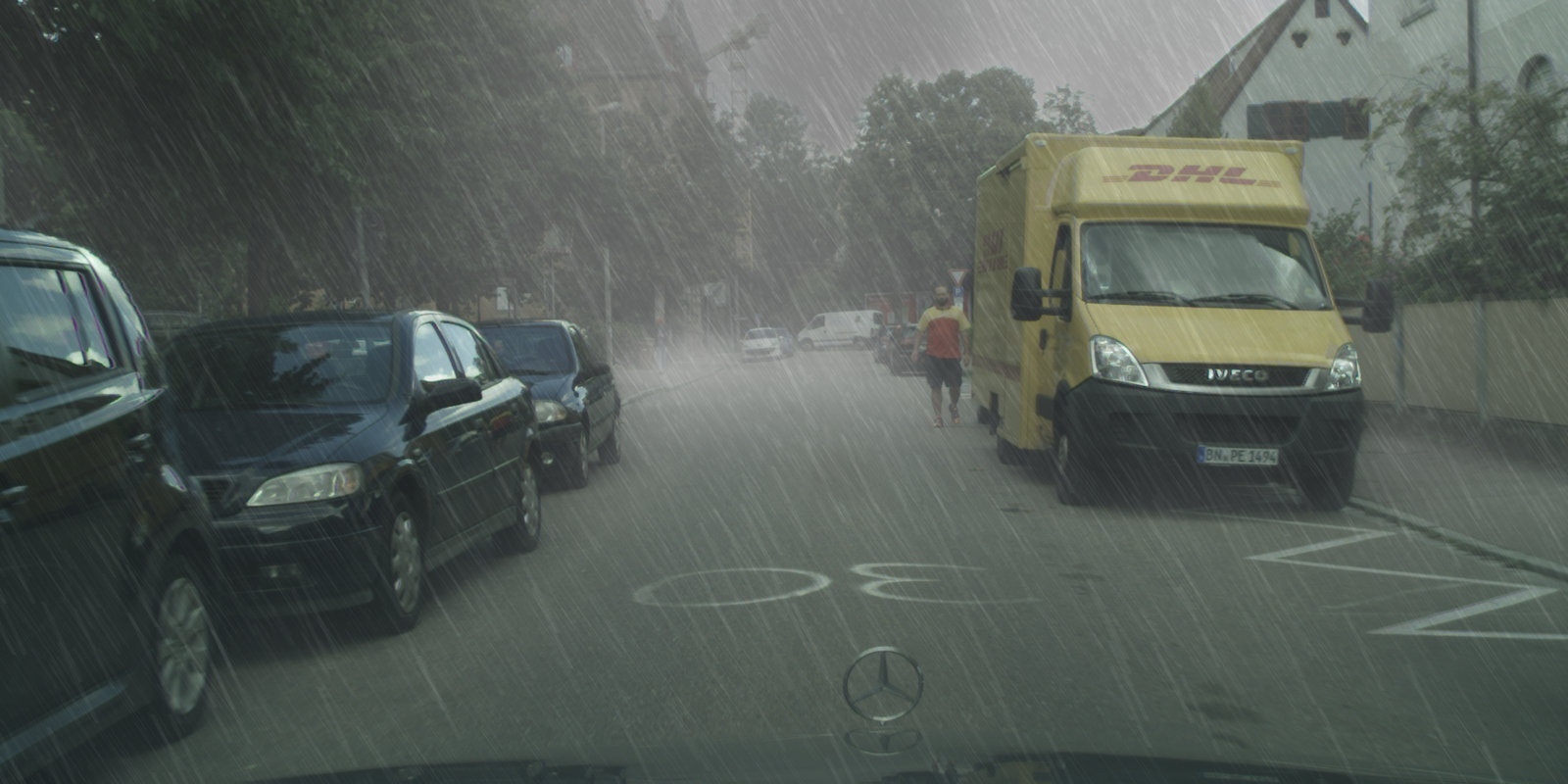} &
\includegraphics[width=\linewidth]{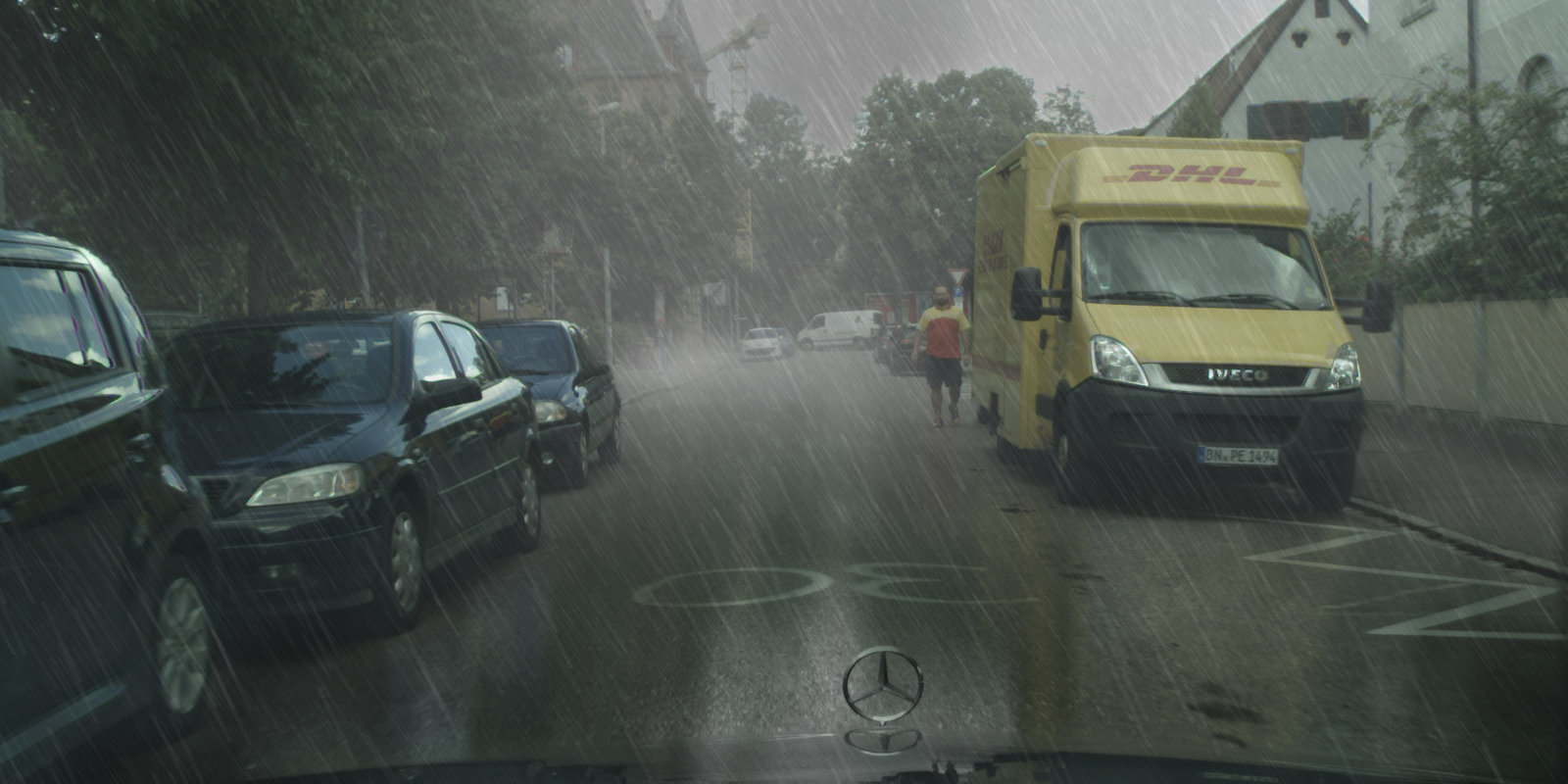} &
\includegraphics[width=\linewidth]{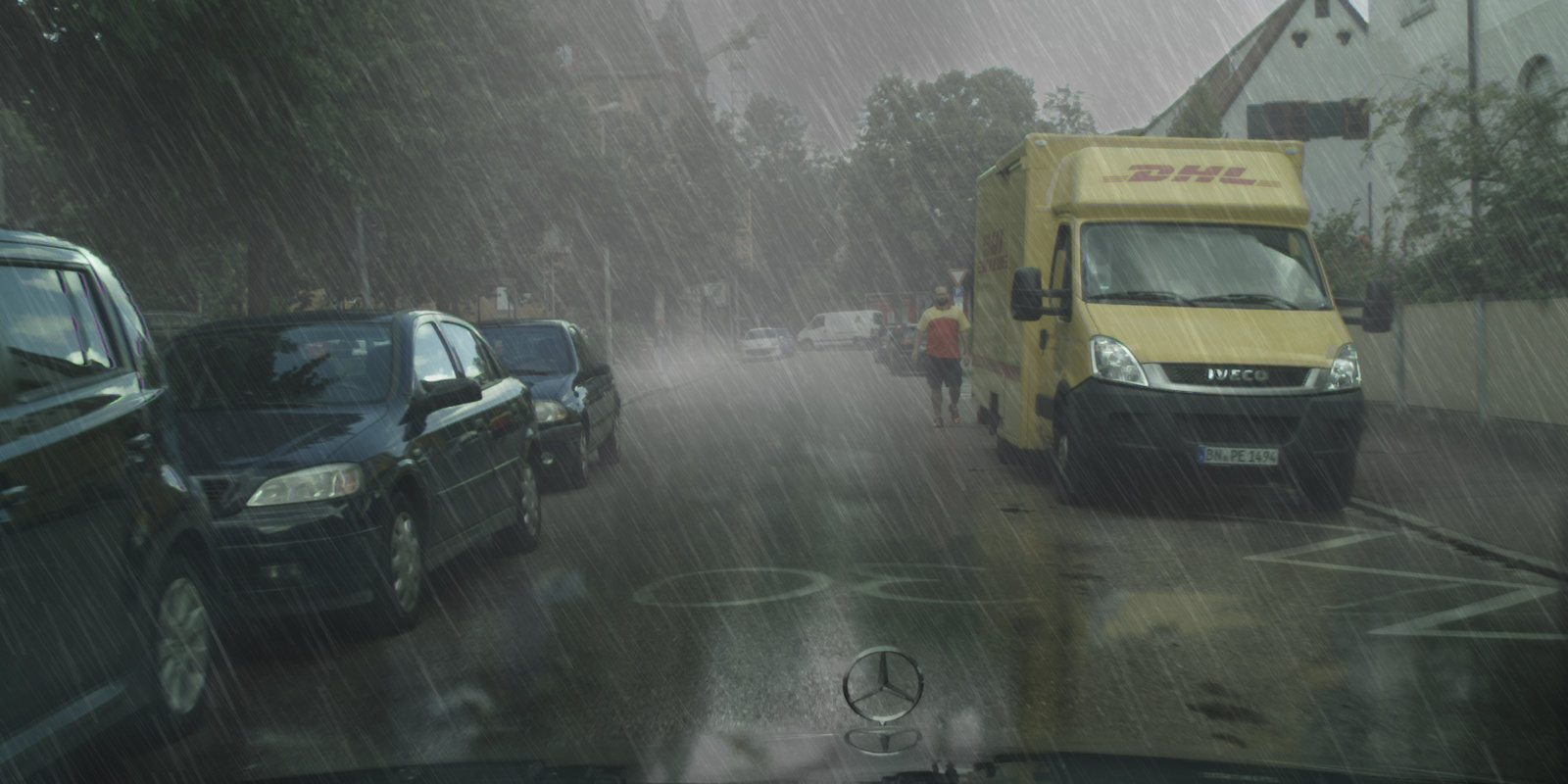} \\[-1mm]

\includegraphics[width=\linewidth]{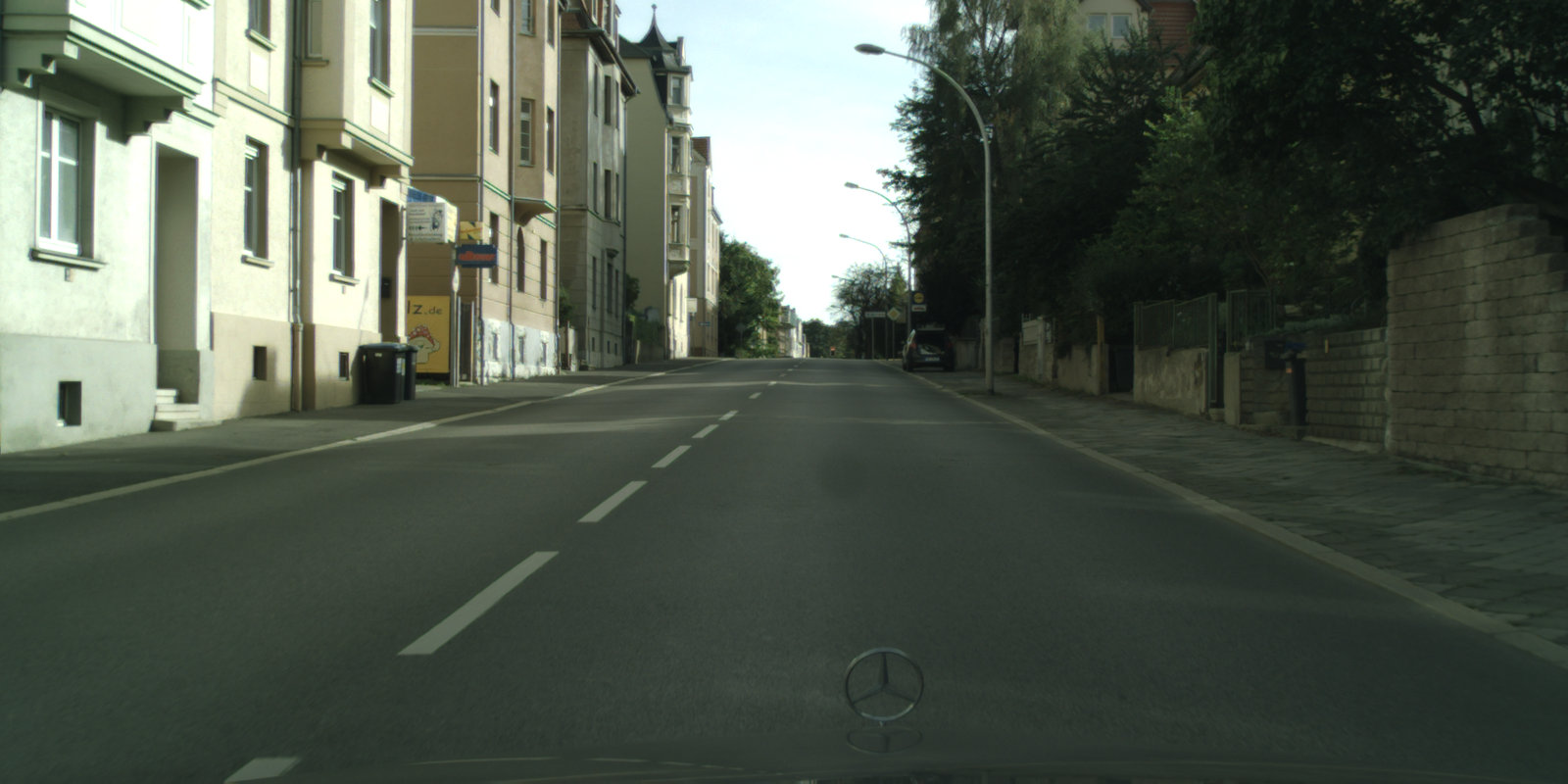} &
\includegraphics[width=\linewidth]{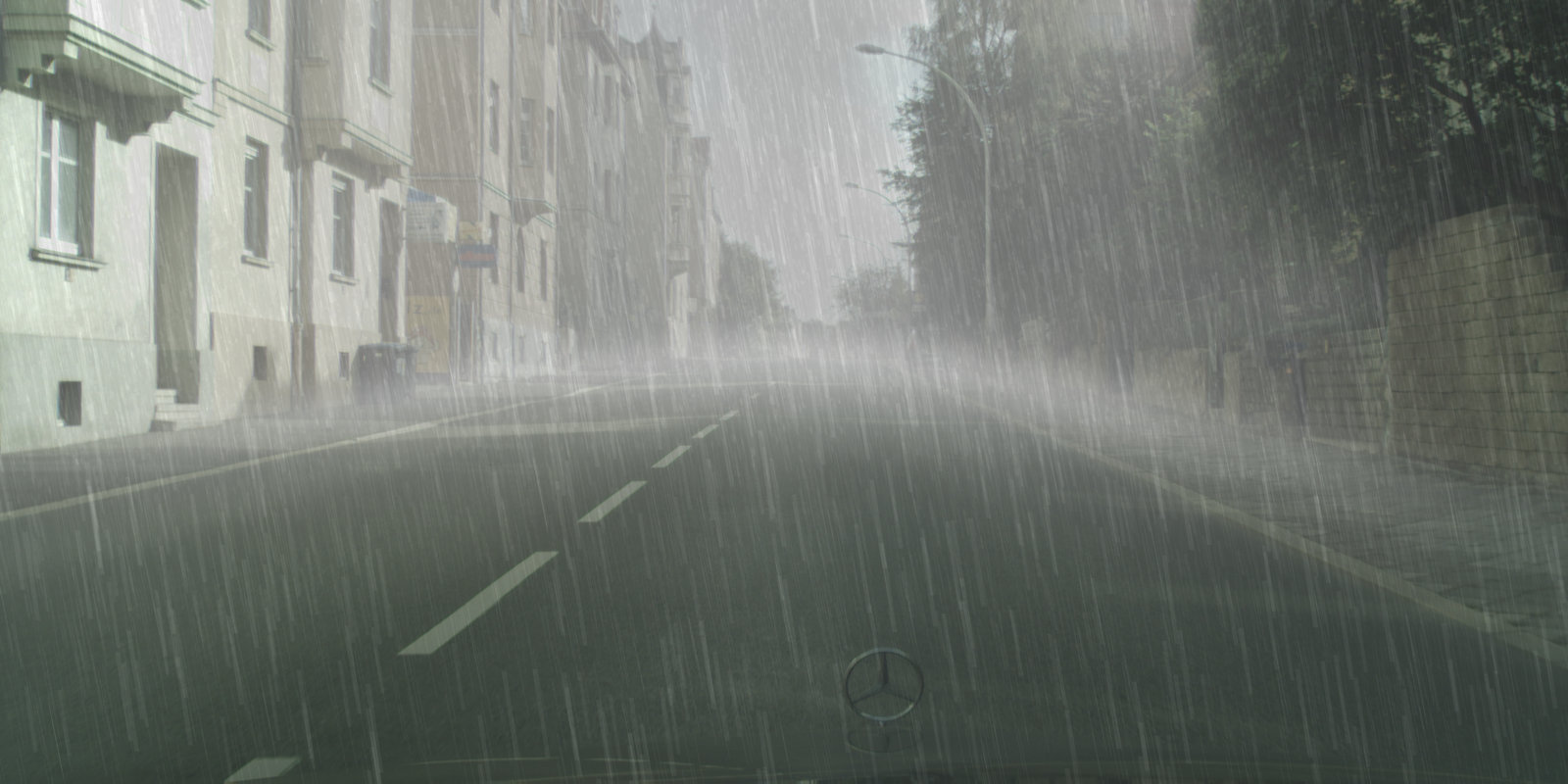} &
\includegraphics[width=\linewidth]{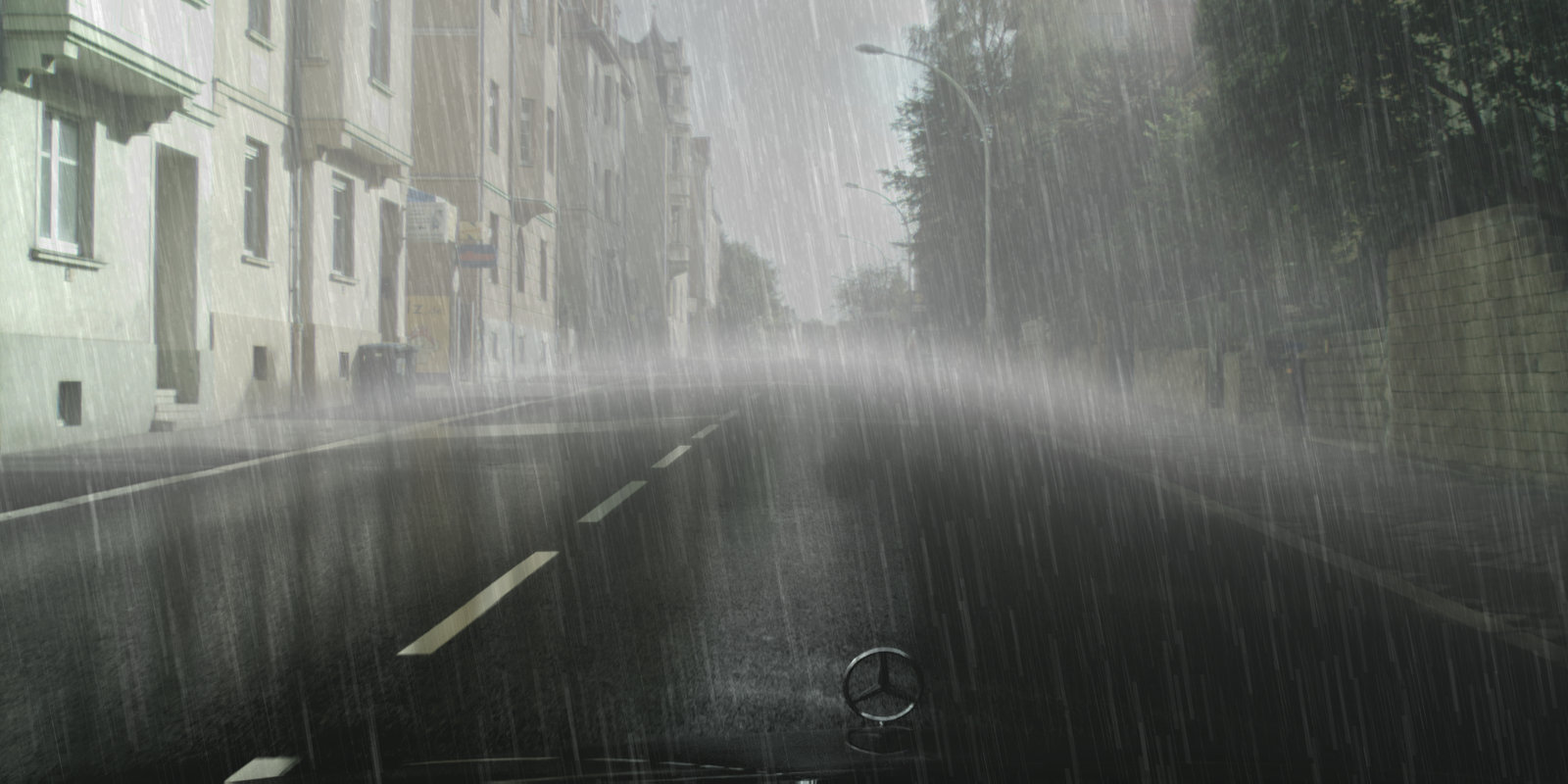} &
\includegraphics[width=\linewidth]{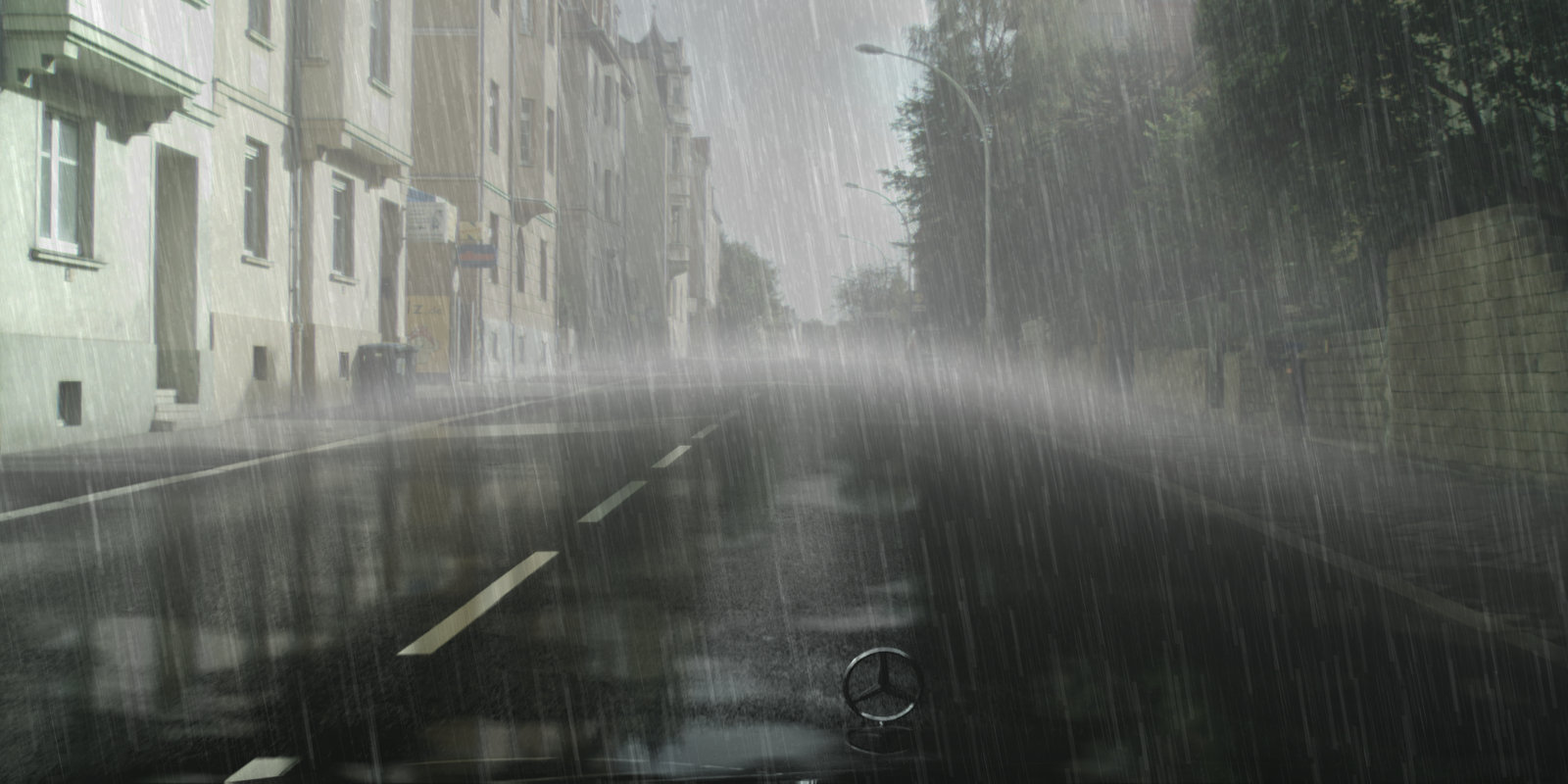} \\[-1mm]

\includegraphics[width=\linewidth]{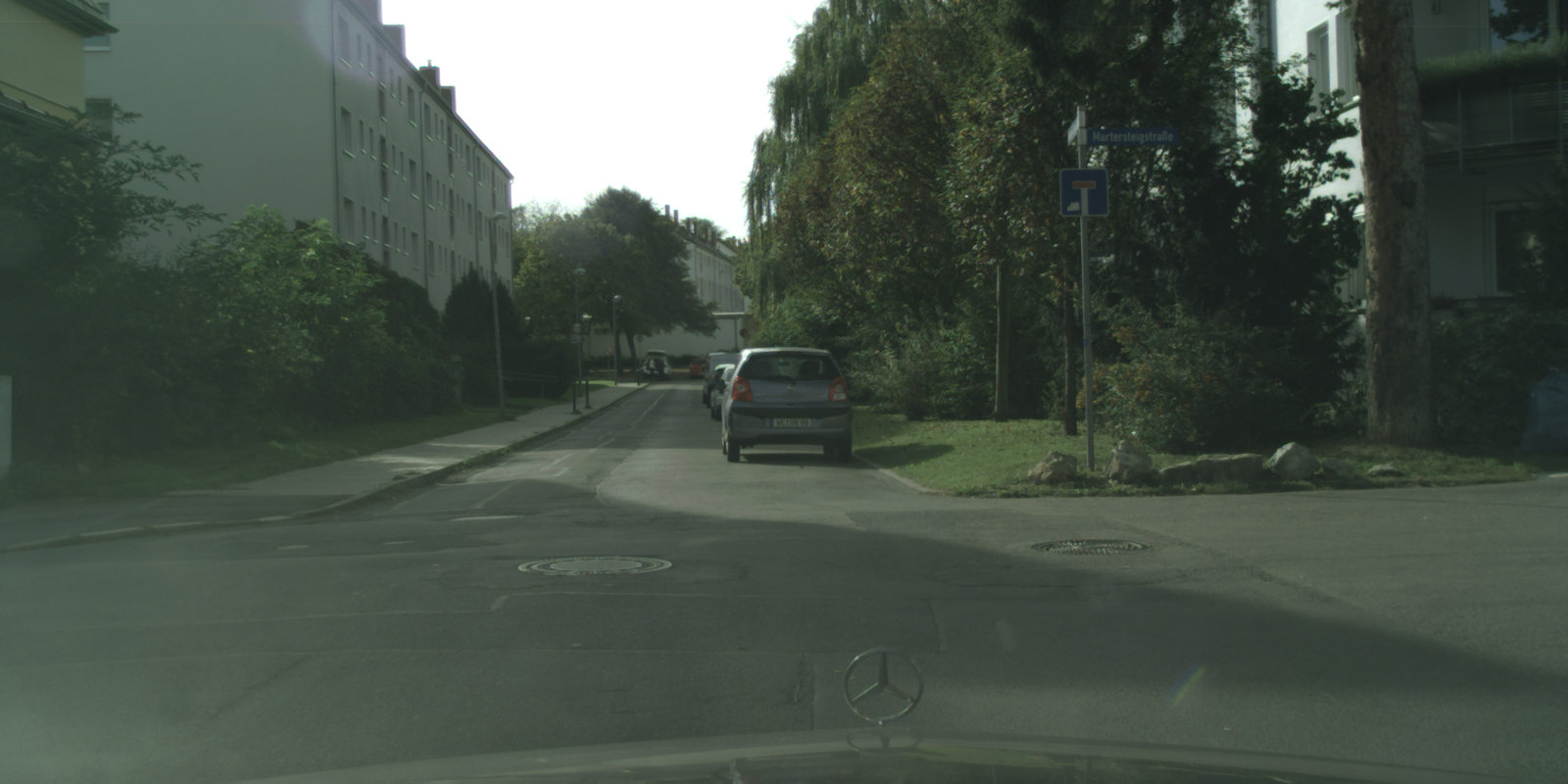} &
\includegraphics[width=\linewidth]{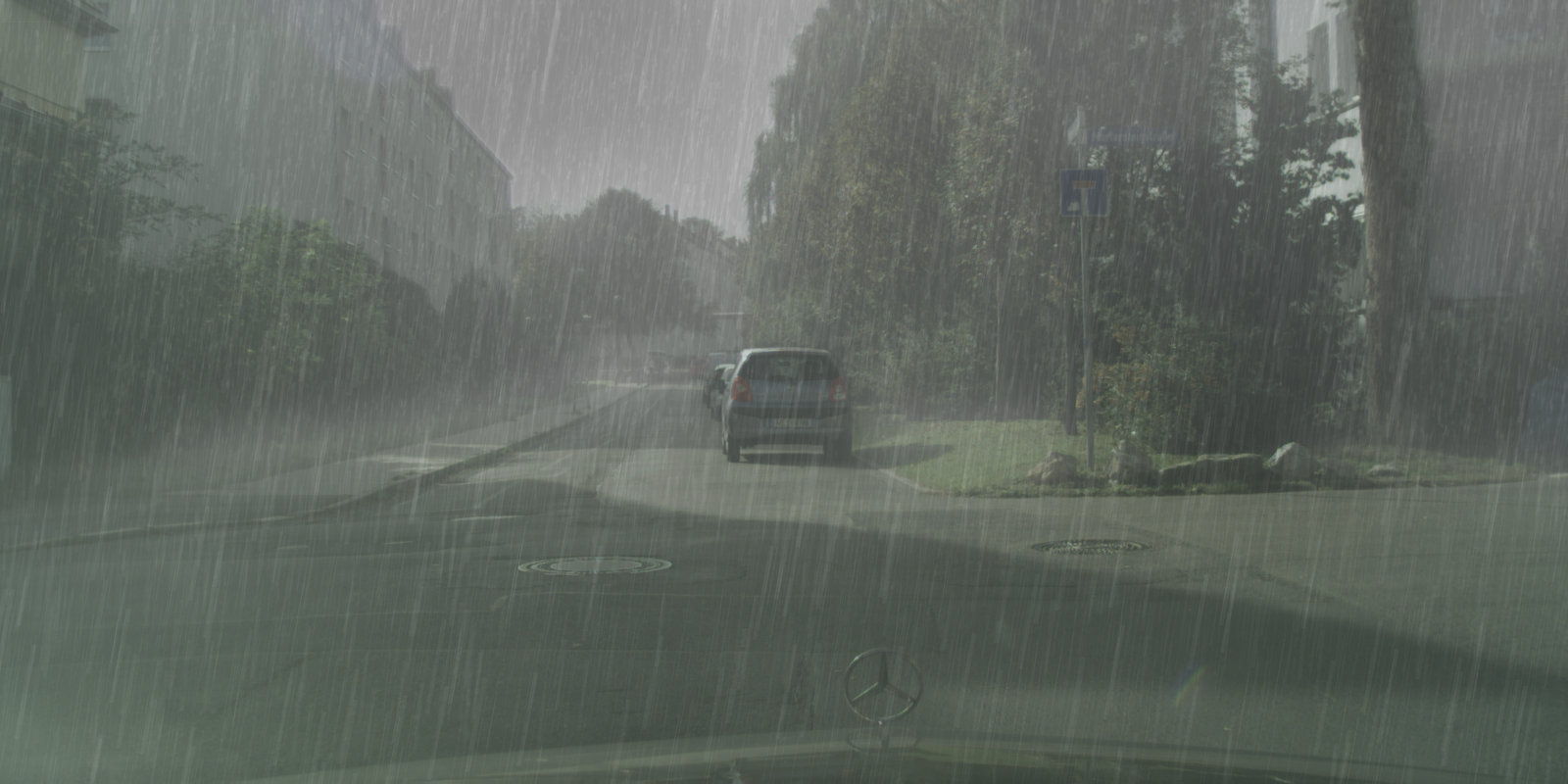} &
\includegraphics[width=\linewidth]{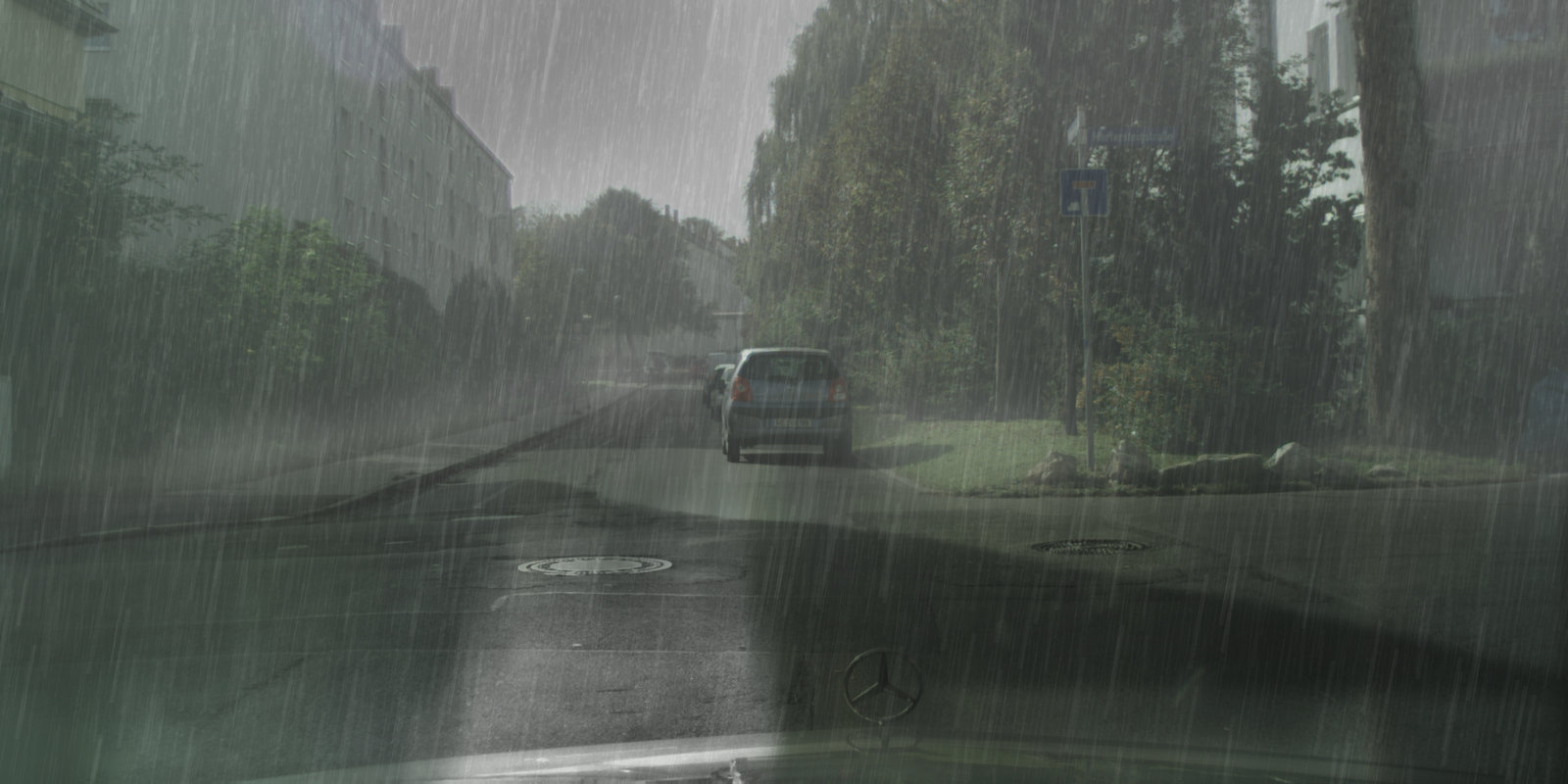} &
\includegraphics[width=\linewidth]{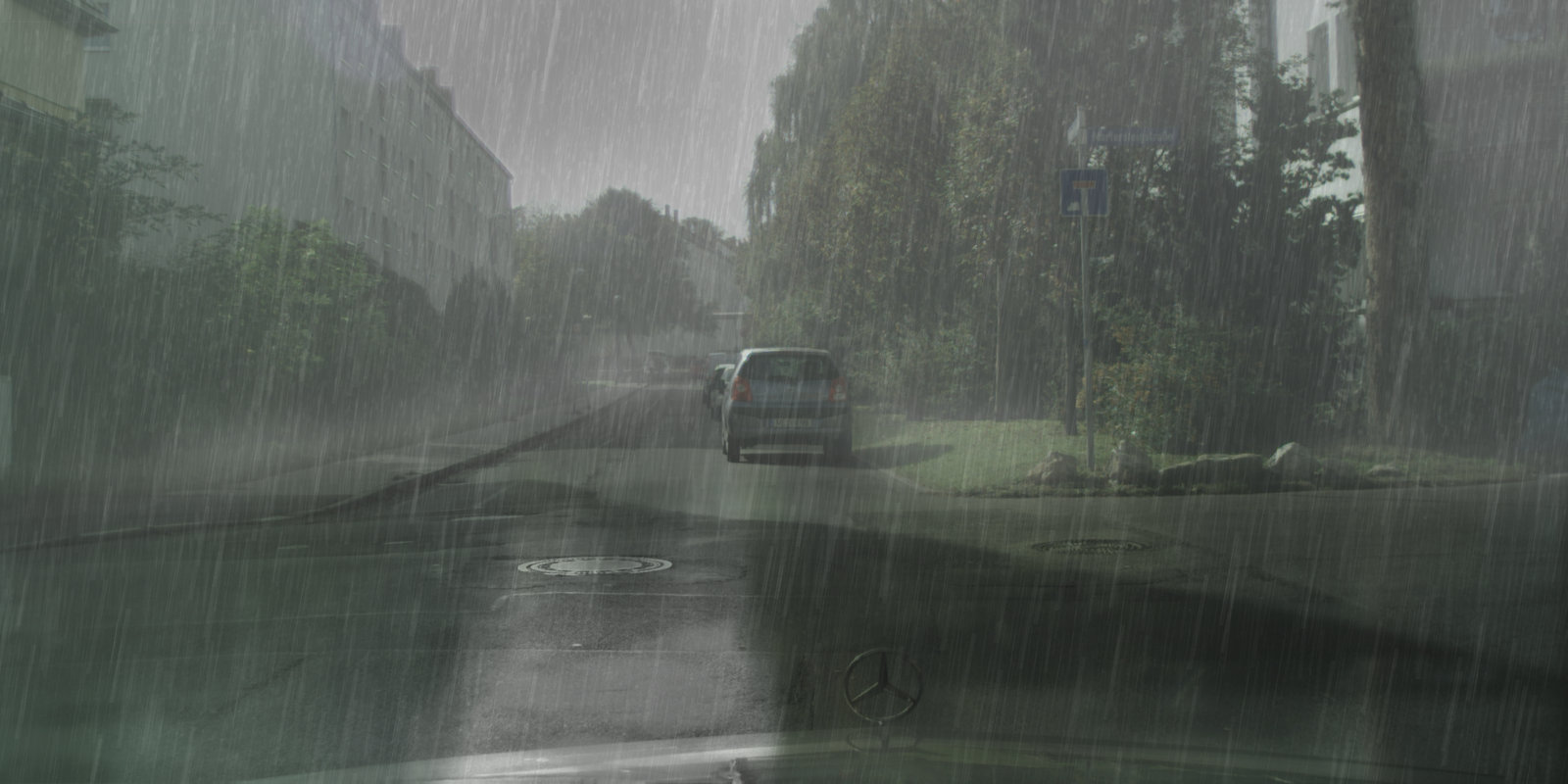} \\[-1mm]

\includegraphics[width=\linewidth]{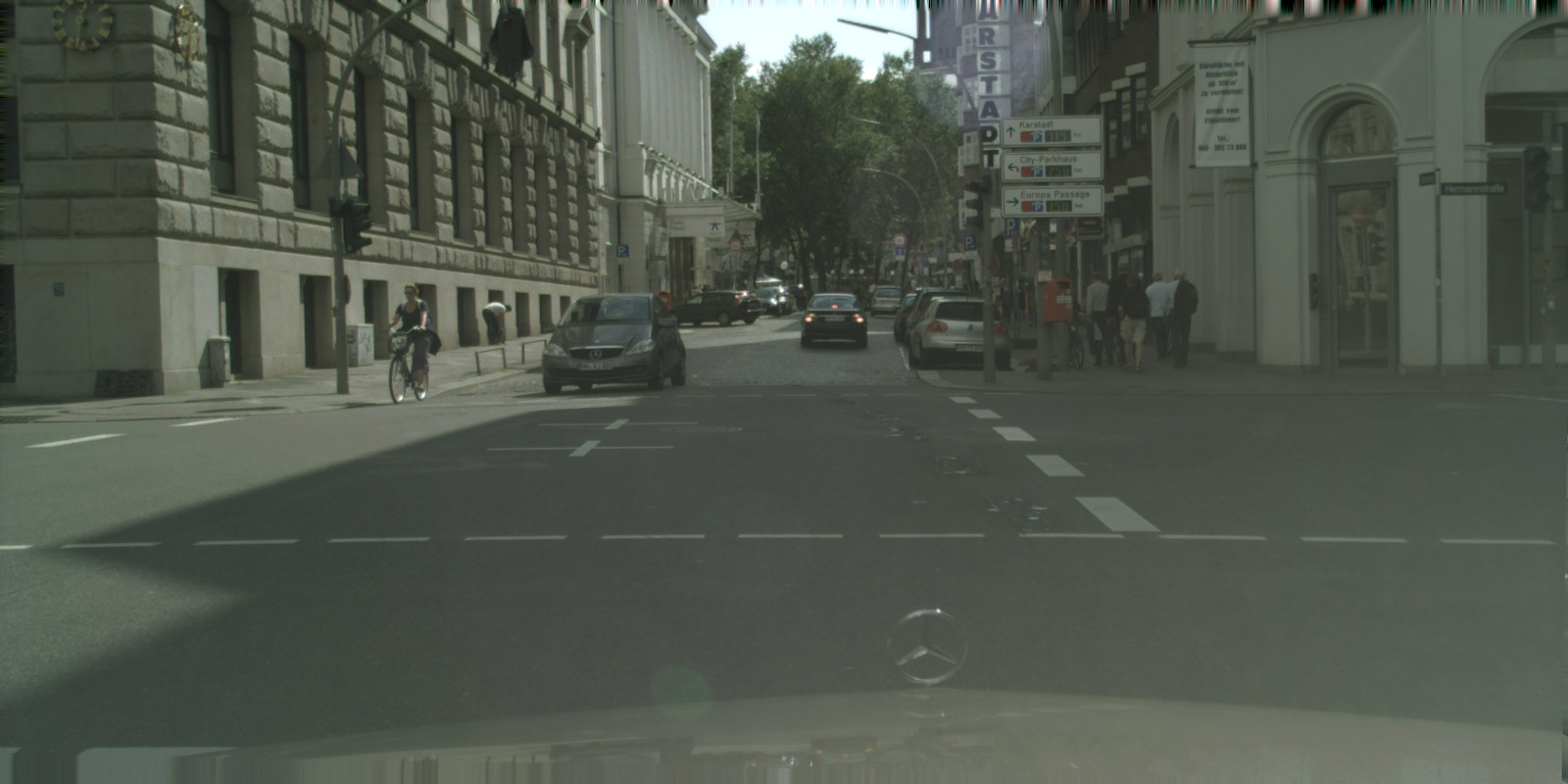} &
\includegraphics[width=\linewidth]{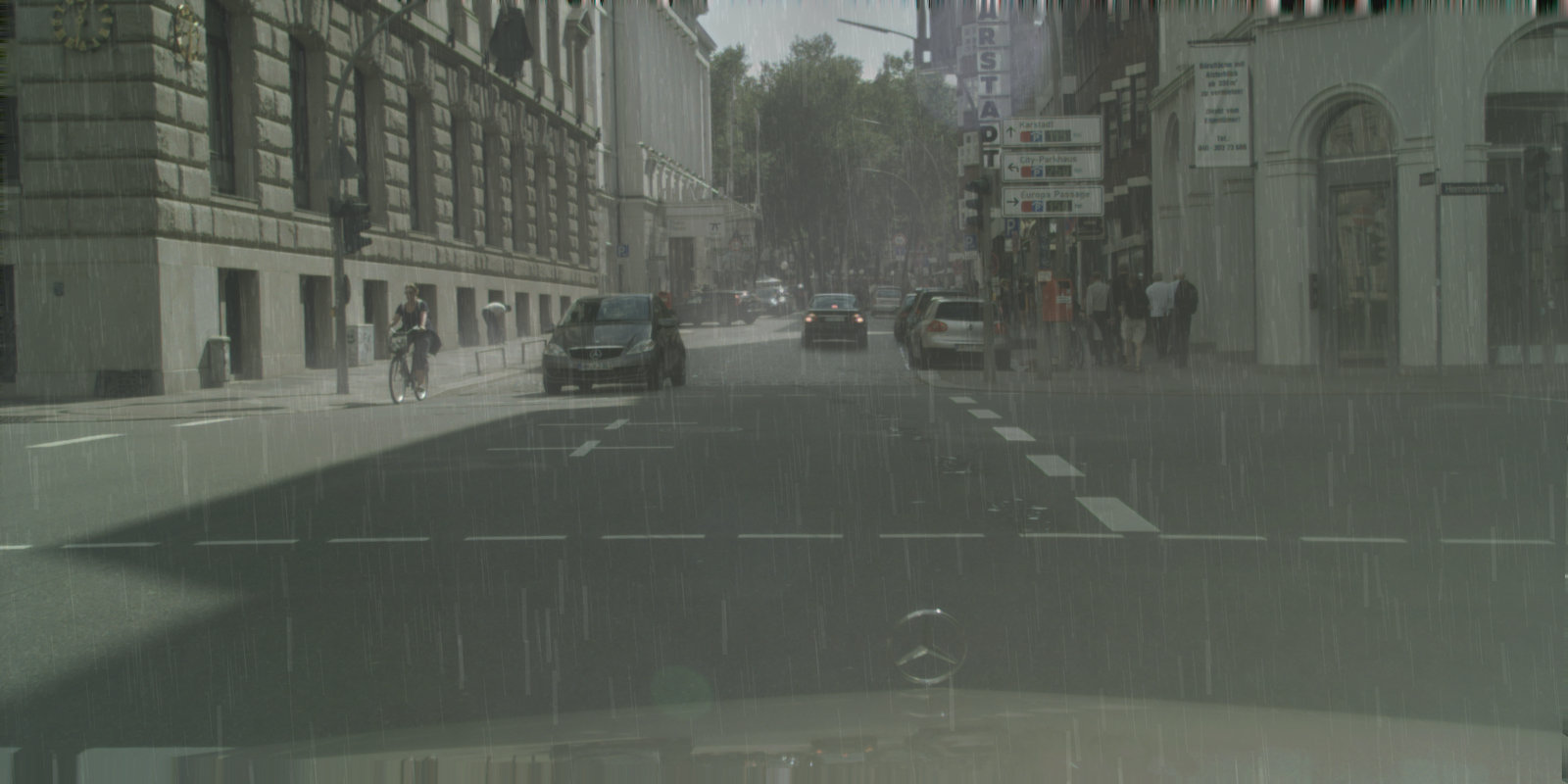} &
\includegraphics[width=\linewidth]{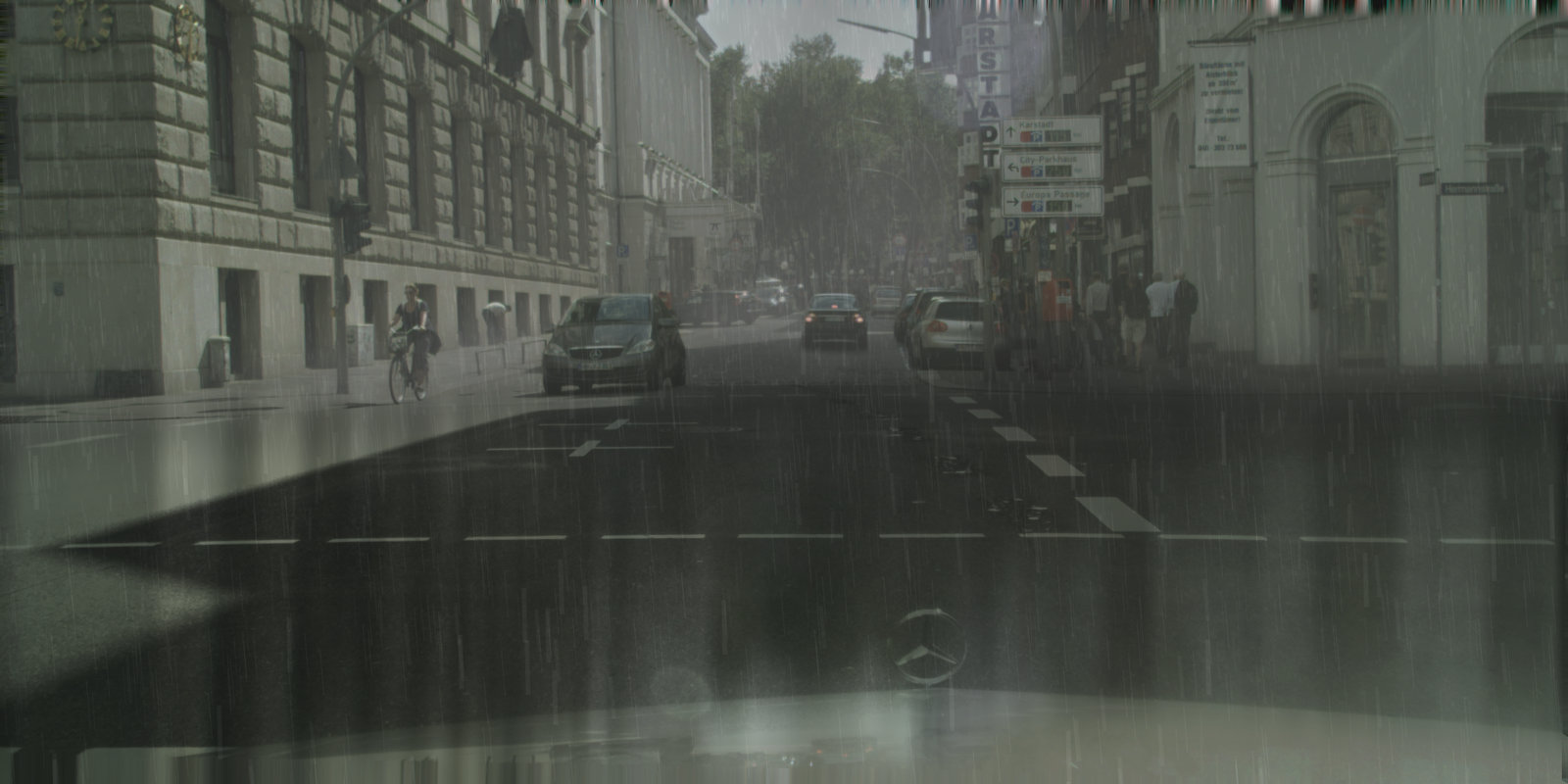} &
\includegraphics[width=\linewidth]{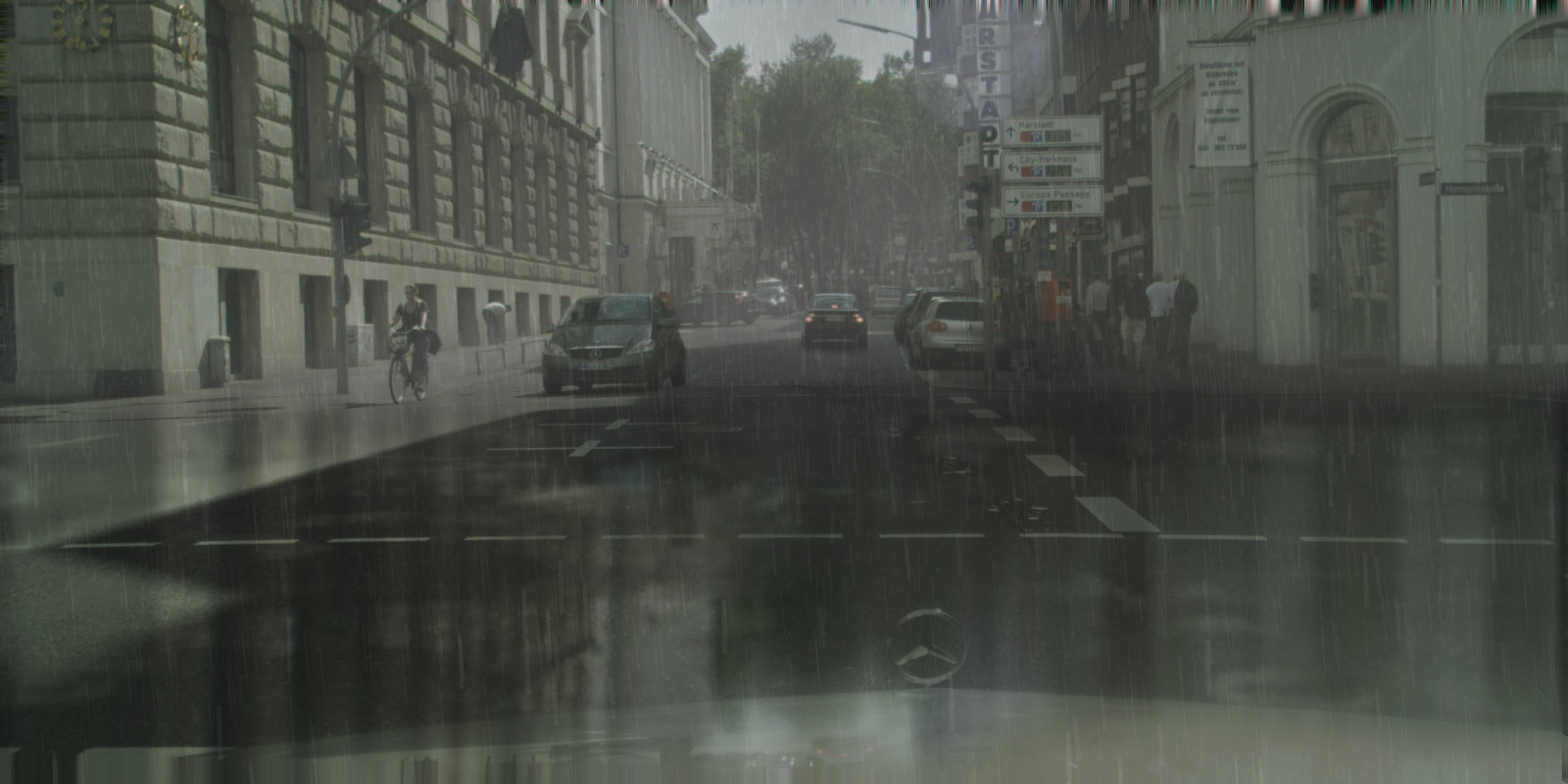} \\[-1mm]

\bottomrule
\end{tabular}

\label{tab:cascaders_multiscene}
\vspace{3mm}
\end{figure*}

\begin{figure*}[p]
\centering
\captionsetup[subfigure]{labelformat=empty}
\setlength{\tabcolsep}{2pt}


\newcommand{\magfactA}{2.5}
\newcommand{\zoomA}{(2.4,-1.3)}
\newcommand{\spyposA}{(-3,-0.1)}

\begin{minipage}{\textwidth}
\centering
\begin{tabular}{cc}
\begin{tikzpicture}[spy using outlines={rectangle, red, line width=0.4pt,
   magnification=\magfactA, connect spies, size=2.0cm}]
  \node[anchor=center] (imgA)
    {\includegraphics[width=0.48\linewidth]{figures/PART3/originals/hamburg_000000_069417_leftImg8bit.jpg}};
  \spy on \zoomA in node [right] at \spyposA;
\end{tikzpicture}
&
\begin{tikzpicture}[spy using outlines={rectangle, red, line width=0.4pt,
   magnification=\magfactA, connect spies, size=2.0cm}]
  \node[anchor=center] (imgB)
    {\includegraphics[width=0.48\linewidth]{figures/PART3/base/hamburg_000000_069417_leftImg8bit_rain.jpg}};
  \spy on \zoomA in node [right] at \spyposA;
\end{tikzpicture}
\\[-2pt]

\begin{tikzpicture}[spy using outlines={rectangle, red, line width=0.4pt,
   magnification=\magfactA, connect spies, size=2.0cm}]
  \node[anchor=center] (imgC)
    {\includegraphics[width=0.48\linewidth]{figures/PART3/wet/hamburg_000000_069417_leftImg8bit_rain_w.jpg}};
  \spy on \zoomA in node [right] at \spyposA;
\end{tikzpicture}
&
\begin{tikzpicture}[spy using outlines={rectangle, red, line width=0.4pt,
   magnification=\magfactA, connect spies, size=2.0cm}]
  \node[anchor=center] (imgD)
    {\includegraphics[width=0.48\linewidth]{figures/PART3/puddle/hamburg_000000_069417_leftImg8bit_rain_wp.jpg}};
  \spy on \zoomA in node [right] at \spyposA;
\end{tikzpicture}
\end{tabular}
\end{minipage}

\raggedright
{\small
\textbf{(a) Effect of reflections and puddles in CascadeRS.}
A zoomed comparison showing how wet-surface reflections are formed (\eg, reflected tree structure) and how puddles further distort and reshape these reflections according to local surface geometry. The progression from rain-only to reflections and finally puddles demonstrates how increasingly detailed surface interactions contribute to the realism of the synthesized rainy scene.}

\vspace{6mm}


\newcommand{\magfactB}{2.3}
\newcommand{\zoomB}{(-2.5,-0.5)}
\newcommand{\spyposB}{(1,-0.2)}

\begin{minipage}{\textwidth}
\centering
\begin{tabular}{cc}
\begin{tikzpicture}[spy using outlines={rectangle, red, line width=0.4pt,
   magnification=\magfactB, connect spies, size=2.0cm}]
  \node[anchor=center] (imgA2)
    {\includegraphics[width=0.48\linewidth]{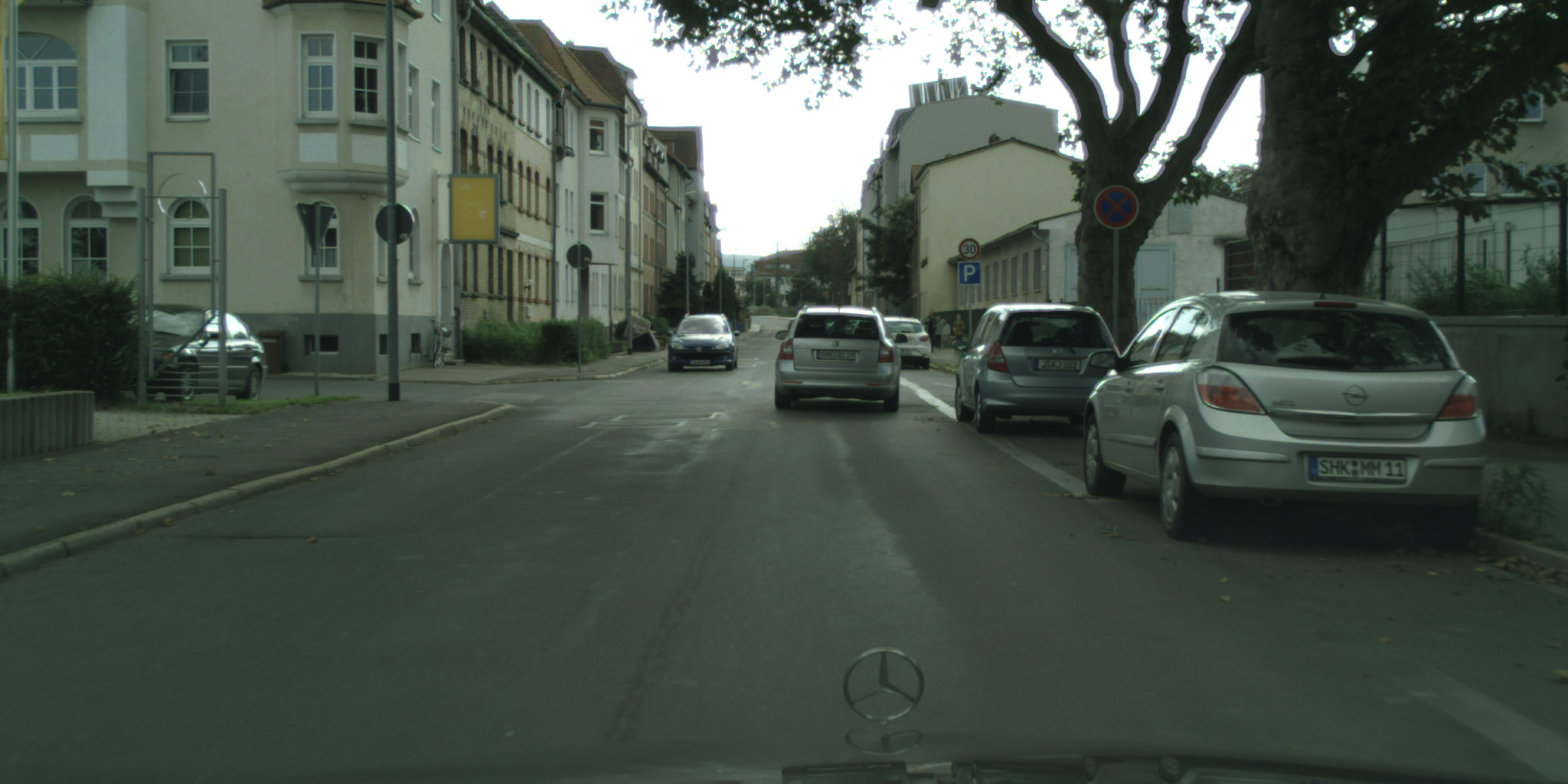}};
  \spy on \zoomB in node [right] at \spyposB;
\end{tikzpicture}
&
\begin{tikzpicture}[spy using outlines={rectangle, red, line width=0.4pt,
   magnification=\magfactB, connect spies, size=2.0cm}]
  \node[anchor=center] (imgB2)
    {\includegraphics[width=0.48\linewidth]{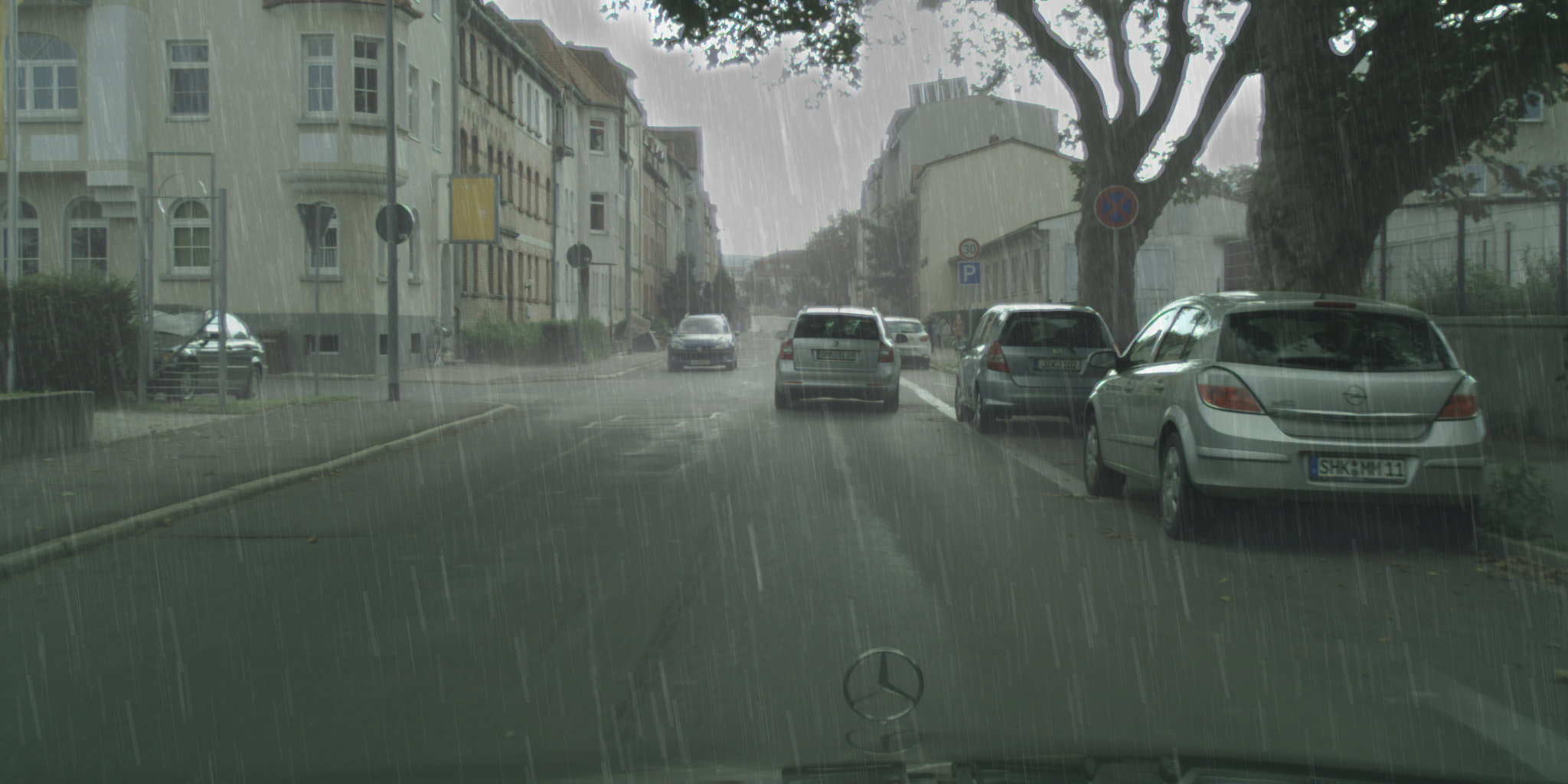}};
  \spy on \zoomB in node [right] at \spyposB;
\end{tikzpicture}
\\[-2pt]

\begin{tikzpicture}[spy using outlines={rectangle, red, line width=0.4pt,
   magnification=\magfactB, connect spies, size=2.0cm}]
  \node[anchor=center] (imgC2)
    {\includegraphics[width=0.48\linewidth]{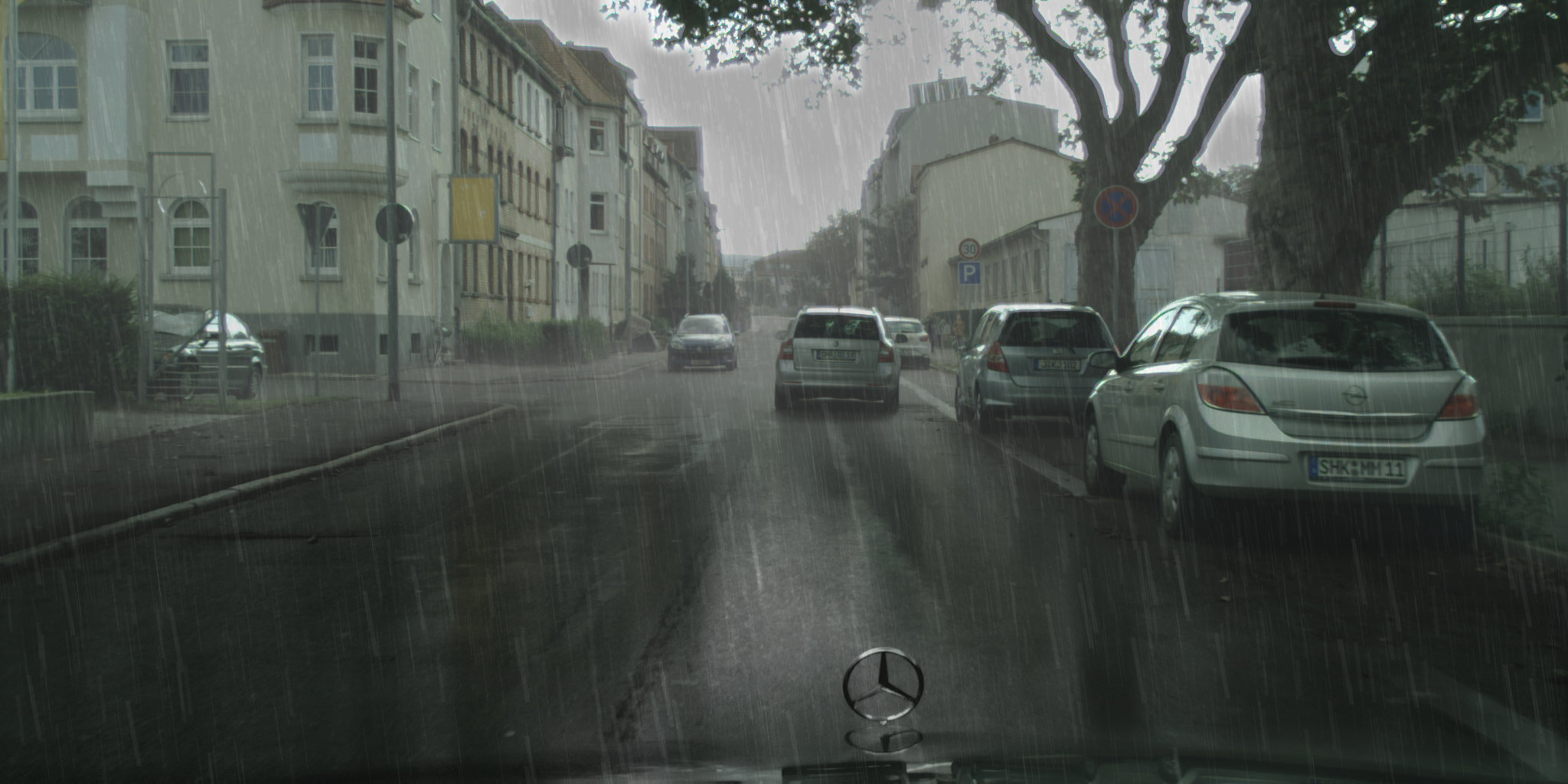}};
  \spy on \zoomB in node [right] at \spyposB;
\end{tikzpicture}
&
\begin{tikzpicture}[spy using outlines={rectangle, red, line width=0.4pt,
   magnification=\magfactB, connect spies, size=2.0cm}]
  \node[anchor=center] (imgD2)
    {\includegraphics[width=0.48\linewidth]{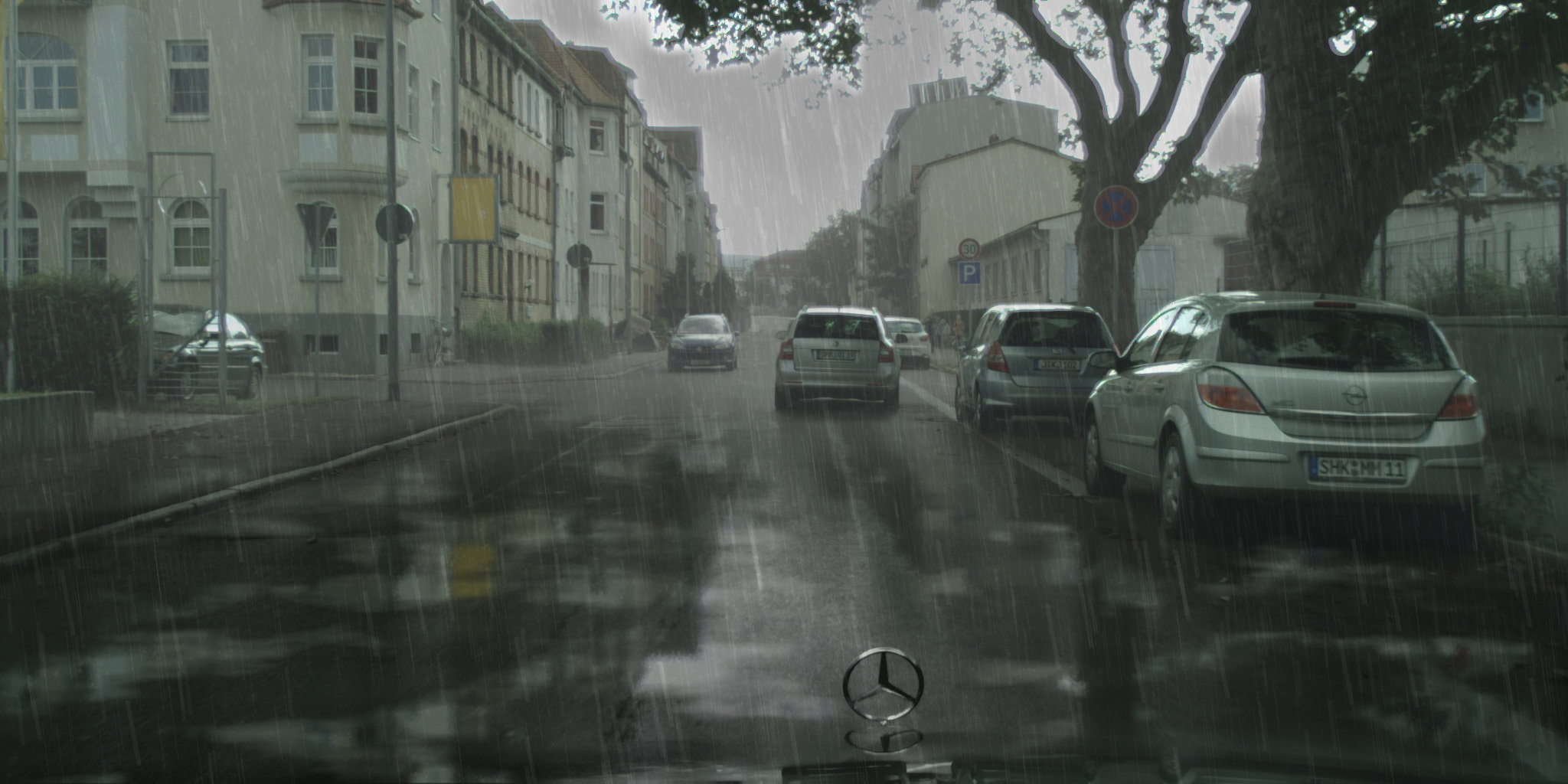}};
  \spy on \zoomB in node [right] at \spyposB;
\end{tikzpicture}
\end{tabular}
\end{minipage}

\vspace{2mm}
\raggedright
{\small
\textbf{(b) Comparison of original, rain-only, wet surfaces, and puddles.} The zoomed region shows how semantic priors ensure physically consistent surface effects, with puddles and wetness  constrained to the road and not crossing onto the sidewalk.}

\caption{
\textbf{Zoomed ablation comparisons across two scenes.}
Each block shows the original image, rain-only rendering, wet-surface reflections, and puddles, together with a magnified region highlighting differences in surface interaction and consistency.
}
\label{fig:combined_zoom_ablations}
\end{figure*}

Figure~\ref{fig:combined_zoom_ablations}-(a) provides a zoomed view of a single scene, highlighting how reflections are formed and how their appearance changes once puddles are added. The zoom reveals the correct reflection of scene structures (e.g., the tree trunk) and how puddle-based distortions modify this appearance in a physically consistent manner.

Figure~\ref{fig:combined_zoom_ablations}-(b) focuses on the role of semantic priors. The zoomed region shows that puddles remain constrained to road surfaces and do not spill onto the sidewalk or other non-ground regions. This demonstrates how depth and semantic segmentation jointly guide the rendering process, ensuring physically plausible placement of wetness and puddle effects. Across all figures, the progression from \emph{Rain streaks} to \emph{Reflections} to \emph{Puddles} consistently increases visual realism and coherence in the synthesized rainy scenes.


\begin{figure*}[!t]
\centering
\renewcommand{\arraystretch}{1.2}
\setlength{\tabcolsep}{2pt}
\caption{
\textbf{Qualitative comparison of rain synthesis methods.}
Each column shows a different rendering method, while each row presents a distinct Cityscapes scene. 
PBR~\cite{halder2019physicsbased} and Cityscapes-Rain~\cite{DepthRainRemoval} provide streak overlays but no surface interactions. 
TPSeNCE~\cite{zheng2023} introduces wetness effects but lacks clear streaks and fog. 
CascadeRS produces depth-aligned streaks, realistic reflections, puddles, and fog.
}
\vspace{3pt}
\scriptsize

\begin{tabular}{
C{0.19\linewidth}
C{0.19\linewidth}
C{0.19\linewidth}
C{0.19\linewidth}
C{0.19\linewidth}
}
\toprule
\textbf{Original~\cite{Cityscapes}} &
\textbf{PBR~\cite{halder2019physicsbased}} &
\textbf{CS-Rain~\cite{DepthRainRemoval}} &
\textbf{TPSeNCE~\cite{zheng2023}} &
\textbf{CascadeRS (ours)} \\
\midrule

\includegraphics[width=\linewidth]{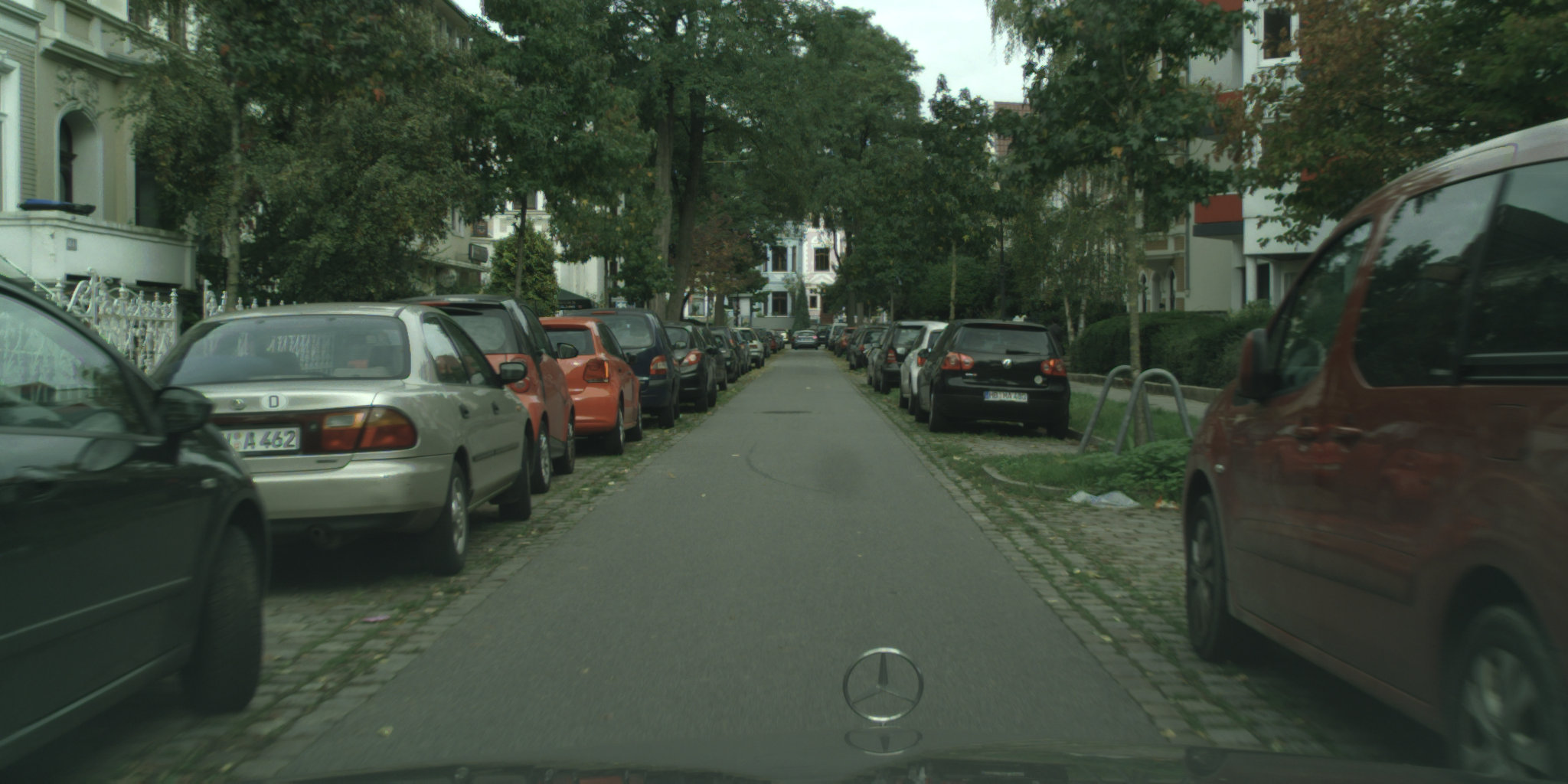} &
\includegraphics[width=\linewidth]{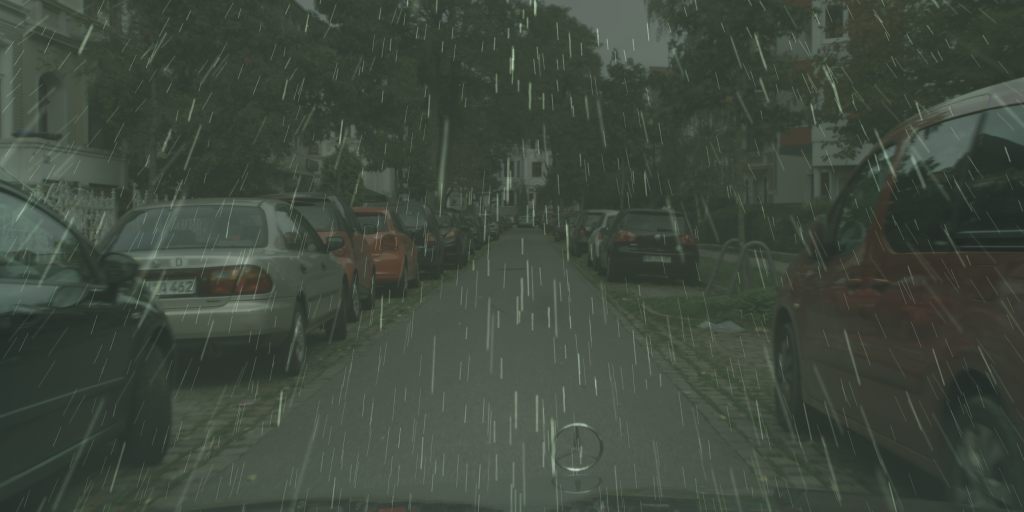} &
\includegraphics[width=\linewidth]{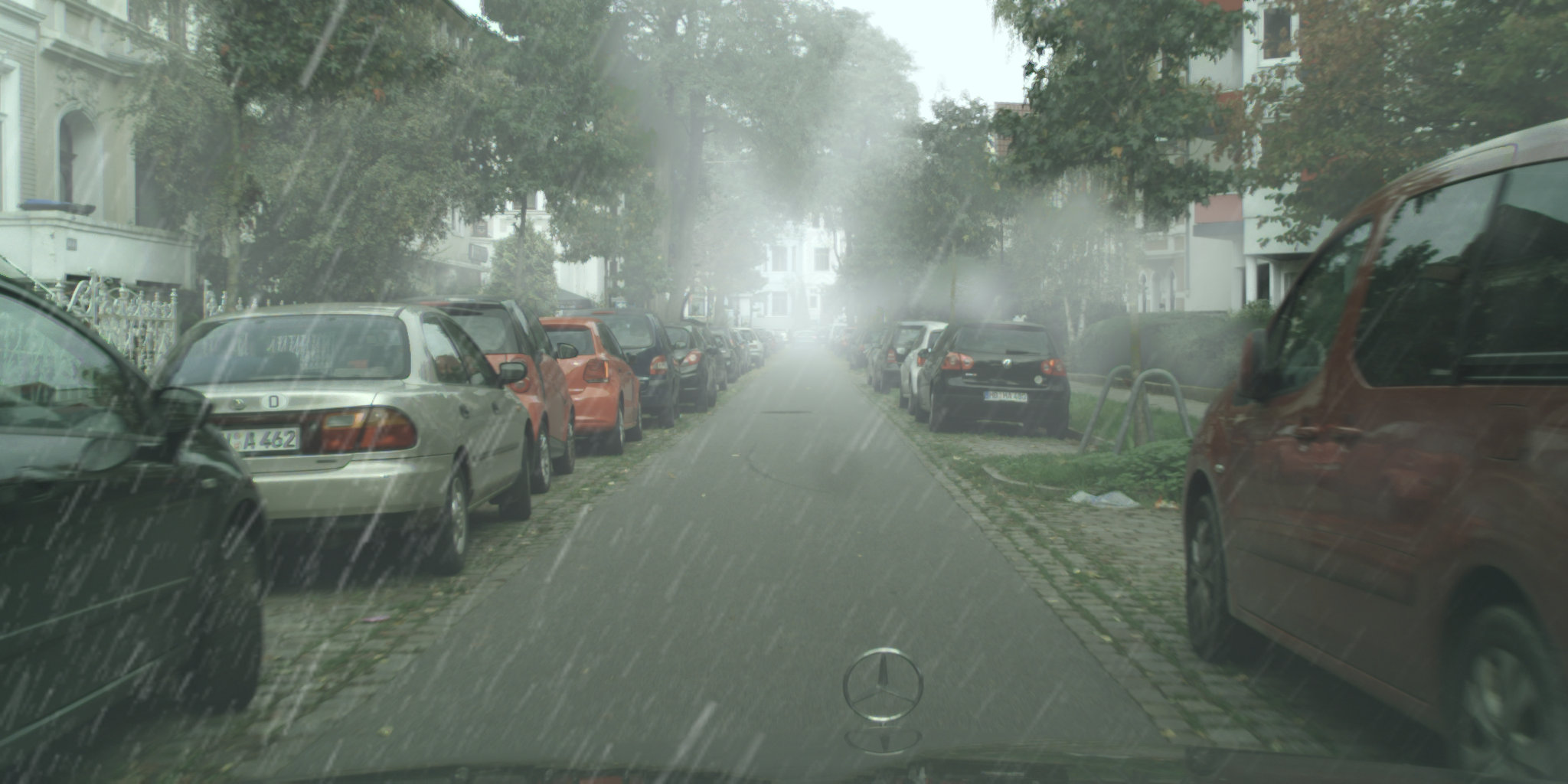} &
\includegraphics[width=\linewidth]{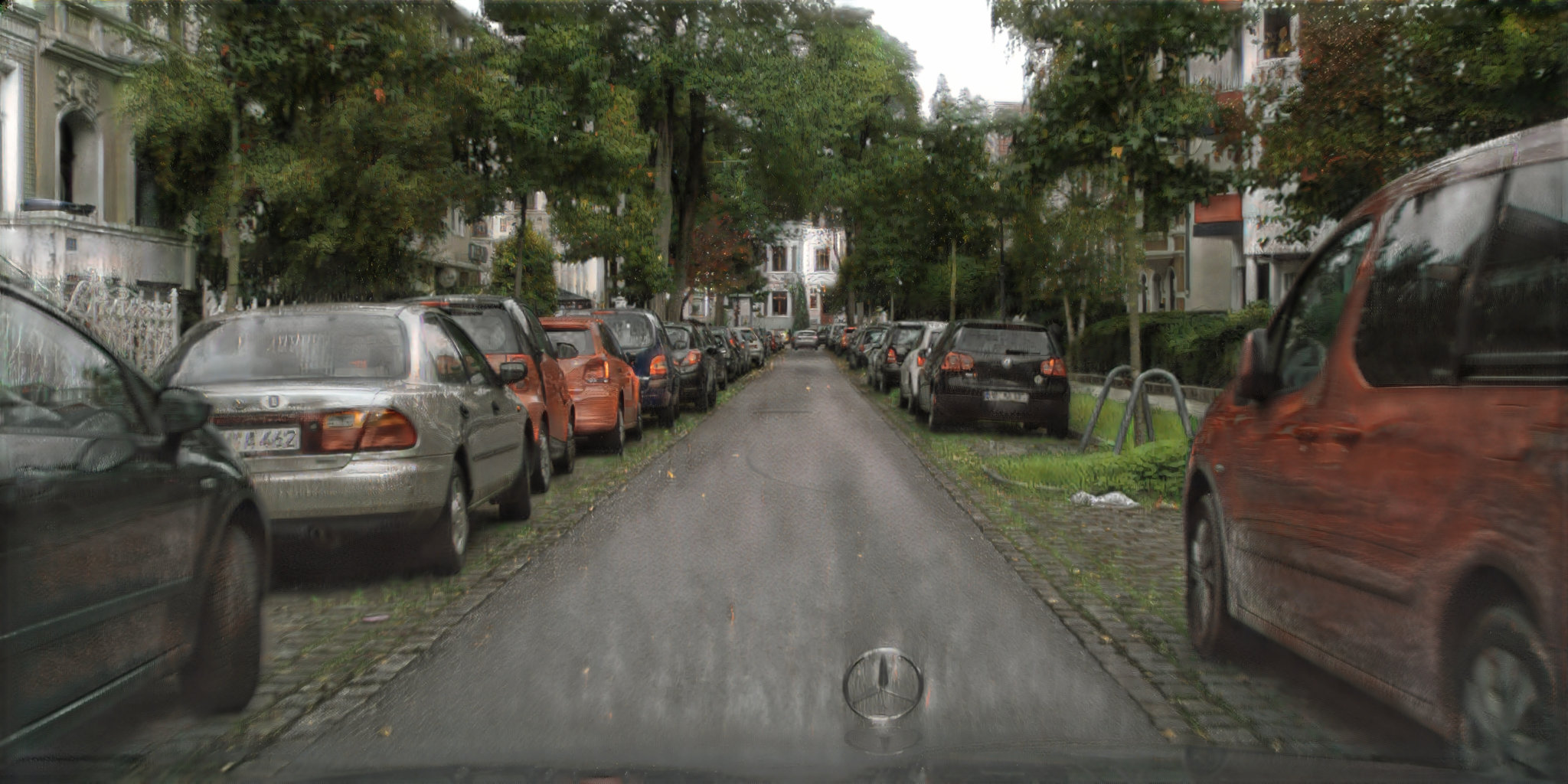} &
\includegraphics[width=\linewidth]{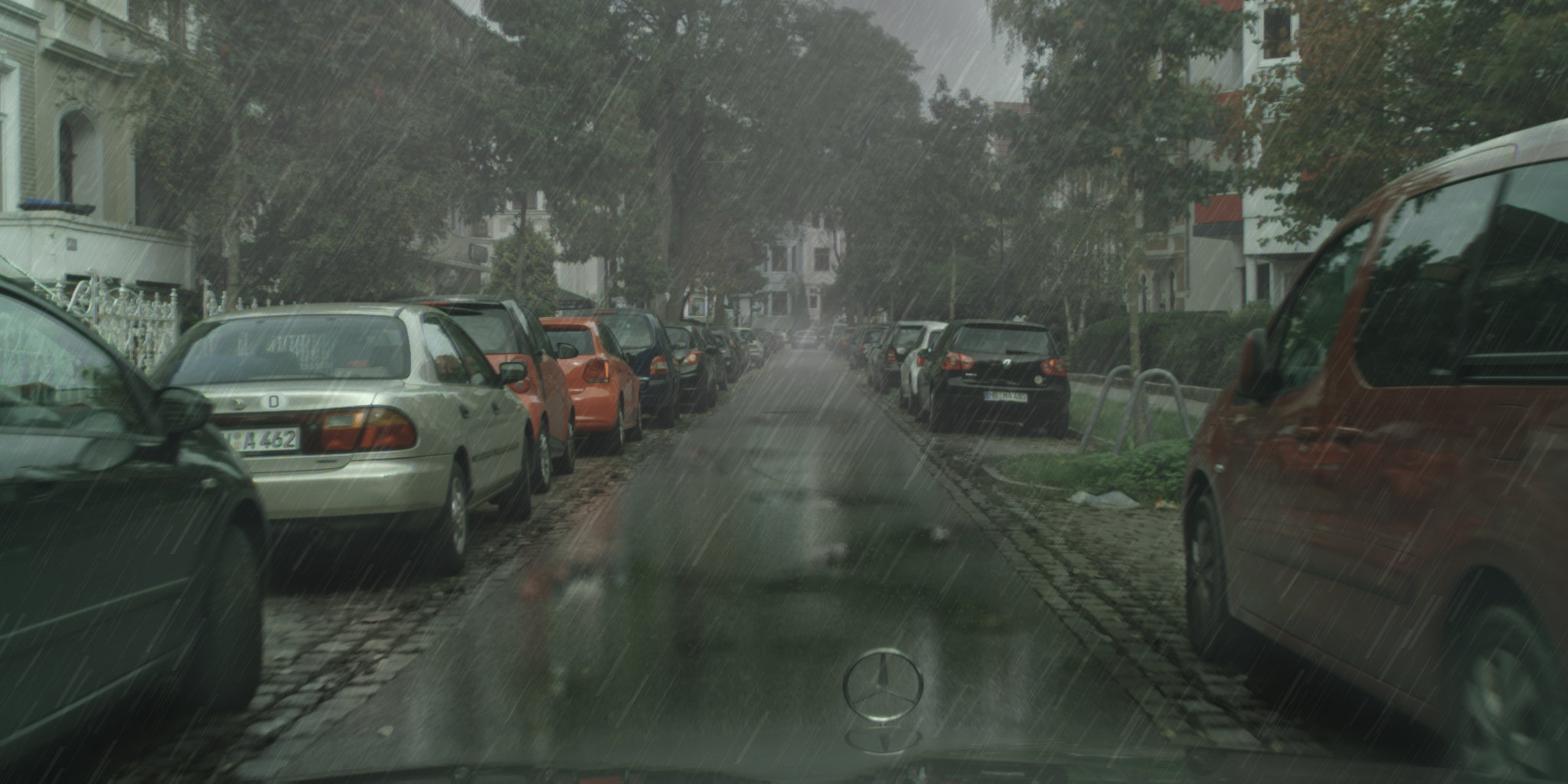} \\[4mm]

\includegraphics[width=\linewidth]{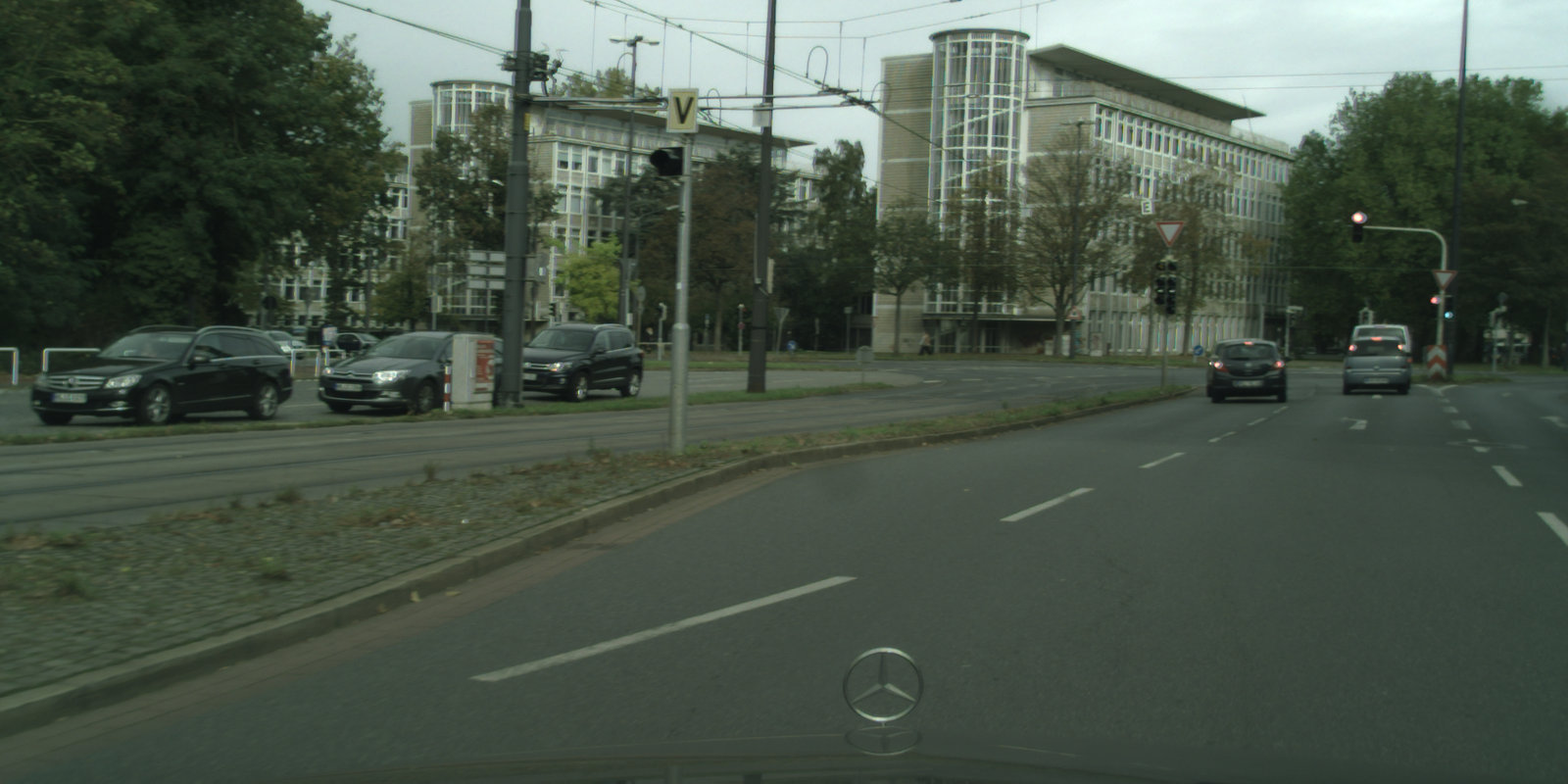} &
\includegraphics[width=\linewidth]{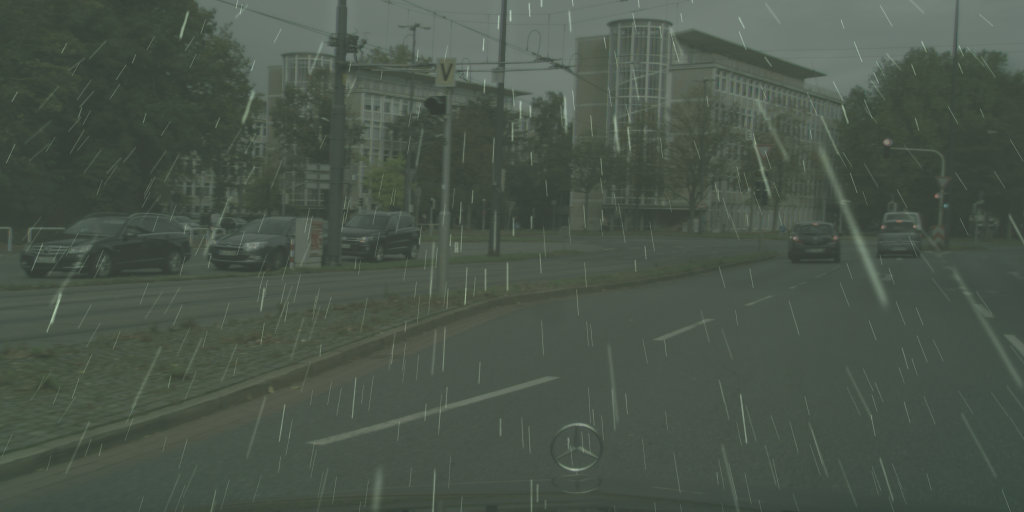} &
\includegraphics[width=\linewidth]{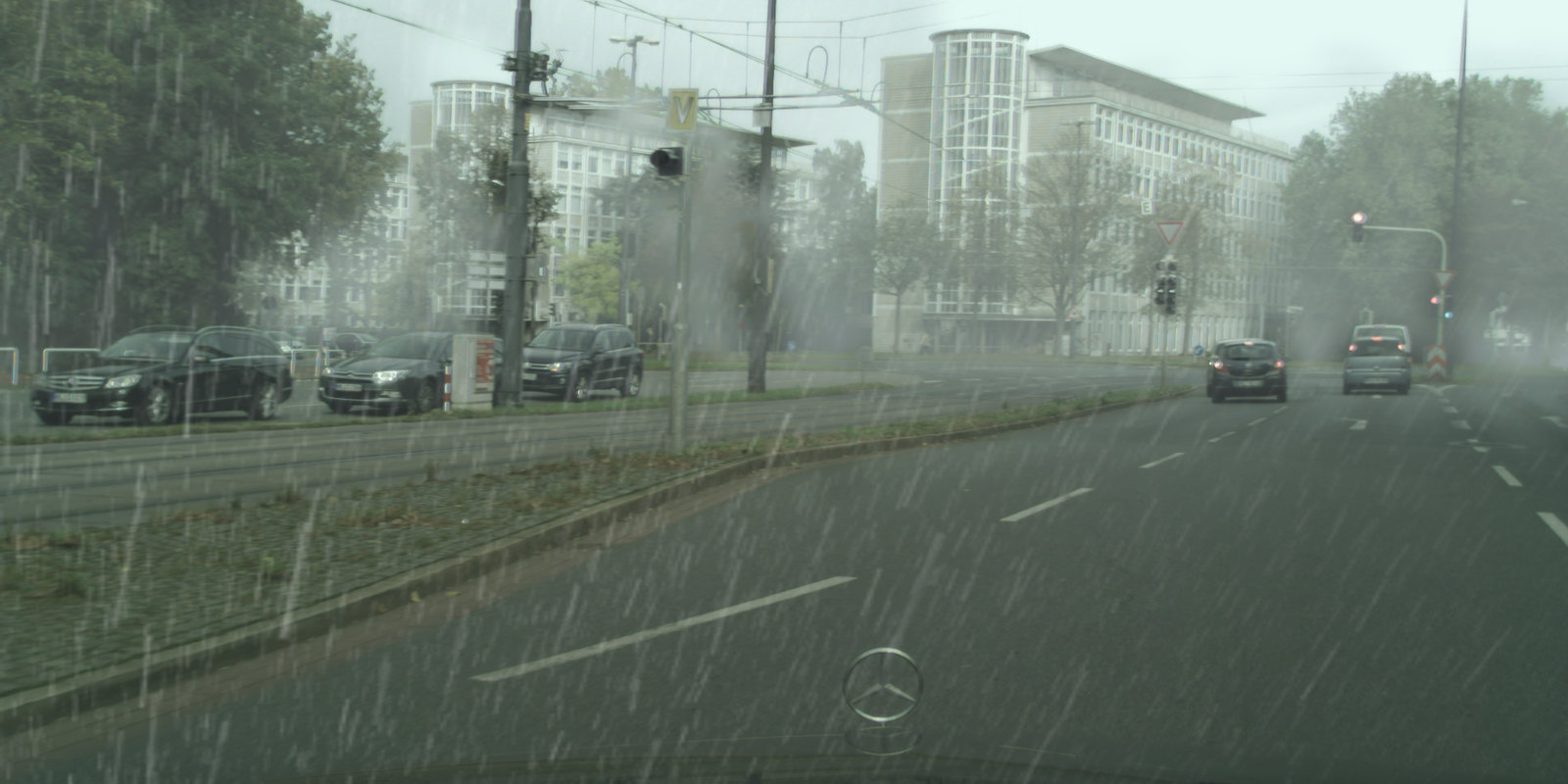} &
\includegraphics[width=\linewidth]{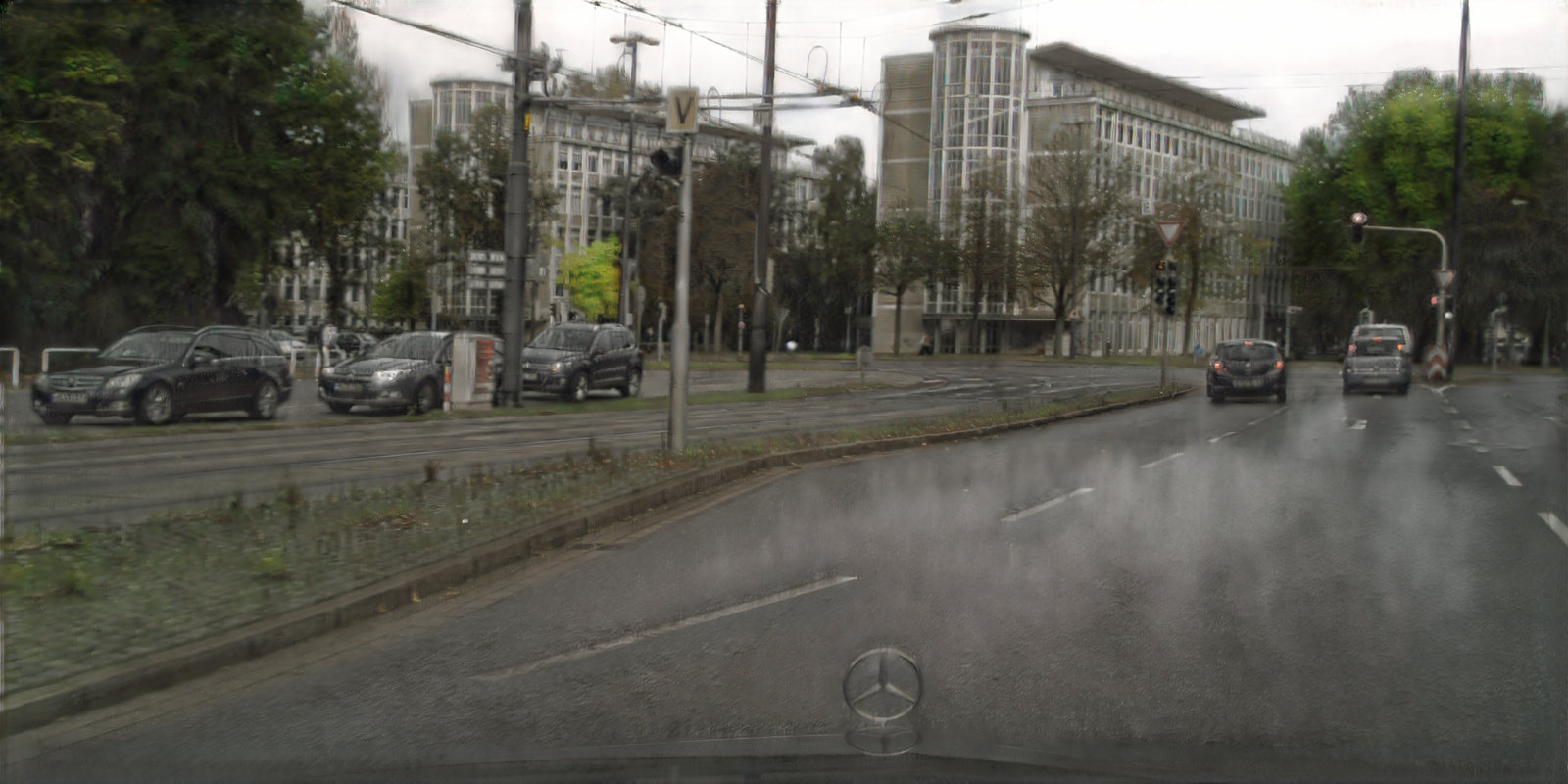} &
\includegraphics[width=\linewidth]{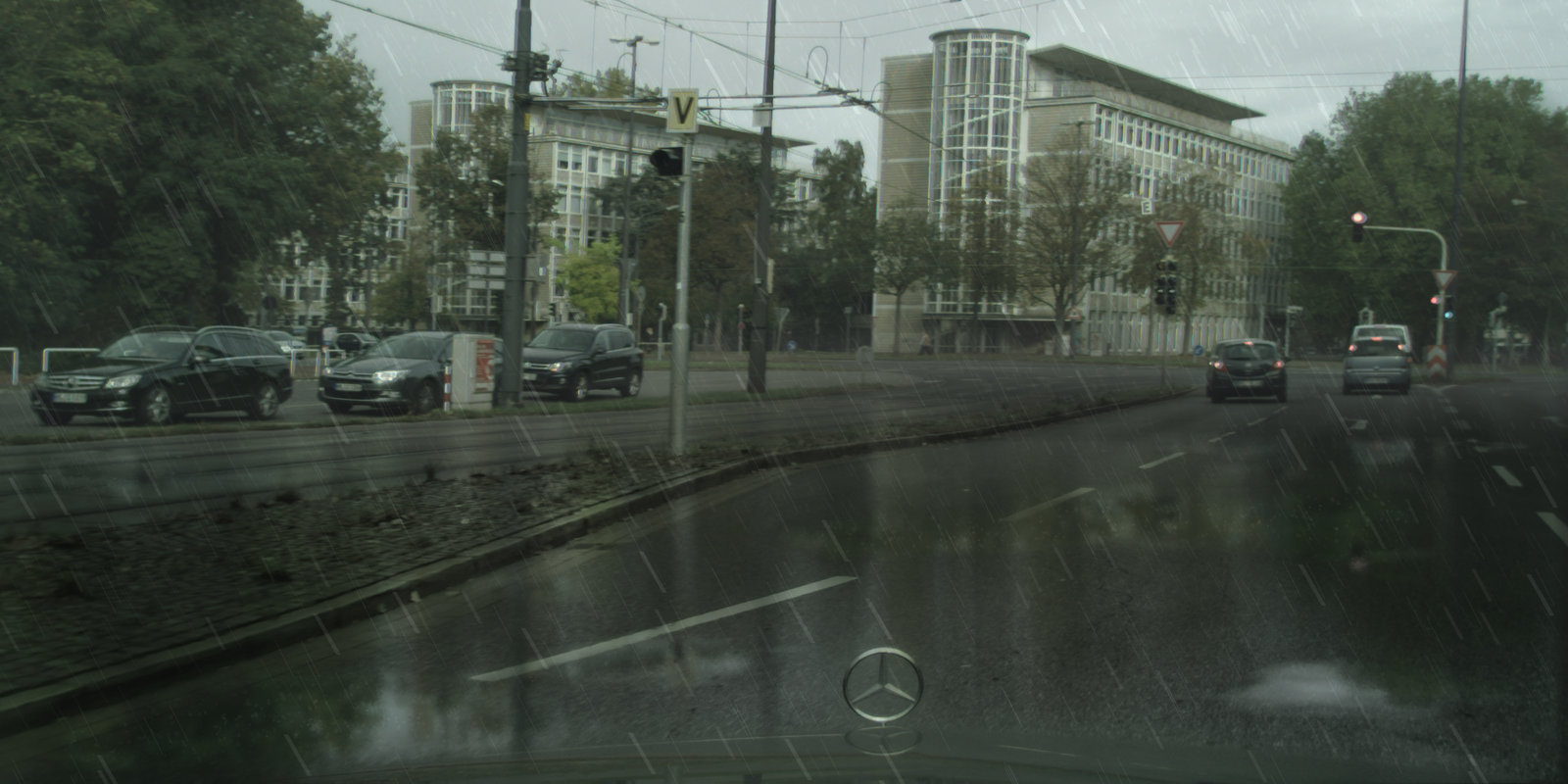} \\[4mm]

\includegraphics[width=\linewidth]{figures/PART2/5/Original.jpg} &
\includegraphics[width=\linewidth]{figures/PART2/5/PBR.jpg} &
\includegraphics[width=\linewidth]{figures/PART2/5/CityscapesRain.jpg} &
\includegraphics[width=\linewidth]{figures/PART2/5/TPSeNCE.jpg} &
\includegraphics[width=\linewidth]{figures/PART2/5/CascadeRS.jpg} \\[4mm]

\includegraphics[width=\linewidth]{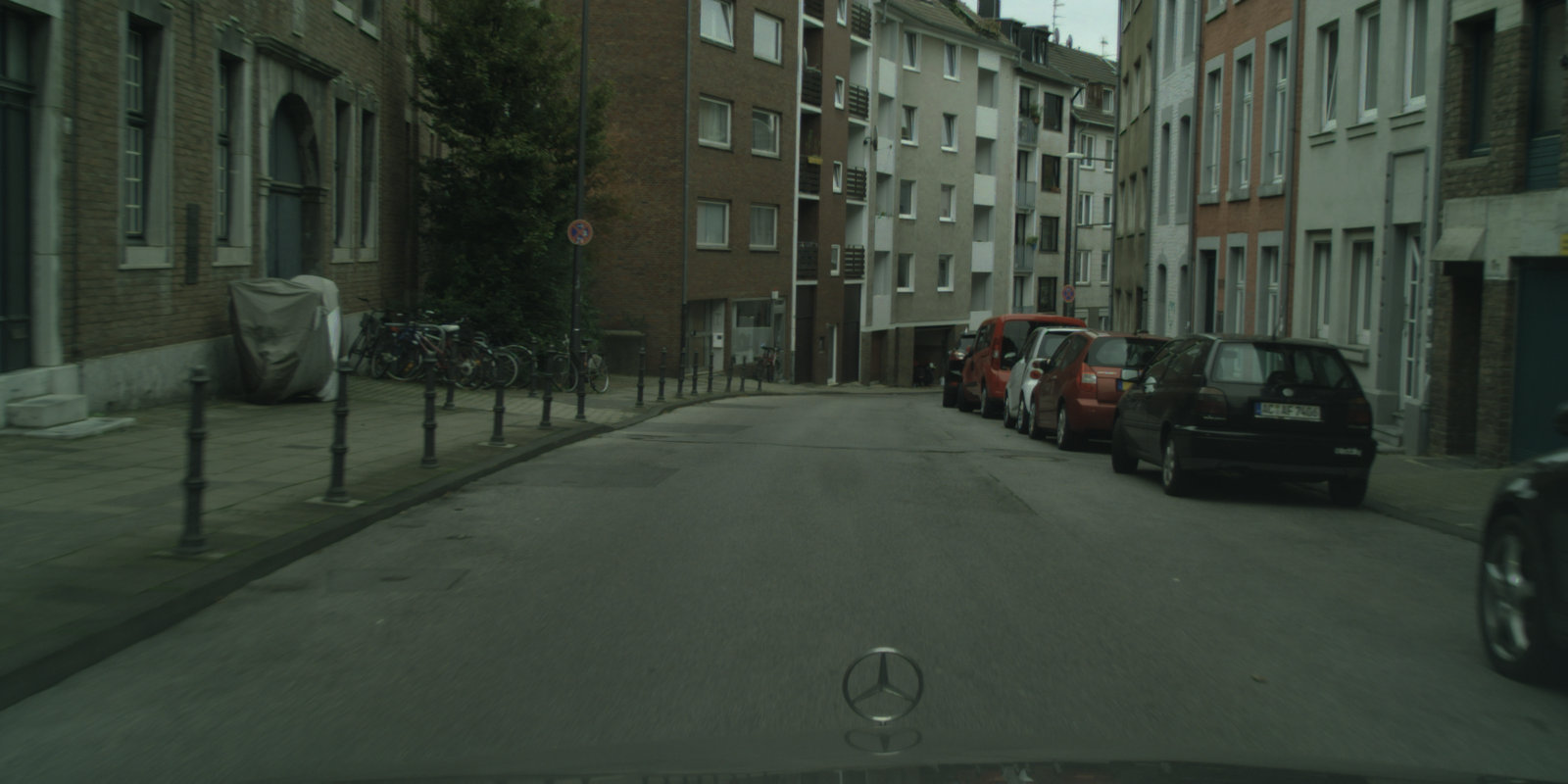} &
\includegraphics[width=\linewidth]{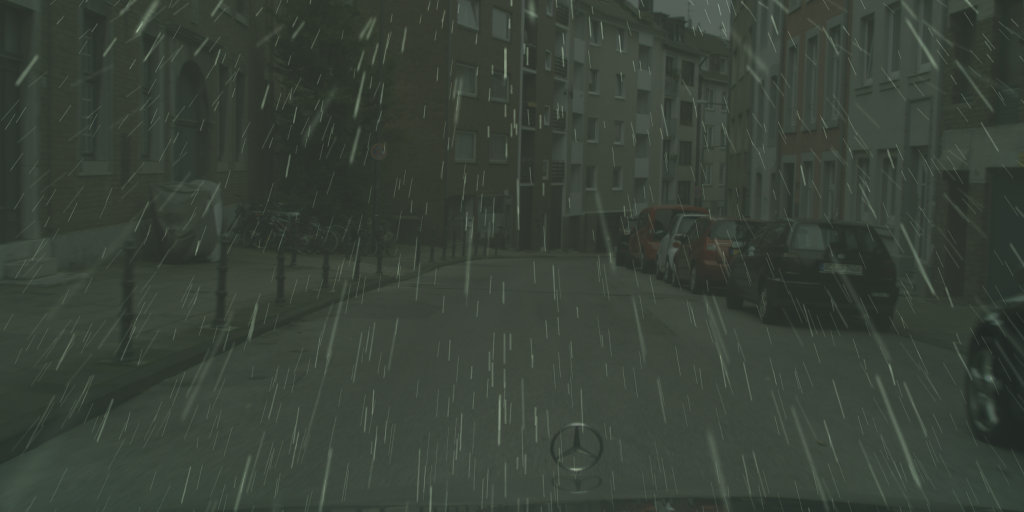} &
\includegraphics[width=\linewidth]{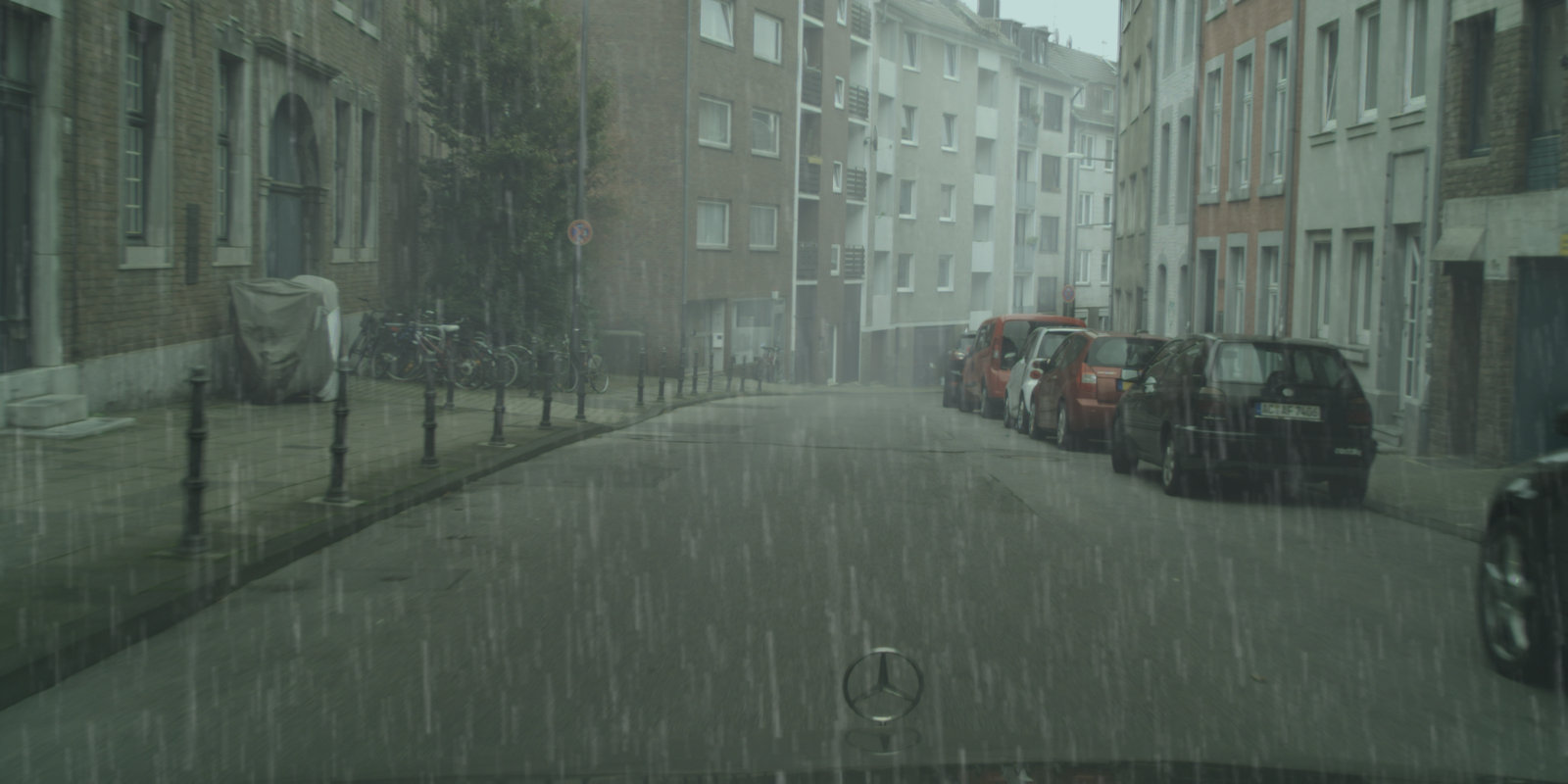} &
\includegraphics[width=\linewidth]{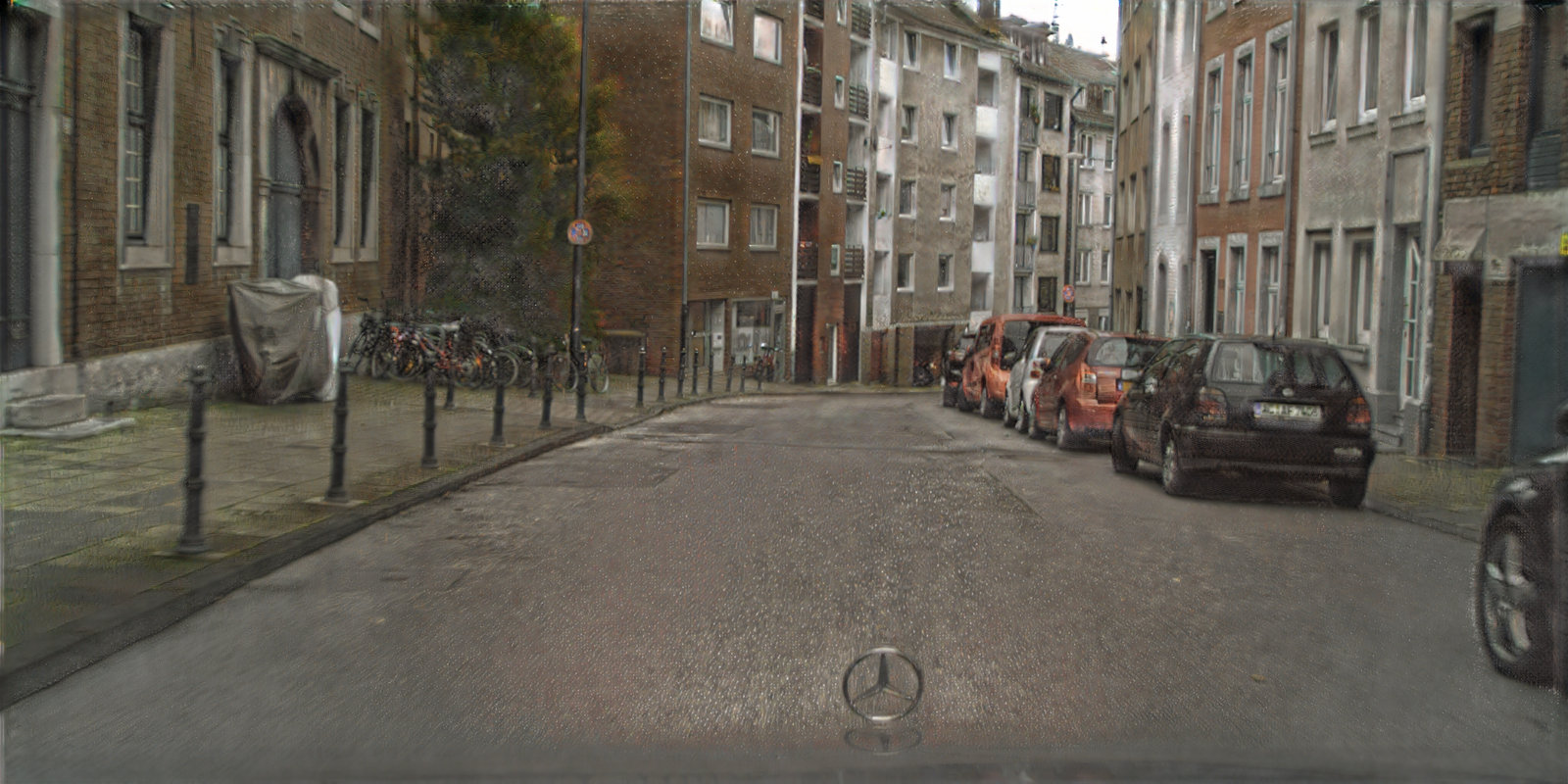} &
\includegraphics[width=\linewidth]{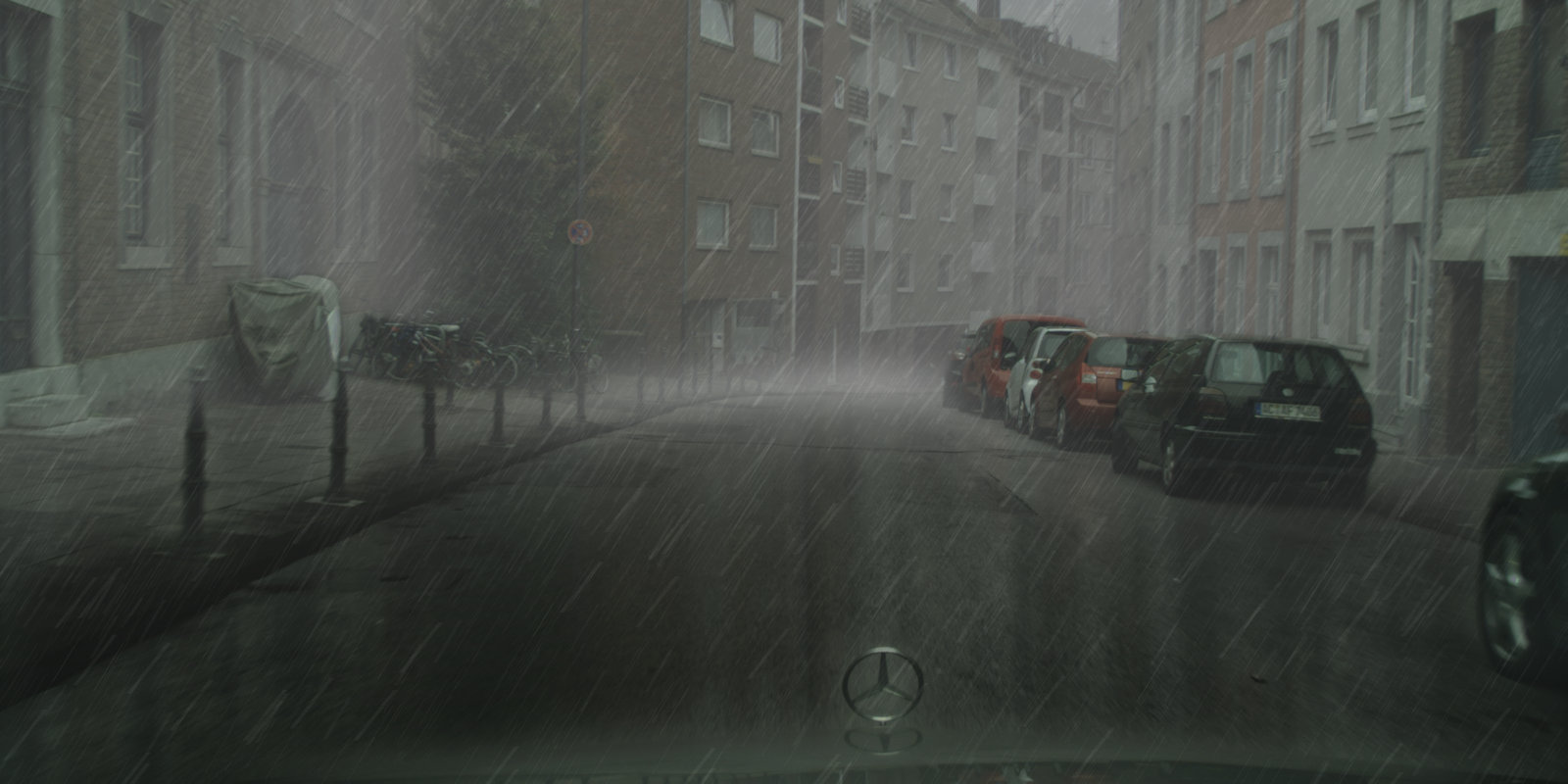} \\[4mm]

\includegraphics[width=\linewidth]{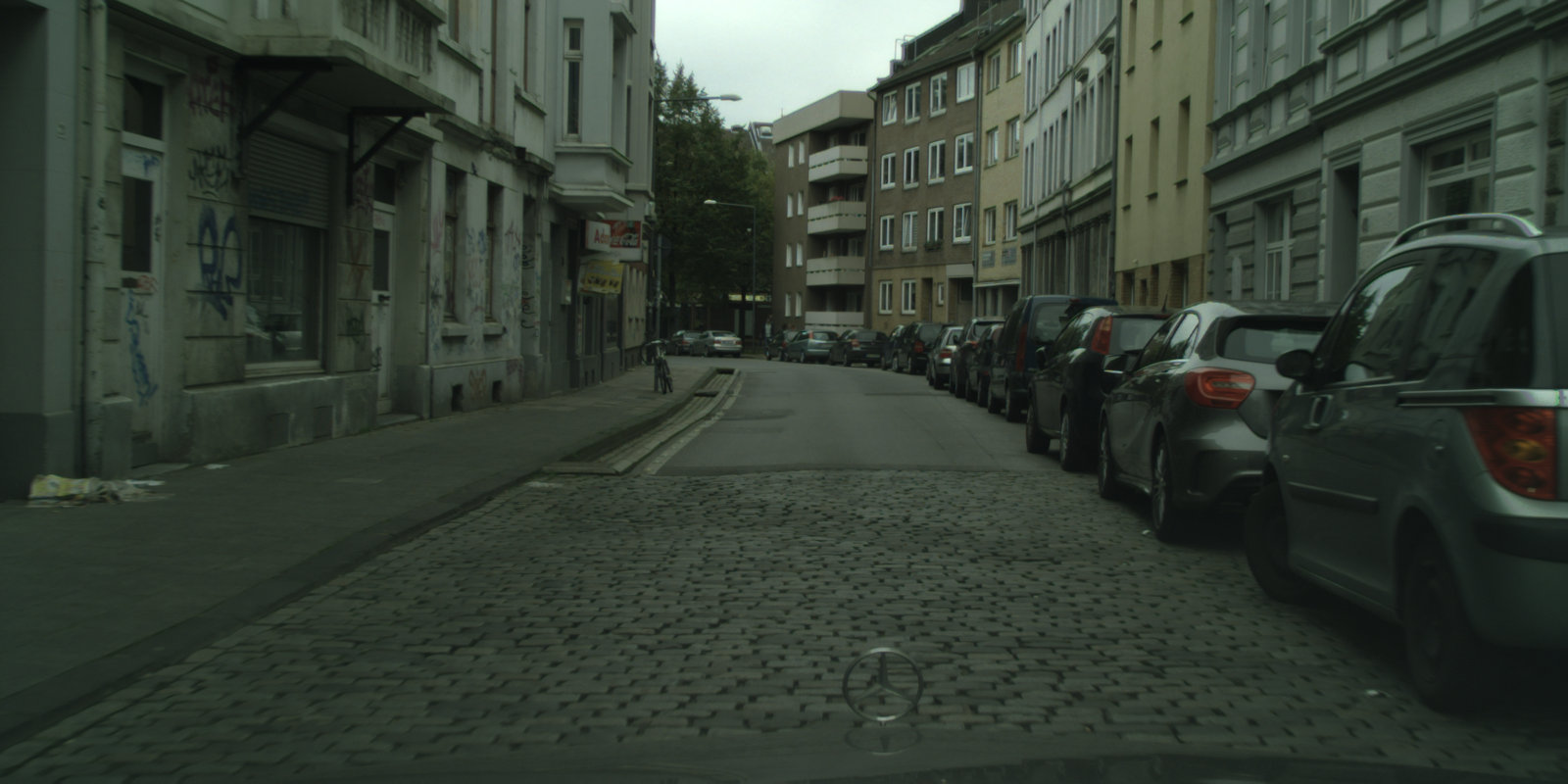} &
\includegraphics[width=\linewidth]{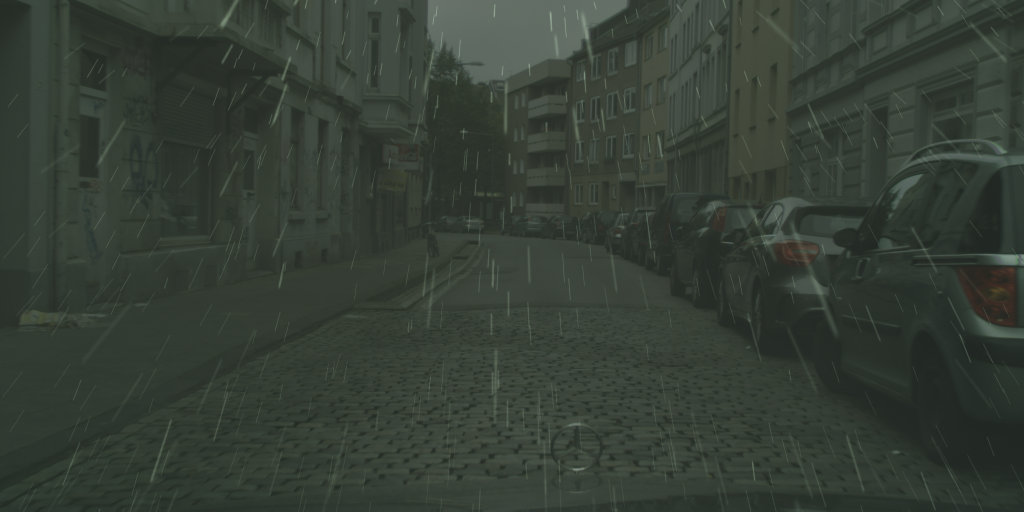} &
\includegraphics[width=\linewidth]{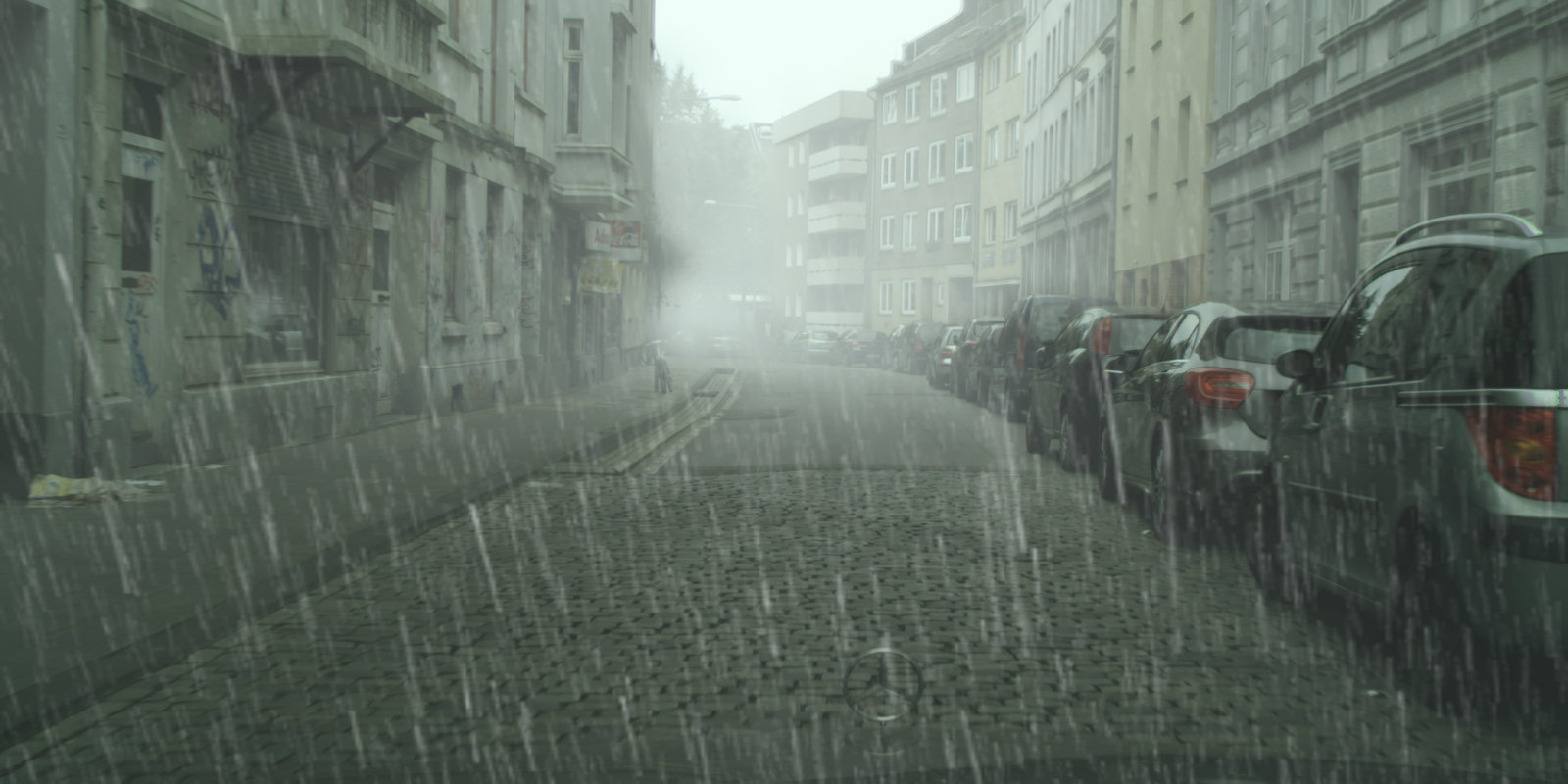} &
\includegraphics[width=\linewidth]{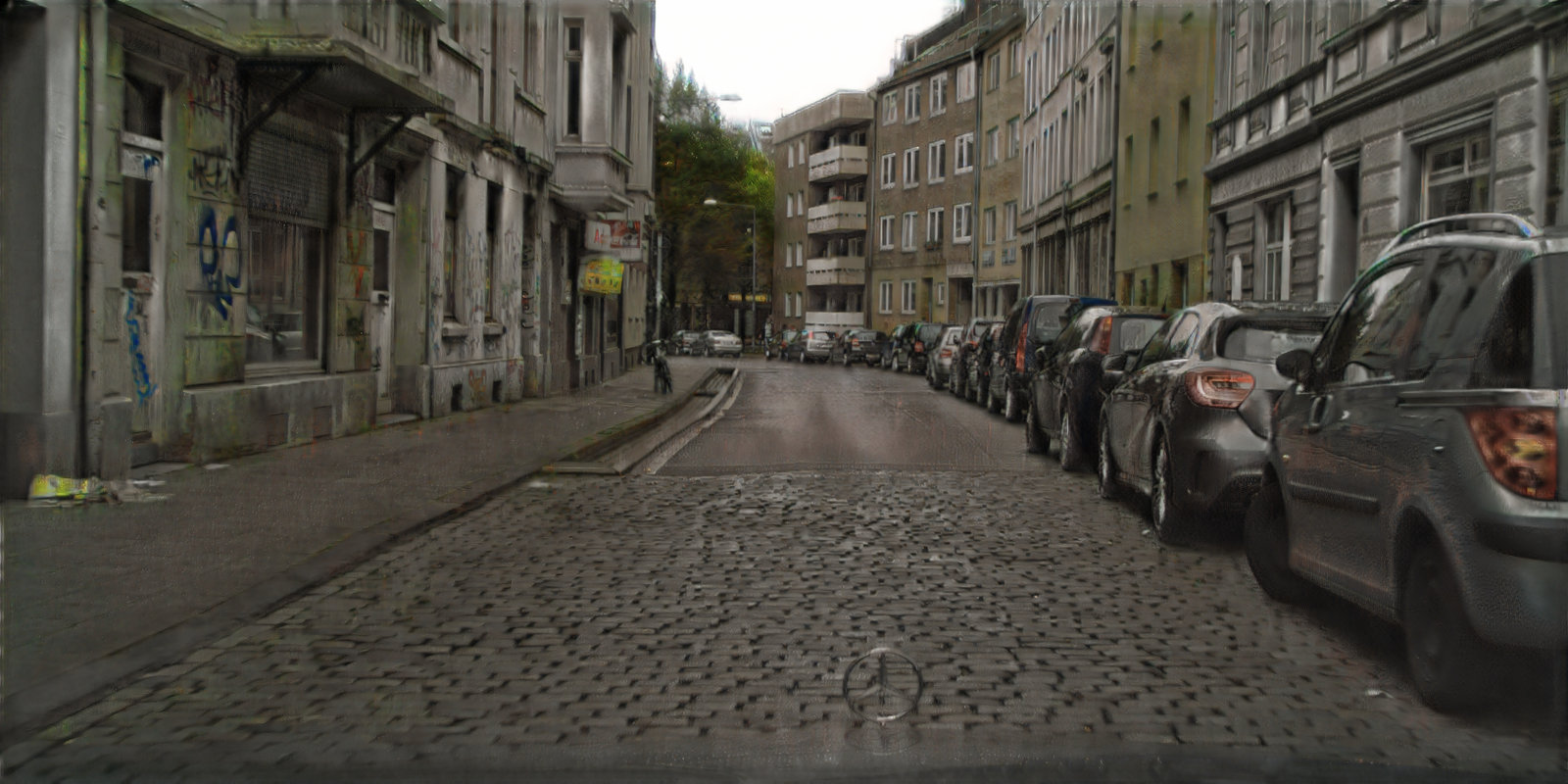} &
\includegraphics[width=\linewidth]{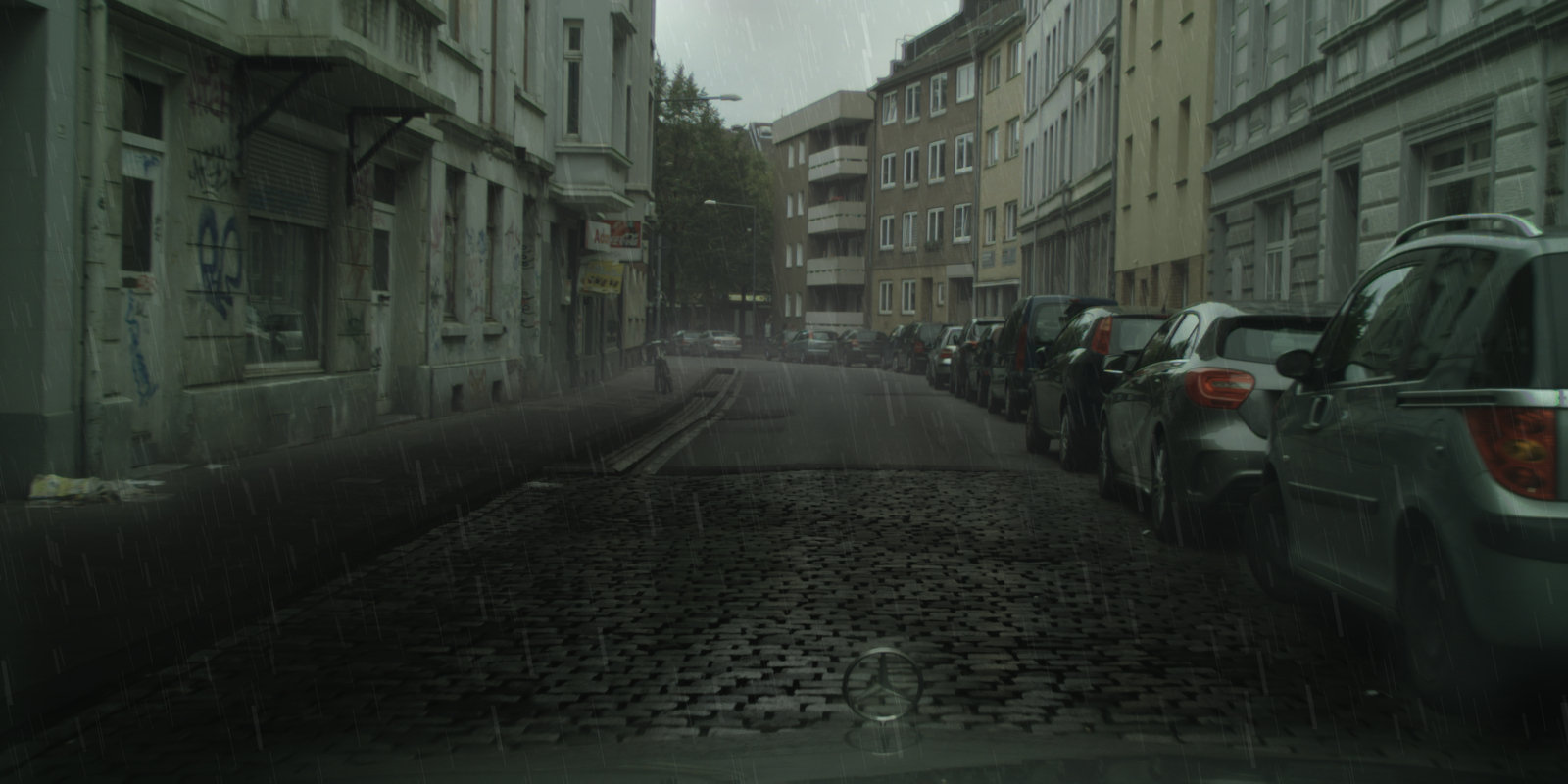} \\[4mm]

\includegraphics[width=\linewidth]{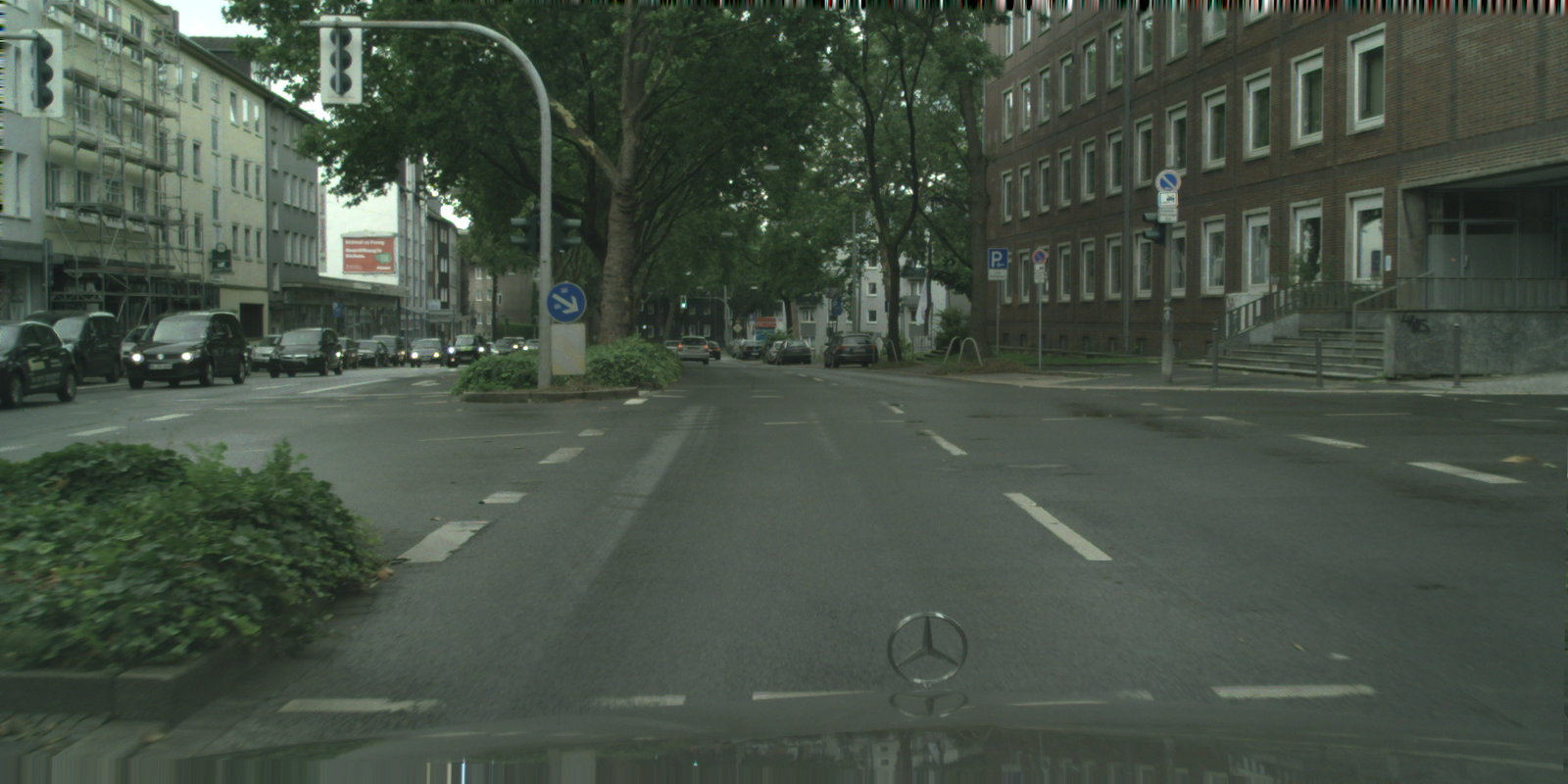} &
\includegraphics[width=\linewidth]{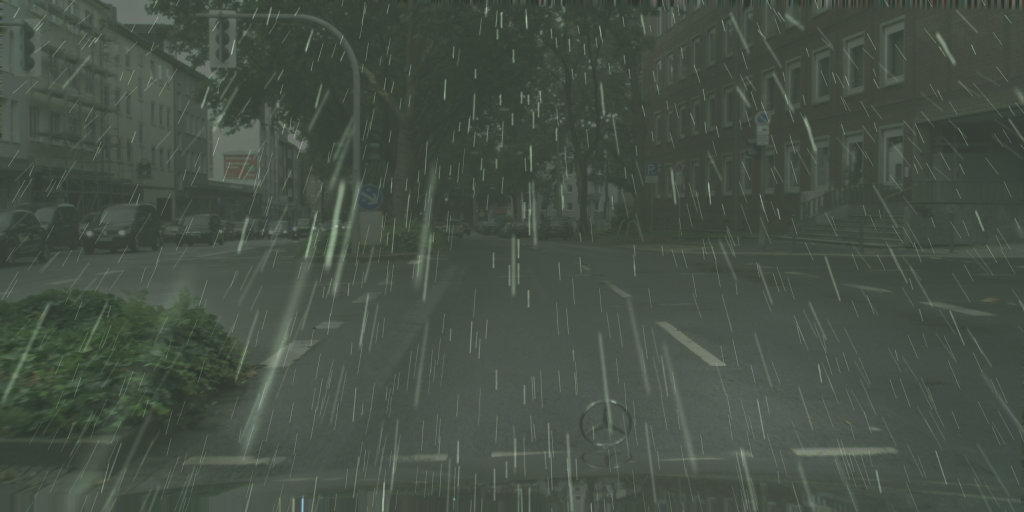} &
\includegraphics[width=\linewidth]{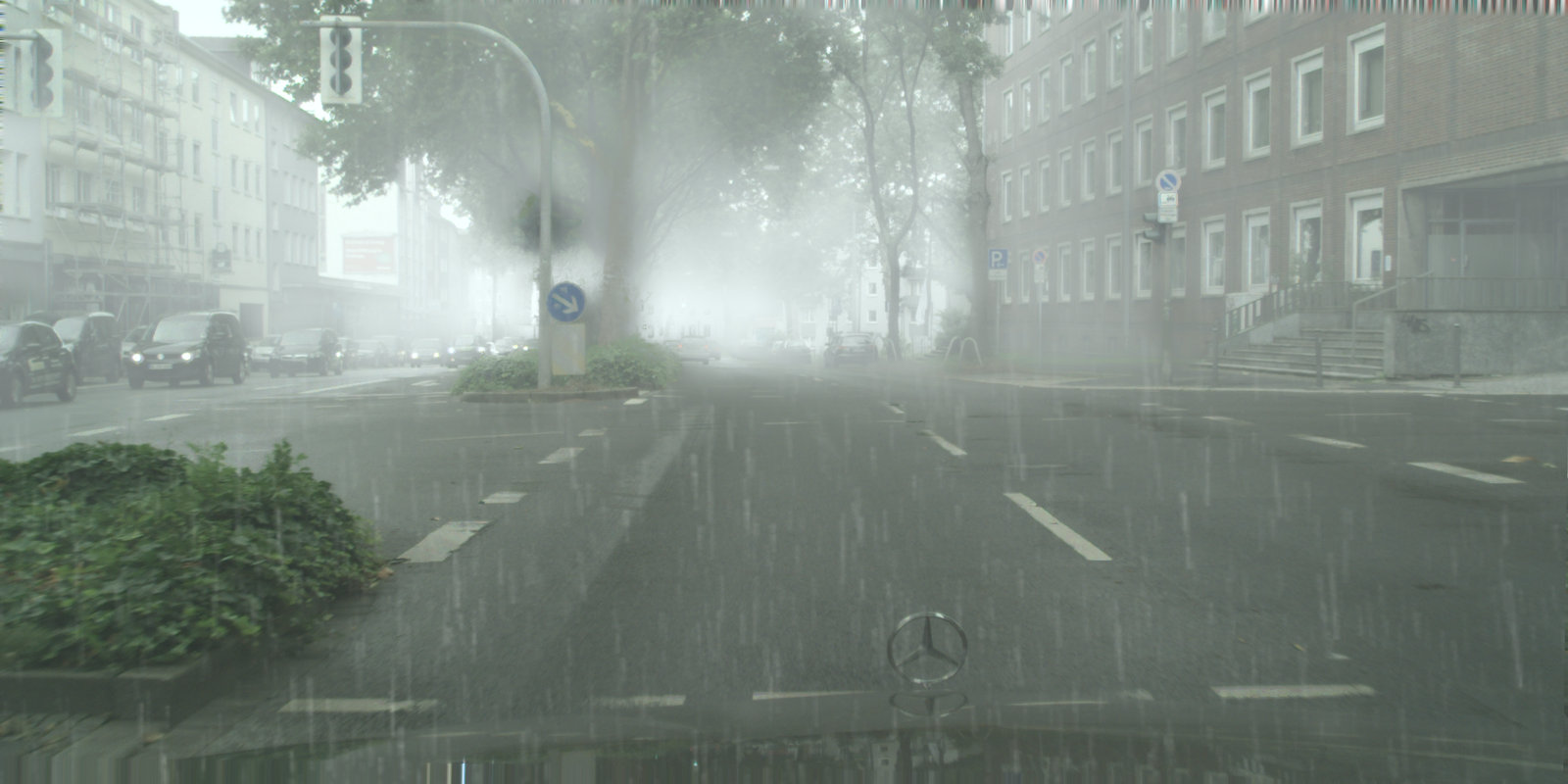} &
\includegraphics[width=\linewidth]{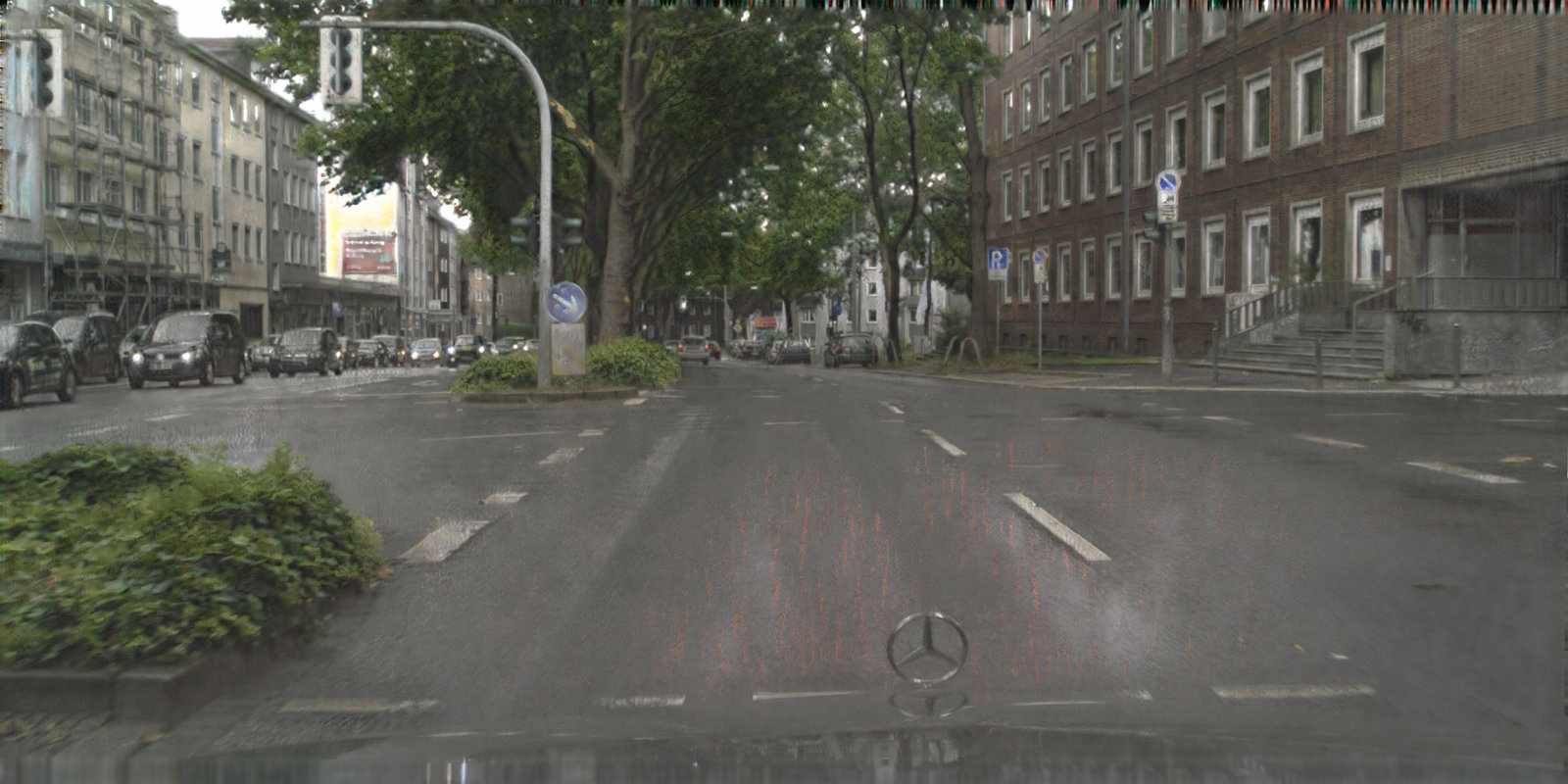} &
\includegraphics[width=\linewidth]{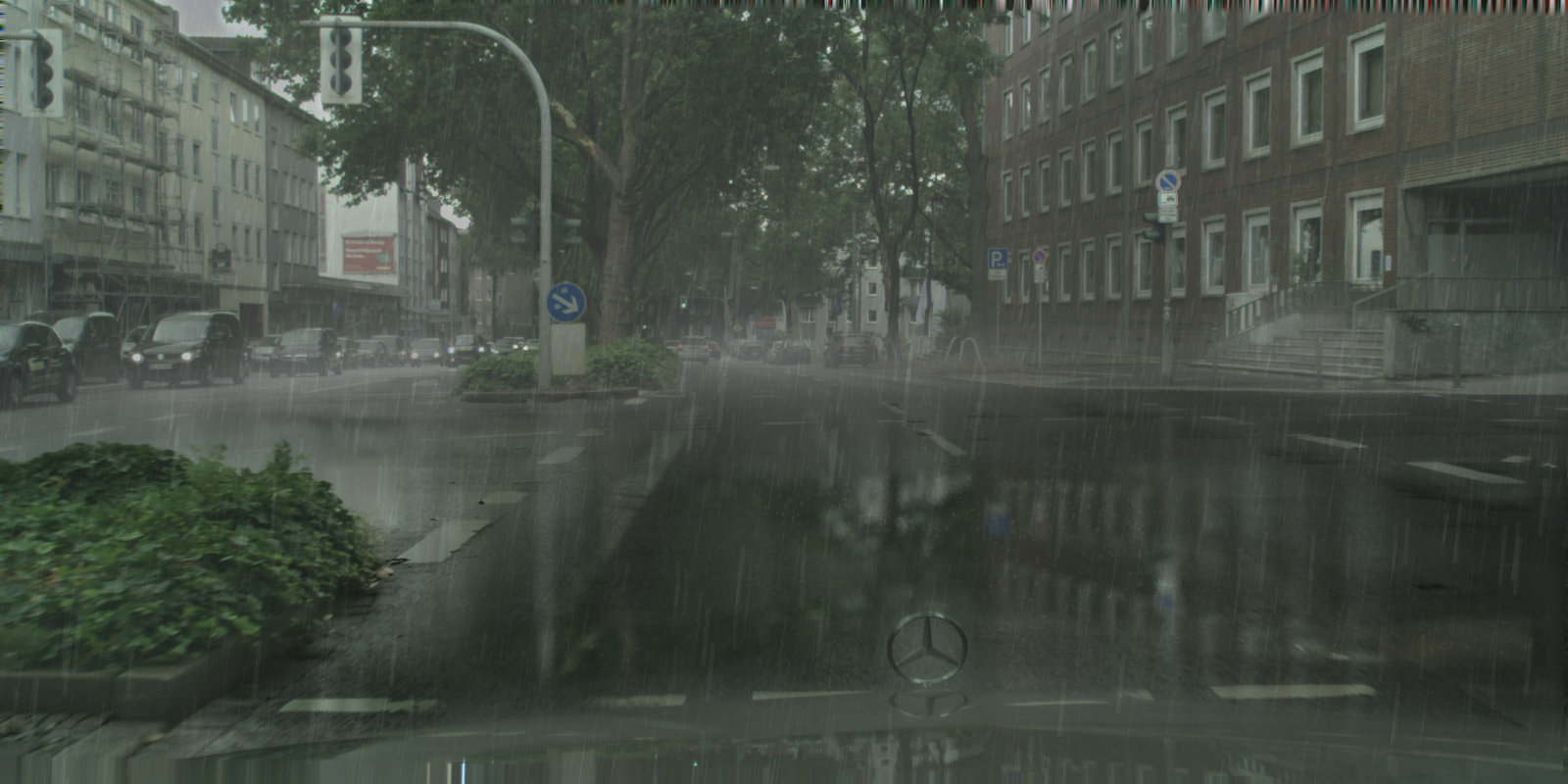} \\[4mm]

\includegraphics[width=\linewidth]{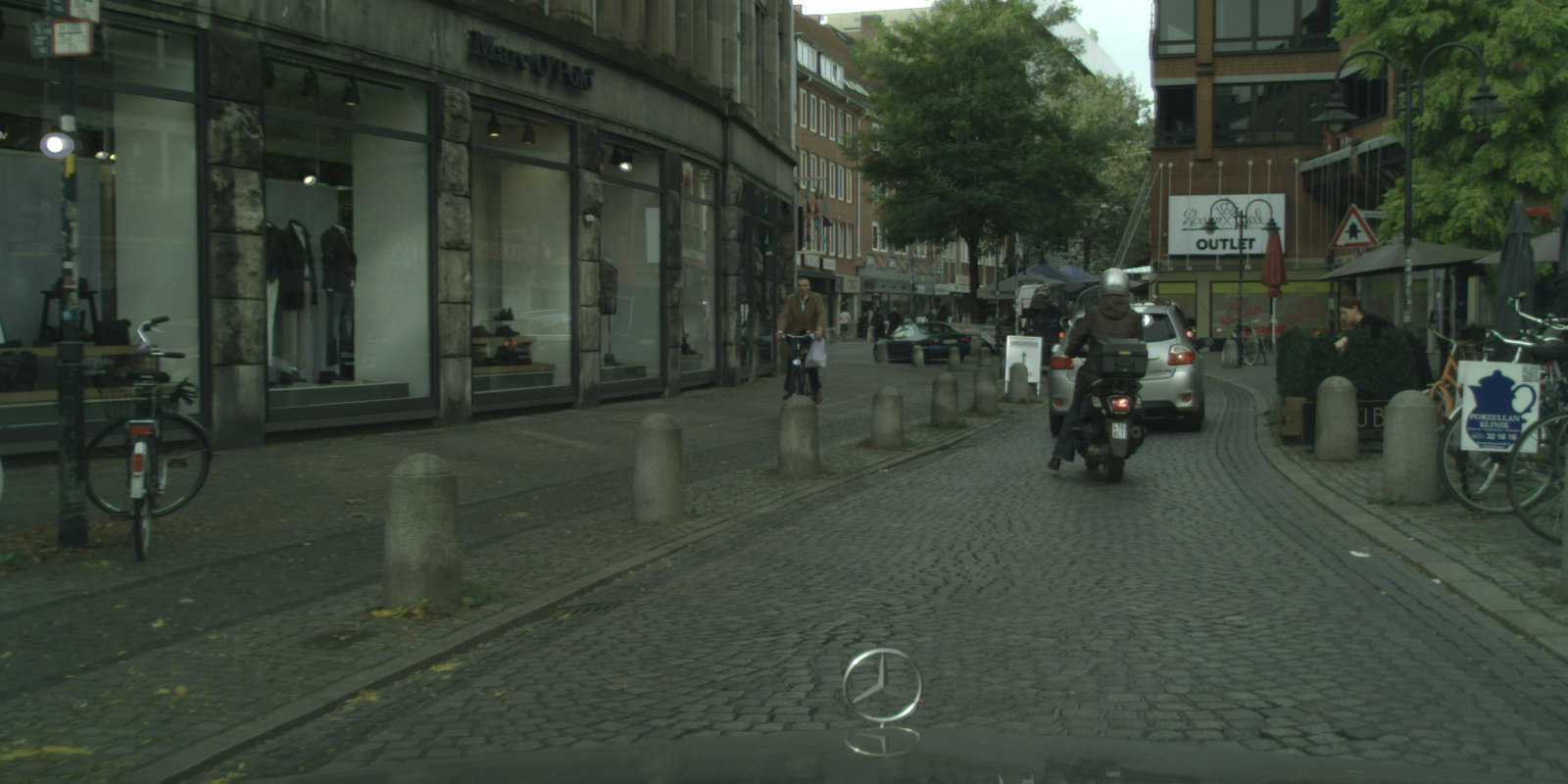} &
\includegraphics[width=\linewidth]{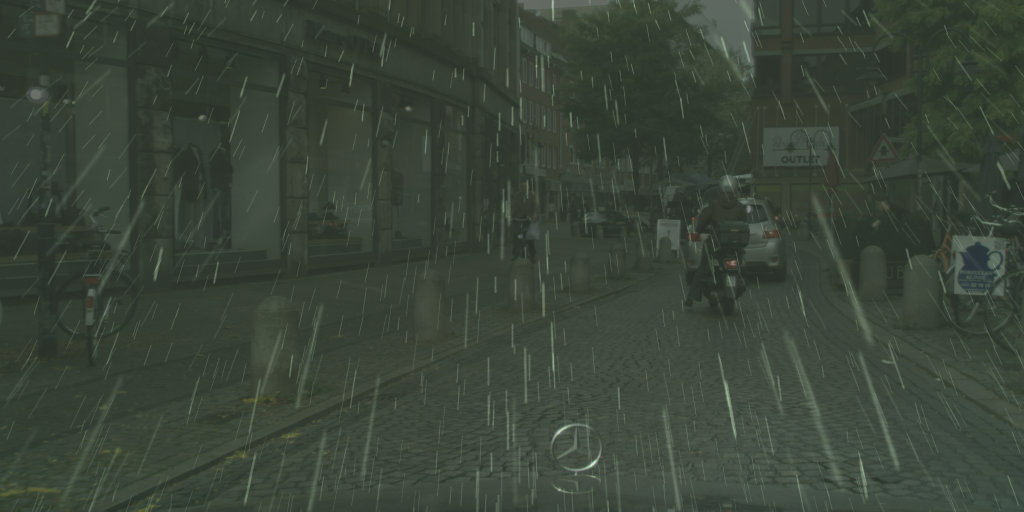} &
\includegraphics[width=\linewidth]{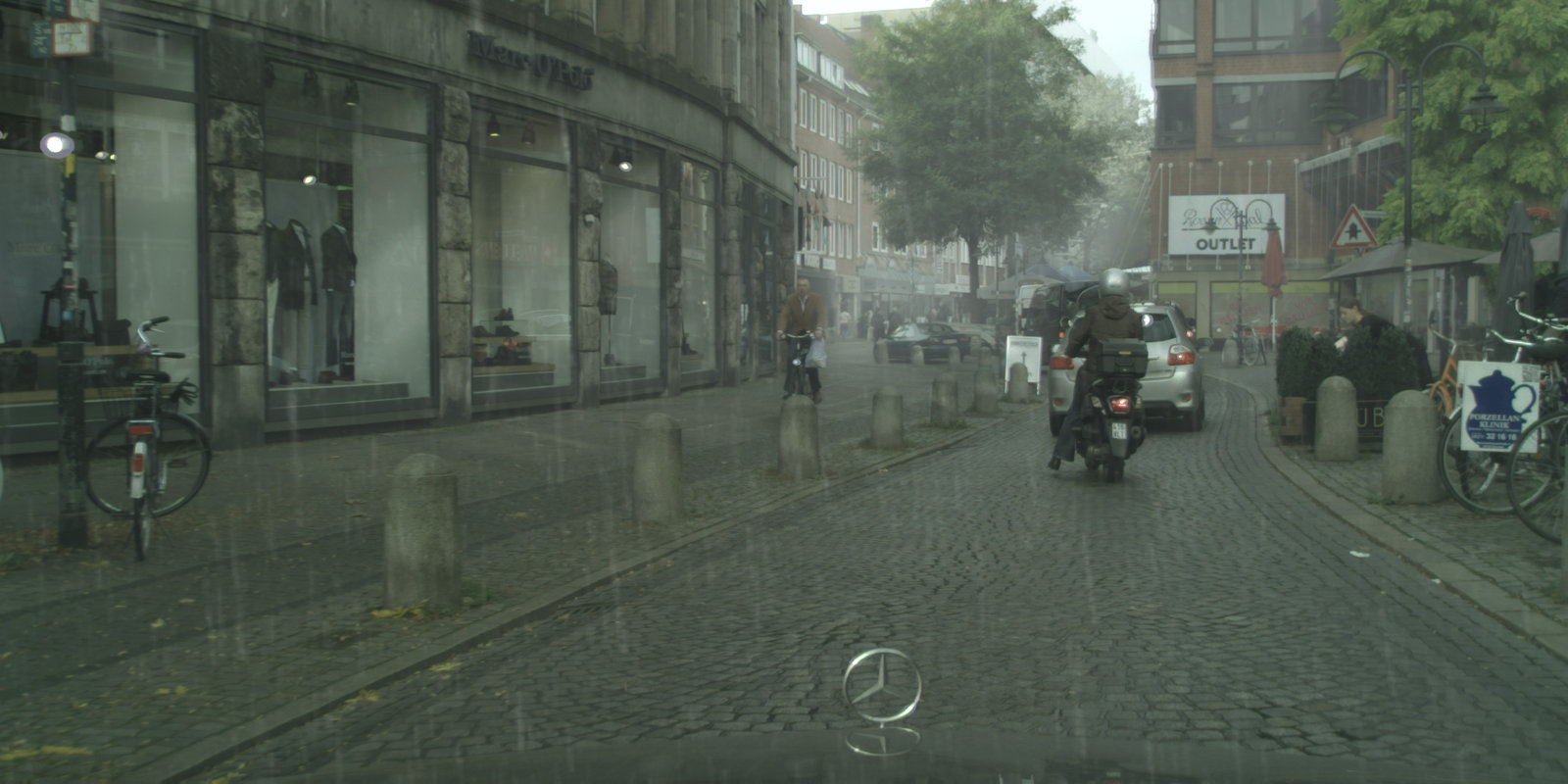} &
\includegraphics[width=\linewidth]{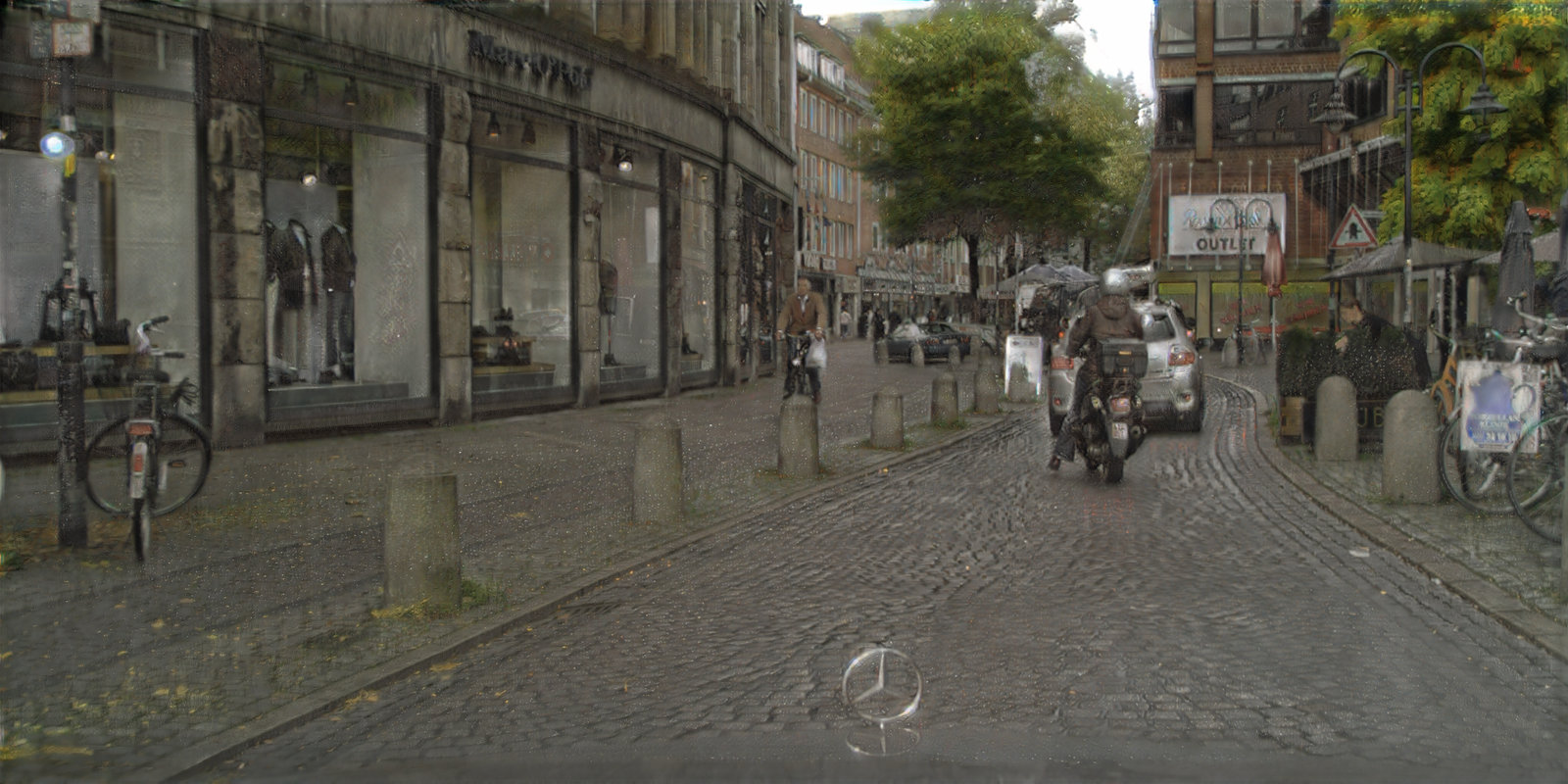} &
\includegraphics[width=\linewidth]{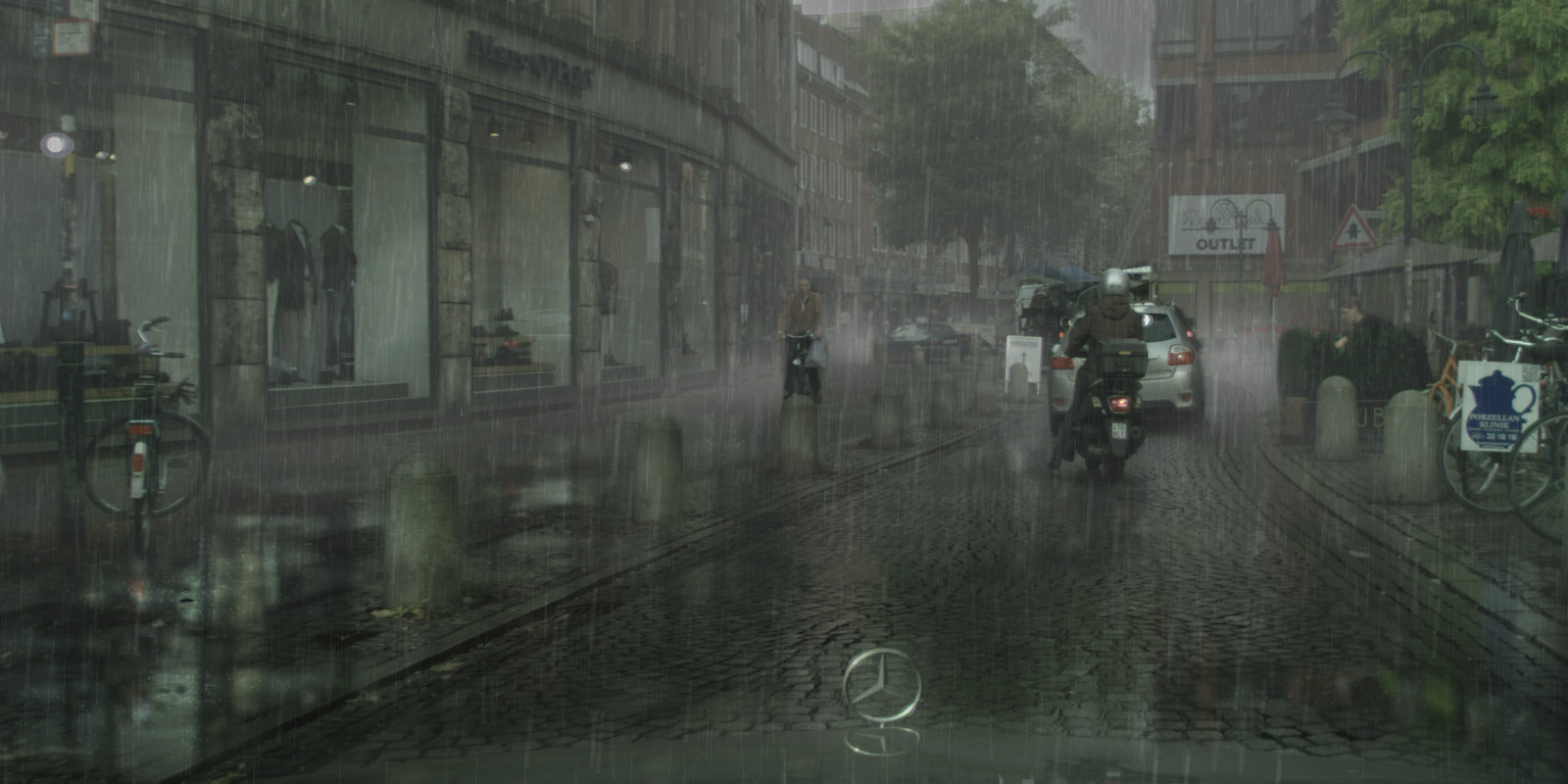} \\[4mm]

\includegraphics[width=\linewidth]{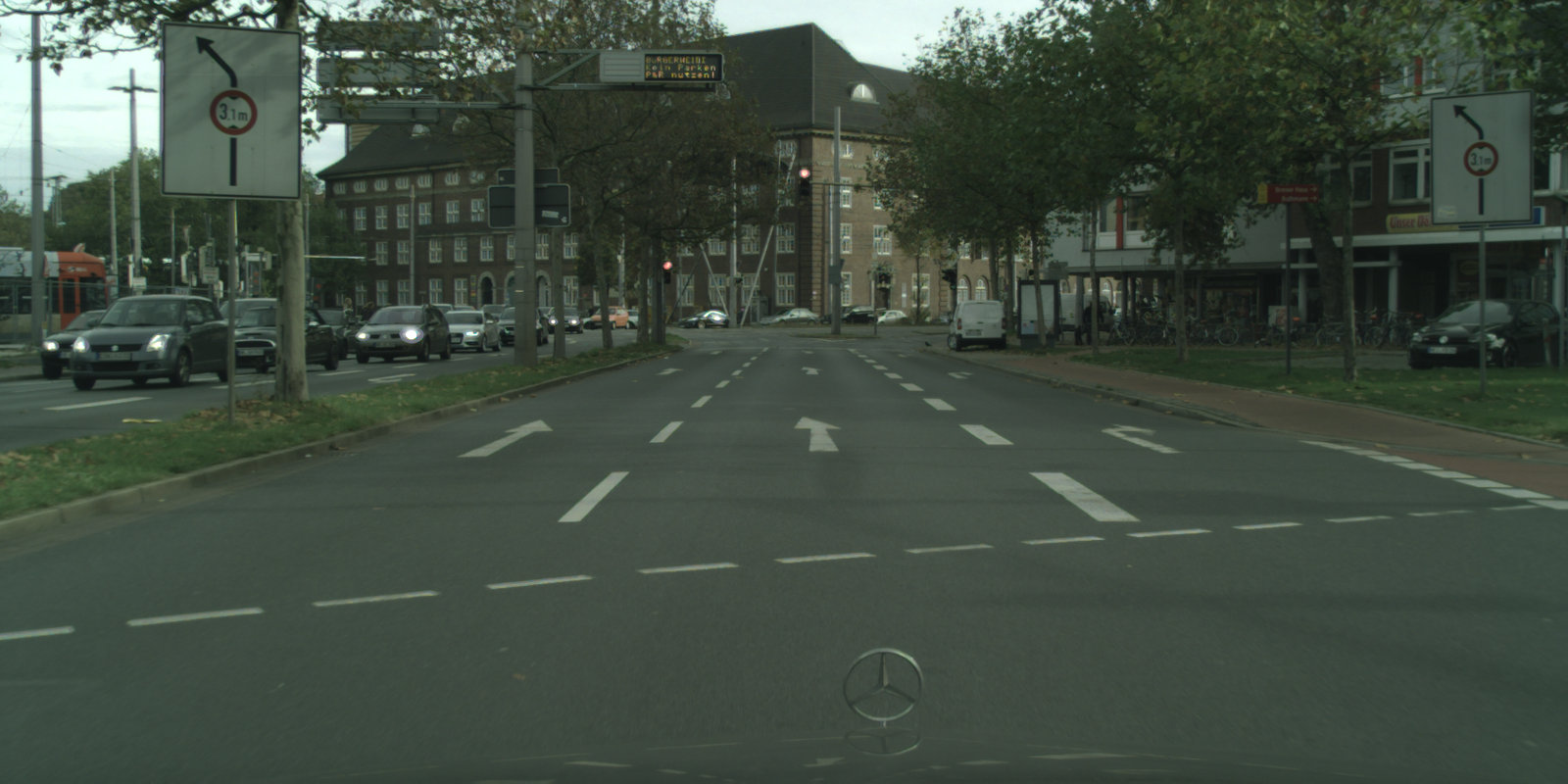} &
\includegraphics[width=\linewidth]{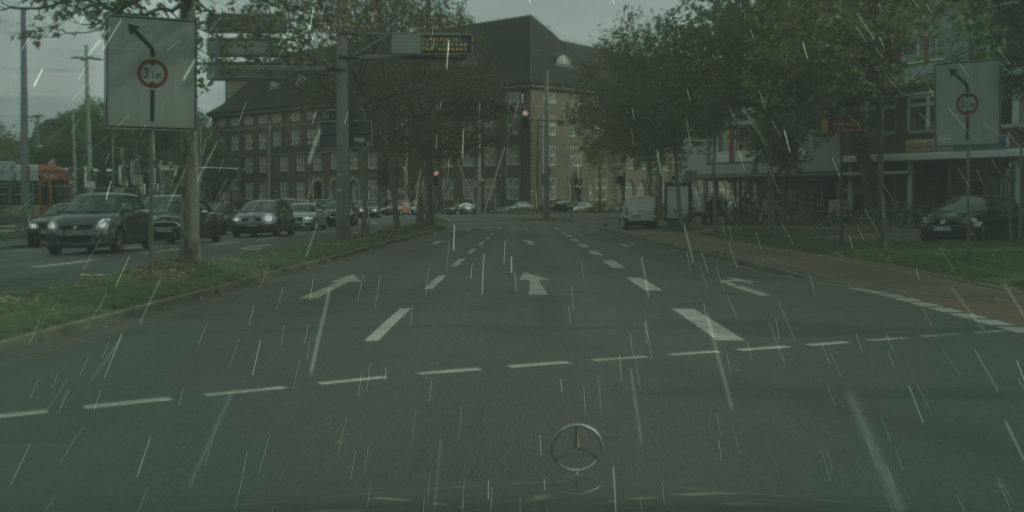} &
\includegraphics[width=\linewidth]{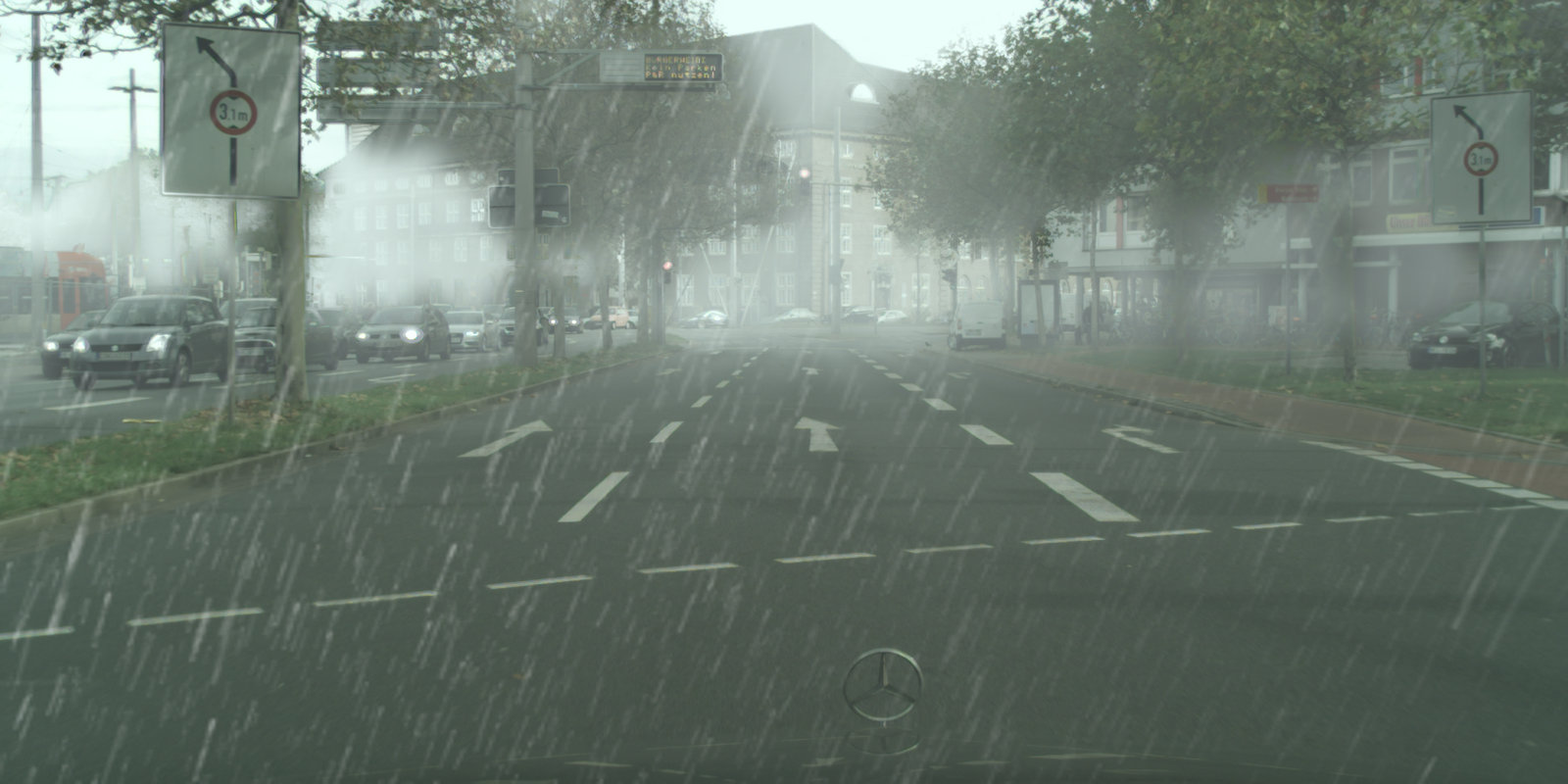} &
\includegraphics[width=\linewidth]{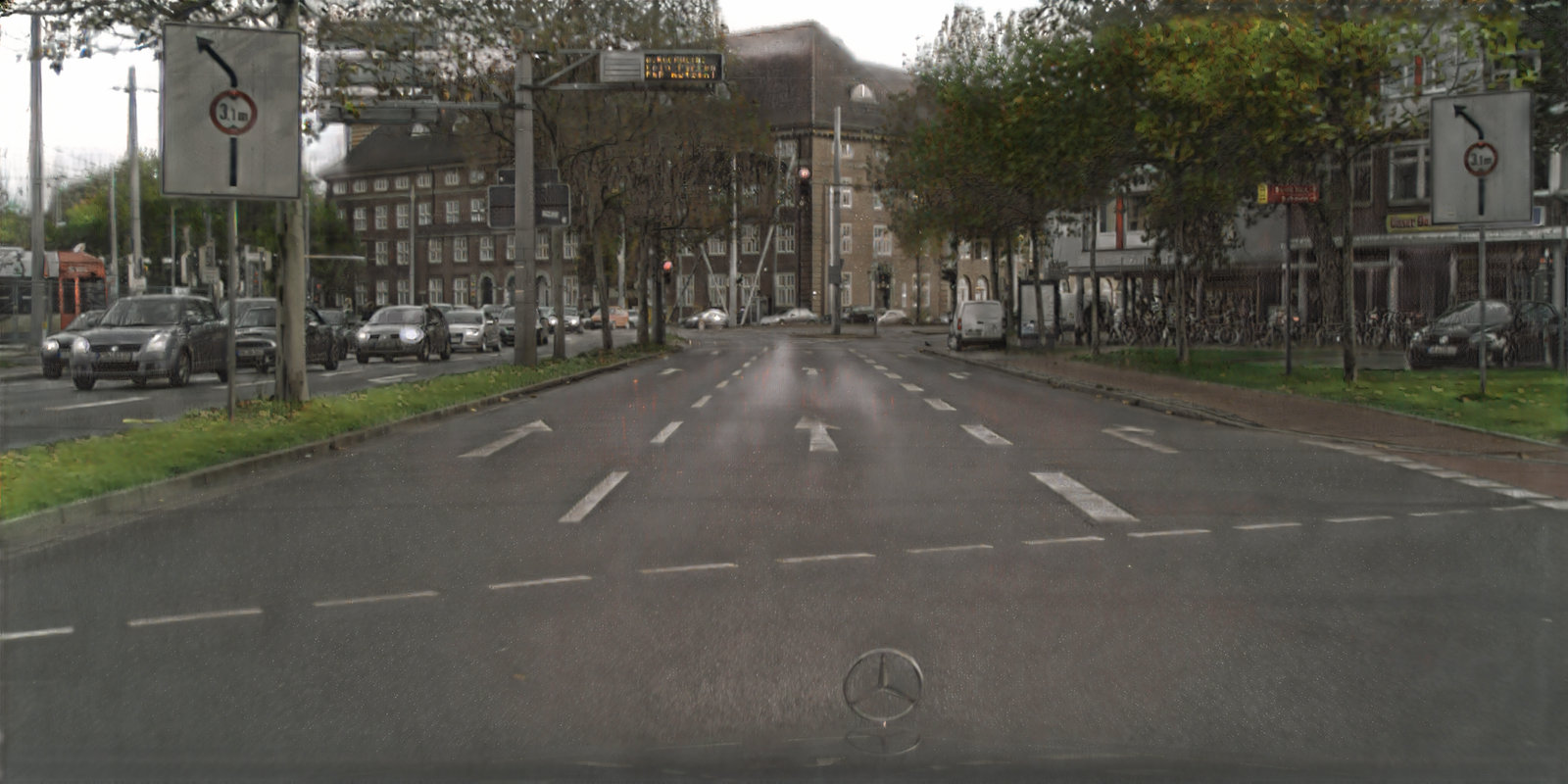} &
\includegraphics[width=\linewidth]{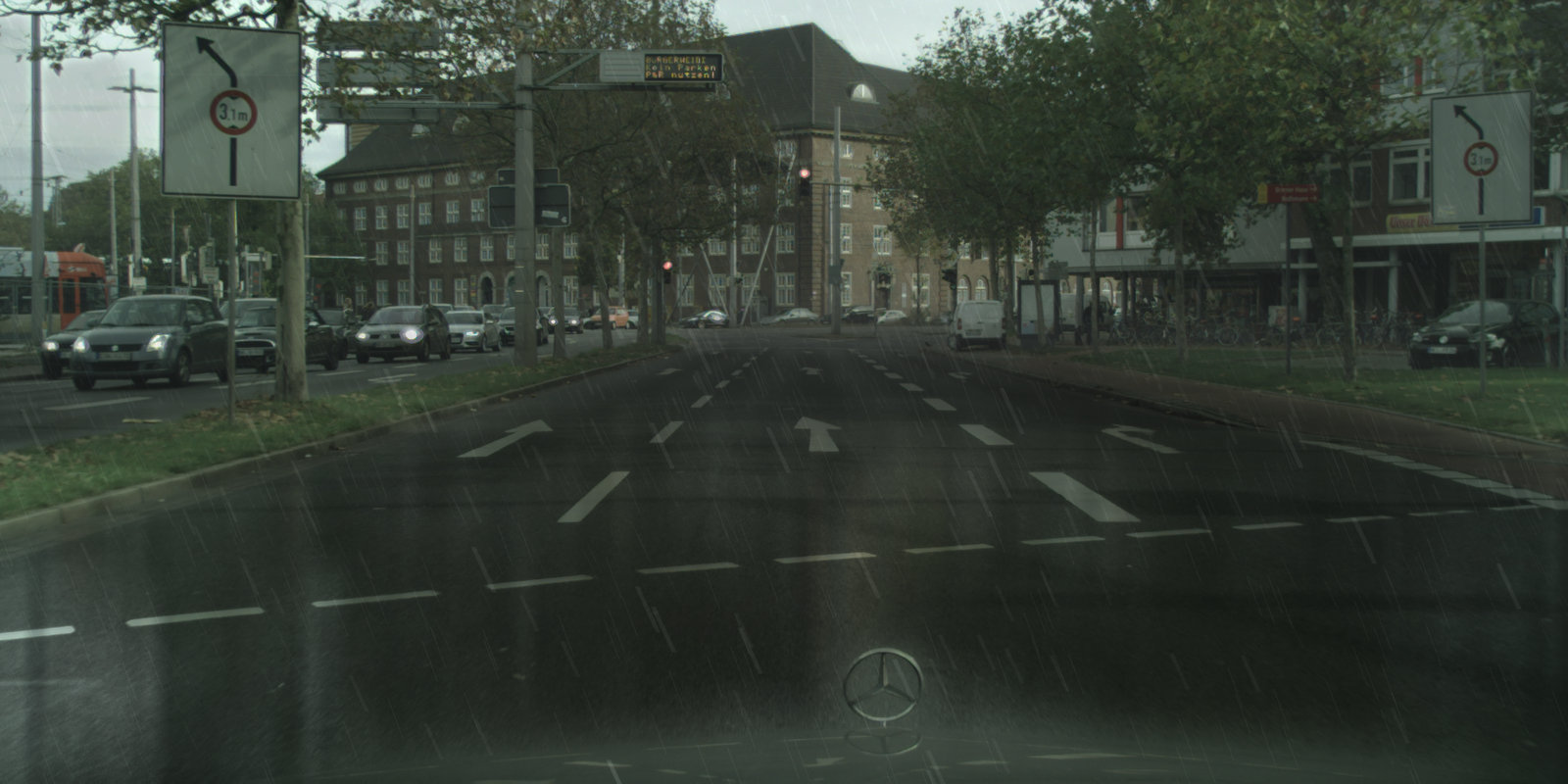} \\[4mm]

\includegraphics[width=\linewidth]{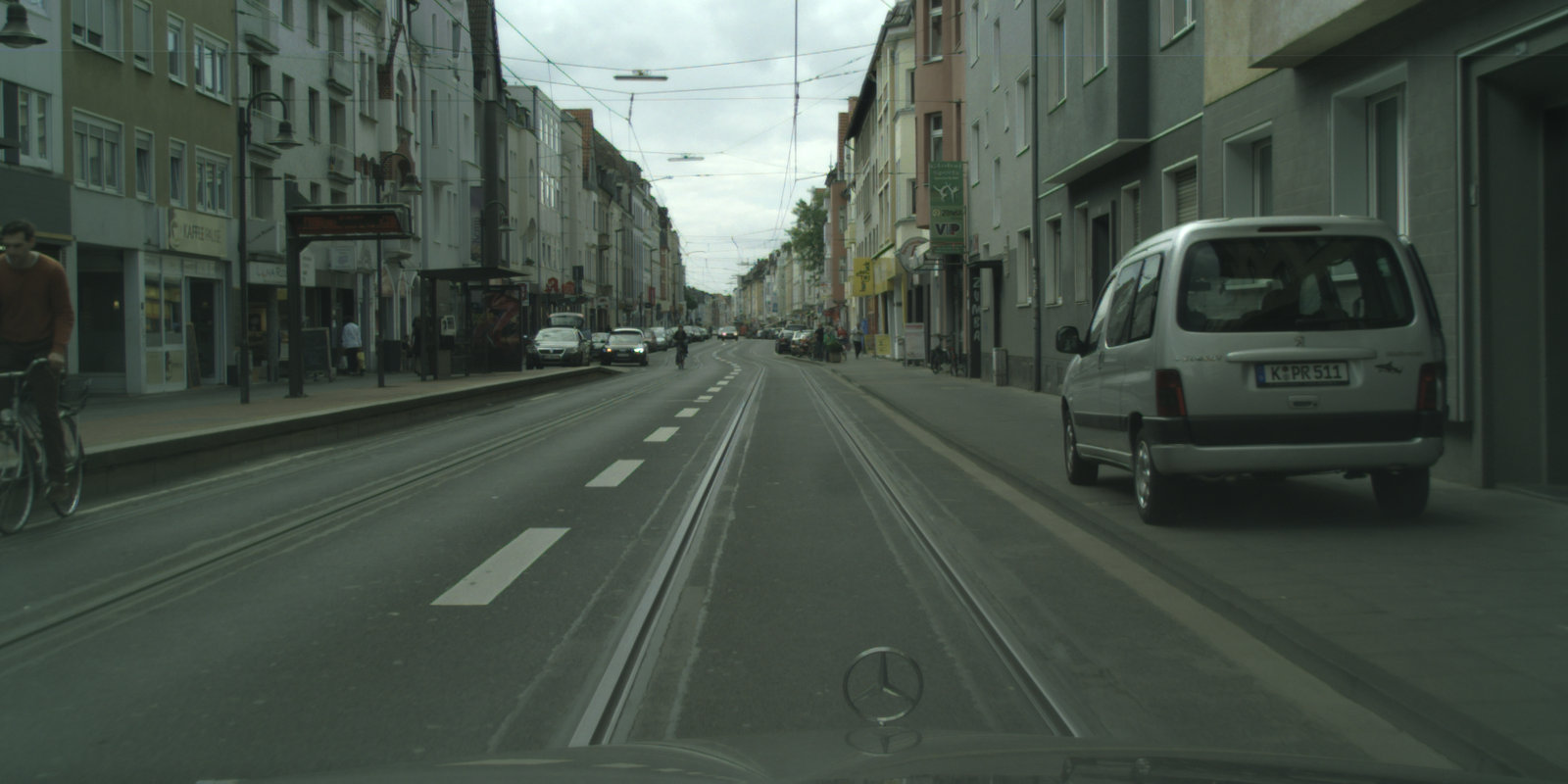} &
\includegraphics[width=\linewidth]{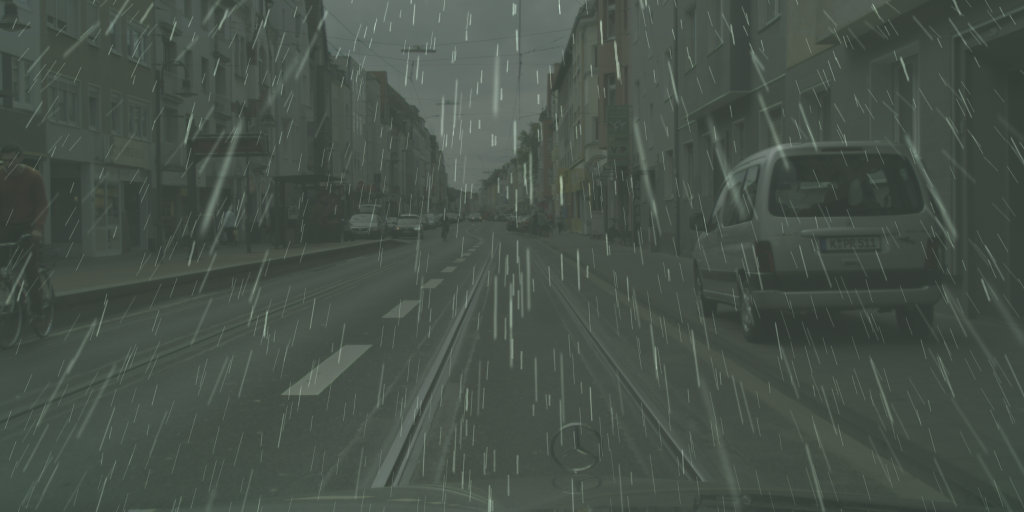} &
\includegraphics[width=\linewidth]{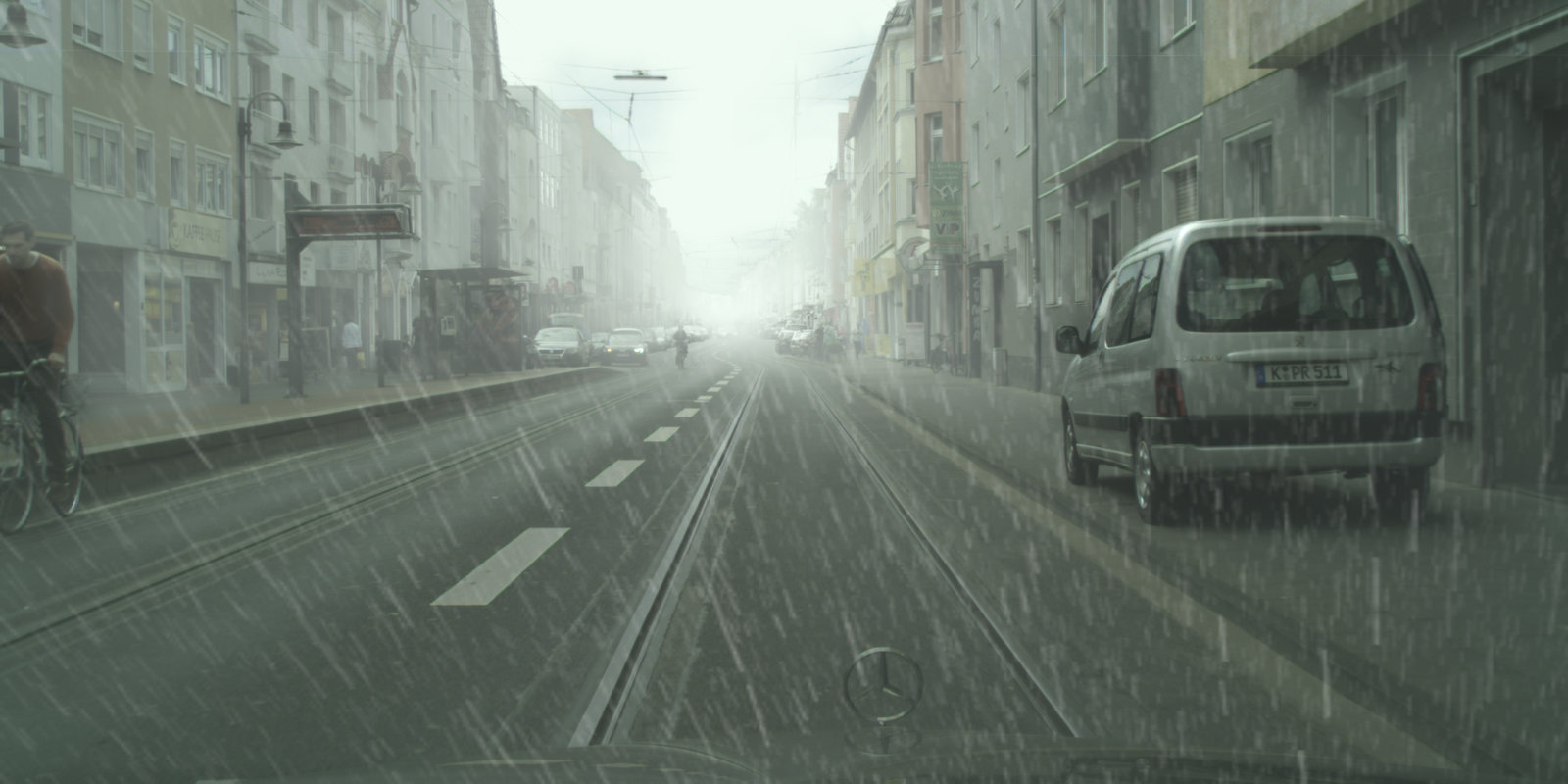} &
\includegraphics[width=\linewidth]{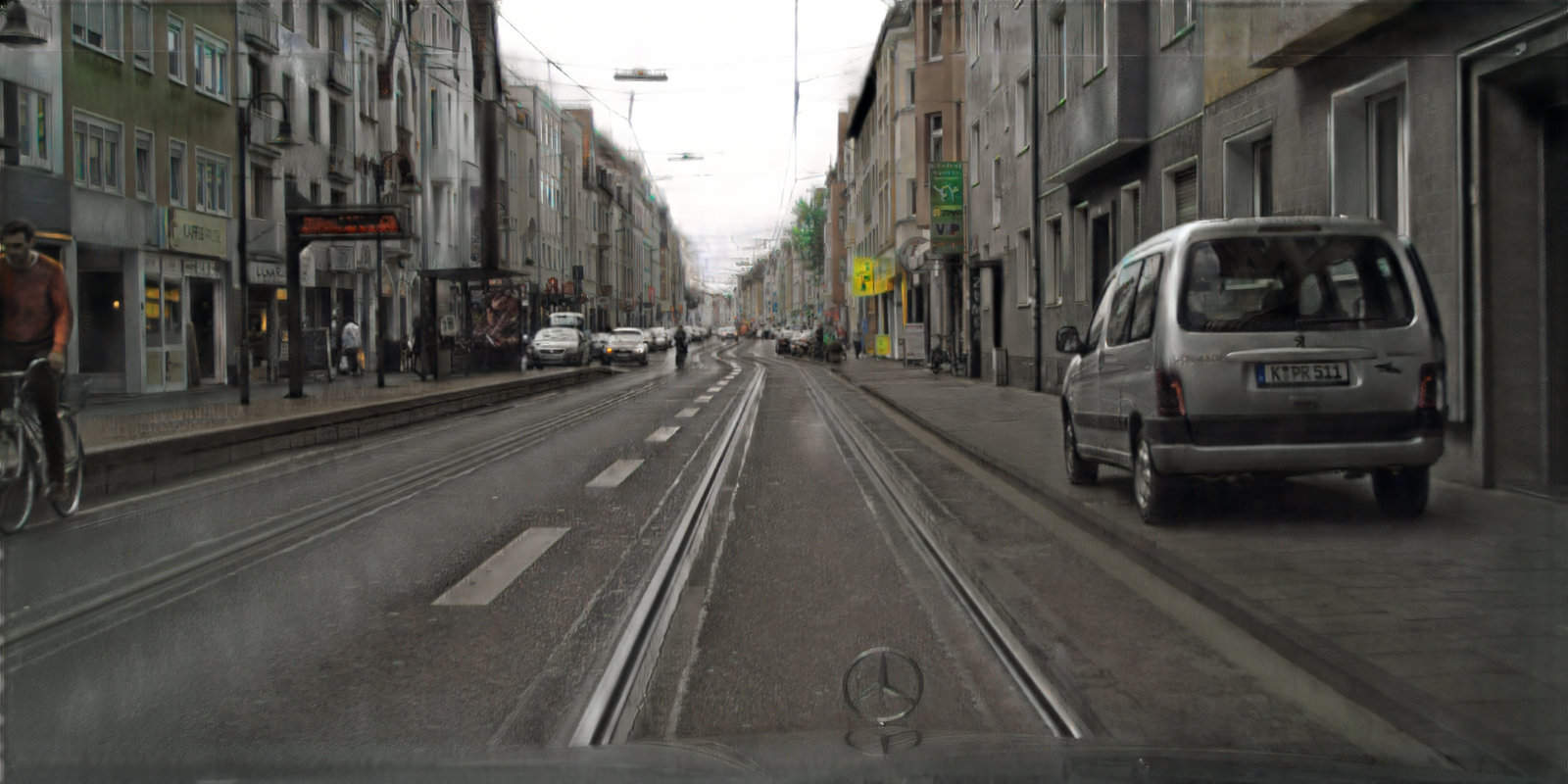} &
\includegraphics[width=\linewidth]{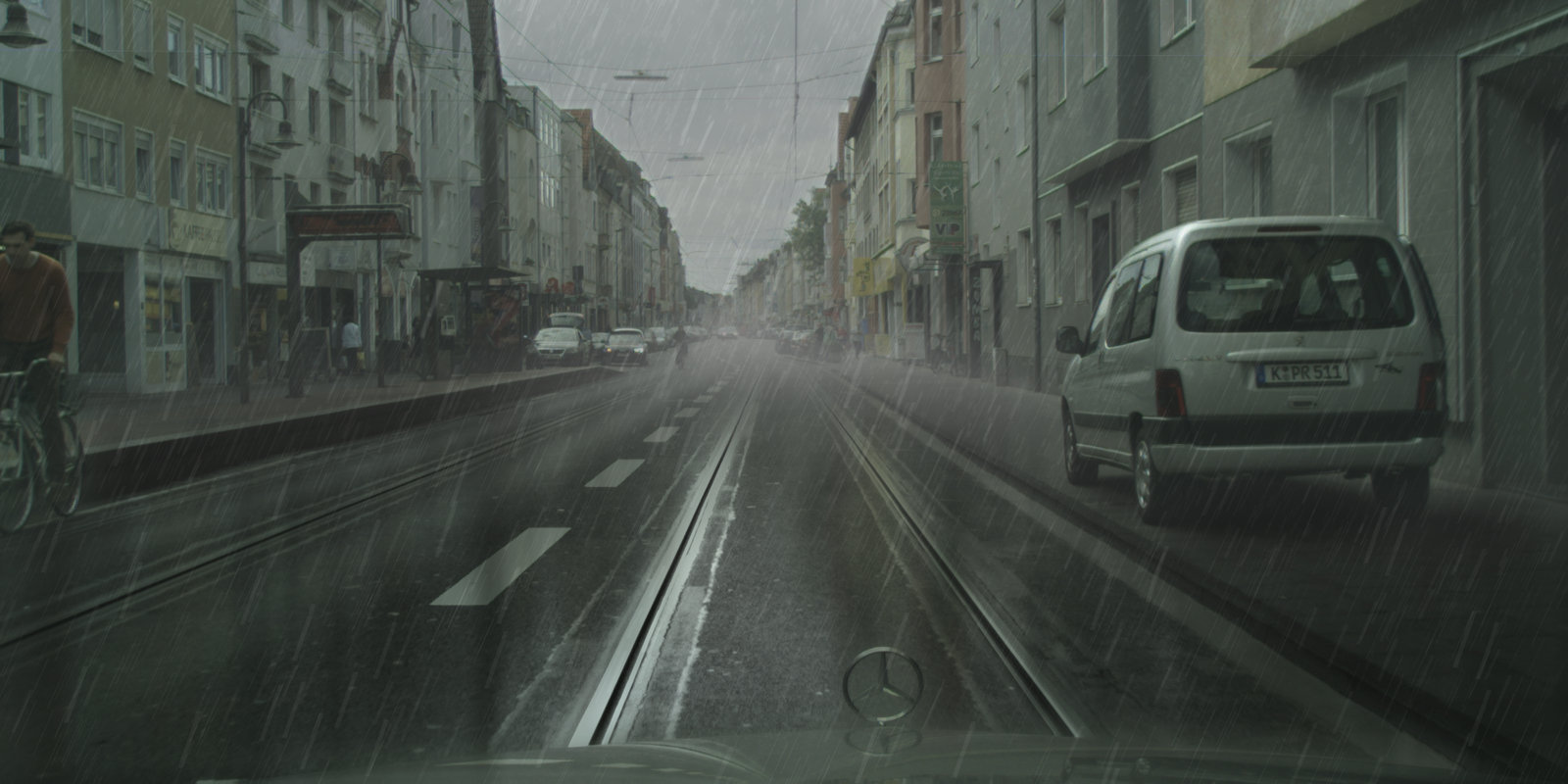} \\[4mm]

\includegraphics[width=\linewidth]{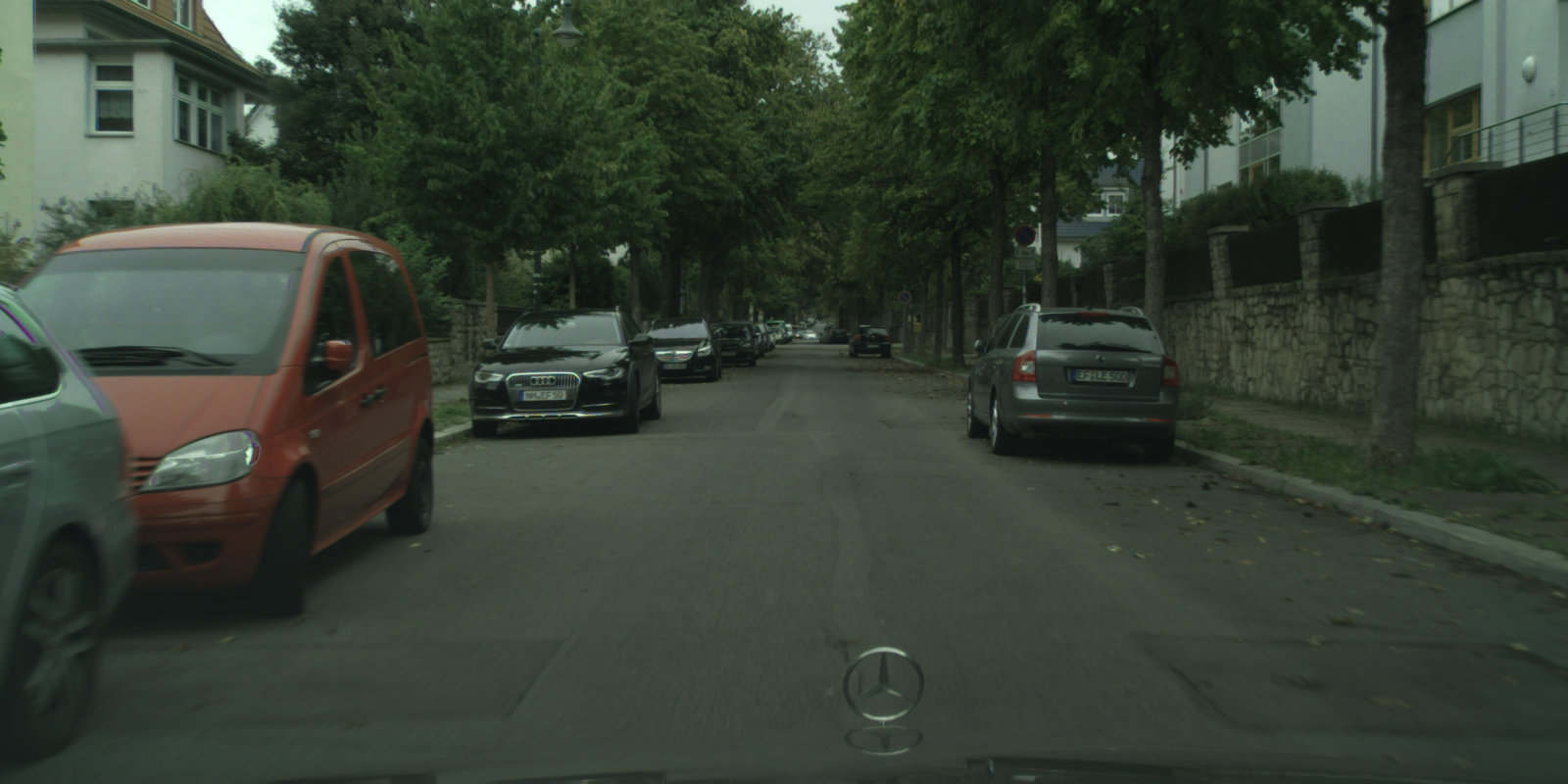} &
\includegraphics[width=\linewidth]{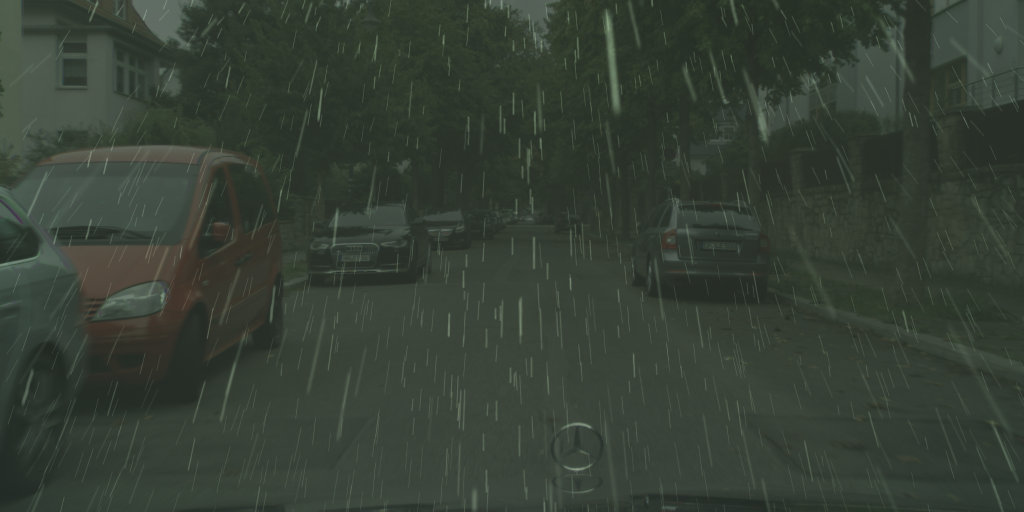} &
\includegraphics[width=\linewidth]{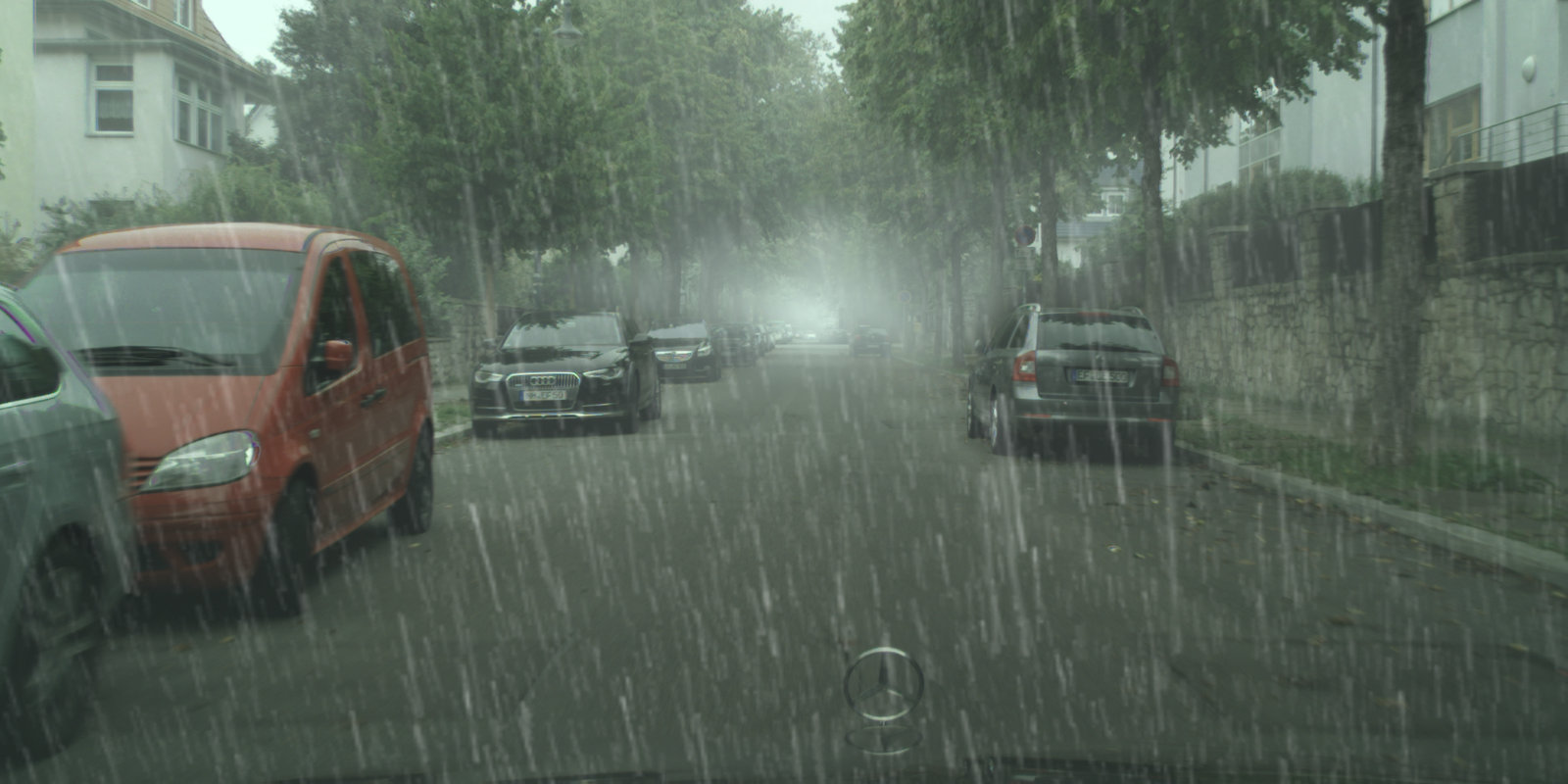} &
\includegraphics[width=\linewidth]{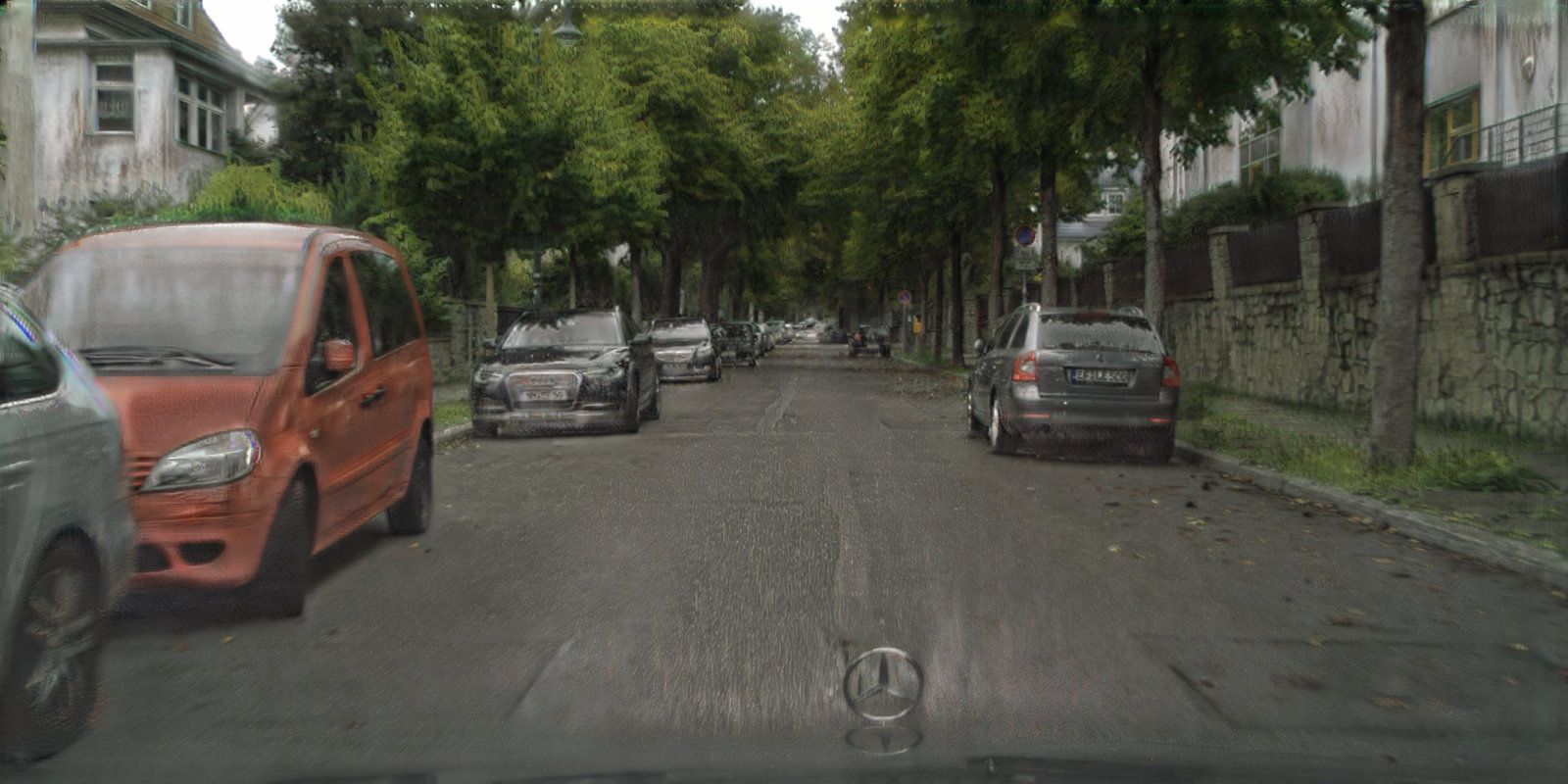} &
\includegraphics[width=\linewidth]{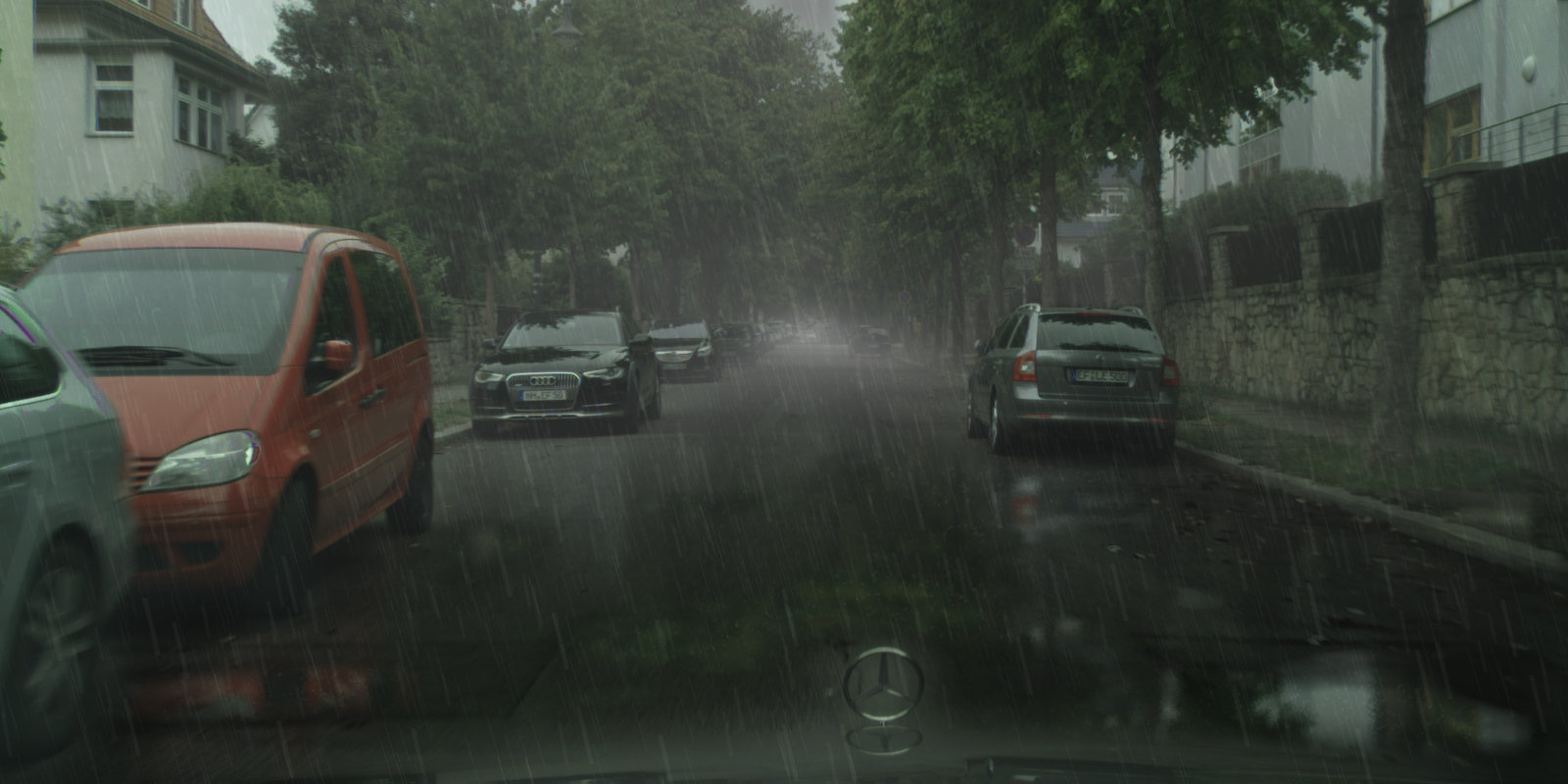} \\

\bottomrule
\end{tabular}

\label{tab:rain_methods_comparison_transposed}
\vspace{-3mm}
\end{figure*}

\section{Qualitative Comparison of Rain Synthesis Methods \label{supmat: qual. comparison of rain synth methods}}
Figure~\ref{tab:rain_methods_comparison_transposed} presents a side-by-side visual comparison of the existing rain synthesis approaches evaluated in this work. Each column corresponds to a different Cityscapes scene, while each row shows the output produced by one of the baseline methods. This layout highlights the characteristic artifacts and stylistic tendencies of each approach under identical scene conditions. In line with the main paper, the examples illustrate the limitations of prior methods, such as depth-inconsistent streaks, oversaturated fog, or missing surface effects, and the improved realism achieved by CascadeRS. In particular, our method generates depth-aligned streaks, scene-consistent reflections, and physically plausible puddles across diverse backgrounds and scene layouts, yielding more coherent and perceptually convincing rainy scenes.

\begin{figure*}[!t]
\centering
\renewcommand{\arraystretch}{1.2}
\setlength{\tabcolsep}{2pt}
\caption{
\textbf{Rain synthesis with GenAI methods.}
The original image (first column) is processed with Stable Diffusion XL (second column) and three times with Gemini (Nano Banana) to synthesize rain effects. For all generations, the same text prompt was used.
}
\vspace{3pt}
\scriptsize

\begin{tabular}{
C{0.19\linewidth}
C{0.19\linewidth}
C{0.19\linewidth}
C{0.19\linewidth}
C{0.19\linewidth}
}
\toprule
\textbf{Original} &
\textbf{CascadeRS} &
\textbf{SDXL} &
\multicolumn{2}{c}{\textbf{Gemini (Nano Banana})} \\
\midrule
\includegraphics[width=\linewidth]{figures/PART2/9/Original.jpg} & 
\includegraphics[width=\linewidth]{figures/PART2/9/CascadeRS.jpg} &
\includegraphics[width=\linewidth,height=11.5mm]{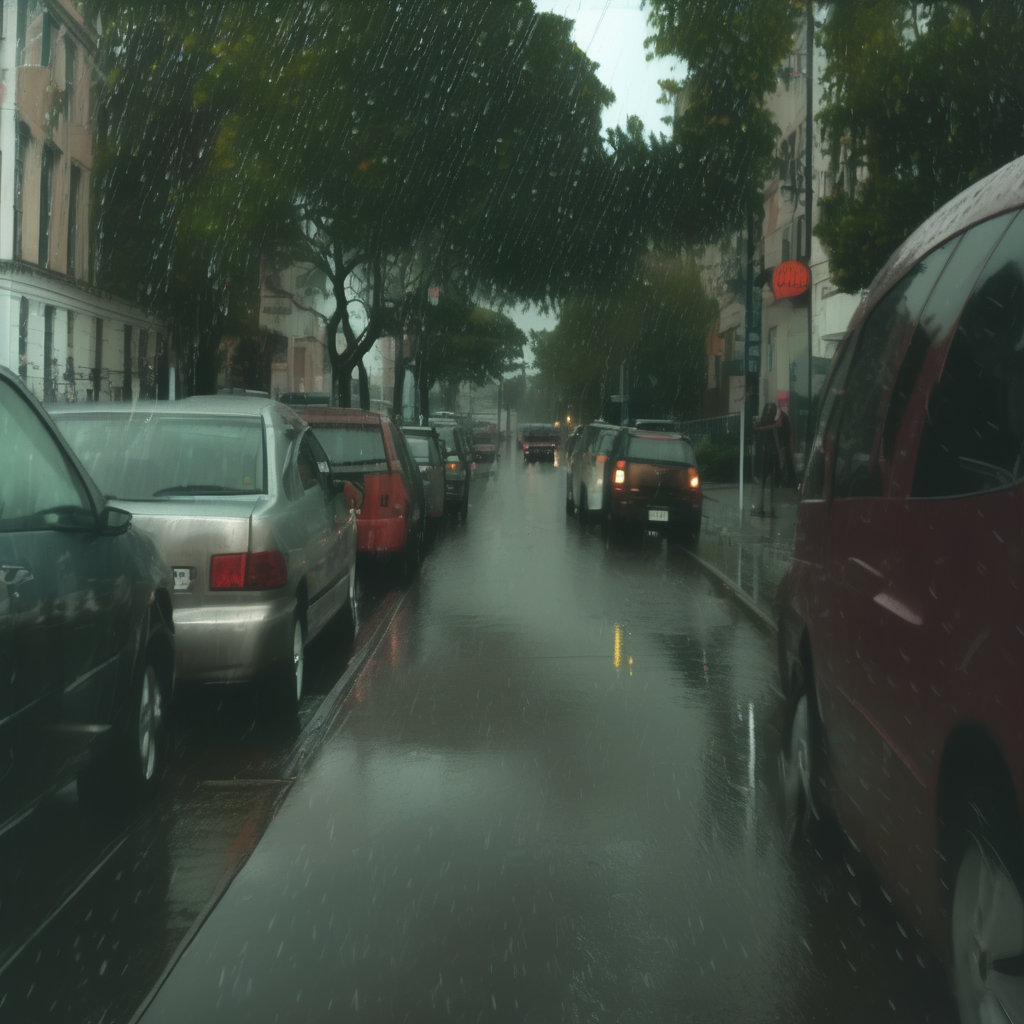} &
\includegraphics[width=\linewidth]{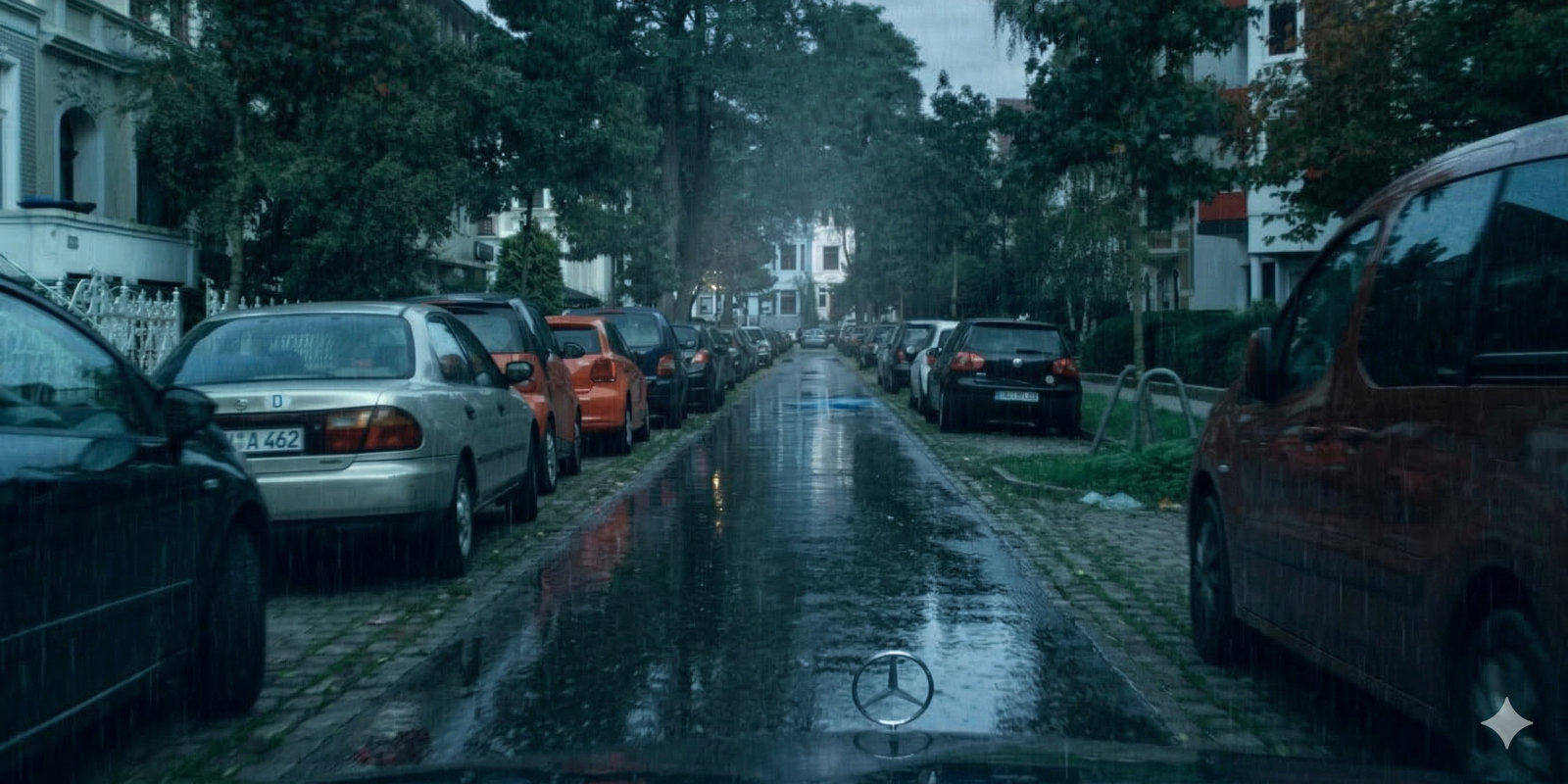} &
\includegraphics[width=\linewidth]{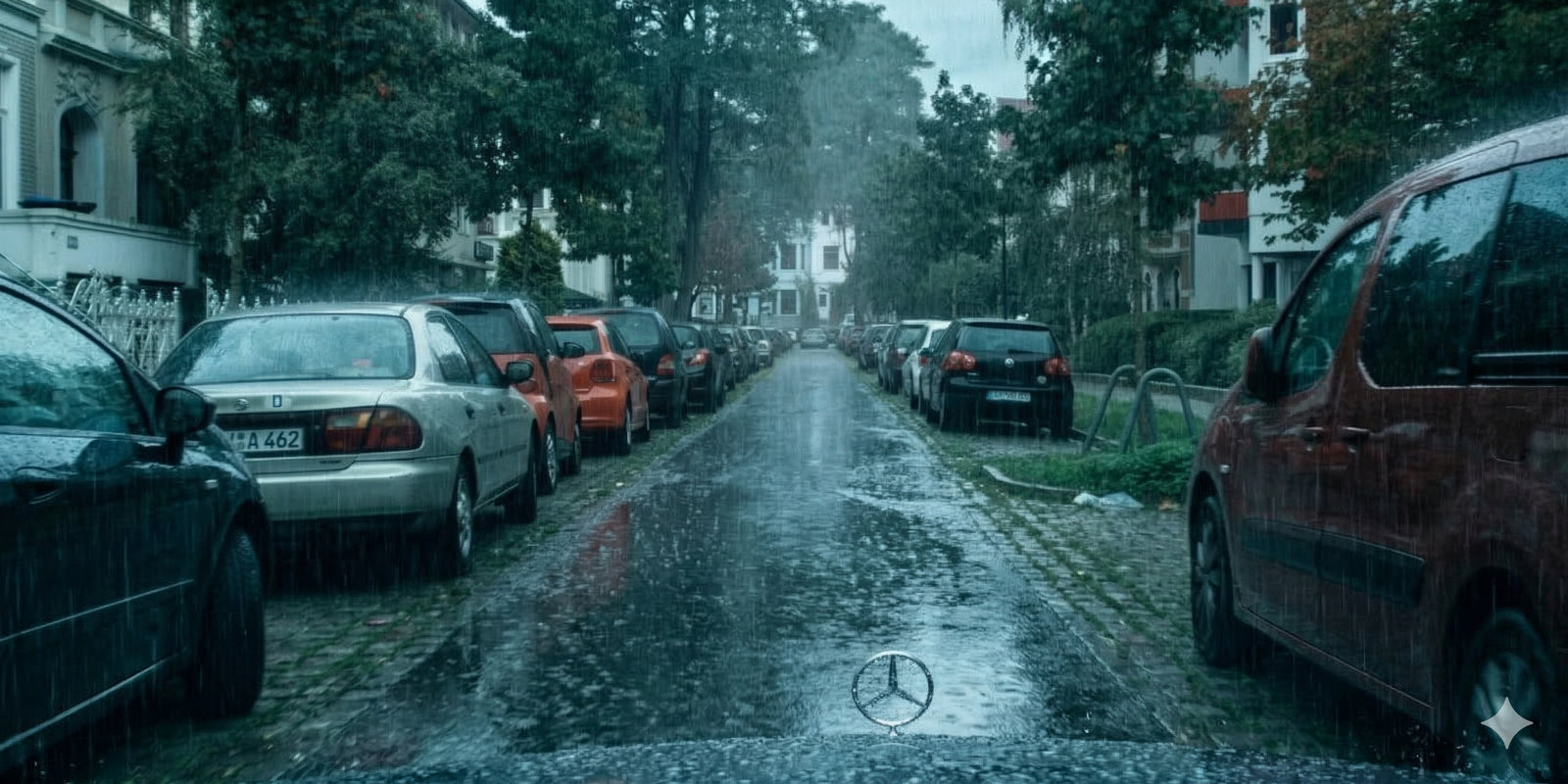}\\[4mm]

\includegraphics[width=\linewidth]{figures/PART2/11/Original.jpg} &
\includegraphics[width=\linewidth]{figures/PART2/11/CascadeRS.jpg} &
\includegraphics[width=\linewidth,height=11.5mm]{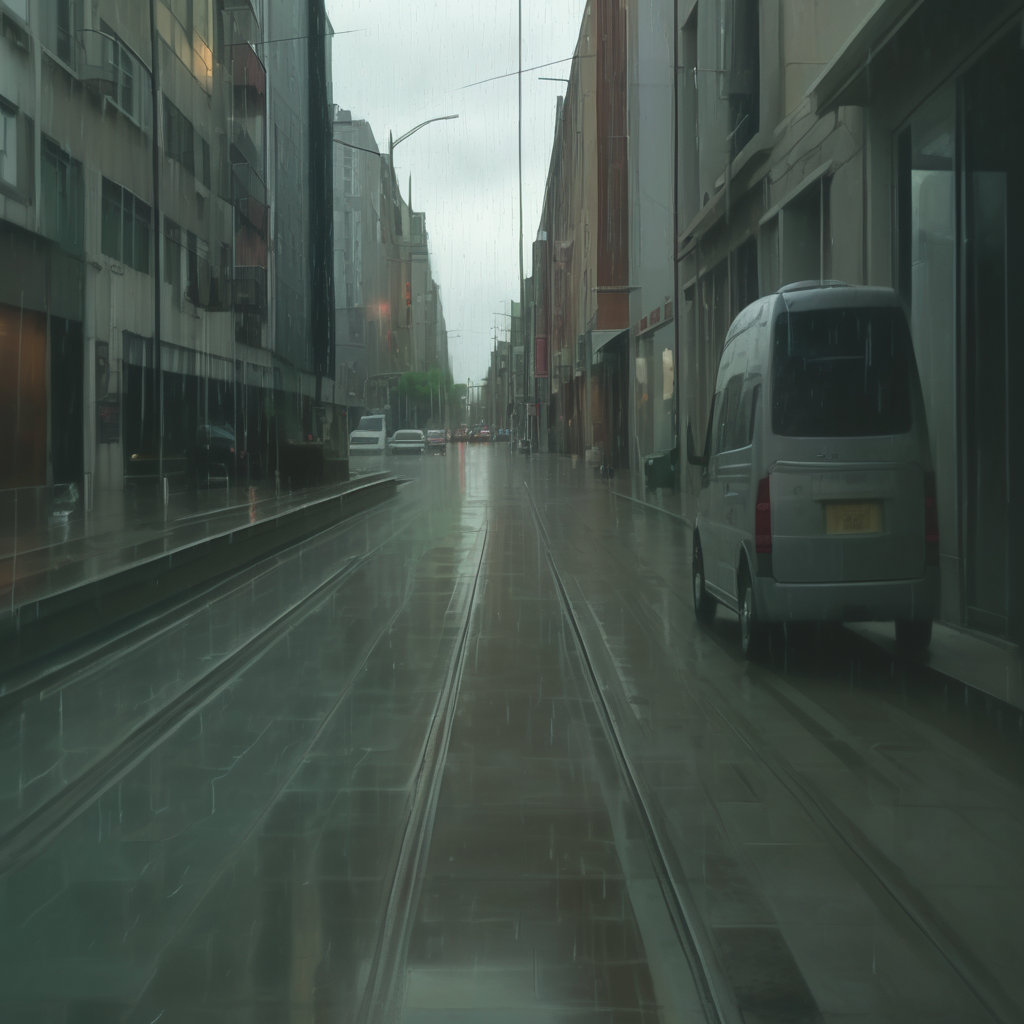} &
\includegraphics[width=\linewidth]{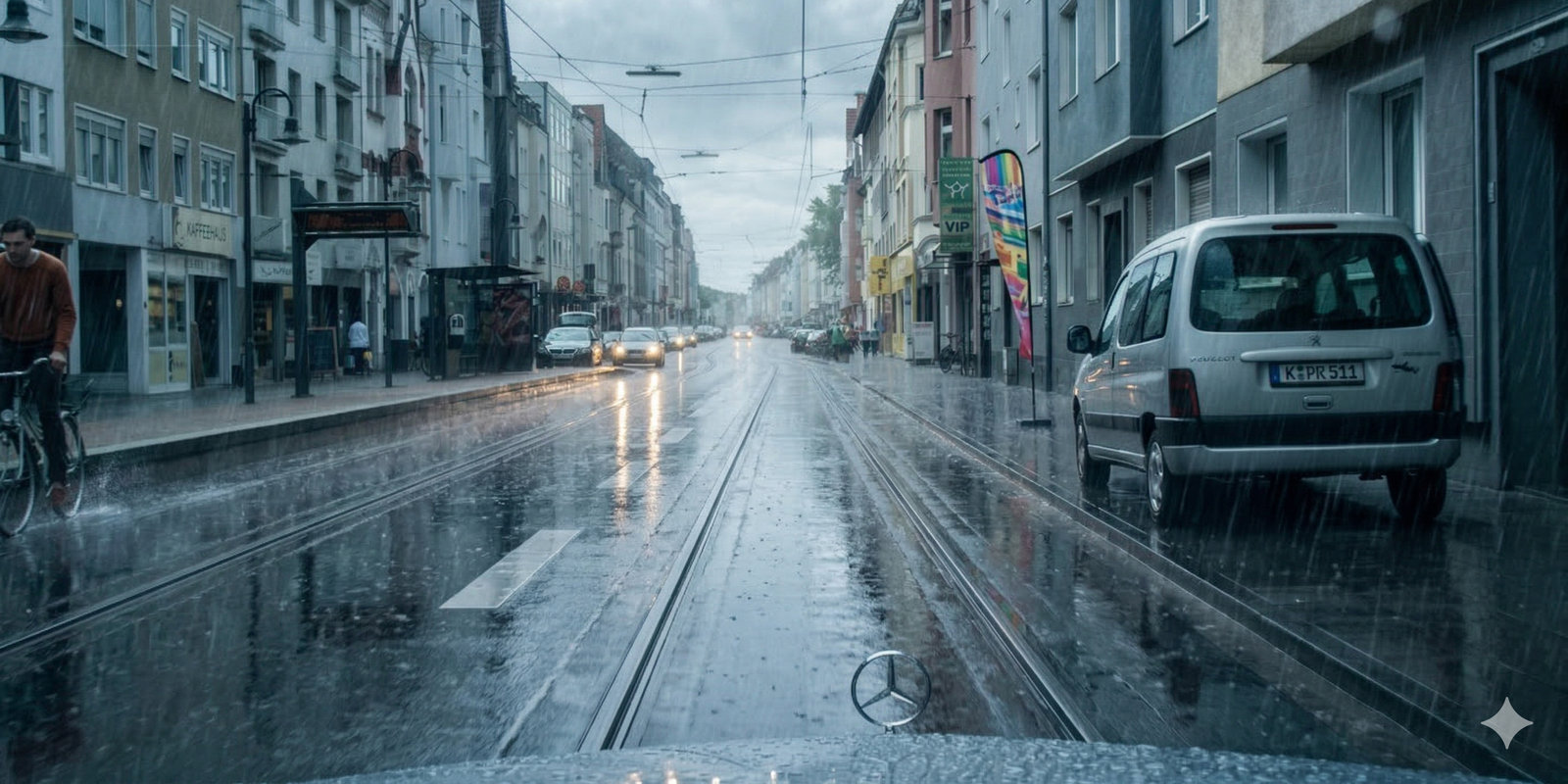} &
\includegraphics[width=\linewidth]{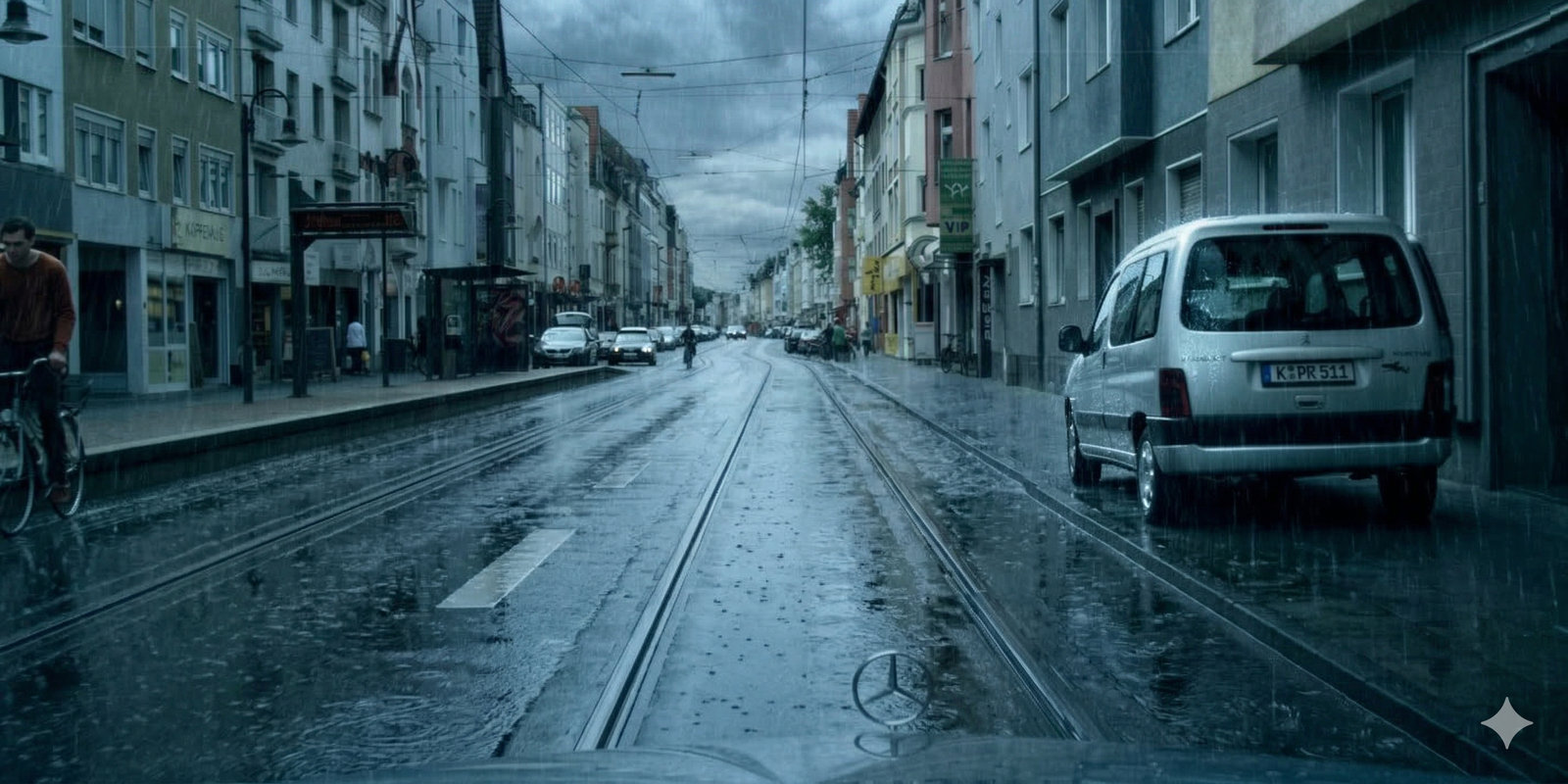}\\[4mm]

\includegraphics[width=\linewidth]{figures/PART2/5/Original.jpg} &
\includegraphics[width=\linewidth]{figures/PART2/5/CascadeRS.jpg} &
\includegraphics[width=\linewidth,height=11.5mm]{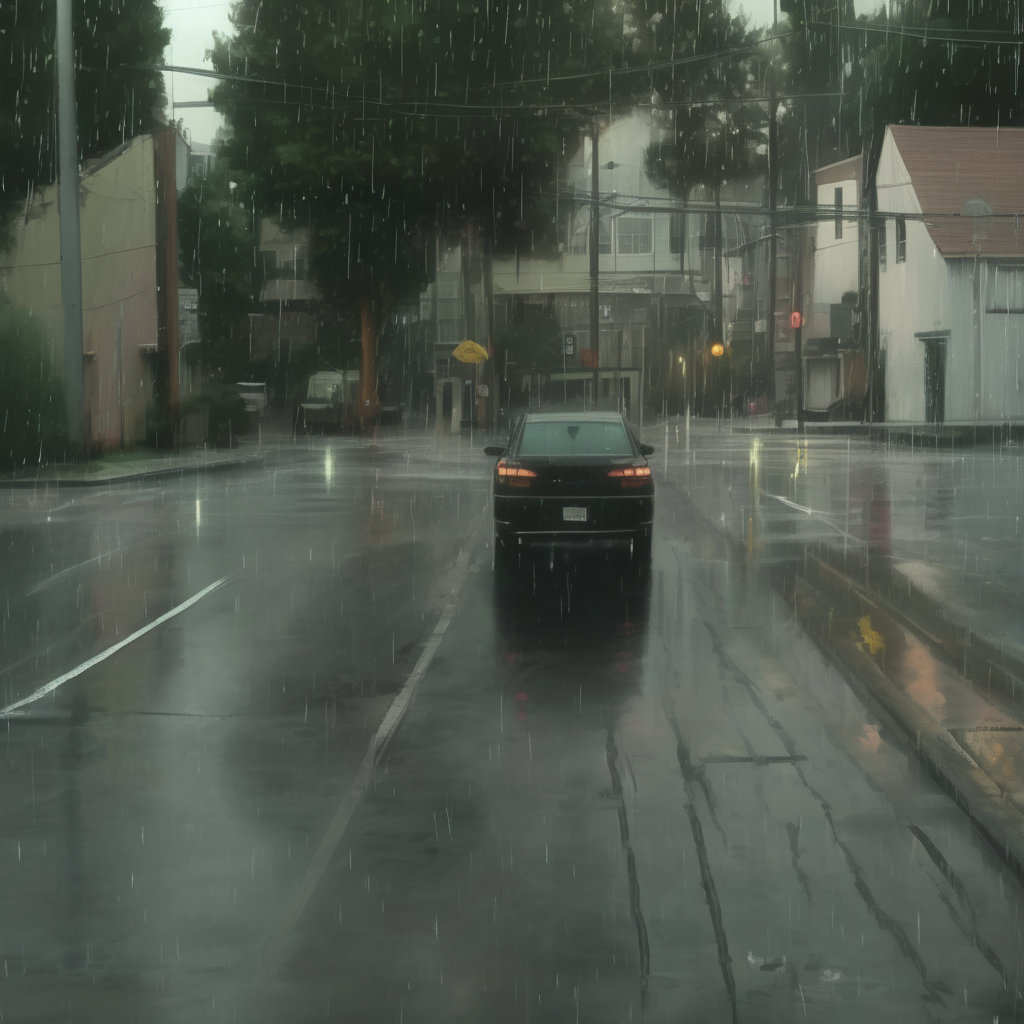} &
\includegraphics[width=\linewidth]{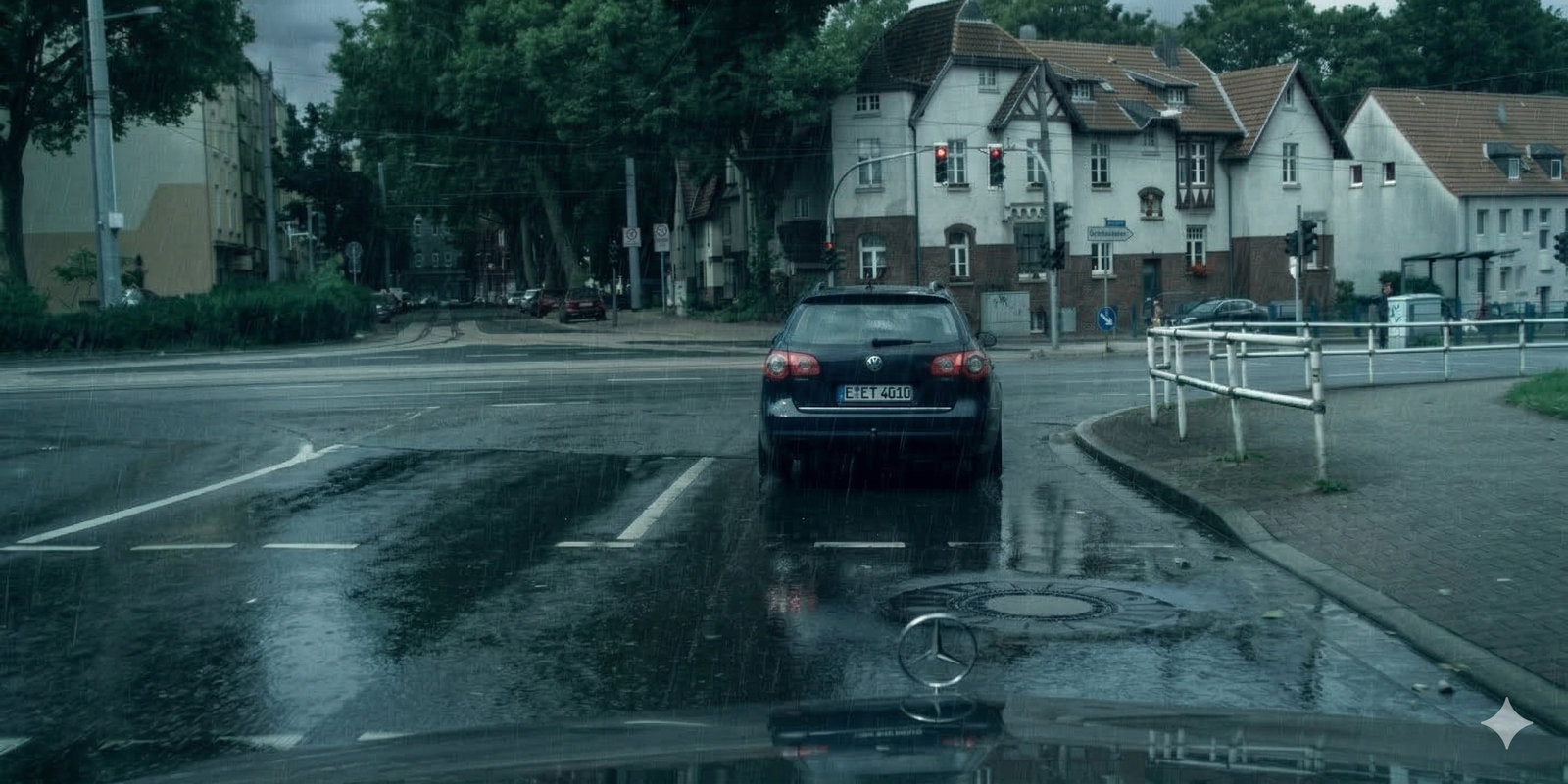} &
\includegraphics[width=\linewidth]{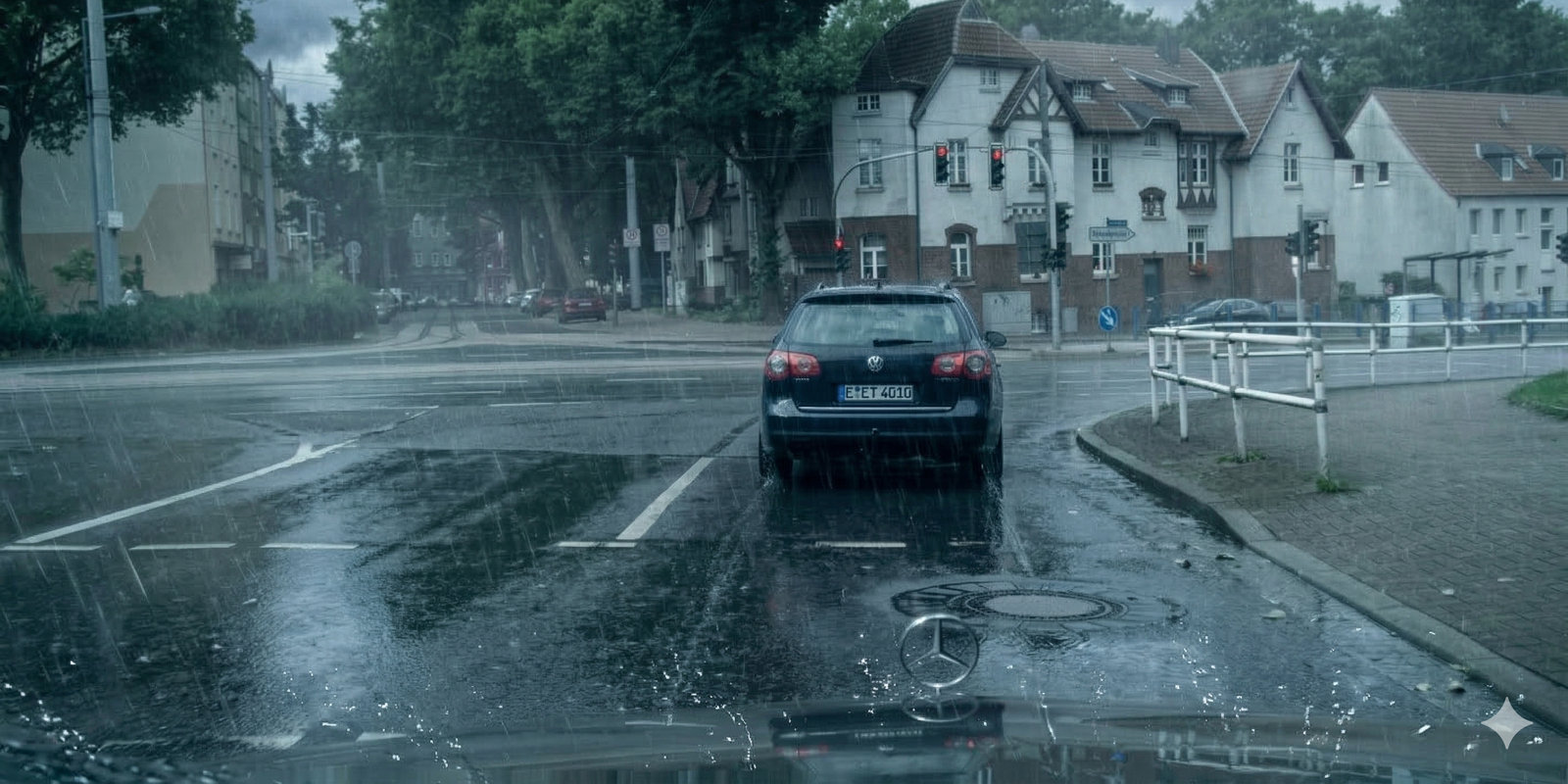}\\[4mm]

\includegraphics[width=\linewidth]{figures/PART2/6/Original.jpg} &
\includegraphics[width=\linewidth]{figures/PART2/6/CascadeRS.jpg} &
\includegraphics[width=\linewidth,height=11.5mm]{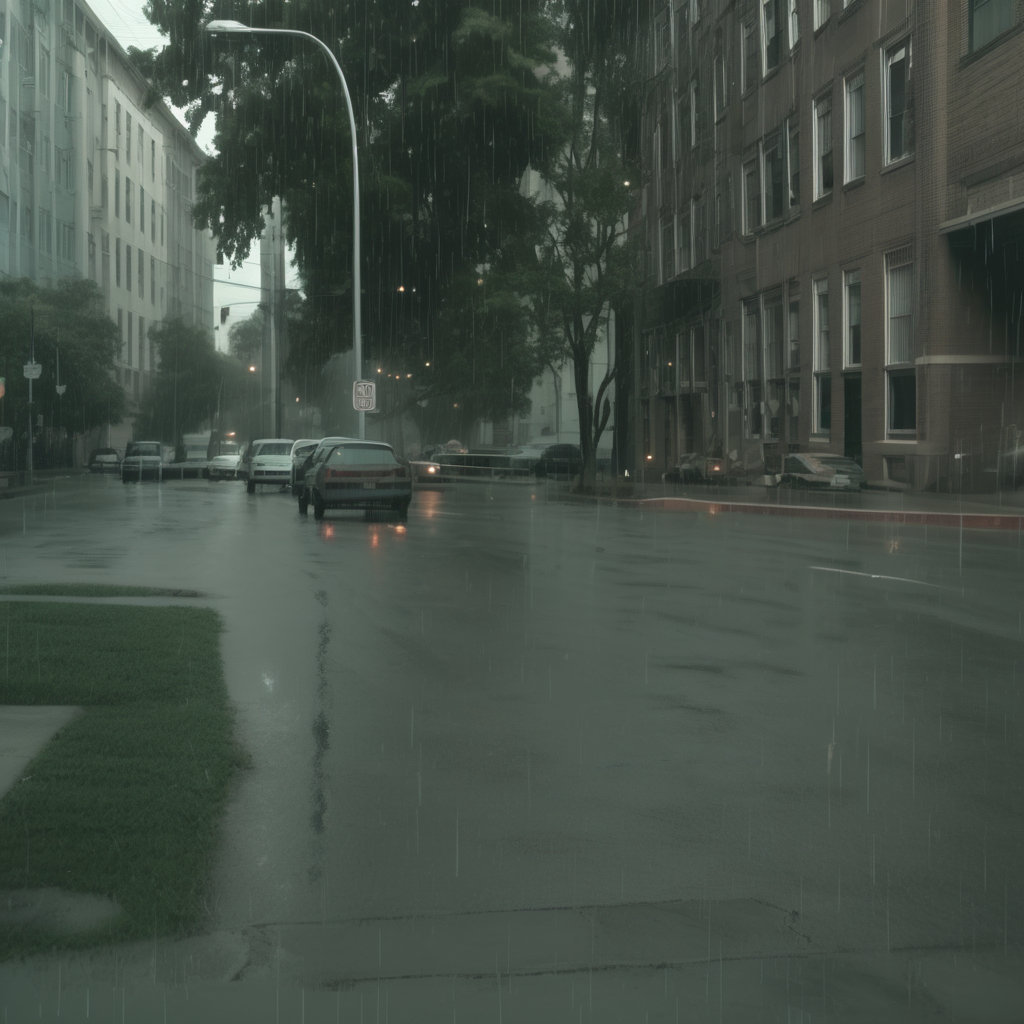} &
\includegraphics[width=\linewidth]{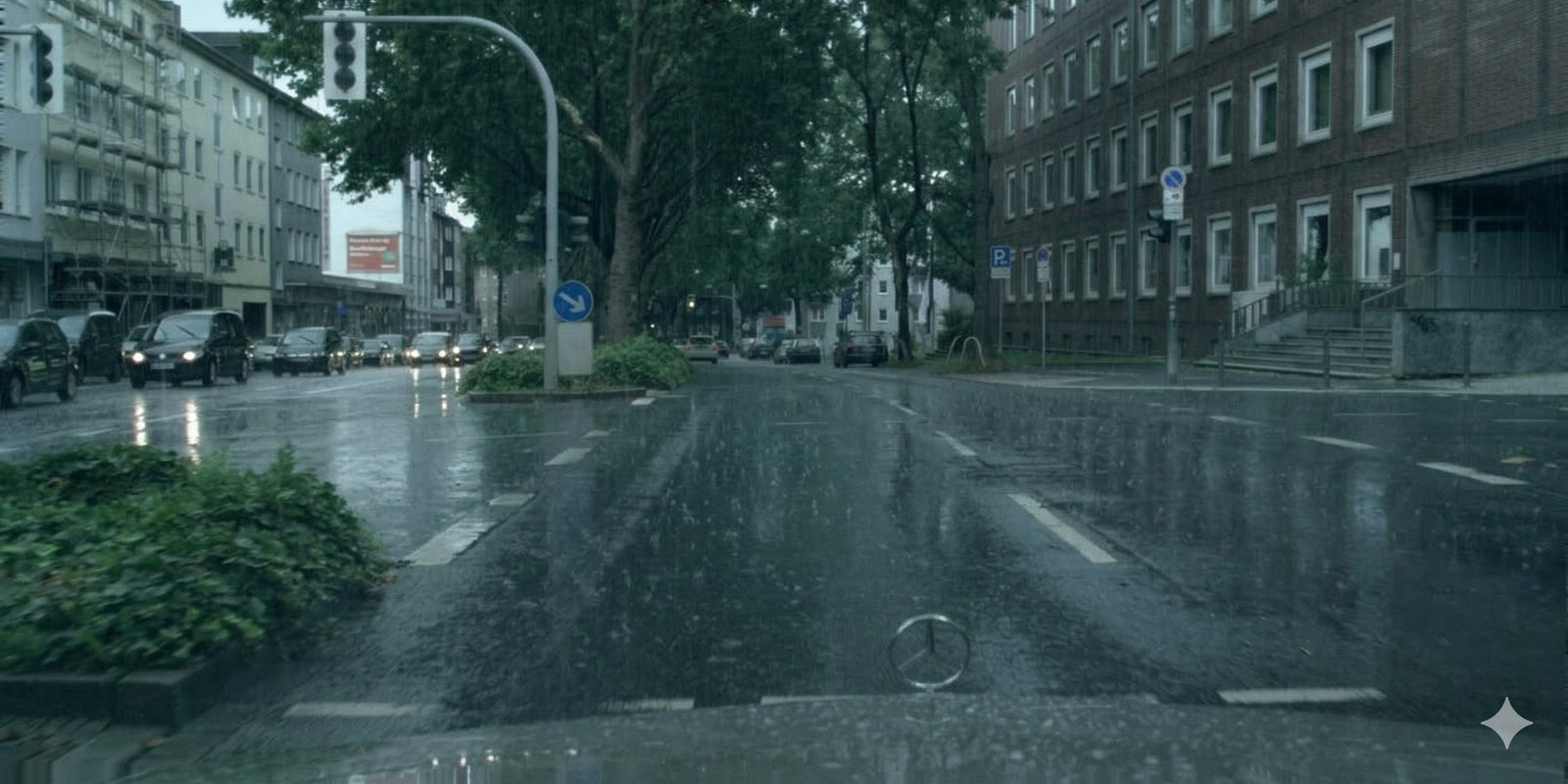} &
\includegraphics[width=\linewidth]{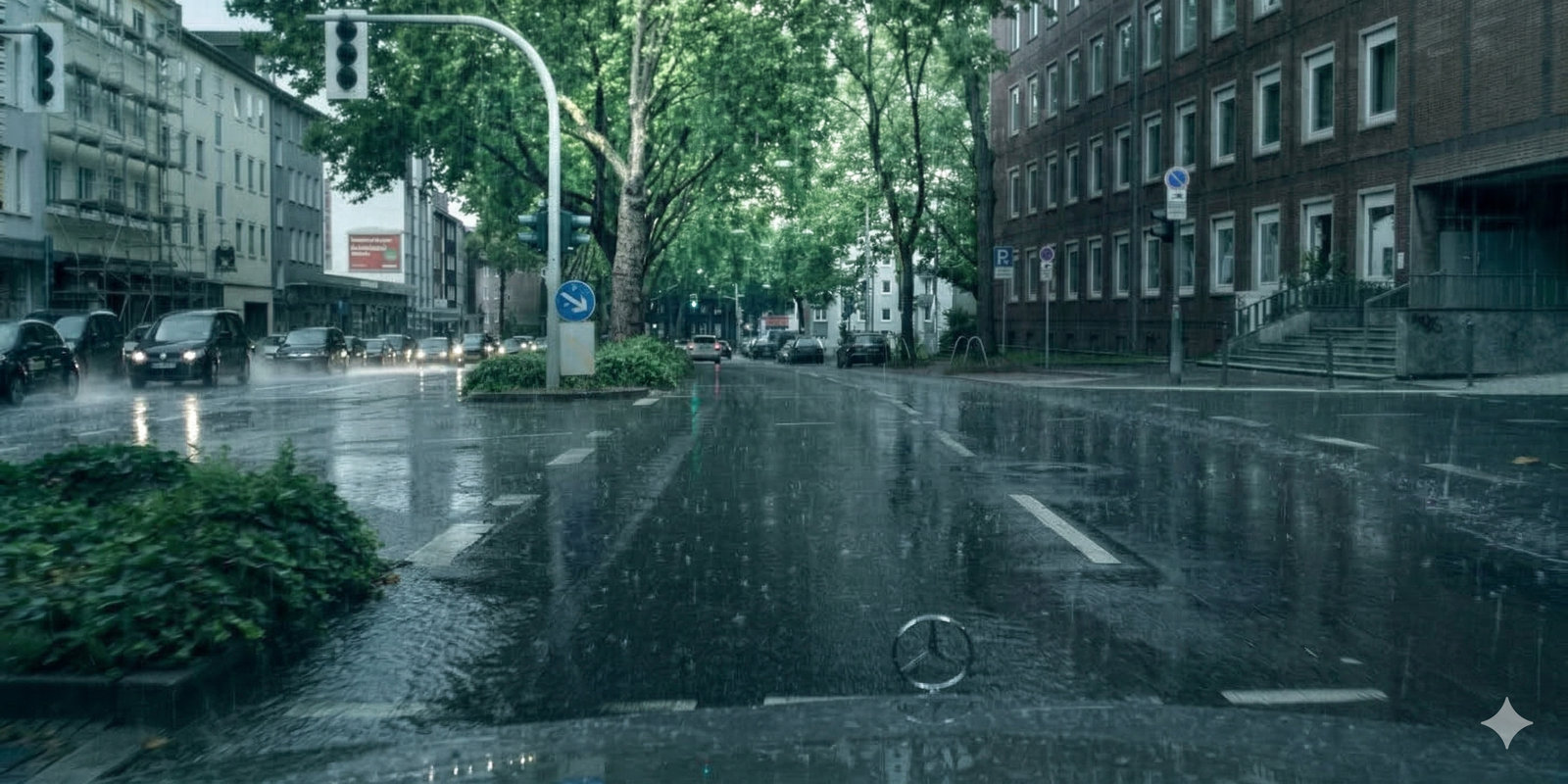}\\[4mm]

\includegraphics[width=\linewidth]{figures/PART2/7/Original.jpg} &
\includegraphics[width=\linewidth]{figures/PART2/7/CascadeRS.jpg} &
\includegraphics[width=\linewidth,height=11.5mm]{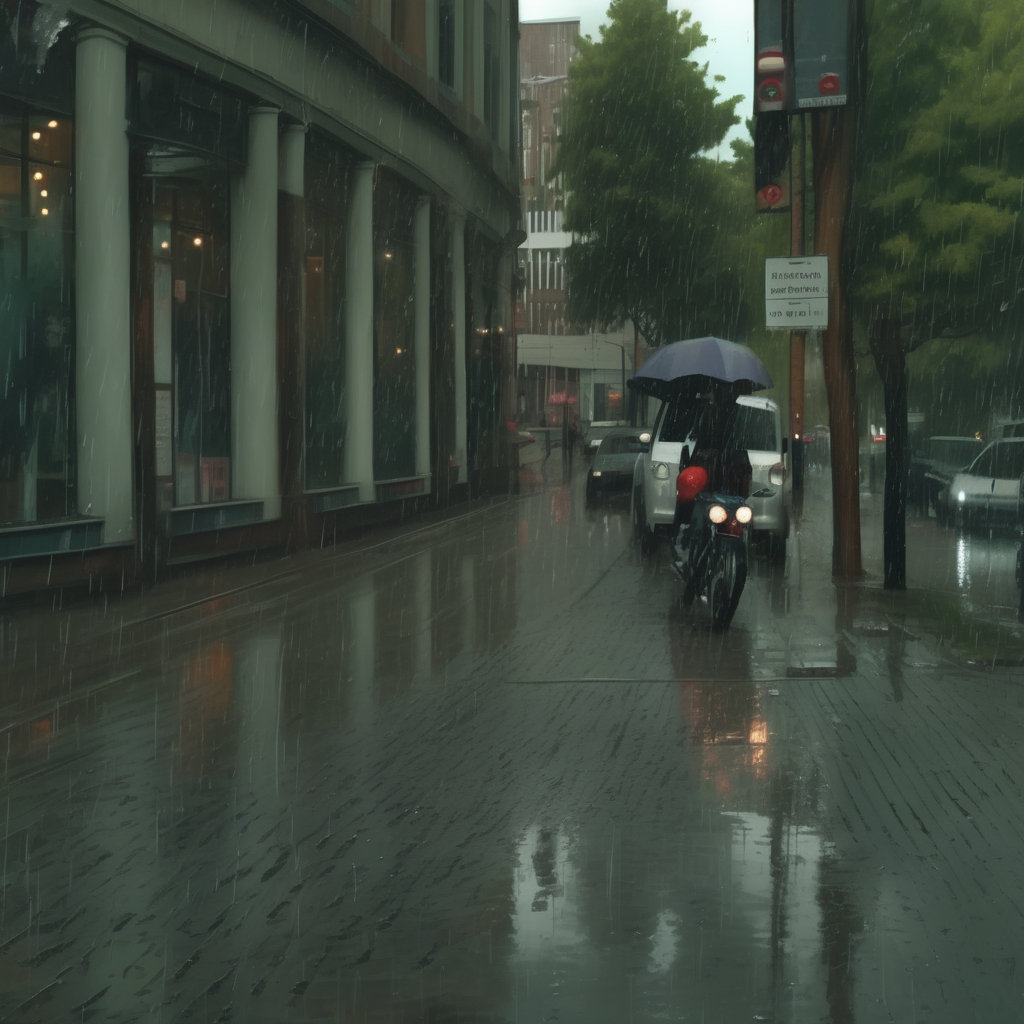} &
\includegraphics[width=\linewidth]{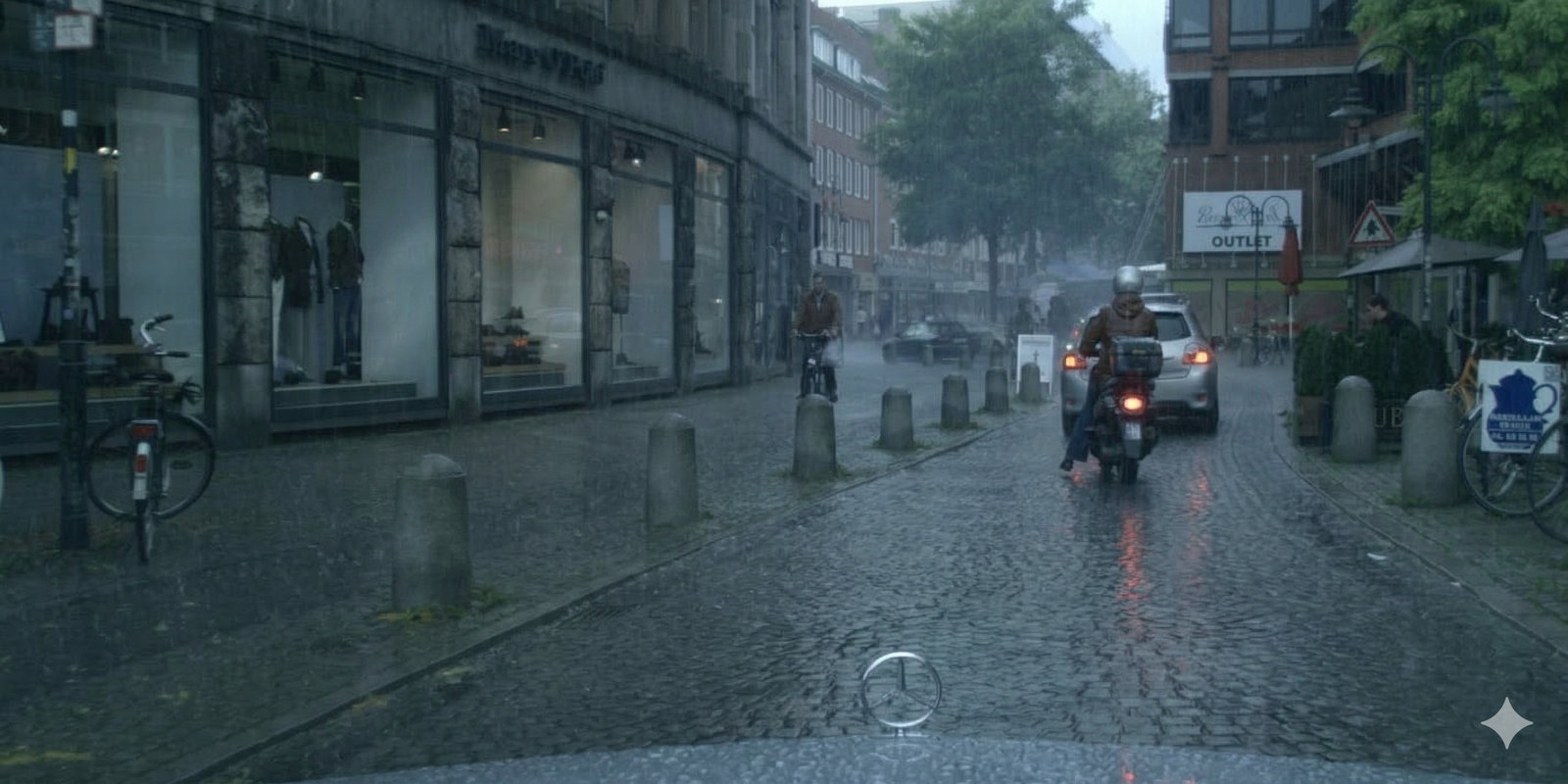} &
\includegraphics[width=\linewidth]{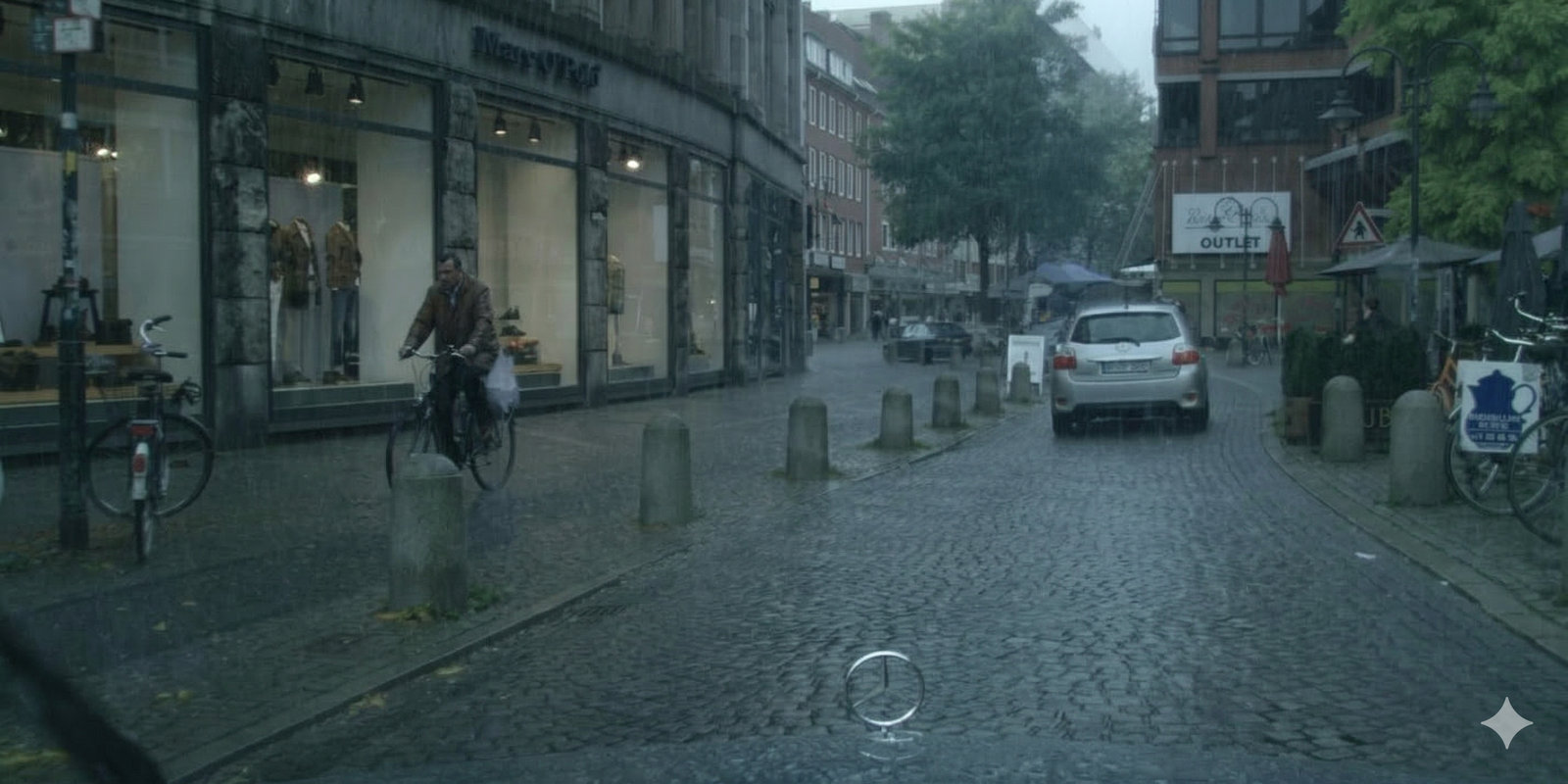}\\[4mm]

\includegraphics[width=\linewidth]{figures/PART2/10/Original.jpg} &
\includegraphics[width=\linewidth]{figures/PART2/10/CascadeRS.jpg} &
\includegraphics[width=\linewidth,height=11.5mm]{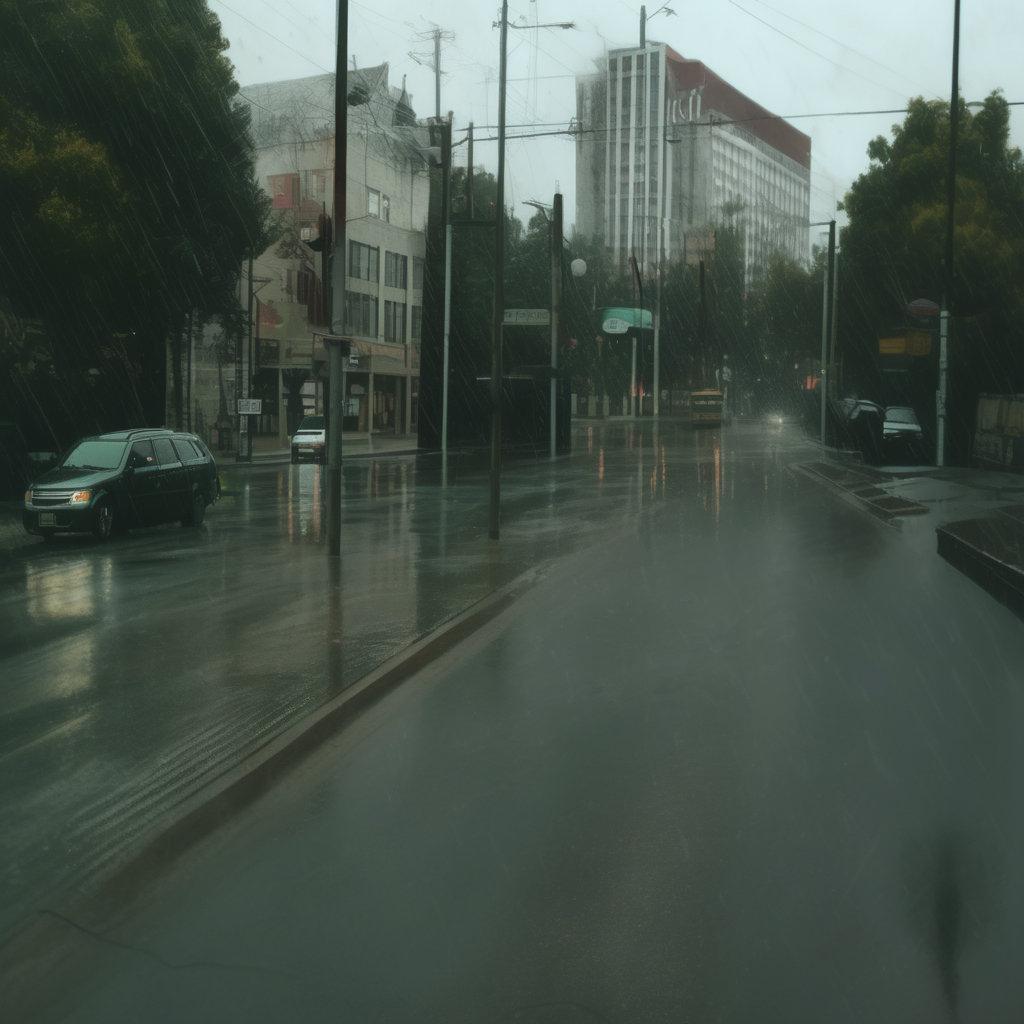} &
\includegraphics[width=\linewidth]{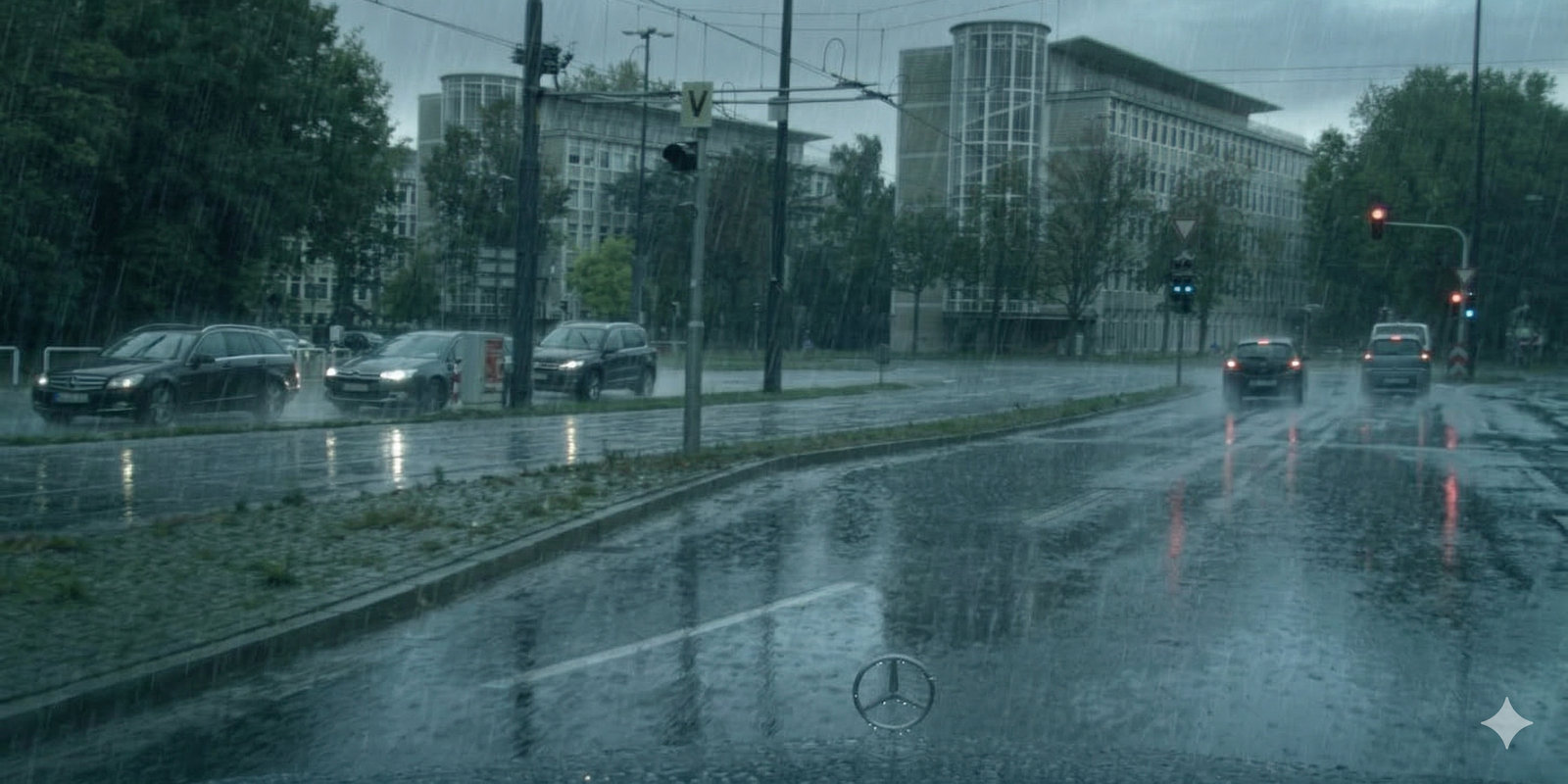} &
\includegraphics[width=\linewidth]{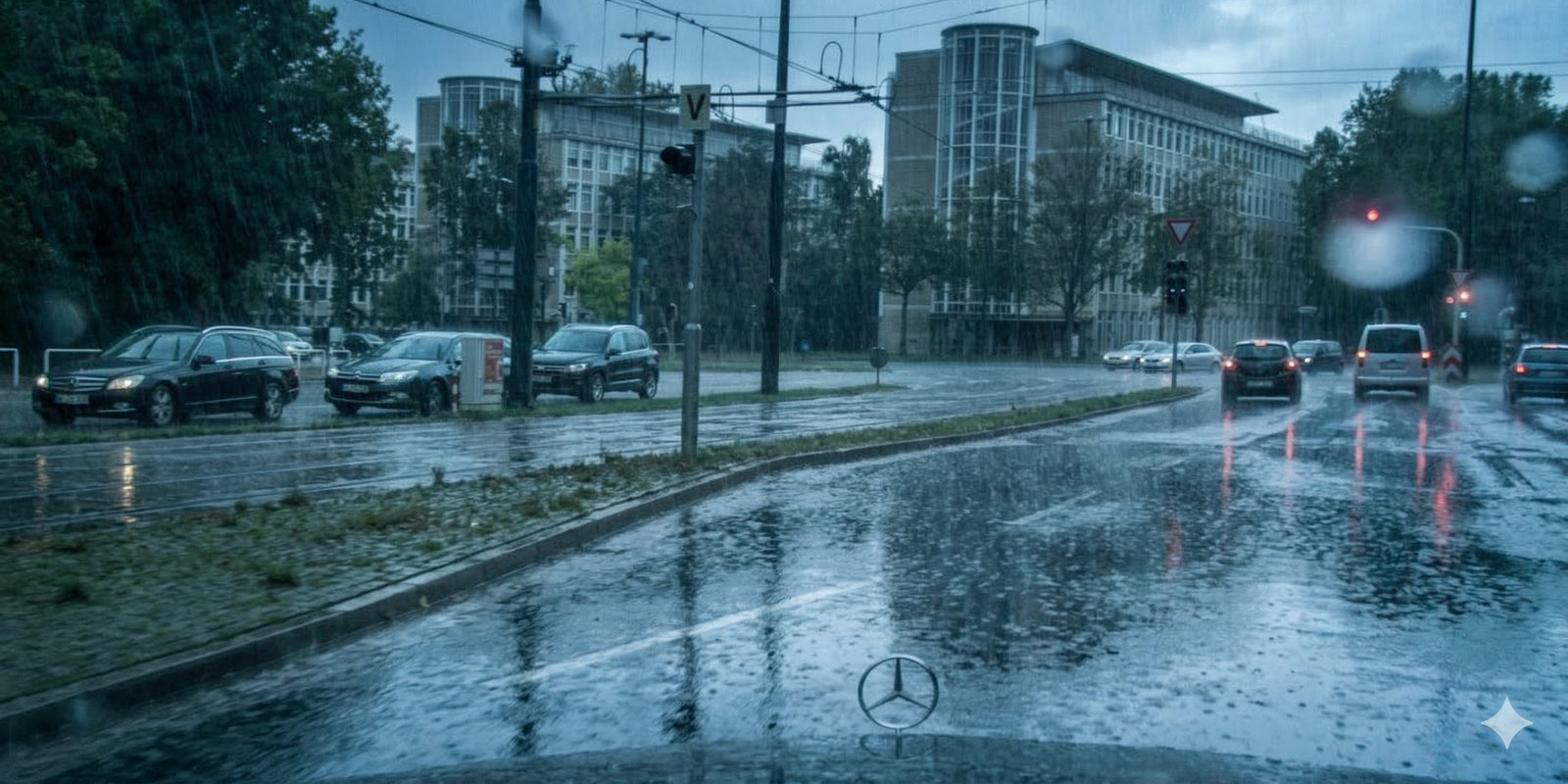}\\[4mm]

\includegraphics[width=\linewidth]{figures/PART2/14/Original.jpg} &
\includegraphics[width=\linewidth]{figures/PART2/14/CascadeRS.jpg} &
\includegraphics[width=\linewidth,height=11.5mm]{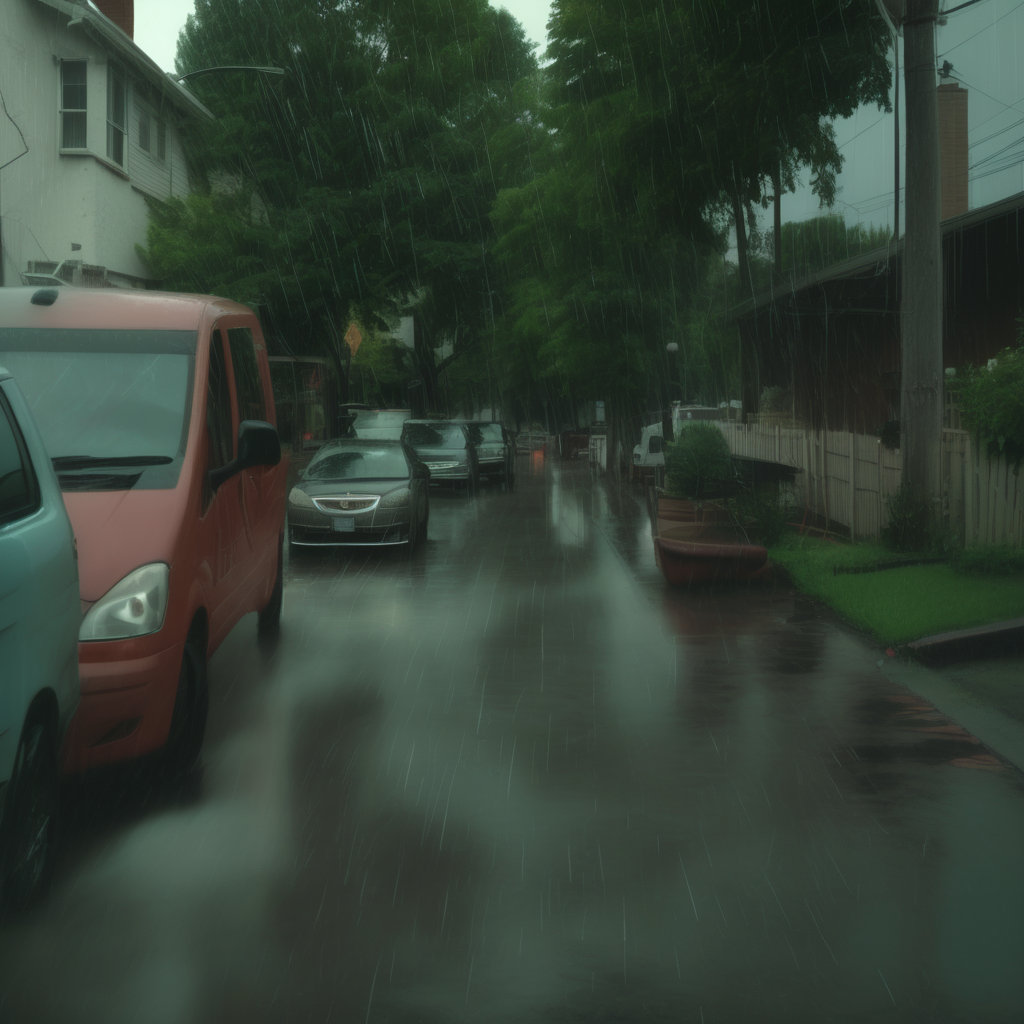} &
\includegraphics[width=\linewidth]{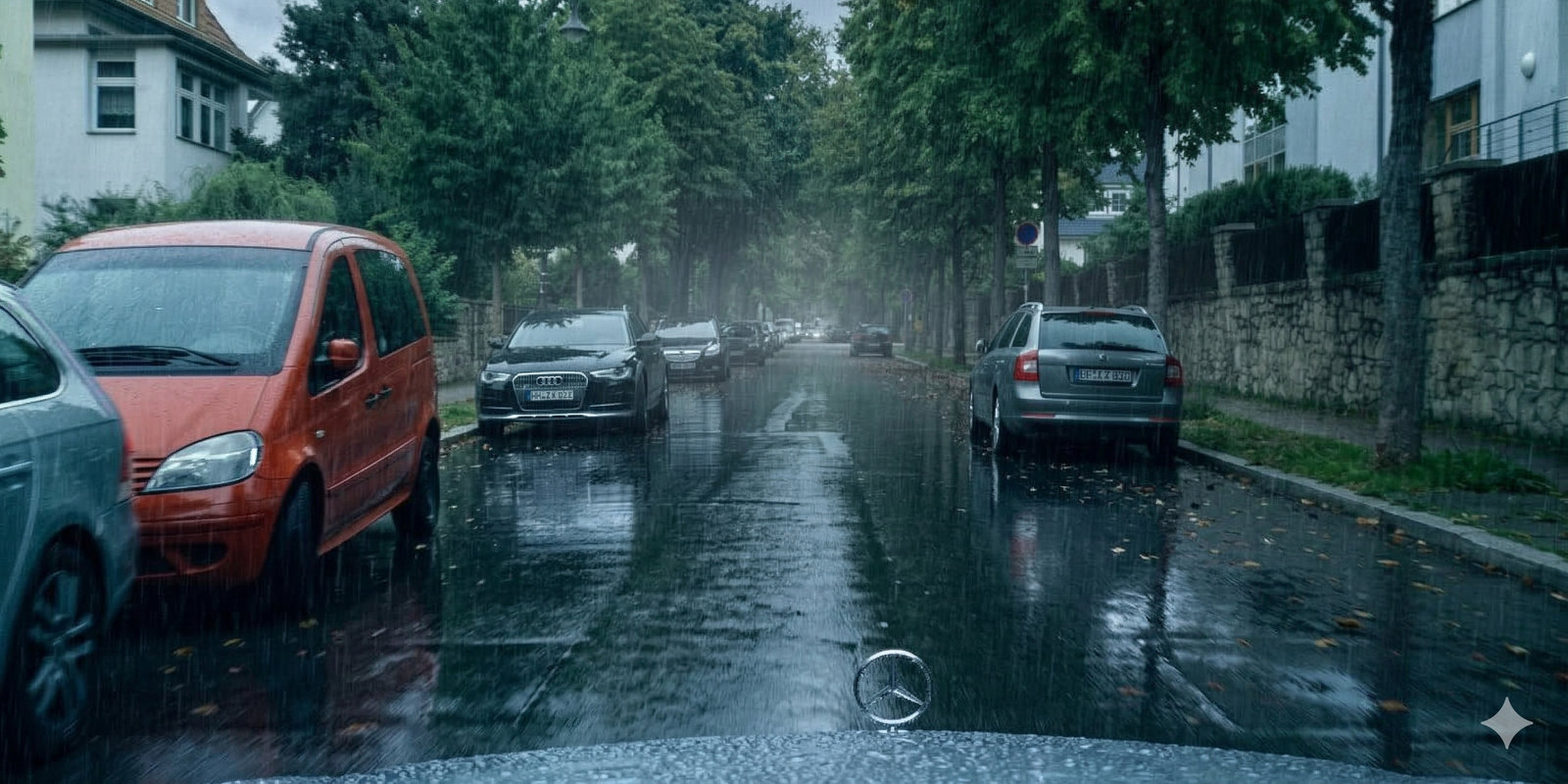} &
\includegraphics[width=\linewidth]{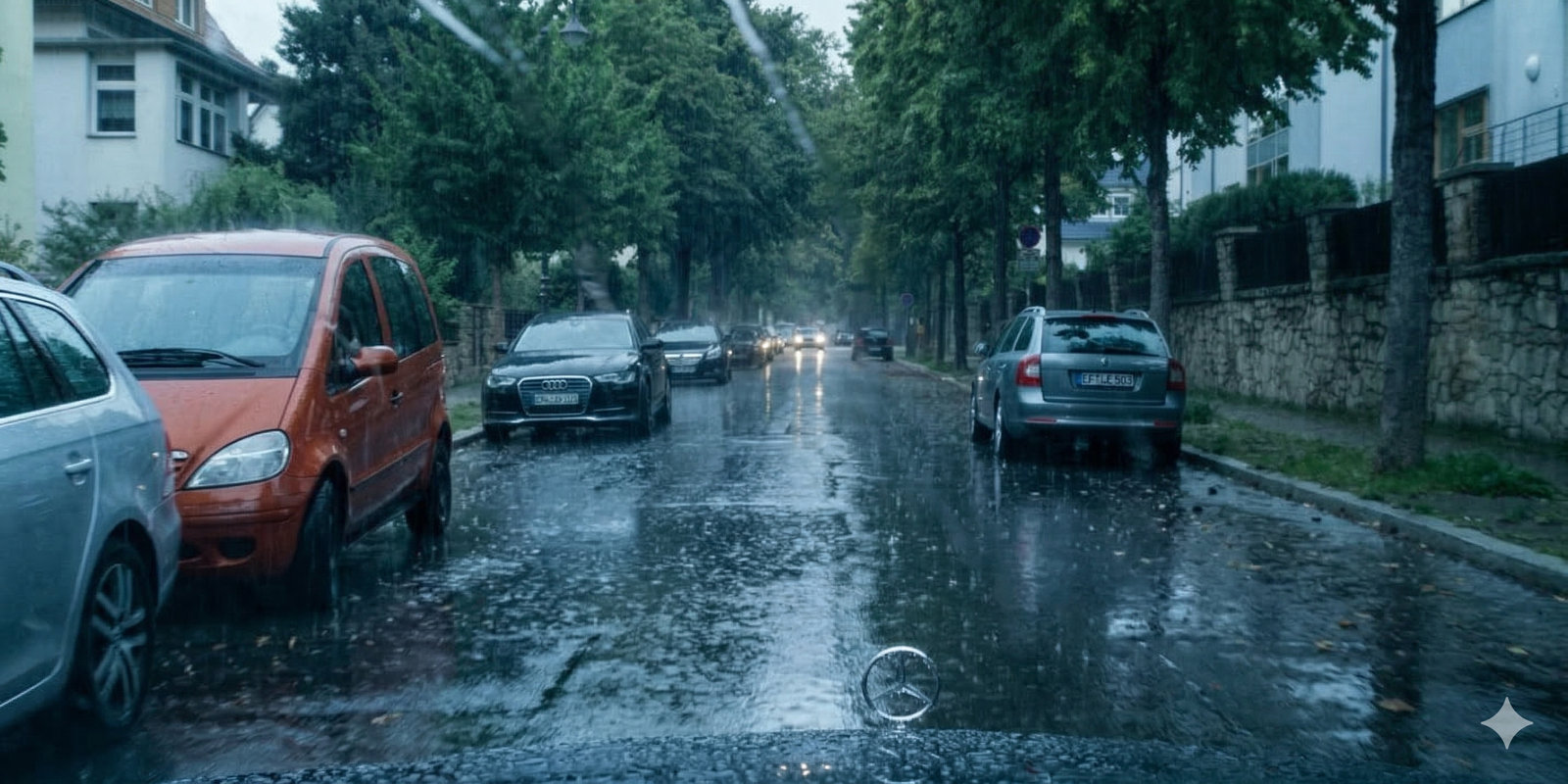} \\[4mm]

\bottomrule
\end{tabular}

\label{tab:rain_methods_genai}
\vspace{-3mm}
\end{figure*}

\section{Rain synthesis with GenAI methods \label{supmat: rain synth with genai}}
We use SDXL and Gemini Nano Banana to synthesize rain weather effects on images from the Cityscapes dataset, and report examples in Figure~\ref{tab:rain_methods_genai}. The rain effects are synthesized on original images using the same text prompt \textit{"A photograph with realistic rain falling, visible rain streaks, wet surfaces reflecting light, an overcast sky, and a cool, atmospheric color palette."} (with strength euql to 0.5 and guidance scale equal to 7.5, for SDXL). While SDXL loses details and oversmooths the appearance of objects in the images, Nano Banana produces more realistic effects although with a low level of controllability. In some cases, it is noticeable how Nano Banana also removes dynamic objects from the scene, which limits the usage of those images as not fully paired with the original iamges for training downstream vision tasks. In all cases, explicitly controlling rain effects properties, such as intensity or direction of the rain is difficult and not easily realizable.


\section{Quantitative Evaluation of Rain Synthesis Methods \label{supmat: quant. eval of rain synth methods}}
Tables~\ref{tab:rain_image_metrics} and \ref{tab:cascaders_effects_fid} report quantiative metrics for comparison of rain synthesis methods. The Fréchet Inception Distance (FID) is computed for all considered rain synthesis methods and evaluated against two benchmarks containing real rainy scenes: ACDC \cite{sakaridis2024acdcadverseconditionsdataset} and RID \cite{Li2019_realrain}. These datasets provide a natural reference distribution for assessing how closely synthetic rain images match real-world images. Baselines such as PBR~\cite{halder2019physicsbased} and Cityscapes-Rain~\cite{DepthRainRemoval} exhibit the highest FID-scores, which is consistent with their limited modeling of surface-level interactions. TPSeNCE~\cite{zheng2023} achieves lower FID due to the inclusion of wetness cues, though its lack of fog and clear streaks limits further gains.

CascadeRS performs competitively across all intensities and rendering variants (Rain streaks, Reflections, and Puddles in Table~\ref{tab:cascaders_effects_fid}), with the lowest overall scores generally observed in the low-to-mid-intensity range (e.g., intensity~30 on RID). At higher intensities, FID increases for all methods. One explanation is that ACDC and RID contain relatively few examples of visually extreme rain. This determines that very heavy synthetic rain generated by CascadeRS, while visually plausible, deviates from the underlying distribution used for FID computation. This aligns with the observed trend that moderate intensities yield the closest match to real data. Even under this distributional mismatch, the CascadeRS \emph{Puddles} configuration achieves the best FID on ACDC and the lowest overall score on RID, suggesting that its combination of streaks, reflections, puddles, and fog aligns well with the statistical characteristics of real rainy scenes. The same result is reflected in the KID (Kernel Inception Distance), where CascadeRS with a light rain intensity achieves lowest overall scores.  These results complement the user study and indicate that CascadeRS maintains strong realism across a range of conditions.

The three content-preservation metrics, Learned Perceptual Image Patch Similarity (LPIPS), Structural Similarity Index Measure (SSIM), and Peak Signal-to-Noise Ratio (PSNR), are included to quantify the extent to which the rain synthesis alters the perceptual and structural characteristics of the original images. Since synthetic rain is deliberately introduced into the scene, some deviation from the original image is expected. The results show that CascadeRS has content-preservation scores comparable to those of the other synthetic rain methods. Moreover, the metrics change progressively with increasing rain intensity: LPIPS increases, while SSIM and PSNR decrease, reflecting the increasing visual difference from the original scene as stronger rain effects are introduced.

\begin{table*}[!t]
\centering
\caption{
\textbf{Comparison of synthetic rain methods using distributional and paired similarity metrics.} Distribution similarity to real rain is evaluated using FID against ACDC~\cite{sakaridis2024acdcadverseconditionsdataset} and RID~\cite{Li2019_realrain}, and KID against ACDC.
Content preservation is evaluated relative to the corresponding original Cityscapes images using LPIPS, SSIM, and PSNR.
Lower FID and KID indicate greater distributional similarity to the real-rain reference dataset, while lower LPIPS and higher SSIM and PSNR indicate greater perceptual and structural similarity to the corresponding original scene.
}
\label{tab:rain_image_metrics}

\renewcommand{\arraystretch}{1.20}
\setlength{\tabcolsep}{4pt}
\scriptsize

\begin{tabular}{llcccccc}
\toprule
Method &
Intensity &
\multicolumn{3}{c}{Distribution similarity} &
\multicolumn{3}{c}{Content preservation} \\
\cmidrule(lr){3-5}
\cmidrule(lr){6-8}
& &
FID $\downarrow$ &
FID $\downarrow$ &
KID $\downarrow$ &
LPIPS $\downarrow$ &
SSIM $\uparrow$ &
PSNR $\uparrow$ \\
& &
\textit{ACDC} &
\textit{RID} &
\textit{ACDC} &
& & \\
\midrule

No Rain
& -- 
& 117.80 & 131.96 & 0.0708
& 0.000 & 1.000 & $\infty$ \\
\midrule

\multirow{5}{*}{PBR~\cite{halder2019physicsbased}}
& 20 mm  & 115.58 & 102.76 & 0.0725 & 0.428 & 0.827 & 21.71 \\
& 40 mm  & 146.61 & 117.35 & 0.1219 & 0.550 & 0.695 & 19.01 \\
& 60 mm  & 194.80 & 154.65 & 0.1962 & 0.621 & 0.586 & 17.61 \\
& 80 mm  & 244.88 & 201.93 & 0.2697 & 0.665 & 0.503 & 16.75 \\
& 100 mm & 299.17 & 260.59 & 0.3435 & 0.708 & 0.421 & 15.92 \\
\midrule

\multirow{3}{*}{CS-Rain~\cite{DepthRainRemoval}}
& Light  & 144.67 & 144.96 & 0.0716 & 0.179 & 0.877 & 19.07 \\
& Medium & 149.51 & 147.46 & 0.0790 & 0.230 & 0.811 & 14.90 \\
& Heavy  & 157.87 & 151.46 & 0.0924 & 0.290 & 0.756 & 12.68 \\
\midrule

TPSeNCE~\cite{zheng2023}
& --
& 111.78 & 113.64 & 0.0608
& 0.308 & 0.840 & 22.08 \\
\midrule

\multirow{10}{*}{CascadeRS}
& 10  & 99.08  & 89.82  & 0.0528 & 0.219 & 0.872 & 20.51 \\
& 20  & 99.60  & 88.69  & 0.0533 & 0.260 & 0.860 & 20.40 \\
& 30  & 101.72 & 87.58  & 0.0562 & 0.300 & 0.841 & 19.88 \\
& 40  & 106.74 & 89.23  & 0.0640 & 0.339 & 0.821 & 19.20 \\
& 50  & 113.42 & 92.08  & 0.0722 & 0.375 & 0.801 & 18.49 \\
& 60  & 122.43 & 95.67  & 0.0844 & 0.408 & 0.781 & 17.77 \\
& 70  & 132.34 & 102.03 & 0.0984 & 0.436 & 0.763 & 17.17 \\
& 80  & 143.39 & 108.02 & 0.1130 & 0.459 & 0.746 & 16.61 \\
& 90  & 155.80 & 116.15 & 0.1300 & 0.481 & 0.730 & 16.10 \\
& 100 & 164.76 & 119.91 & 0.1416 & 0.496 & 0.717 & 15.67 \\

\bottomrule
\end{tabular}
\end{table*}

\begin{table}[!t]
\centering
\caption{
FID comparison of the individual CascadeRS rain effects.
FID is evaluated against real rain from ACDC~\cite{sakaridis2024acdcadverseconditionsdataset} and RID~\cite{Li2019_realrain}.
Lower values indicate greater distributional similarity to real rain.
}
\label{tab:cascaders_effects_fid}

\renewcommand{\arraystretch}{1.20}
\setlength{\tabcolsep}{6pt}
\scriptsize

\begin{tabular}{llcc}
\toprule
Effect &
Intensity &
FID (ACDC) $\downarrow$ &
FID (RID) $\downarrow$ \\
\midrule

\multirow{10}{*}{Streaks}
& 10  & 112.36 & 121.73 \\
& 20  & 110.29 & 113.64 \\
& 30  & 108.71 & 107.41 \\
& 40  & 108.66 & 102.72 \\
& 50  & 111.25 & 100.15 \\
& 60  & 115.27 & 99.94 \\
& 70  & 121.55 & 101.43 \\
& 80  & 128.04 & 104.09 \\
& 90  & 135.83 & 107.63 \\
& 100 & 145.57 & 113.02 \\
\midrule

\multirow{10}{*}{Reflections}
& 10  & 102.11 & 92.26 \\
& 20  & 103.40 & 89.89 \\
& 30  & 105.72 & 88.94 \\
& 40  & 111.37 & 90.14 \\
& 50  & 117.87 & 92.45 \\
& 60  & 125.83 & 96.01 \\
& 70  & 135.57 & 101.90 \\
& 80  & 145.71 & 108.17 \\
& 90  & 157.43 & 115.75 \\
& 100 & 166.66 & 121.67 \\
\midrule

\multirow{10}{*}{Puddles}
& 10  & 99.08  & 89.82 \\
& 20  & 99.60  & 88.69 \\
& 30  & 101.72 & 87.58 \\
& 40  & 106.74 & 89.23 \\
& 50  & 113.42 & 92.08 \\
& 60  & 122.43 & 95.67 \\
& 70  & 132.34 & 102.03 \\
& 80  & 143.39 & 108.02 \\
& 90  & 155.80 & 116.15 \\
& 100 & 164.76 & 119.91 \\

\bottomrule
\end{tabular}
\end{table}

\begin{table*}[!t]
\centering
\scriptsize
\setlength{\tabcolsep}{2.2pt}
\renewcommand{\arraystretch}{1.20}

\caption{\textbf{Cross-dataset segmentation comparison.}
Per-class IoU on ACDC \textit{train} and Cityscapes \textit{validation} sets for
SegFormer-B2~\cite{xie2021segformersimpleefficientdesign} and
GCNet~\cite{yang2025goldencudgel}, trained on Cityscapes (CS) with and without
CascadeRS augmentation. Mean and standard deviation over five random seeds are
reported separately for SegFormer-B2; GCNet results are from a single training
run. Classes are ordered by their occurrence in ACDC; the last column lists
their relative frequency in the Cityscapes training set. The three largest
improvements on ACDC are shown in \textbf{bold}, and the two largest reductions
are \underline{underlined}.}
\label{tab:crossdataset}
\vspace{0.4em}

\begin{tabular}{l|ccccc|ccccc|c}
\hline

\multirow{3}{*}{\textbf{Metric / Class}} &
\multicolumn{5}{c|}{\textbf{ACDC (Train)}} &
\multicolumn{5}{c|}{\textbf{Cityscapes (Validation)}} &
\multirow{2}{*}{\textbf{CS Train}} \\

&
\multirow{2}{*}{\textbf{Occ. [\%]}} &
\multirow{2}{*}{\textbf{CS}} &
\multicolumn{2}{c}{\textbf{CascadeRS}} &
\multirow{2}{*}{\textbf{$\Delta$}} &
\multirow{2}{*}{\textbf{Occ. [\%]}} &
\multirow{2}{*}{\textbf{CS}} &
\multicolumn{2}{c}{\textbf{CascadeRS}} &
\multirow{2}{*}{\textbf{$\Delta$}} &
\multirow{2}{*}{\textbf{Occ. [\%]}} \\

& & & \textbf{Mean} & \textbf{Std.} &
& & & \textbf{Mean} & \textbf{Std.} & & \\


\hline
\multicolumn{12}{c}{\textbf{(a) SegFormer-B2}} \\
\hline

mIoU $\uparrow$
& -- & 0.4259 & 0.4215 & 0.0056 & $-$0.0044
& -- & 0.6813 & 0.6705 & 0.0036 & $-$0.0108 & -- \\

FWIoU $\uparrow$
& -- & 0.7824 & 0.8003 & 0.0036 & +0.0179
& -- & 0.9034 & 0.9004 & 0.0007 & $-$0.0030 & -- \\

\hline

Sky
& 28.35 & 0.9713 & 0.9550 & 0.0065 & $-$0.0163
& 3.35 & 0.9432 & 0.9415 & 0.0008 & $-$0.0017 & 4.019 \\

Vegetation
& 22.54 & 0.8799 & 0.8716 & 0.0058 & $-$0.0083
& 17.32 & 0.9116 & 0.9099 & 0.0003 & $-$0.0017 & 15.929 \\

\textbf{Road}
& 14.81 & 0.6407 & 0.7455 & 0.0136 & \textbf{+0.1048}
& 37.65 & 0.9746 & 0.9729 & 0.0007 & $-$0.0017 & 36.870 \\

Building
& 20.32 & 0.8016 & 0.8074 & 0.0145 & +0.0058
& 21.92 & 0.8990 & 0.8961 & 0.0010 & $-$0.0029 & 22.824 \\

\textbf{Sidewalk}
& 5.74 & 0.3576 & 0.4871 & 0.0115 & \textbf{+0.1295}
& 5.41 & 0.8037 & 0.7953 & 0.0021 & $-$0.0084 & 6.085 \\

\textbf{Car}
& 1.84 & 0.5613 & 0.6284 & 0.0218 & \textbf{+0.0671}
& 6.51 & 0.9211 & 0.9172 & 0.0013 & $-$0.0039 & 6.995 \\

Terrain
& 1.48 & 0.5178 & 0.4743 & 0.0214 & $-$0.0435
& 0.83 & 0.6239 & 0.6217 & 0.0095 & $-$0.0022 & 1.158 \\

Wall
& 1.40 & 0.1543 & 0.1945 & 0.0129 & +0.0402
& 0.73 & 0.5432 & 0.5177 & 0.0185 & $-$0.0255 & 0.655 \\

Fence
& 1.22 & 0.1652 & 0.1713 & 0.0176 & +0.0061
& 0.82 & 0.4552 & 0.4594 & 0.0117 & +0.0042 & 0.877 \\

Pole
& 1.06 & 0.4041 & 0.3481 & 0.0104 & $-$0.0560
& 1.48 & 0.5256 & 0.5101 & 0.0024 & $-$0.0155 & 1.227 \\

Traffic sign
& 0.29 & 0.4020 & 0.3967 & 0.0167 & $-$0.0053
& 0.67 & 0.6411 & 0.6341 & 0.0047 & $-$0.0070 & 0.551 \\

Traffic light
& 0.19 & 0.3759 & 0.3537 & 0.0112 & $-$0.0222
& 0.20 & 0.5363 & 0.5297 & 0.0042 & $-$0.0066 & 0.208 \\

\underline{Bus}
& 0.23 & 0.3487 & 0.2155 & 0.0489 & \underline{$-$0.1332}
& 0.39 & 0.7046 & 0.6737 & 0.0373 & $-$0.0309 & 0.235 \\

Truck
& 0.14 & 0.0162 & 0.0404 & 0.0209 & +0.0242
& 0.30 & 0.5861 & 0.5939 & 0.0517 & +0.0078 & 0.267 \\

Train
& 0.20 & 0.4140 & 0.3927 & 0.0663 & $-$0.0213
& 0.11 & 0.6004 & 0.5651 & 0.0238 & $-$0.0353 & 0.233 \\

\underline{Bicycle}
& 0.08 & 0.3993 & 0.2913 & 0.0210 & \underline{$-$0.1080}
& 0.71 & 0.6640 & 0.6573 & 0.0033 & $-$0.0067 & 0.414 \\

Person
& 0.07 & 0.1390 & 0.0929 & 0.0096 & $-$0.0461
& 1.30 & 0.7090 & 0.7058 & 0.0019 & $-$0.0032 & 1.219 \\

Motorcycle
& 0.05 & 0.1530 & 0.1727 & 0.0783 & +0.0197
& 0.08 & 0.4505 & 0.4059 & 0.0268 & $-$0.0446 & 0.099 \\

Rider
& 0.01 & 0.3899 & 0.3698 & 0.0716 & $-$0.0201
& 0.22 & 0.4515 & 0.4314 & 0.0067 & $-$0.0201 & 0.135 \\


\hline
\multicolumn{12}{c}{\textbf{(b) GCNet}} \\
\hline

mIoU $\uparrow$
& -- & 0.3546 & 0.3925 & -- & +0.0379
& -- & 0.7156 & 0.7031 & -- & $-$0.0125 & -- \\

FWIoU $\uparrow$
& -- & 0.7241 & 0.7861 & -- & +0.0620
& -- & 0.9106 & 0.9075 & -- & $-$0.0031 & -- \\

\hline

Sky
& 28.35 & 0.9224 & 0.9464 & -- & +0.0241
& 3.35 & 0.9387 & 0.9381 & -- & $-$0.0005 & 4.019 \\

Vegetation
& 22.54 & 0.8252 & 0.8391 & -- & +0.0139
& 17.32 & 0.9119 & 0.9096 & -- & $-$0.0023 & 15.929 \\

\textbf{Road}
& 14.81 & 0.5931 & 0.7866 & -- & \textbf{+0.1935}
& 37.65 & 0.9768 & 0.9763 & -- & $-$0.0004 & 36.870 \\

Building
& 20.32 & 0.7259 & 0.7655 & -- & +0.0396
& 21.92 & 0.9080 & 0.9044 & -- & $-$0.0035 & 22.824 \\

\textbf{Sidewalk}
& 5.74 & 0.2530 & 0.4873 & -- & \textbf{+0.2344}
& 5.41 & 0.8197 & 0.8143 & -- & $-$0.0054 & 6.085 \\

Car
& 1.84 & 0.5571 & 0.5616 & -- & +0.0044
& 6.51 & 0.9291 & 0.9242 & -- & $-$0.0048 & 6.995 \\

Terrain
& 1.48 & 0.5140 & 0.5046 & -- & $-$0.0094
& 0.83 & 0.5910 & 0.6053 & -- & +0.0144 & 1.158 \\

\textbf{Wall}
& 1.40 & 0.1052 & 0.2431 & -- & \textbf{+0.1379}
& 0.73 & 0.5432 & 0.5031 & -- & $-$0.0401 & 0.655 \\

\underline{Fence}
& 1.22 & 0.1519 & 0.1092 & -- & \underline{$-$0.0427}
& 0.82 & 0.5333 & 0.5088 & -- & $-$0.0245 & 0.877 \\

Pole
& 1.06 & 0.2056 & 0.2181 & -- & +0.0125
& 1.48 & 0.5687 & 0.5541 & -- & $-$0.0145 & 1.227 \\

Traffic sign
& 0.29 & 0.4035 & 0.4483 & -- & +0.0448
& 0.67 & 0.7072 & 0.6683 & -- & $-$0.0390 & 0.551 \\

Traffic light
& 0.19 & 0.4299 & 0.4704 & -- & +0.0406
& 0.20 & 0.6063 & 0.5952 & -- & $-$0.0111 & 0.208 \\

Bus
& 0.23 & 0.2689 & 0.3323 & -- & +0.0634
& 0.39 & 0.7336 & 0.7265 & -- & $-$0.0071 & 0.235 \\

Truck
& 0.14 & 0.0419 & 0.0983 & -- & +0.0564
& 0.30 & 0.6479 & 0.6793 & -- & +0.0314 & 0.267 \\

Train
& 0.20 & 0.2658 & 0.2554 & -- & $-$0.0104
& 0.11 & 0.6610 & 0.6346 & -- & $-$0.0264 & 0.233 \\

\underline{Bicycle}
& 0.08 & 0.2288 & 0.1880 & -- & \underline{$-$0.0409}
& 0.71 & 0.7032 & 0.6839 & -- & $-$0.0194 & 0.414 \\

Person
& 0.07 & 0.0239 & 0.0297 & -- & +0.0058
& 1.30 & 0.7569 & 0.7506 & -- & $-$0.0063 & 1.219 \\

Motorcycle
& 0.05 & 0.0352 & 0.0180 & -- & $-$0.0172
& 0.08 & 0.5172 & 0.4482 & -- & $-$0.0690 & 0.099 \\

Rider
& 0.01 & 0.1856 & 0.1551 & -- & $-$0.0305
& 0.22 & 0.5432 & 0.5342 & -- & $-$0.0089 & 0.135 \\

\hline
\end{tabular}

\end{table*}
\section{Cross-Dataset Segmentation Analysis \label{supmat: cross-dataset segm analys.}}
Table~\ref{tab:crossdataset} reports the per-class results for the cross-dataset semantic segmentation experiment. We evaluate two segmentation architectures, SegFormer-B2~\cite{xie2021segformersimpleefficientdesign} and GCNet~\cite{yang2025goldencudgel}. Both models are trained on Cityscapes~\cite{Cityscapes}, augmenting 30\% of the training images (randomly selected) with CascadeRS-generated rainy versions, using ten rain intensity values in the range $[10, 20, \dots, 100]$. To account for variability introduced by training, we train the CascadeRS-augmented SegFormer-B2 model five times using different random seeds and report the mean and standard deviation across these runs. GCNet results are obtained from a single training run. The models are evaluated on real rainy scenes from ACDC~\cite{sakaridis2024acdcadverseconditionsdataset} and on the clean Cityscapes validation set.

We report mean the Intersection-over-Union (mIoU) and the frequency-weighted IoU (FWIoU). Let $p_{ij}$ denote the number of pixels of ground-truth class $i$ predicted as class $j$, $g_i=\sum_j p_{ij}$ the total number of pixels belonging to class $i$, and  $n=\sum_i g_i$ the total number of labeled pixels: using the per-class Intersection-over-Union $\mathrm{IoU}_i$, the metrics are defined as:

\begin{equation}
    \mathrm{mIoU} = \frac{1}{K}\sum_{i=1}^{K} \mathrm{IoU}_i,
\end{equation}

\begin{equation}
    \mathrm{FWIoU} = \sum_{i=1}^{K} \frac{g_i}{n}\,\mathrm{IoU}_i,
\end{equation}

where $K$ is the number of semantic classes. Because both ACDC and Cityscapes exhibit strong class imbalance, FWIoU provides an important complementary view that stabilizes the contribution of high-frequency classes, while mIoU highlights behavior on rare categories.

For SegFormer-B2, CascadeRS augmentation increases ACDC FWIoU from 0.7824 to $0.8003 \pm 0.0036$, corresponding to an improvement of 0.0179. The mIoU changes from 0.4259 to $0.4215 \pm 0.0056$, a small mean difference of $-$0.0044 that falls within one standard deviation across the five training runs. On the clean Cityscapes validation set, mIoU decreases from 0.6813 to $0.6705 \pm 0.0036$, while FWIoU remains nearly unchanged, from 0.9034 to $0.9004 \pm 0.0007$. 

GCNet shows a stronger improvement on real rain. On ACDC, mIoU increases from 0.3546 to 0.3925 (+0.0379), while FWIoU increases from 0.7241 to 0.7861 (+0.0620). In comparison, performance on clean Cityscapes changes only moderately: mIoU decreases from 0.7156 to 0.7031 ($-$0.0125), while FWIoU changes from 0.9106 to 0.9075 ($-$0.0031). Thus, for both architectures, CascadeRS augmentation improves frequency-weighted performance on real rainy scenes while largely retaining performance on clean scenes.

The per-class results reveal a consistent benefit for ground-related categories across the two architectures. For SegFormer-B2, the three largest gains on ACDC are observed for \textit{sidewalk} (+0.1295 IoU), \textit{road} (+0.1048 IoU), and \textit{car} (+0.0671 IoU). For GCNet, the largest gains occur for \textit{sidewalk} (+0.2344 IoU), \textit{road} (+0.1935 IoU), and \textit{wall} (+0.1379 IoU). In particular, the substantial improvements for \textit{road} and \textit{sidewalk} across both architectures suggest that the observed effect is not specific to SegFormer-B2. These classes are strongly affected by wet surfaces, reflections, puddles, and other rain-induced appearance changes modeled by CascadeRS.

For both segmentation architectures, large homogeneous regions such as \textit{sky} and \textit{vegetation} show relatively small changes after CascadeRS augmentation. For SegFormer-B2, their IoU changes by $-$0.0163 and $-$0.0083, respectively, while for GCNet the changes are +0.0241 and +0.0139. These classes occupy large spatial extents in ACDC (28.35\% and 22.54\% of pixels, respectively). One possible explanation is that their large, relatively homogeneous appearance may make them less sensitive to rain-related changes, and therefore less likely to benefit substantially from rain augmentation. Because these large and relatively stable classes contribute substantially to FWIoU, the observed FWIoU gains appear to be primarily driven by improvements in other frequently occurring classes, particularly ground-related categories.

For SegFormer-B2, the largest performance drops on ACDC occur for \textit{bus} ($-$0.1332 IoU) and \textit{bicycle} ($-$0.1080 IoU), which represent only 0.235\% and 0.414\% of Cityscapes training pixels, respectively. Other low-frequency classes also exhibit considerable variability across seeds; for example, \textit{motorcycle}, \textit{rider}, and \textit{train} have standard deviations of 0.0783, 0.0716, and 0.0663 IoU, respectively. For GCNet, the largest reductions are smaller and occur for \textit{fence} ($-$0.0427) and \textit{bicycle} ($-$0.0409). Overall, these results suggest that performance changes for low-frequency categories can be less stable. Since mIoU weights all classes equally, changes in these rare categories have a larger influence on mIoU than on FWIoU. This helps explain the stronger improvements observed in FWIoU, particularly for SegFormer-B2, where FWIoU increases even as mIoU decreases slightly.

On the original Cityscapes validation set (clear weather), most high-frequency classes remain stable for both architectures. For SegFormer-B2, \textit{road}, \textit{vegetation}, \textit{building}, and \textit{car} change by only $-$0.0017, $-$0.0017, $-$0.0029, and $-$0.0039 IoU, respectively, with small standard deviations between 0.0003 and 0.0013 across the five training runs. Similarly, for GCNet, these classes change by $-$0.0004, $-$0.0023, $-$0.0035, and $-$0.0048 IoU. Consequently, the small changes in Cityscapes FWIoU ($-$0.0030 for SegFormer-B2 and $-$0.0031 for GCNet) indicate that performance on the dominant classes is largely retained after introducing CascadeRS augmentation.


Overall, CascadeRS augmentation improves segmentation robustness on real rainy scenes across both segmentation architectures, with consistent gains for road-surface and ground-related classes. The multi-seed SegFormer-B2 results additionally reveal greater variability for rare categories, while performance on clean-weather scenes remains largely preserved for both models. These findings support the conclusion that physically grounded, perspective- and semantics-aware rain synthesis can offer meaningful benefits for downstream vision robustness across different segmentation architectures.

\section{Object Detection Analysis \label{supmat: object detection}}
To evaluate object detection, we use the DINO Teacher model \cite{lavoie2025large}, pretrained on the Cityscapes dataset to detect eight object classes: person, truck, motorcycle, rider, bus, bicycle, car, and train. The model uses a frozen DINOv2 backbone; specifically, we use the ViT-L backbone, which achieves an AP$_{50}$ of 61.3 on Cityscapes. We evaluate object detection on Cityscapes, despite it primarily being a semantic segmentation dataset, because its scenes and annotation format are similar to those of the ACDC dataset \cite{sakaridis2024acdcadverseconditionsdataset}, which contains real rainy driving scenes. Table~\ref{tab:rain_detection_performance} shows the performance of the pretrained DINO Teacher model on different synthetic rainy datasets, with ACDC and clean Cityscapes included as references.

For CascadeRS, detection performance generally decreases with increasing rain intensity, indicating that heavier synthesized rain increasingly degrades object detection. Considering the commonly reported AP and AP$_{50}$ metrics, CascadeRS at higher rain intensities (80--100) and PBR at medium intensity achieve performance similar to that observed on real rain. In particular, CascadeRS at intensity 100 achieves an AP of 19.9, close to the 19.1 AP obtained on ACDC. Considering performance across object sizes, the closest correspondence with ACDC is obtained at different rain intensities: CascadeRS at intensity 40 approaches the AP$_s$ of ACDC, while intensity 60 closely matches its AP$_m$. For large objects, performance remains higher than on ACDC, even at the strongest CascadeRS rain intensity. Overall, these results show that increasing the rain intensity in CascadeRS produces a gradual shift in detection performance toward that observed under real rainy conditions, rather than an abrupt degradation.

\begin{table*}[!t] 
\centering 
\caption{\textbf{Object detection performance under rainy conditions.} Performance of the pretrained DINO Teacher model \cite{lavoie2025large} on real and synthetic rainy images. Results are reported using COCO AP metrics.} 
\label{tab:rain_detection_performance} 
\renewcommand{\arraystretch}{1.12} 
\setlength{\tabcolsep}{5pt} 
\scriptsize 
\begin{tabular}{llrrrrrr} 
\toprule \textbf{Method} & \textbf{Intensity} & \textbf{AP} & \textbf{AP$_{50}$} & \textbf{AP$_{75}$} & \textbf{AP$_s$} & \textbf{AP$_m$} & \textbf{AP$_l$} \\ 
\midrule ACDC (real rain) & Reference & 19.0782 & 41.1327 & 15.1855 & 3.7629 & 21.4794 & 41.7014 \\ 
Cityscapes & Clean & 32.1823 & 61.2716 & 30.4232 & 5.8718 & 30.9993 & 57.3393 \\
\midrule \multirow{10}{*}{CascadeRS} & 10 & 30.8019 & 59.6321 & 27.9950 & 5.3069 & 28.9772 & 55.4775 \\ 
& 20 & 29.7713 & 56.8627 & 28.9478 & 4.6032 & 26.9596 & 55.5098 \\
& 30 & 28.6653 & 54.8198 & 27.3550 & 4.0520 & 25.6022 & 54.0948 \\ 
& 40 & 27.3669 & 52.9954 & 24.2306 & 3.3321 & 24.1872 & 52.7352 \\ 
& 50 & 26.3697 & 50.5434 & 24.8679 & 2.8203 & 23.0049 & 52.2393 \\ 
& 60 & 25.2625 & 48.6628 & 23.6714 & 2.4927 & 21.3783 & 51.5903 \\ 
& 70 & 23.8791 & 44.9074 & 22.2546 & 1.9817 & 19.7072 & 50.9353 \\ 
& 80 & 22.2288 & 42.1751 & 20.7218 & 1.3707 & 17.4095 & 48.8865 \\ 
& 90 & 21.2932 & 41.1711 & 19.8807 & 1.1397 & 16.2526 & 48.2297 \\ 
& 100 & 19.8647 & 39.1930 & 17.7936 & 1.1214 & 14.1573 & 46.8020 \\ 
\midrule TPSeNCE & -- & 27.9949 & 54.3466 & 25.0665 & 4.0987 & 24.8023 & 54.1581 \\ \midrule \multirow{5}{*}{PBR} & 20 & 28.0241 & 54.6321 & 24.4086 & 4.8630 & 25.7020 & 53.8531 \\ 
& 40 & 24.4029 & 49.5736 & 20.9074 & 2.7467 & 21.3865 & 49.7195 \\ 
& 60 & 20.5142 & 42.2528 & 17.3787 & 1.7456 & 15.1857 & 45.3921 \\ 
& 80 & 16.4448 & 35.1255 & 13.5973 & 1.0665 & 10.6226 & 39.9877 \\ 
& 100 & 11.6607 & 24.9516 & 9.7240 & 0.6110 & 5.8049 & 29.4542 \\ 
\midrule \multirow{3}{*}{Cityscapes-Rain} & Light & 35.6371 & 65.2971 & 34.1892 & 13.2824 & 41.0203 & 62.4778 \\ 
& Medium & 34.8440 & 61.9113 & 35.4455 & 12.0518 & 39.5374 & 65.1275 \\ 
& Heavy & 29.7464 & 56.8541 & 28.8734 & 6.5435 & 32.1306 & 61.3575 \\ 
\bottomrule 
\end{tabular} 
\end{table*}

\section{Qualitative Results Across Multiple Datasets \label{supmat: qual. results across datasets}}
Figure~\ref{tab:dataset_renders} showcases the visual consistency of CascadeRS when applied to images from four different, widely used autonomous-driving datasets: Cityscapes~\cite{Cityscapes}, BDD100K~\cite{yu2020bdd100kdiversedrivingdataset}, KITTI~\cite{Geiger2013IJRR}, and Roadmarks~\cite{Wu2012Roadmarks}. Despite substantial variation in scene layout, illumination, camera geometry, and image quality across these datasets, CascadeRS produces coherent rain streaks, wet-surface reflections, and puddles that remain aligned with local depth and semantic structure. The examples illustrate that our method generalizes well beyond Cityscapes, without tuning or adapting its inner parameters, generating physically meaningful rain effects even in nighttime conditions (BDD100K), wide-baseline highway images (KITTI), and low-texture road-surface data (Roadmarks). This demonstrates that CascadeRS can be deployed as a dataset-agnostic augmentation tool without requiring dataset-specific tuning.

\begin{figure*}[!t]
\centering
\renewcommand{\arraystretch}{1.9}
\setlength{\tabcolsep}{2pt}
\caption{\textbf{Qualitative results of rain synthesis across different datasets.} Each row shows four example renderings from our CascadeRS pipeline for the given dataset.}
\scriptsize

\begin{tabular}{C{1.7cm} C{0.2\linewidth} C{0.2\linewidth} C{0.2\linewidth} C{0.2\linewidth}}
\toprule
\textbf{Dataset} & \multicolumn{4}{c}{\textbf{Rendered Examples}} \\
\midrule

\multirow{2}{=}{\centering \textbf{Cityscapes}} &
\includegraphics[width=\linewidth]{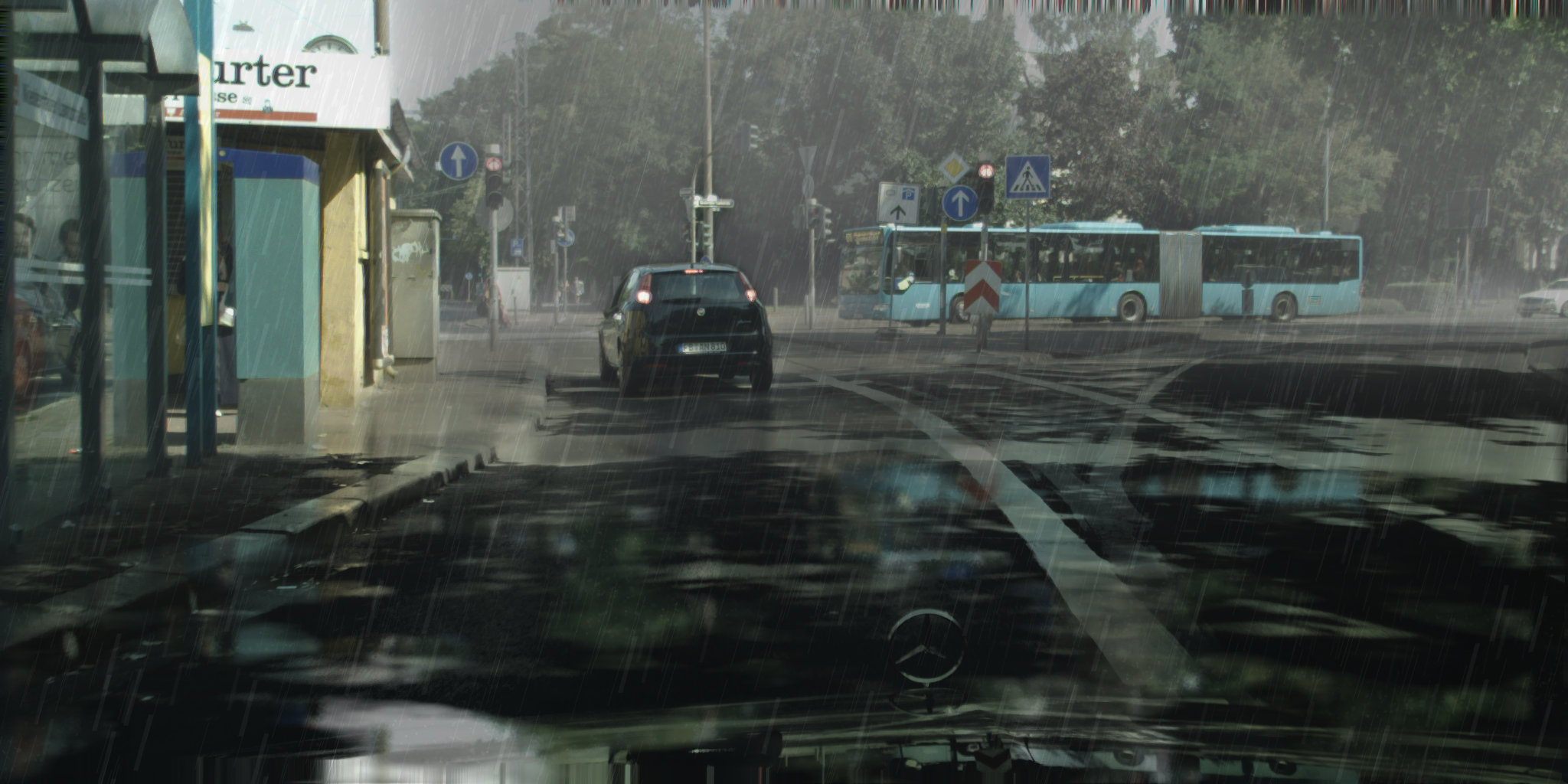} &
\includegraphics[width=\linewidth]{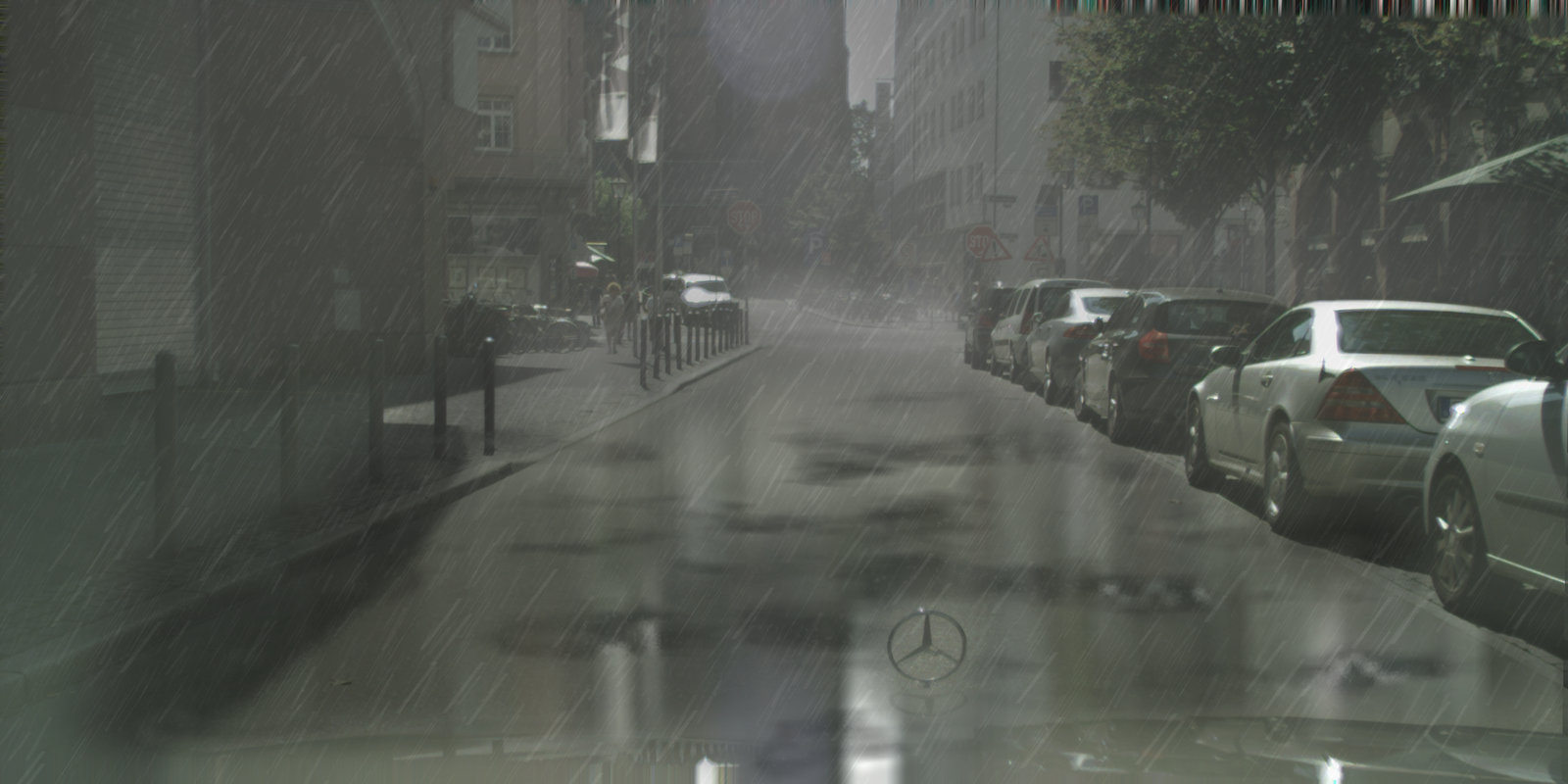} &
\includegraphics[width=\linewidth]{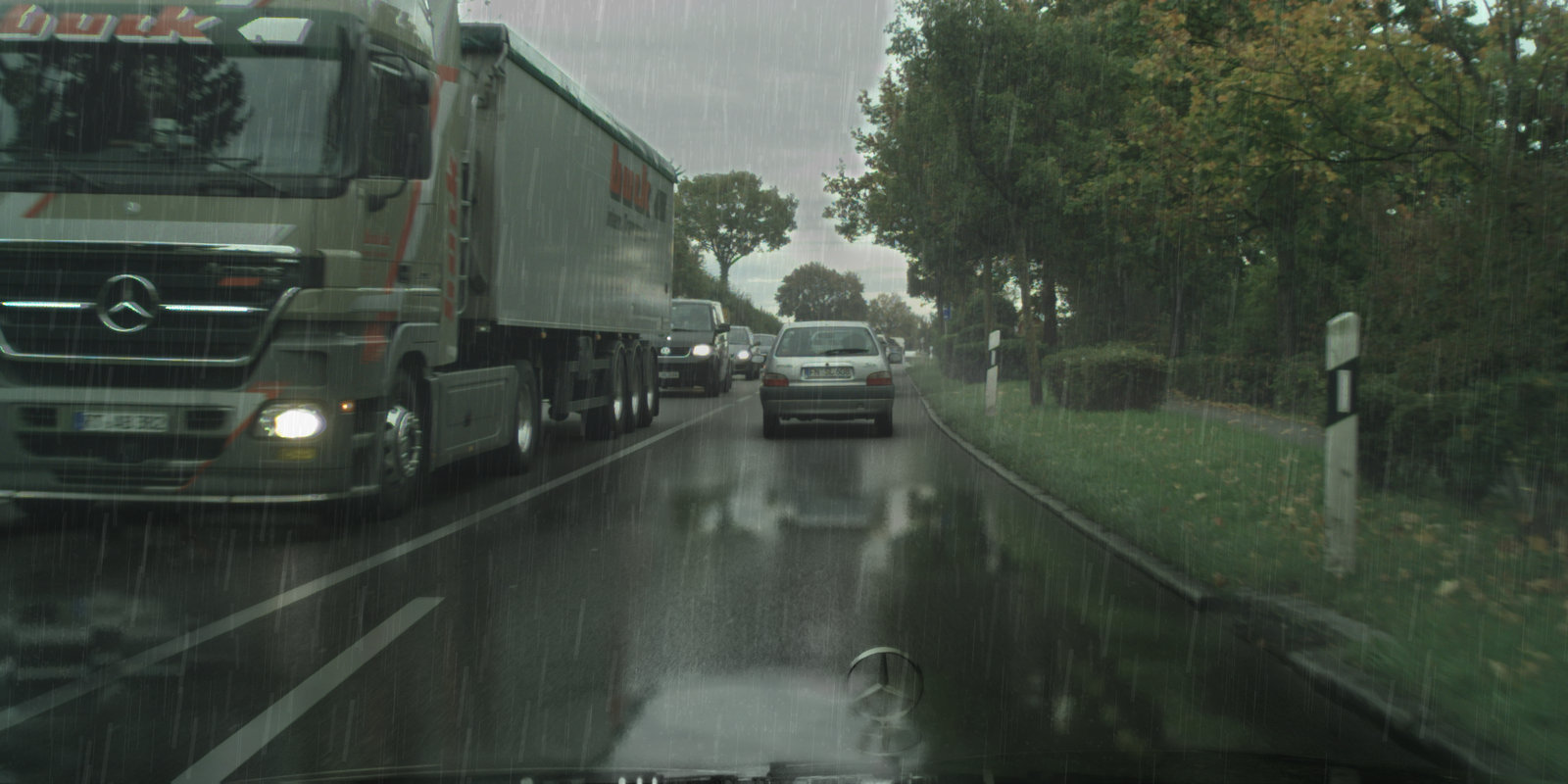} &
\includegraphics[width=\linewidth]{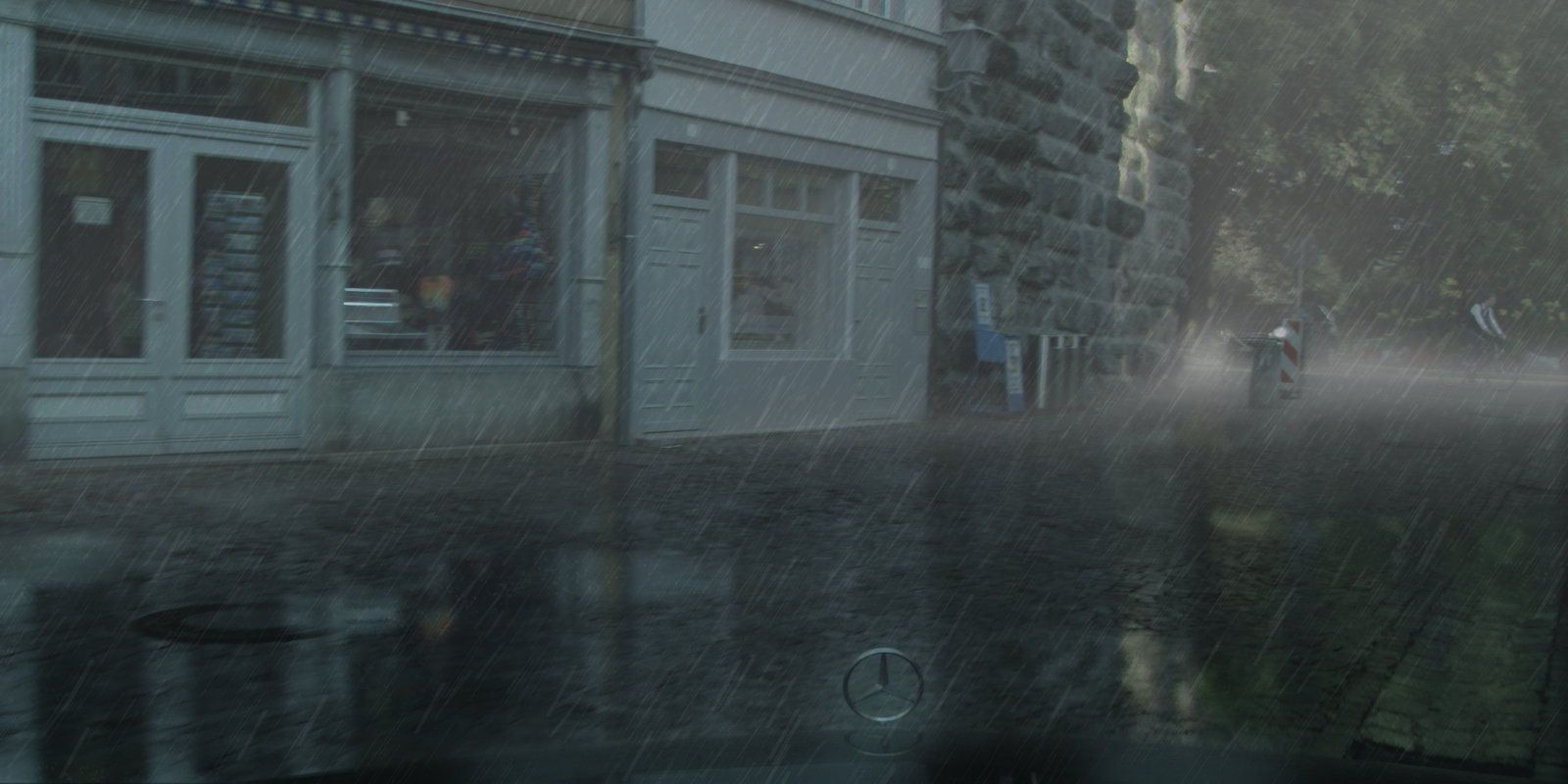} \\
&
\includegraphics[width=\linewidth]{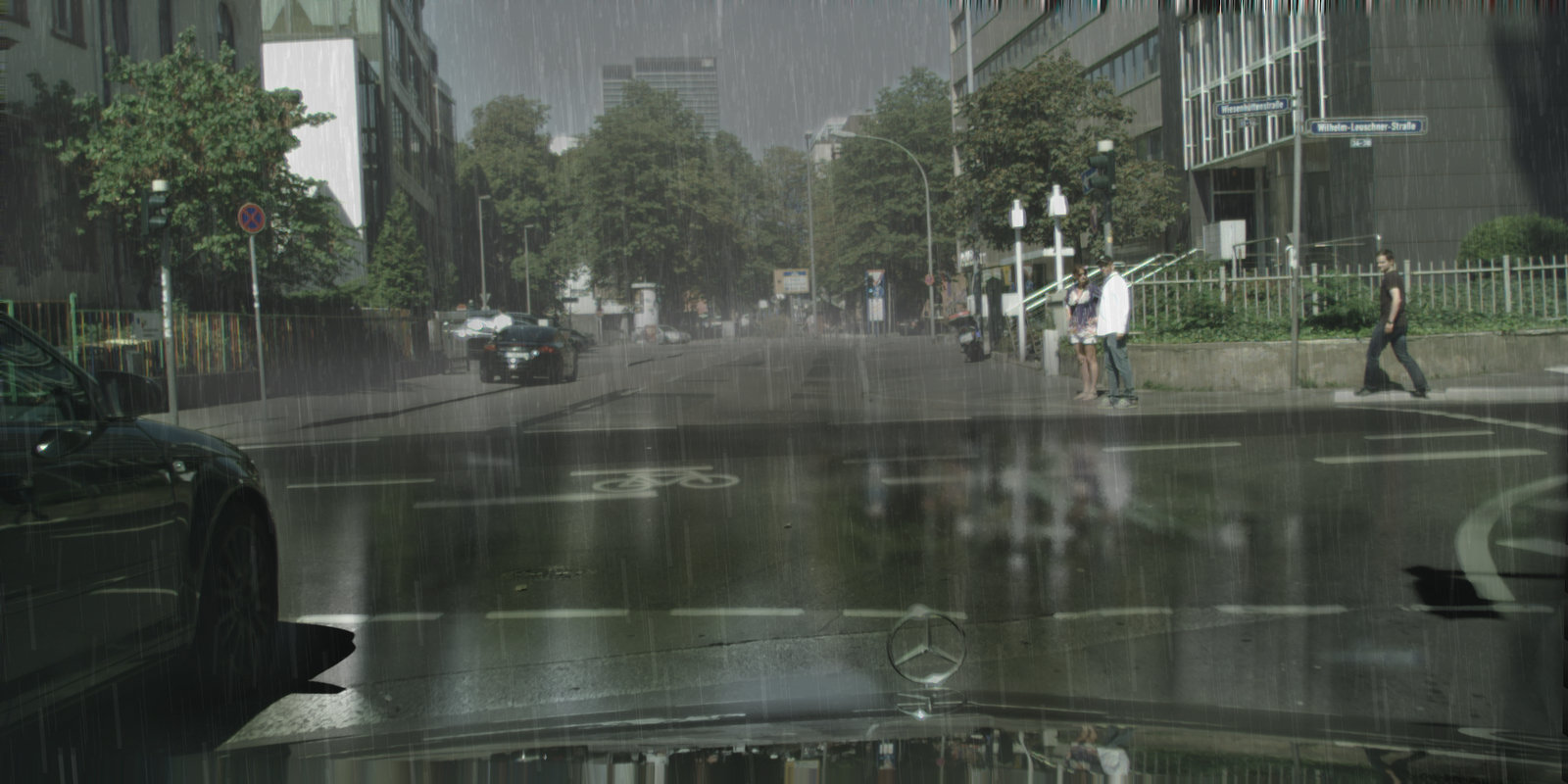} &
\includegraphics[width=\linewidth]{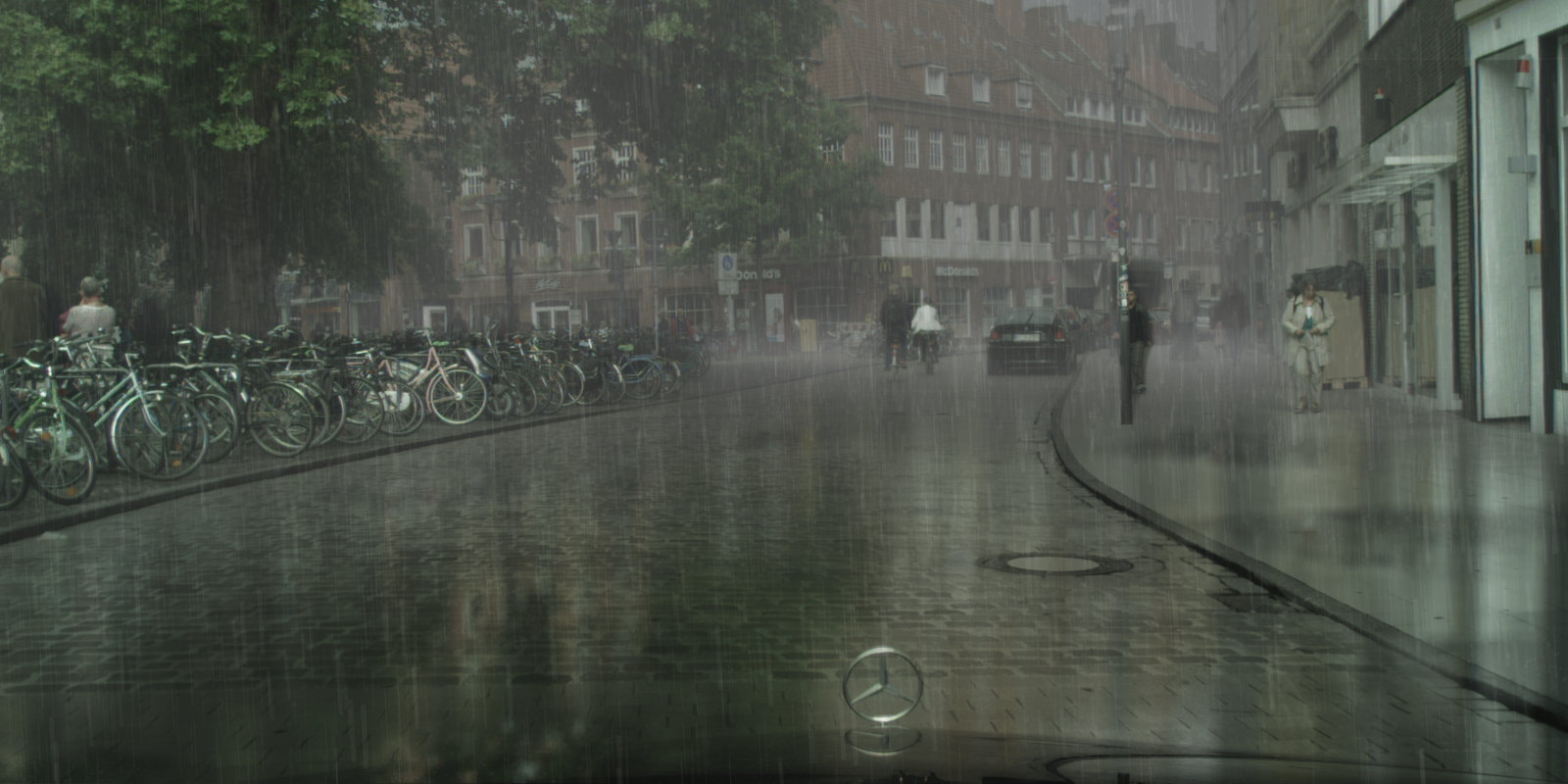} &
\includegraphics[width=\linewidth]{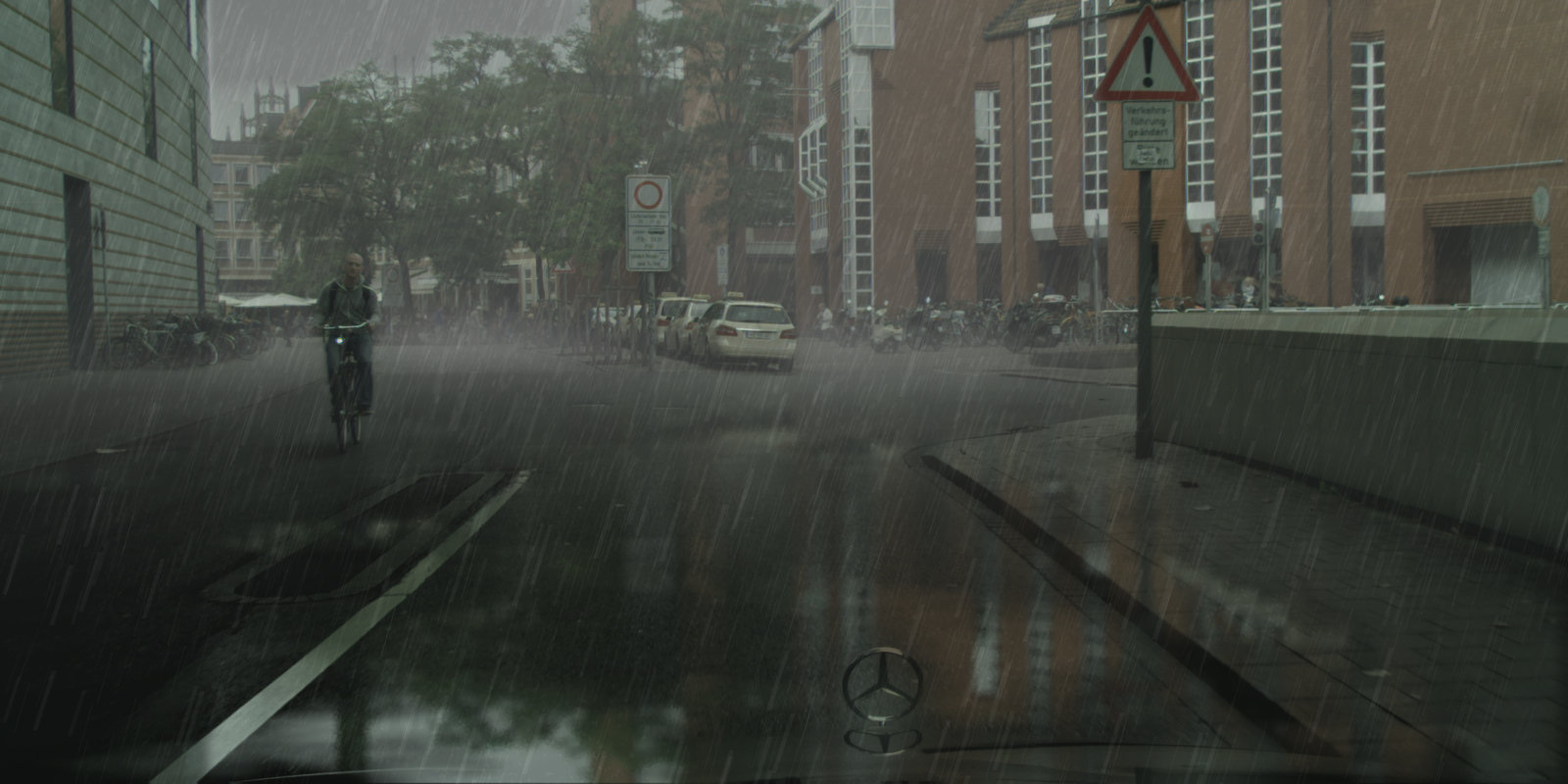} &
\includegraphics[width=\linewidth]{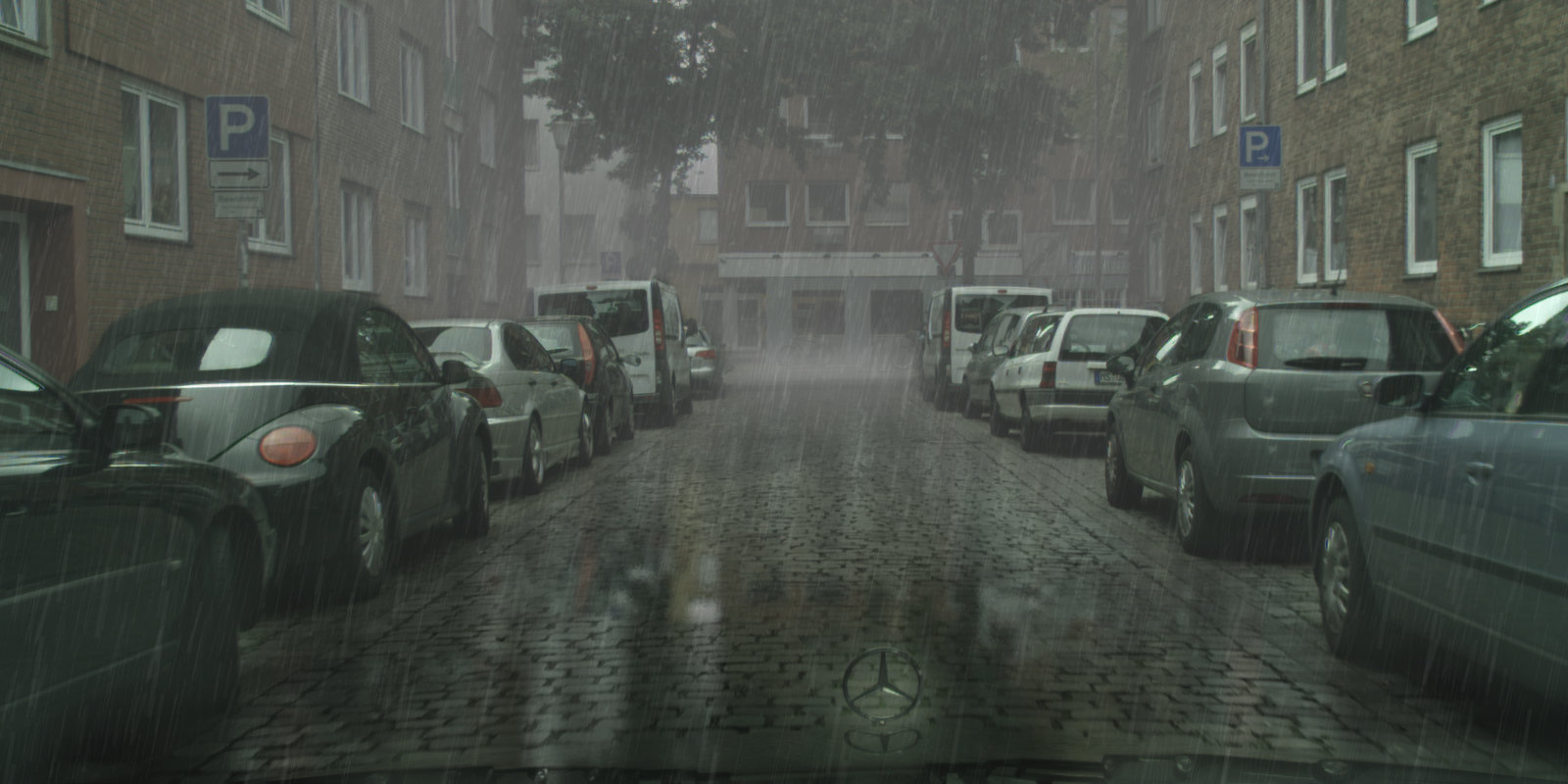} \\[6pt]

\midrule 
\addlinespace[6pt]

\multirow{2}{=}{\centering \textbf{BDD100K}} &
\includegraphics[width=\linewidth]{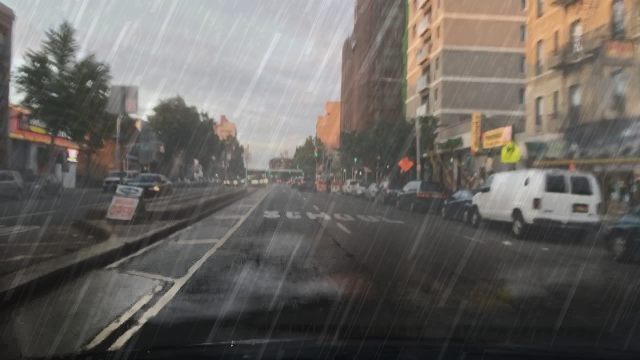} &
\includegraphics[width=\linewidth]{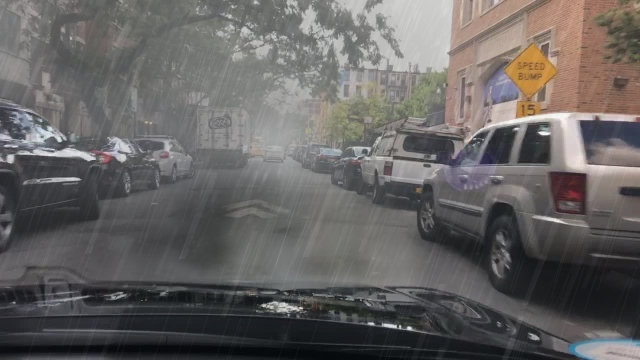} &
\includegraphics[width=\linewidth]{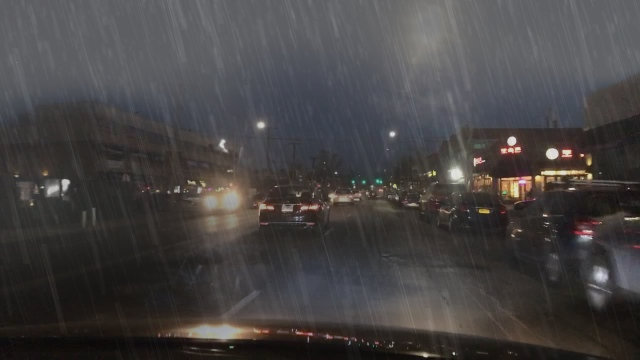} &
\includegraphics[width=\linewidth]{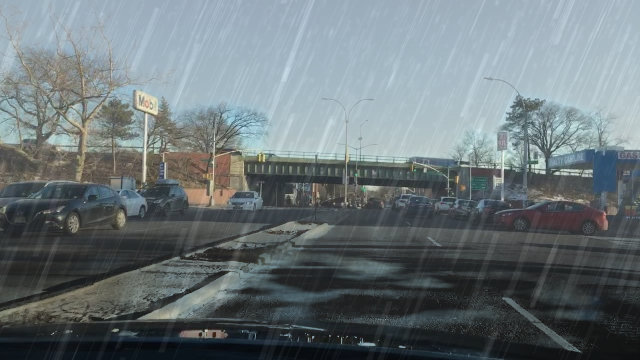} \\
&
\includegraphics[width=\linewidth]{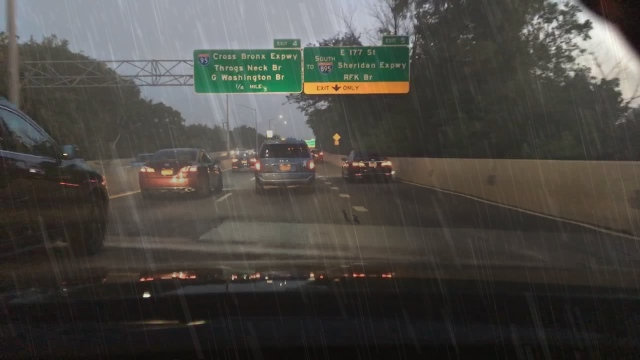} &
\includegraphics[width=\linewidth]{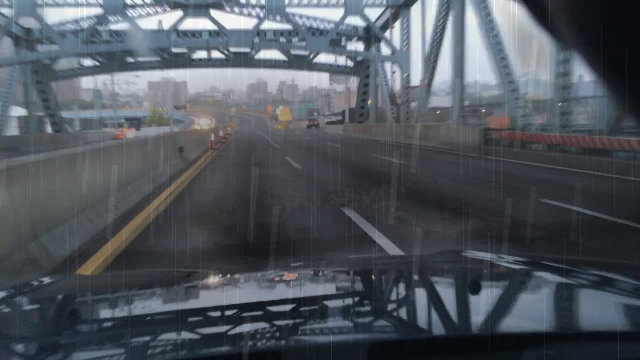} &
\includegraphics[width=\linewidth]{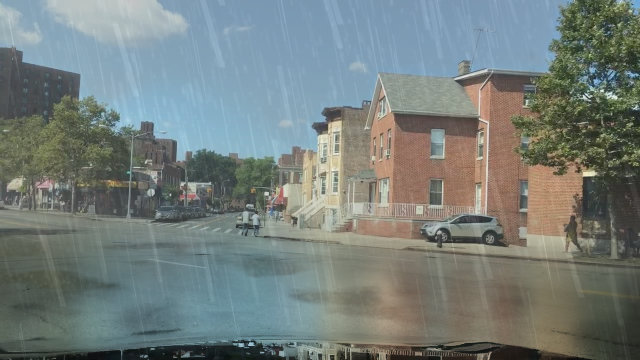} &
\includegraphics[width=\linewidth]{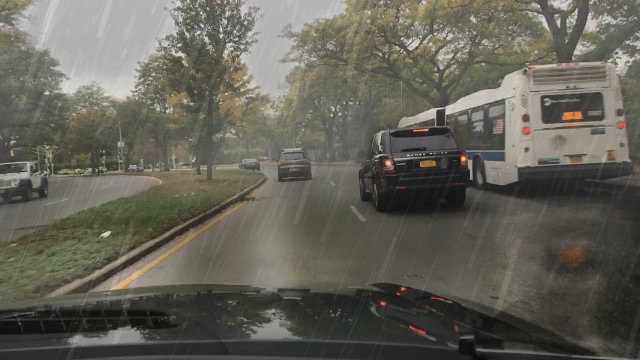} \\[6pt]

\midrule
\addlinespace[6pt]

\multirow{2}{=}{\centering \textbf{KITTI}} &
\includegraphics[width=\linewidth]{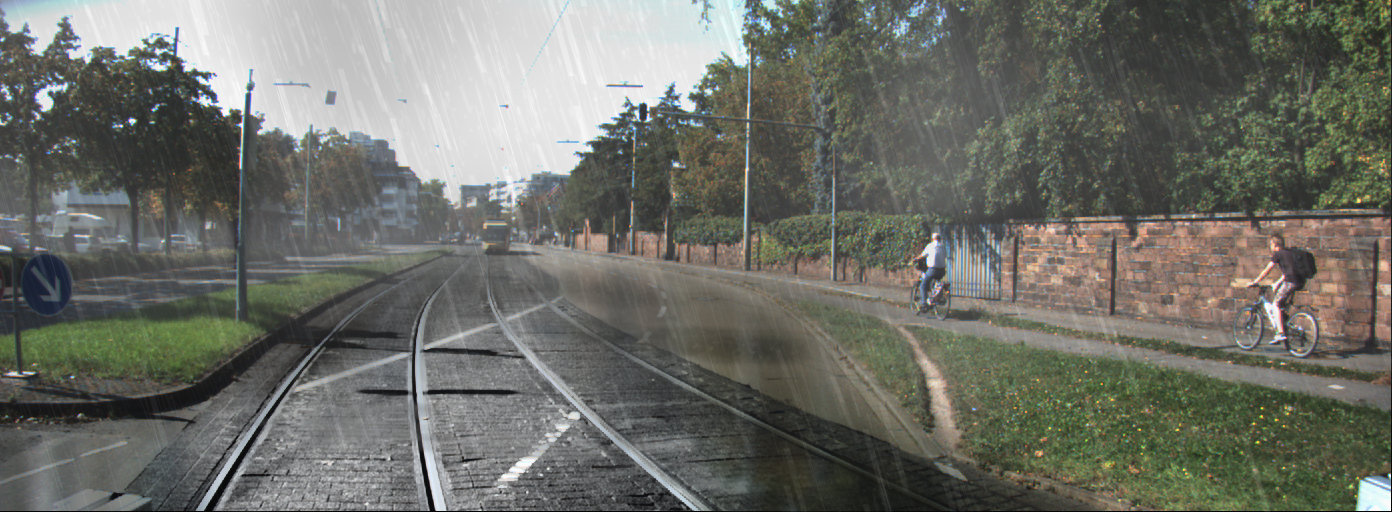} &
\includegraphics[width=\linewidth]{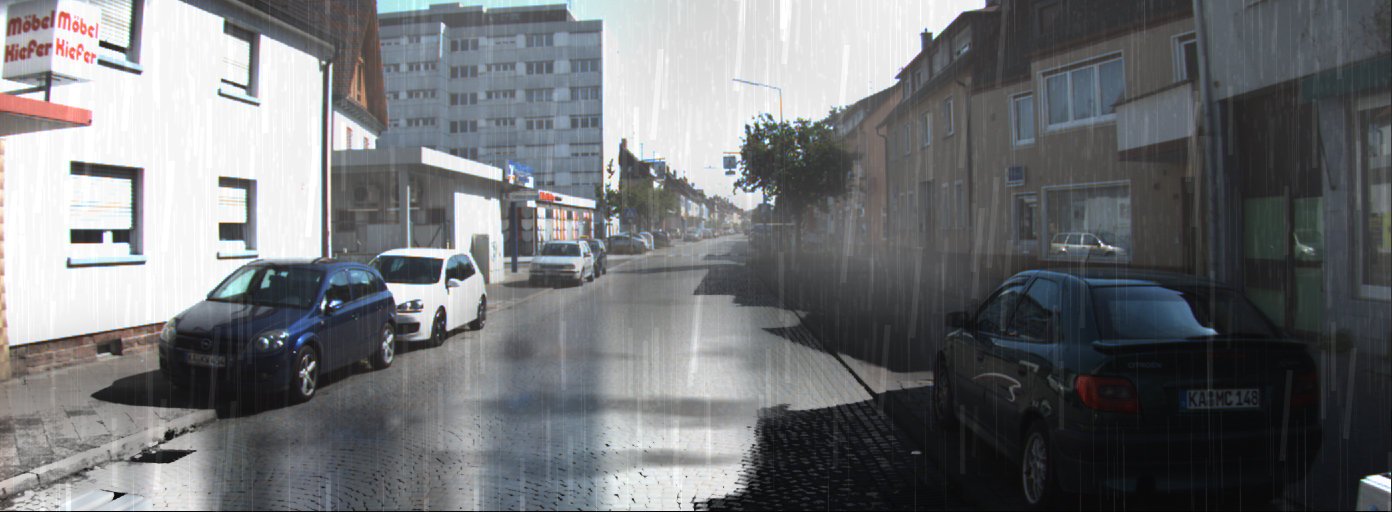} &
\includegraphics[width=\linewidth]{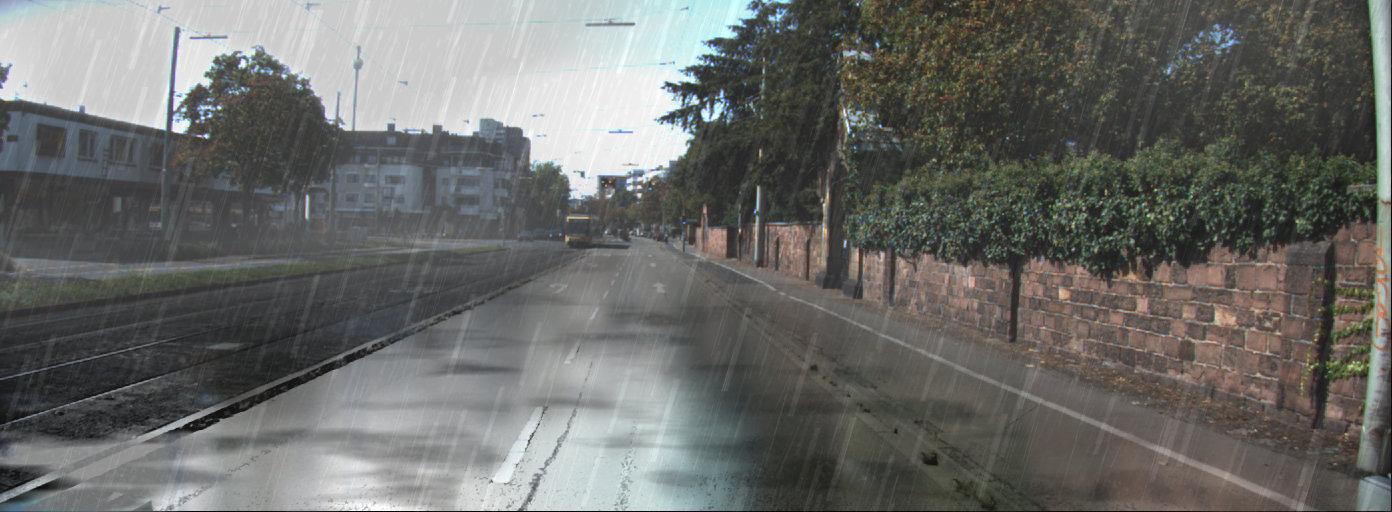} &
\includegraphics[width=\linewidth]{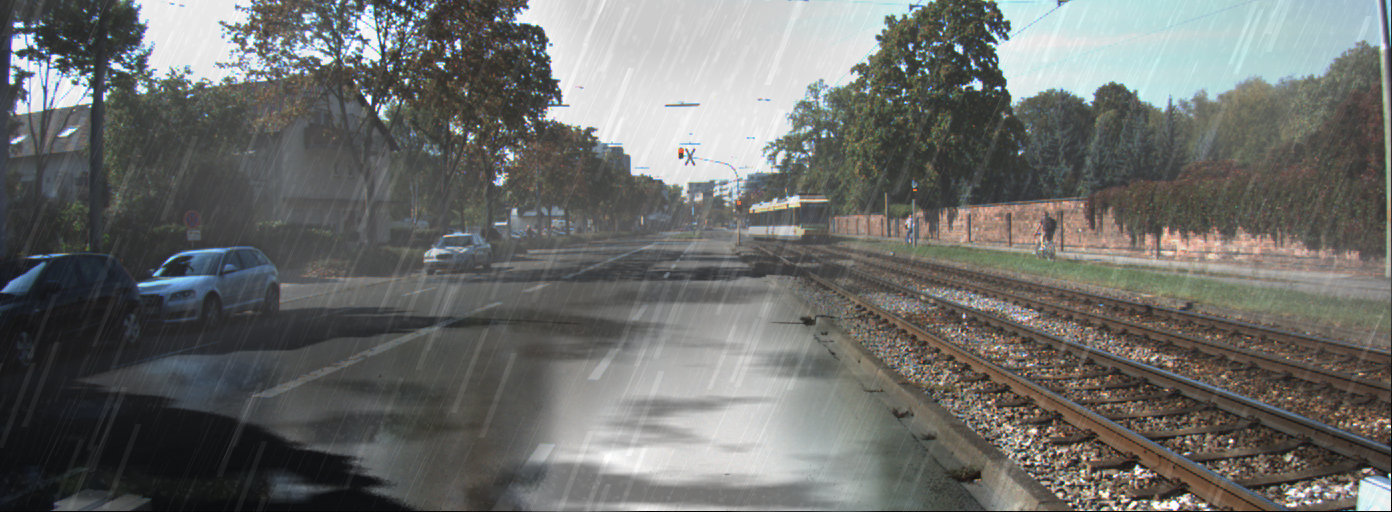} \\
&
\includegraphics[width=\linewidth]{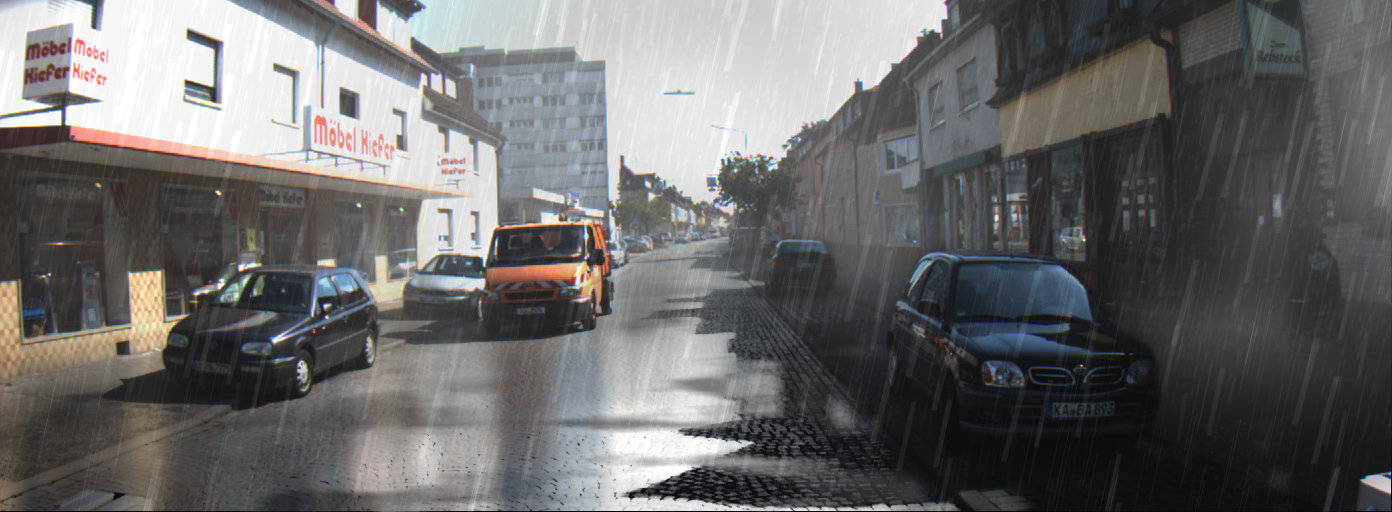} &
\includegraphics[width=\linewidth]{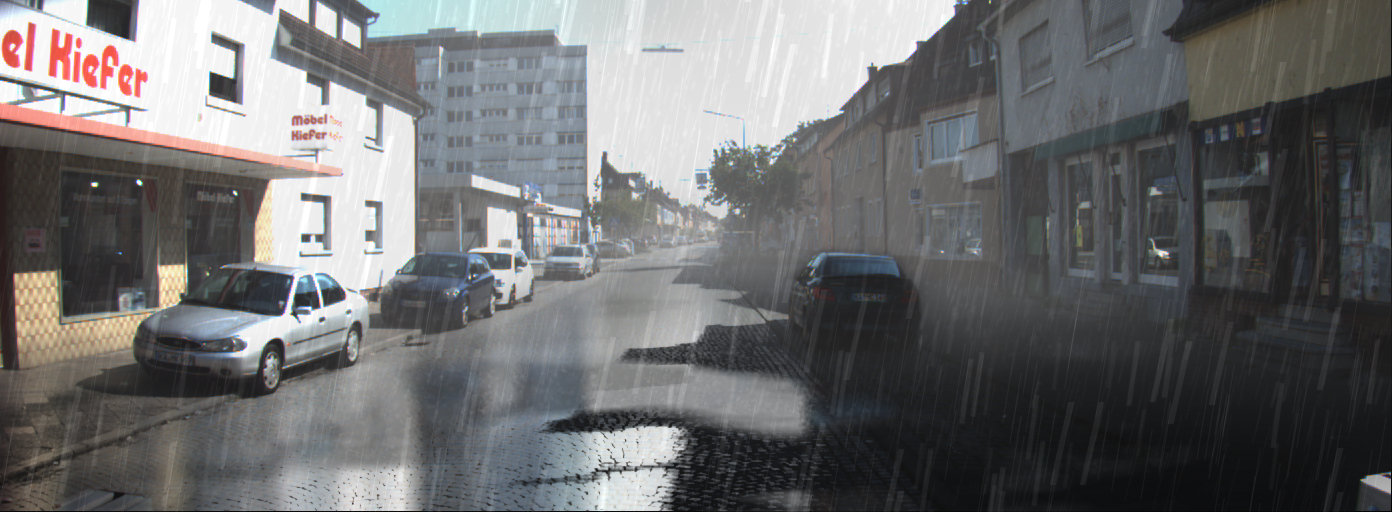} &
\includegraphics[width=\linewidth]{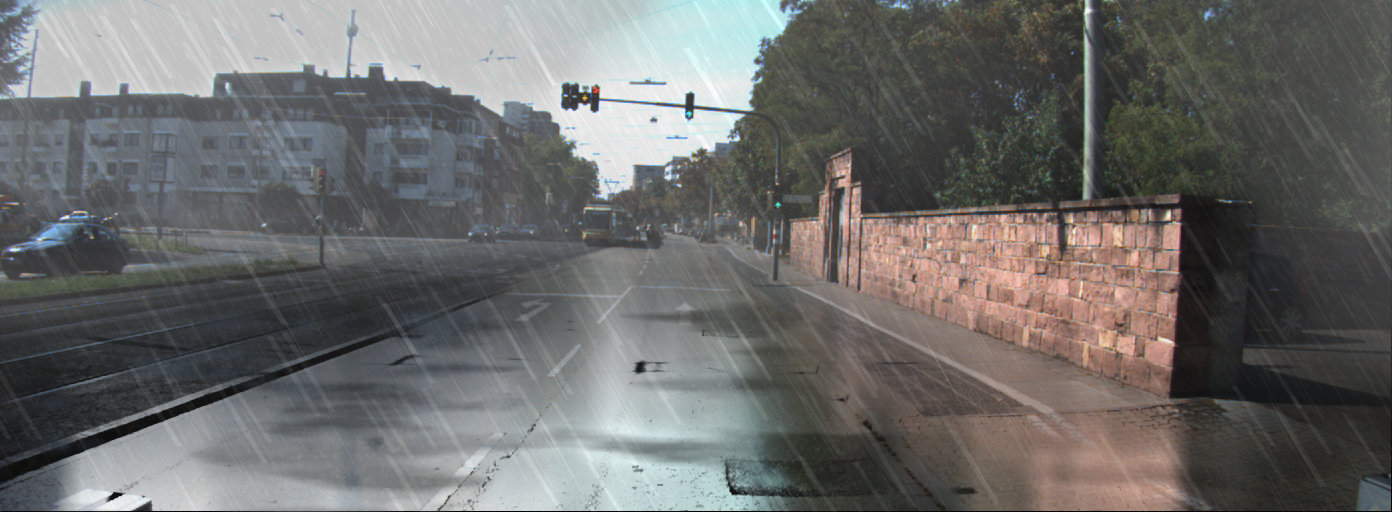} &
\includegraphics[width=\linewidth]{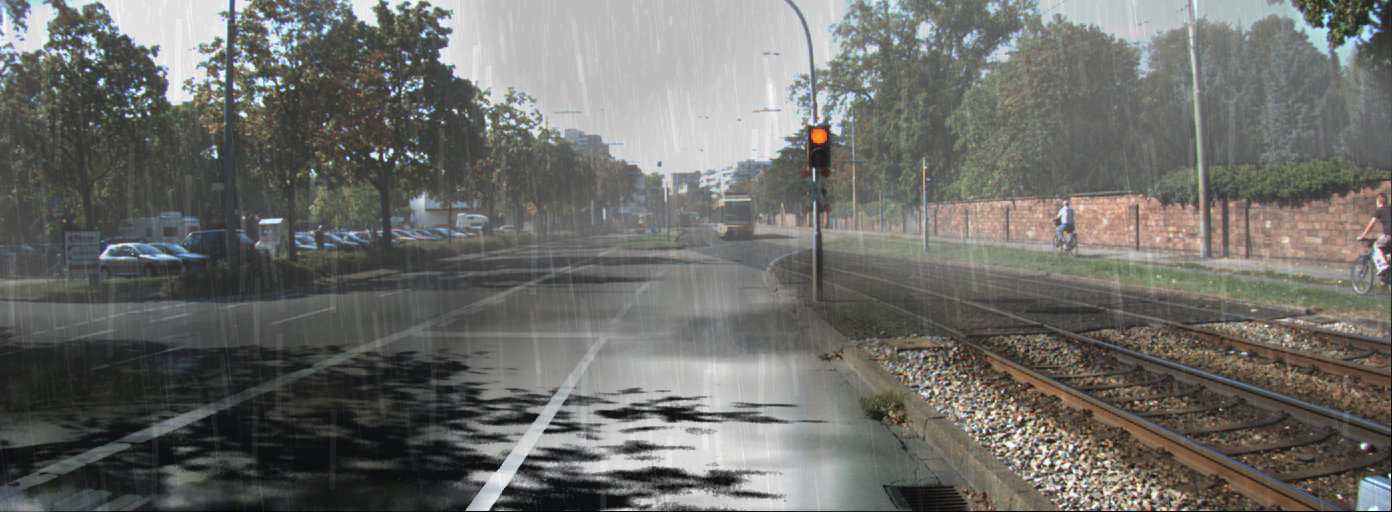} \\[6pt]

\midrule 

\addlinespace[6pt]

\multirow{2}{=}{\centering \textbf{Roadmarks}} &
\includegraphics[width=\linewidth]{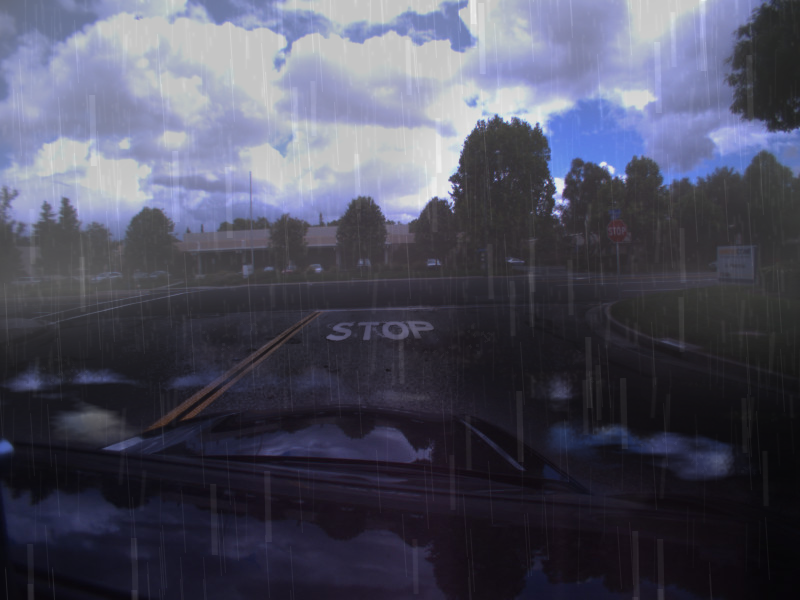} &
\includegraphics[width=\linewidth]{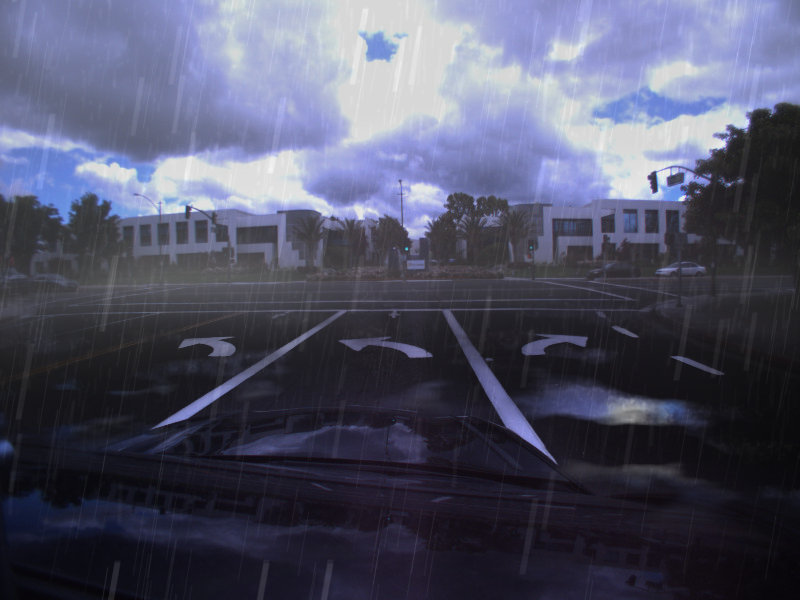} &
\includegraphics[width=\linewidth]{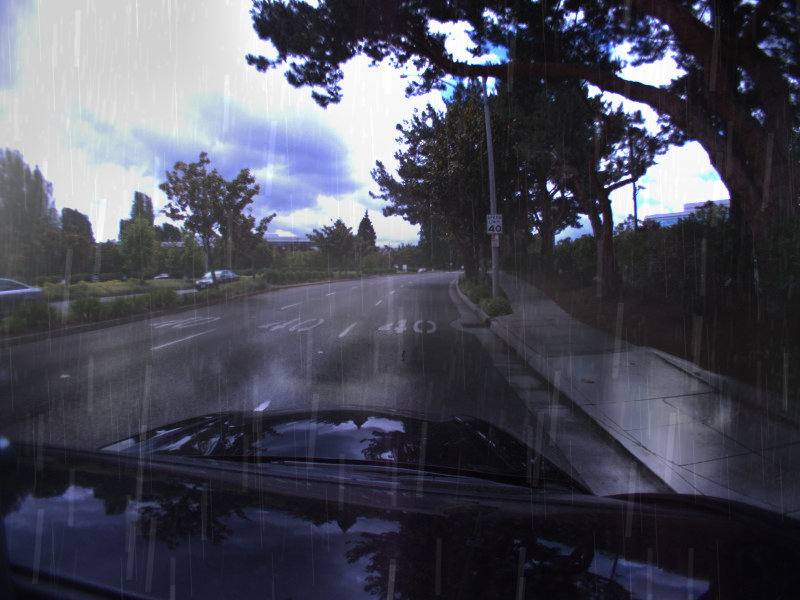} &
\includegraphics[width=\linewidth]{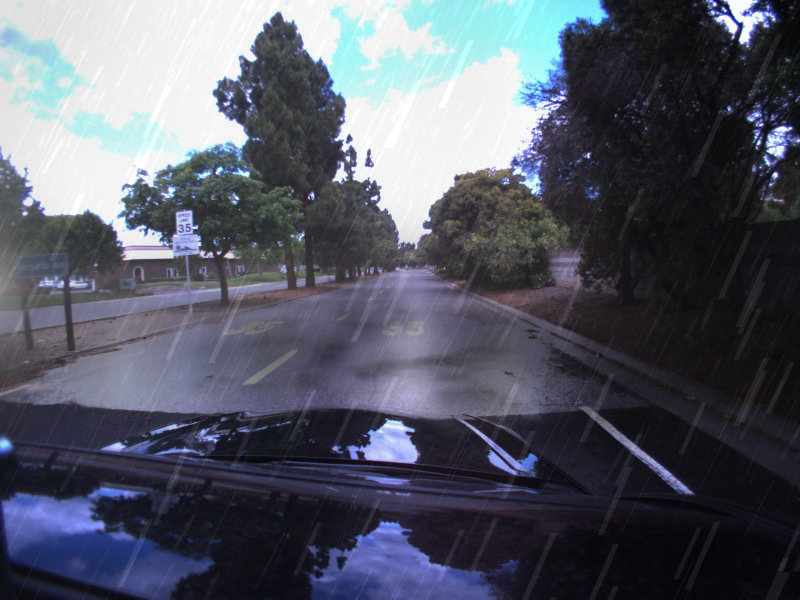} \\
&
\includegraphics[width=\linewidth]{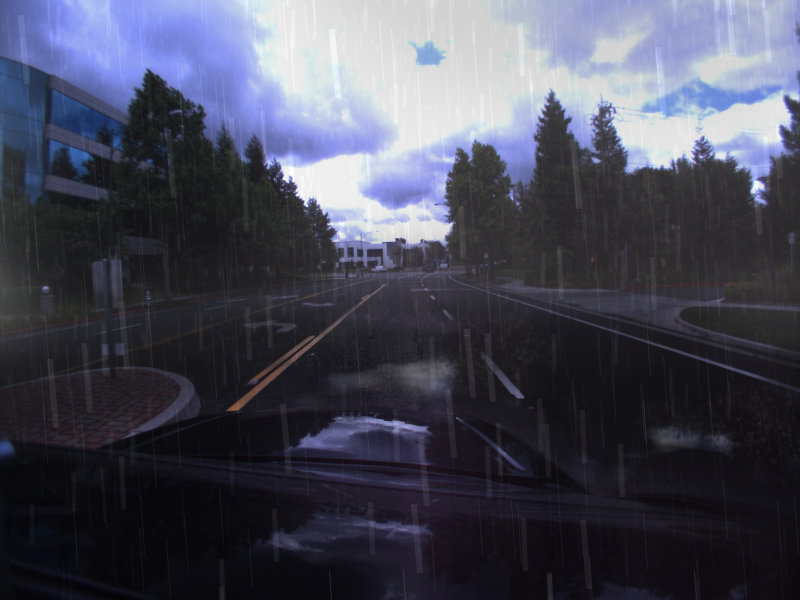} &
\includegraphics[width=\linewidth]{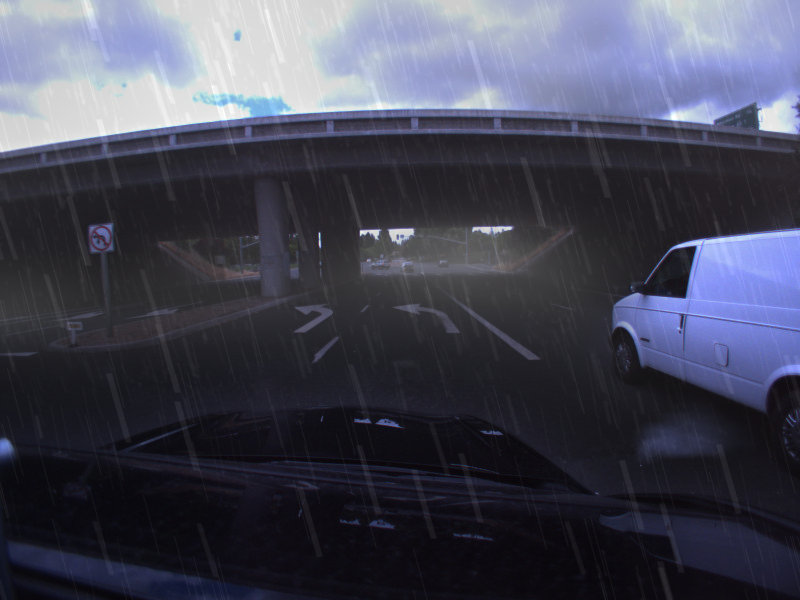} &
\includegraphics[width=\linewidth]{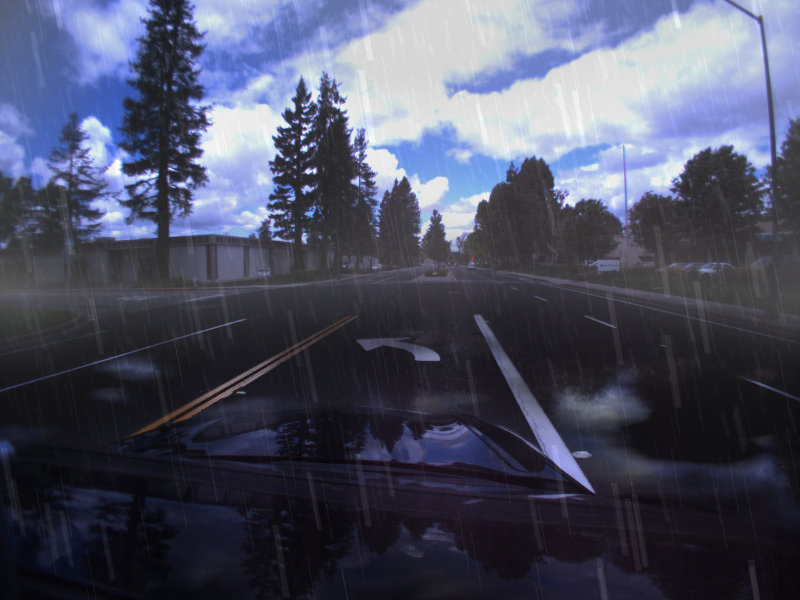} &
\includegraphics[width=\linewidth]{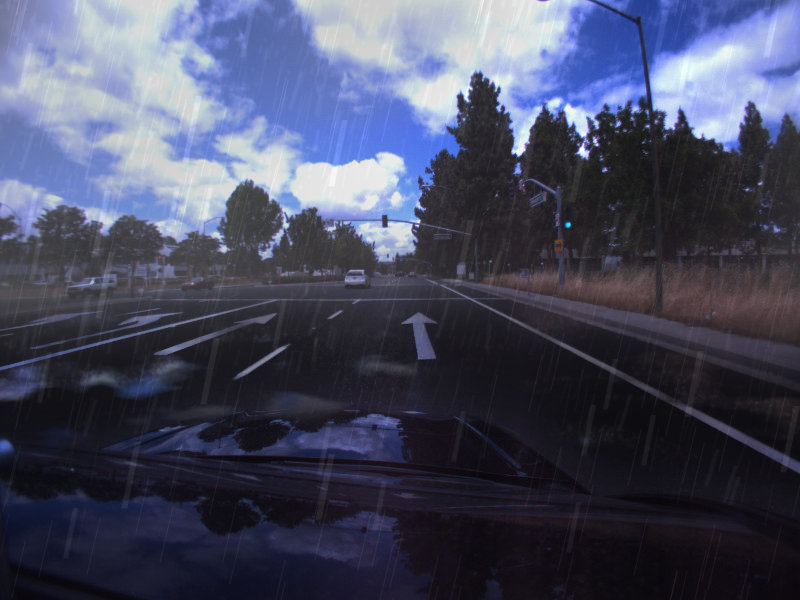} \\

\bottomrule
\end{tabular}
\label{tab:dataset_renders}
\vspace{-3mm}
\end{figure*}

\section{Effect Pretained Depth and Segmentation Models \label{supmat: effect pretrained priors}}
In Table~\ref{tab:cascaders_puddles_module_fid} we show the tabular data of the effect of changing the pretrained depth model MiDaS \cite{midas} and segmentation model SegFormer \cite{xie2021segformersimpleefficientdesign} by DepthAnything \cite{depthanything} and FastSeg \cite{zhang2020fastseg} on the FID scores. Changing the depth model affects FID scores, most likely due to fog rendering, which depends on depth levels. This is visible in the qualitative examples we show in Figure \ref{fig:modularity_examples}, where clearly more and denser fog is included in the images rendered using DepthAnything. It is also clear that no semantic information is changed and the puddle and reflection effects are rendered as intended, following perspective and semantic cues. We further demonstrate this in Figure \ref{fig:umap_embeddings}, a UMAP of embeddings extracted by the DinoV3 model \cite{simeoni2025dinov3}. UMAP visualizes the high-dimensional DINOv3 feature embeddings in two dimensions, where nearby samples correspond to images with similar semantic representations. The embedding distributions of all four implementations exhibit strong overlap, indicating that the generated images occupy the same high-level feature space regardless of the selected models. This suggests that the module substitutions primarily affect appearance statistics captured by FID, while preserving the semantic content and overall image distribution.

\begin{table*}[!t]
\centering
\caption{\textbf{FID comparison for CascadeRS module variants.}
Fr\'echet Inception Distance (FID) is reported across rain intensities, using ACDC and RID as reference datasets. Lower values indicate greater distributional similarity to the reference dataset.}
\label{tab:cascaders_puddles_module_fid}
\renewcommand{\arraystretch}{1.18}
\setlength{\tabcolsep}{3.5pt}
\scriptsize
\begin{tabular}{lccccccccccc}
\toprule
\textbf{Implementation} & \textbf{No rain} & \textbf{10} & \textbf{20} & \textbf{30} & \textbf{40} & \textbf{50} & \textbf{60} & \textbf{70} & \textbf{80} & \textbf{90} & \textbf{100} \\
\midrule
\multicolumn{12}{l}{\textit{FID-ACDC (500 reference images)}} \\
CascadeRS original (MiDaS + SegFormer) & 117.80 & 99.08 & 99.60 & 101.72 & 106.74 & 113.42 & 122.43 & 132.34 & 143.39 & 155.80 & 164.76 \\
CascadeRS (DepthAnything + SegFormer) & 117.80 & 97.58 & 99.00 & 103.52 & 112.81 & 124.04 & 139.24 & 154.97 & 170.45 & 184.76 & 198.45 \\
CascadeRS (MiDaS + FastSeg) & 117.80 & 97.09 & 97.59 & 100.44 & 105.25 & 111.70 & 120.07 & 131.43 & 142.64 & 153.50 & 164.54 \\
CascadeRS (DepthAnything + FastSeg) & 117.80 & 95.79 & 96.85 & 102.04 & 111.22 & 123.59 & 139.01 & 154.52 & 171.26 & 187.08 & 201.27 \\
\addlinespace[0.6em]
\midrule
\multicolumn{12}{l}{\textit{FID-RID (2495 reference images)}} \\
CascadeRS original (MiDaS + SegFormer) & 131.96 & 89.82 & 88.69 & 87.58 & 89.23 & 92.08 & 95.67 & 102.03 & 108.02 & 116.15 & 119.91 \\
CascadeRS (DepthAnything + SegFormer) & 131.96 & 90.83 & 89.94 & 91.50 & 94.73 & 100.38 & 108.31 & 117.69 & 126.51 & 135.71 & 145.03 \\
CascadeRS (MiDaS + FastSeg) & 131.96 & 88.96 & 87.42 & 87.28 & 88.76 & 90.92 & 96.49 & 102.02 & 108.72 & 115.82 & 120.42 \\
CascadeRS (DepthAnything + FastSeg) & 131.96 & 90.51 & 90.37 & 91.59 & 95.35 & 101.64 & 109.41 & 118.00 & 128.24 & 138.45 & 148.05 \\
\bottomrule
\end{tabular}
\end{table*}

\begin{figure*}[t]
\centering
\renewcommand{\arraystretch}{1.6}
\setlength{\tabcolsep}{2pt}

\scriptsize

\begin{tabular}{C{1.1cm} C{0.22\linewidth} C{0.22\linewidth} C{0.22\linewidth} C{0.22\linewidth}}
\toprule
\textbf{Image} &
\textbf{M + SegF} &
\textbf{DA + SegF} &
\textbf{M + FS} &
\textbf{DA + FS} \\
\midrule

\textbf{1} &
\includegraphics[width=\linewidth]{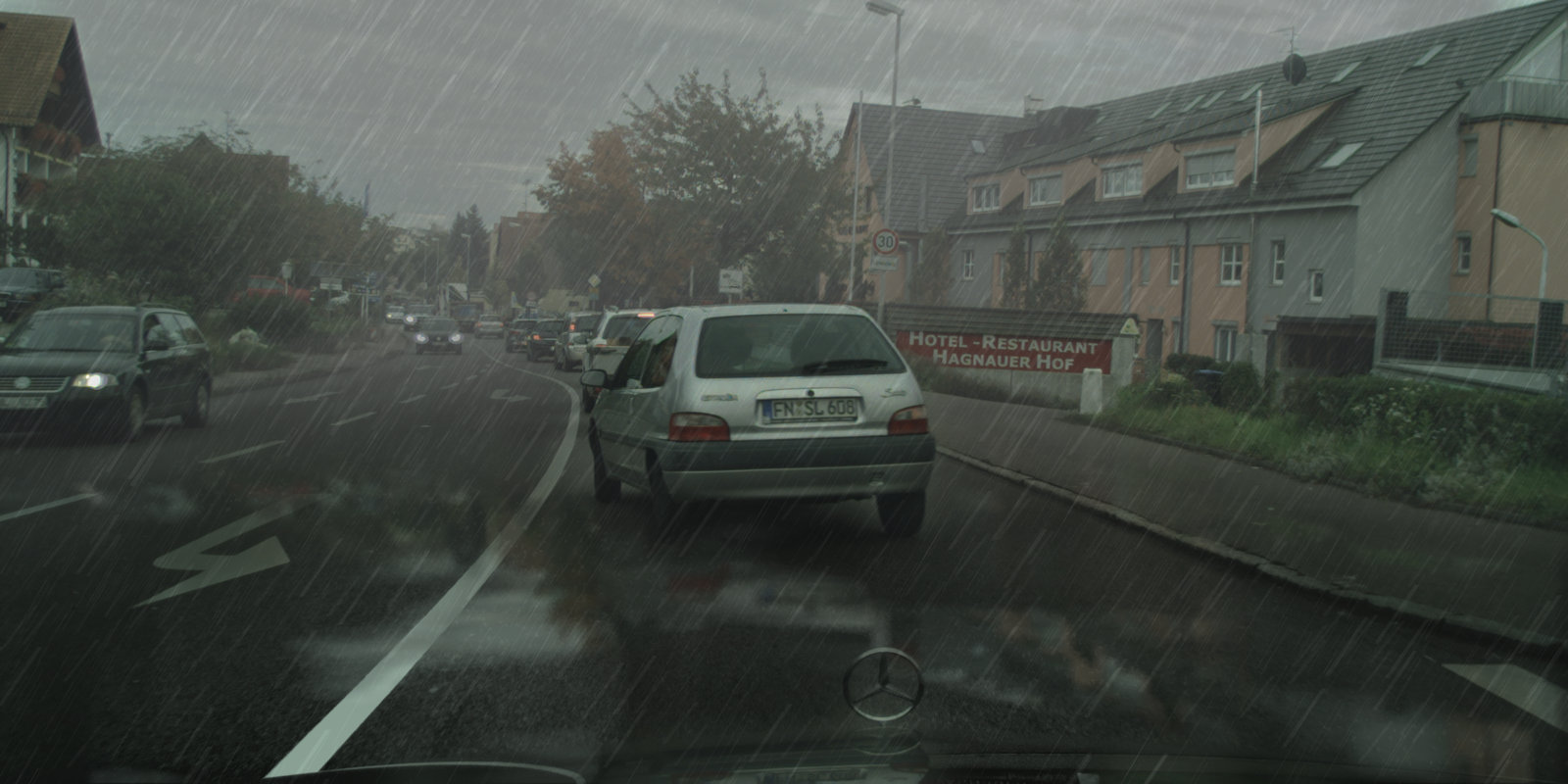} &
\includegraphics[width=\linewidth]{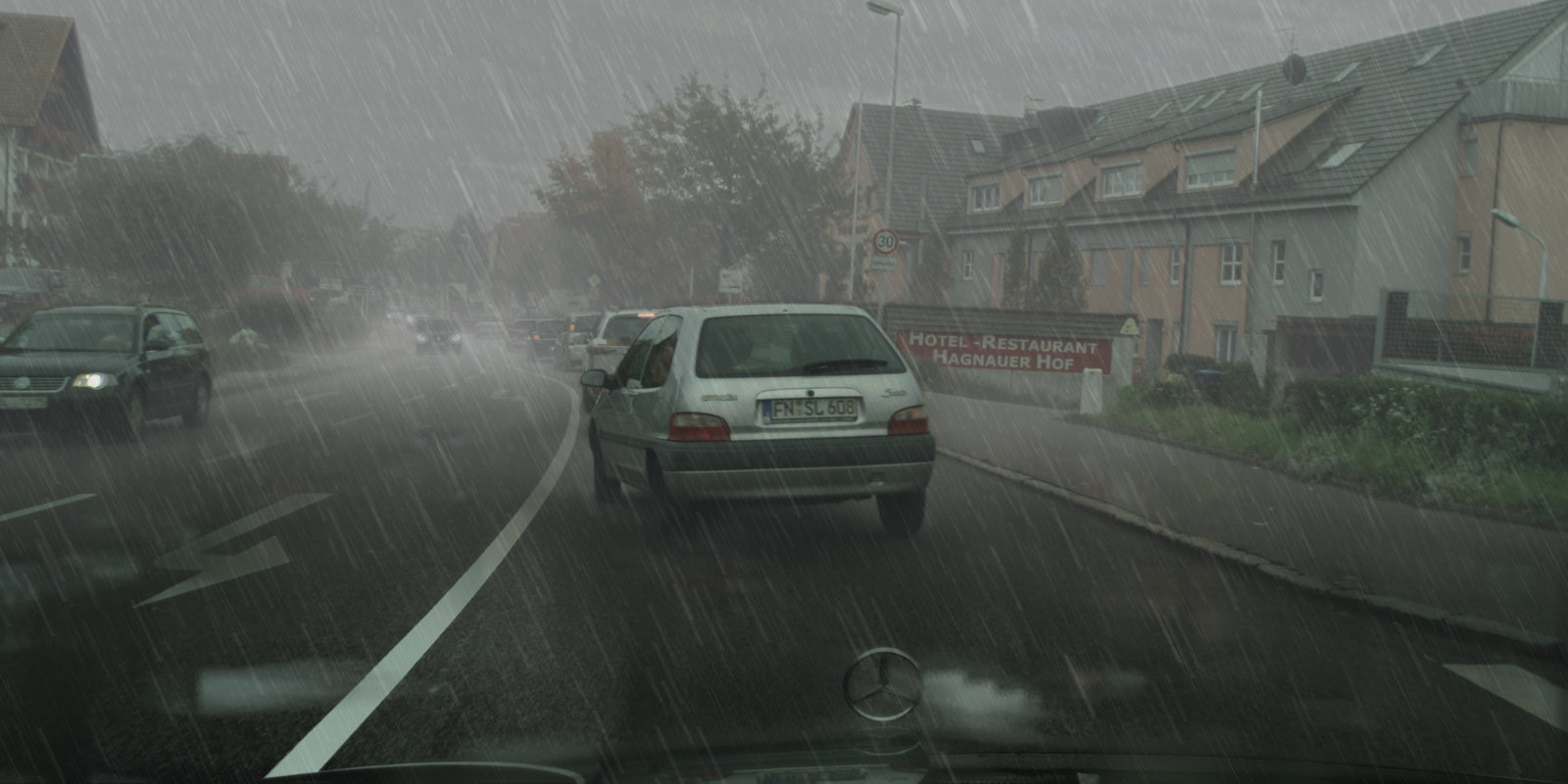} &
\includegraphics[width=\linewidth]{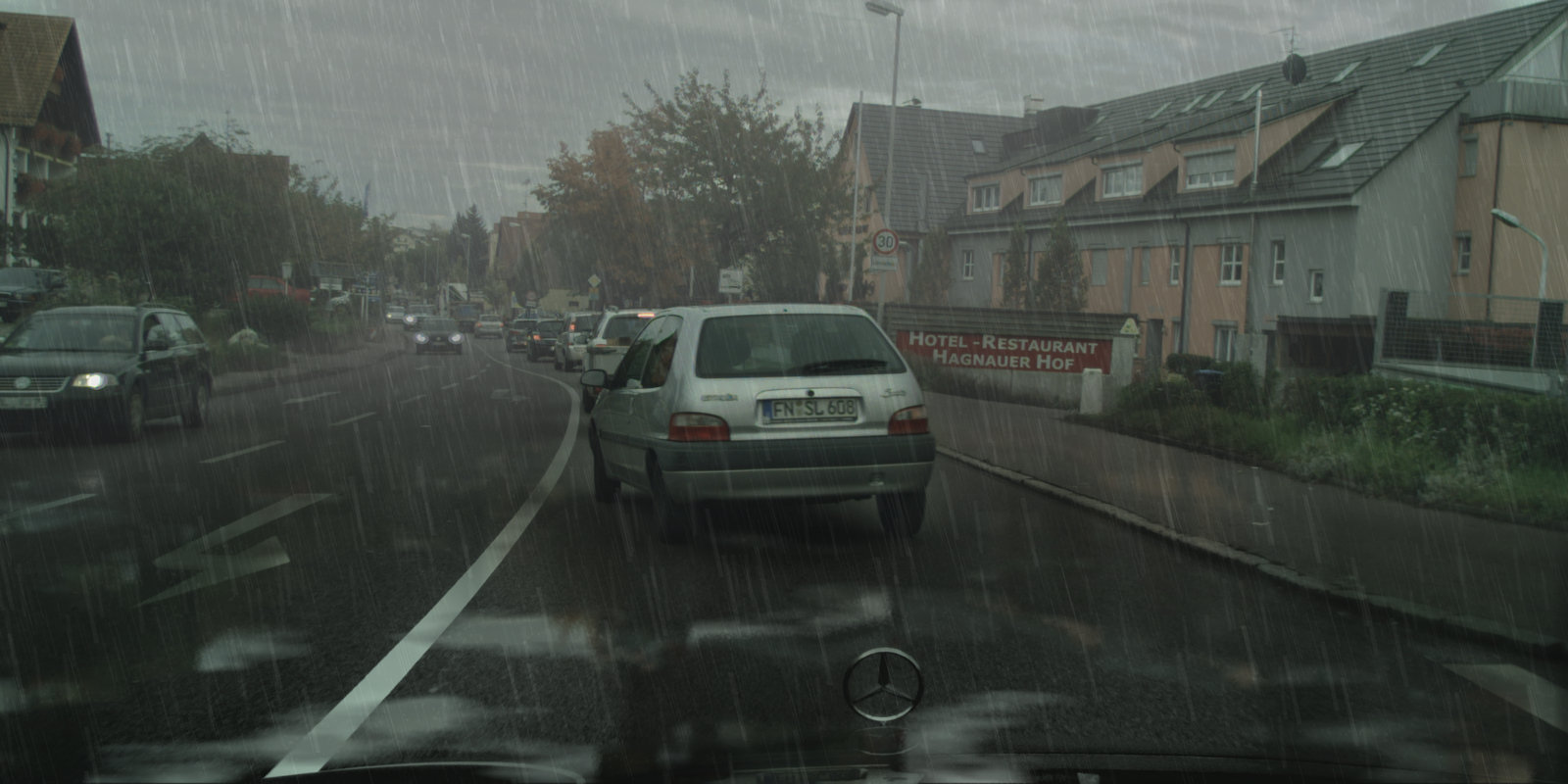} &
\includegraphics[width=\linewidth]{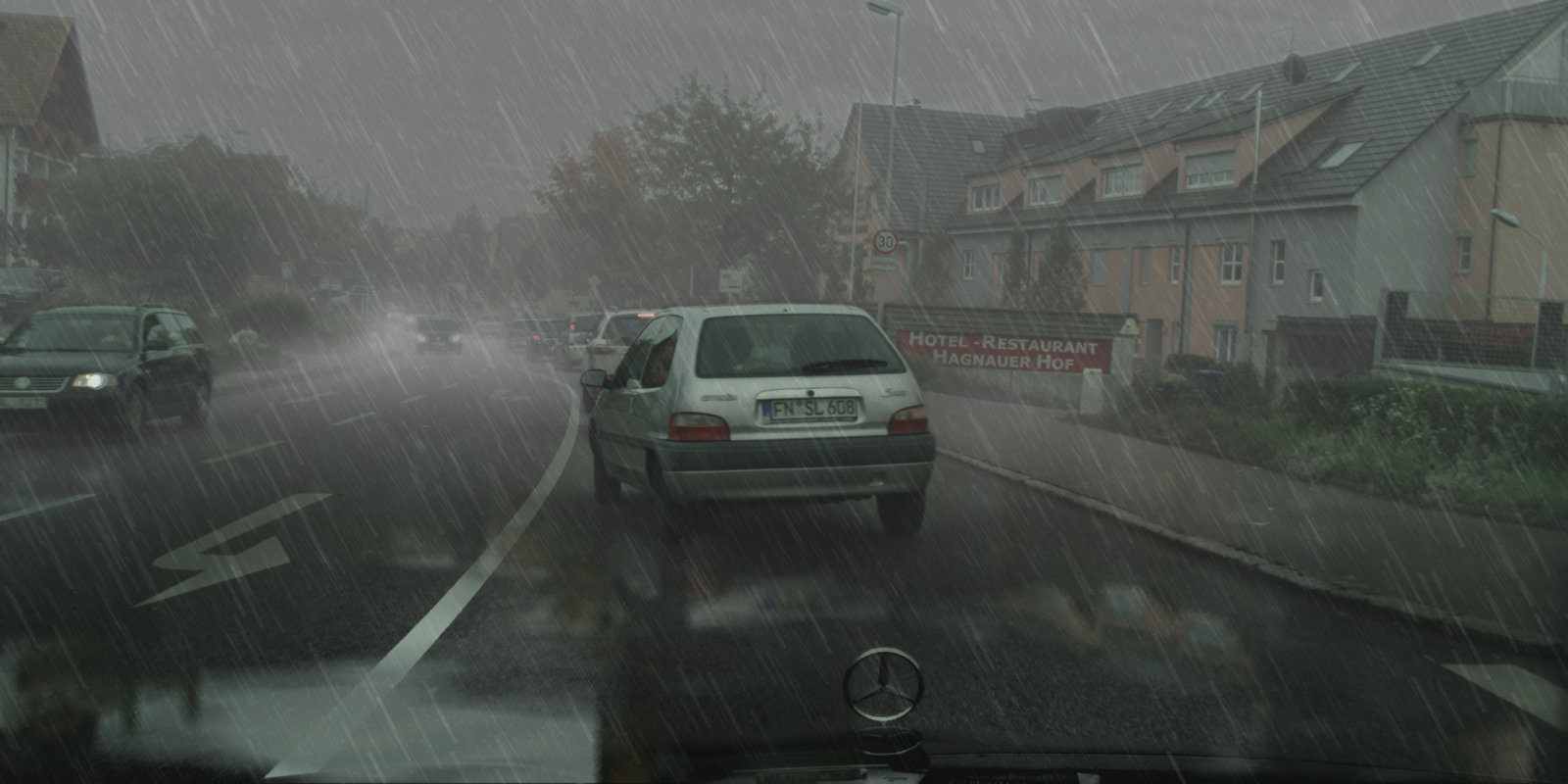} \\

\textbf{2} &
\includegraphics[width=\linewidth]{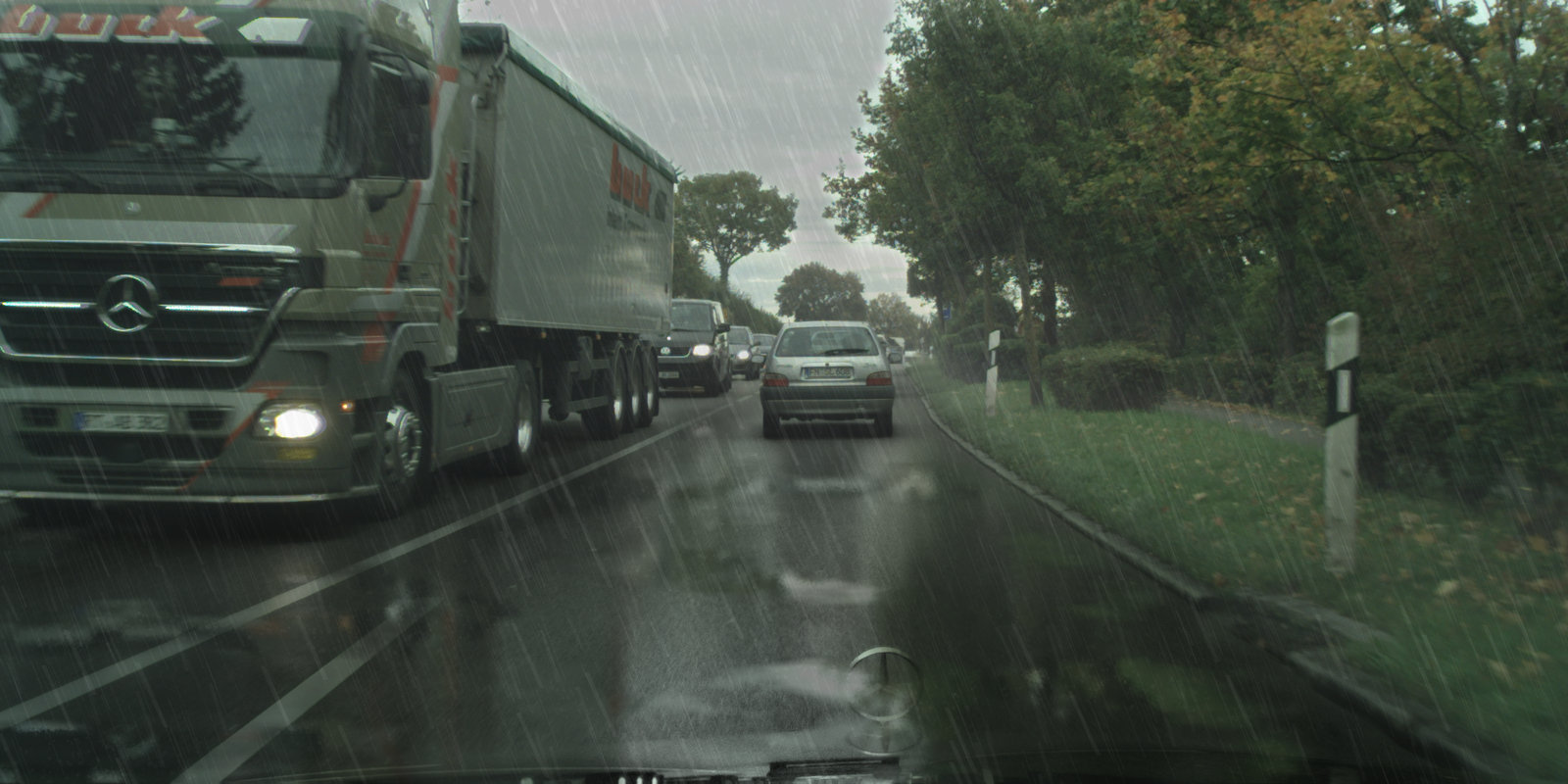} &
\includegraphics[width=\linewidth]{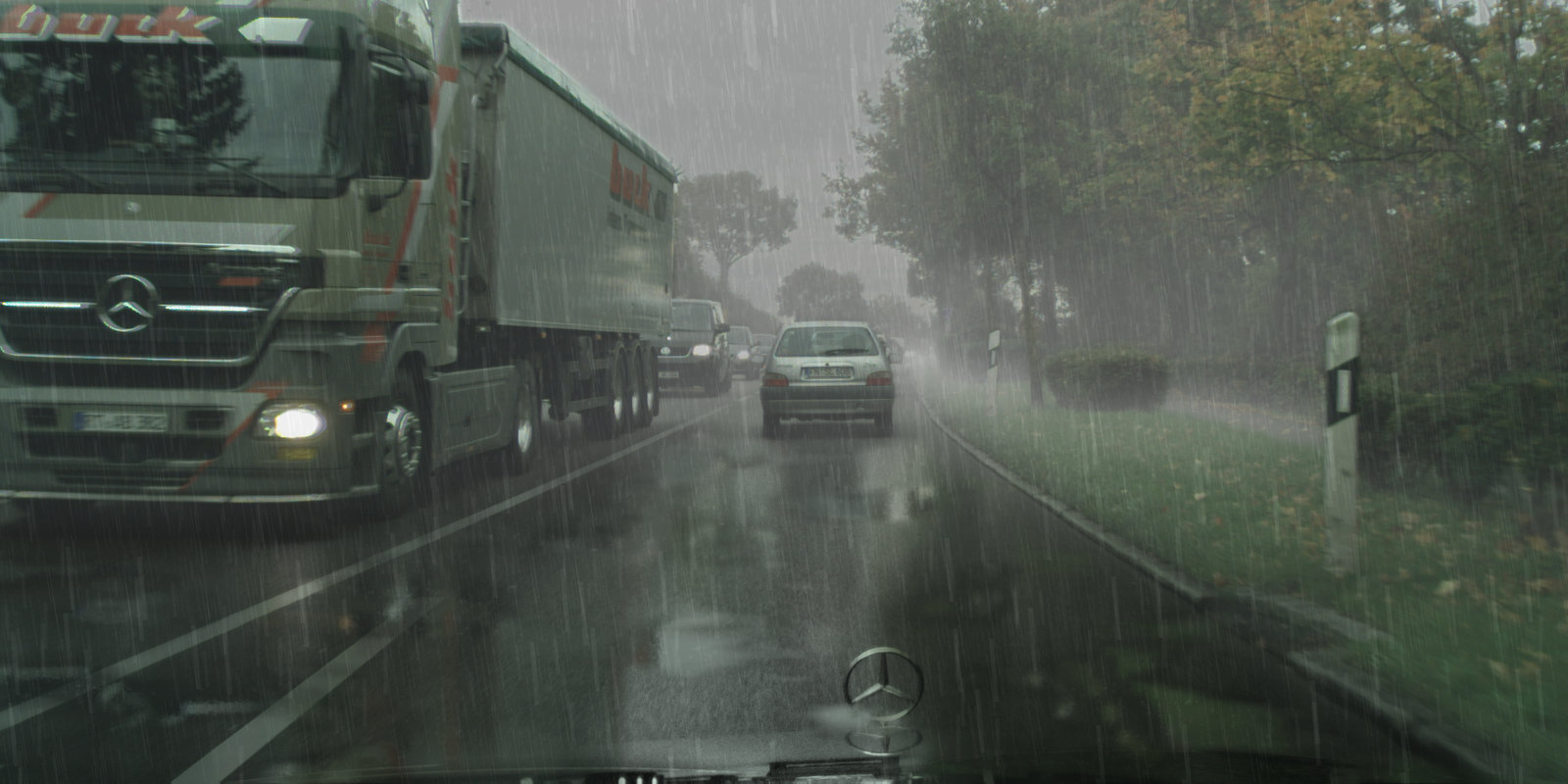} &
\includegraphics[width=\linewidth]{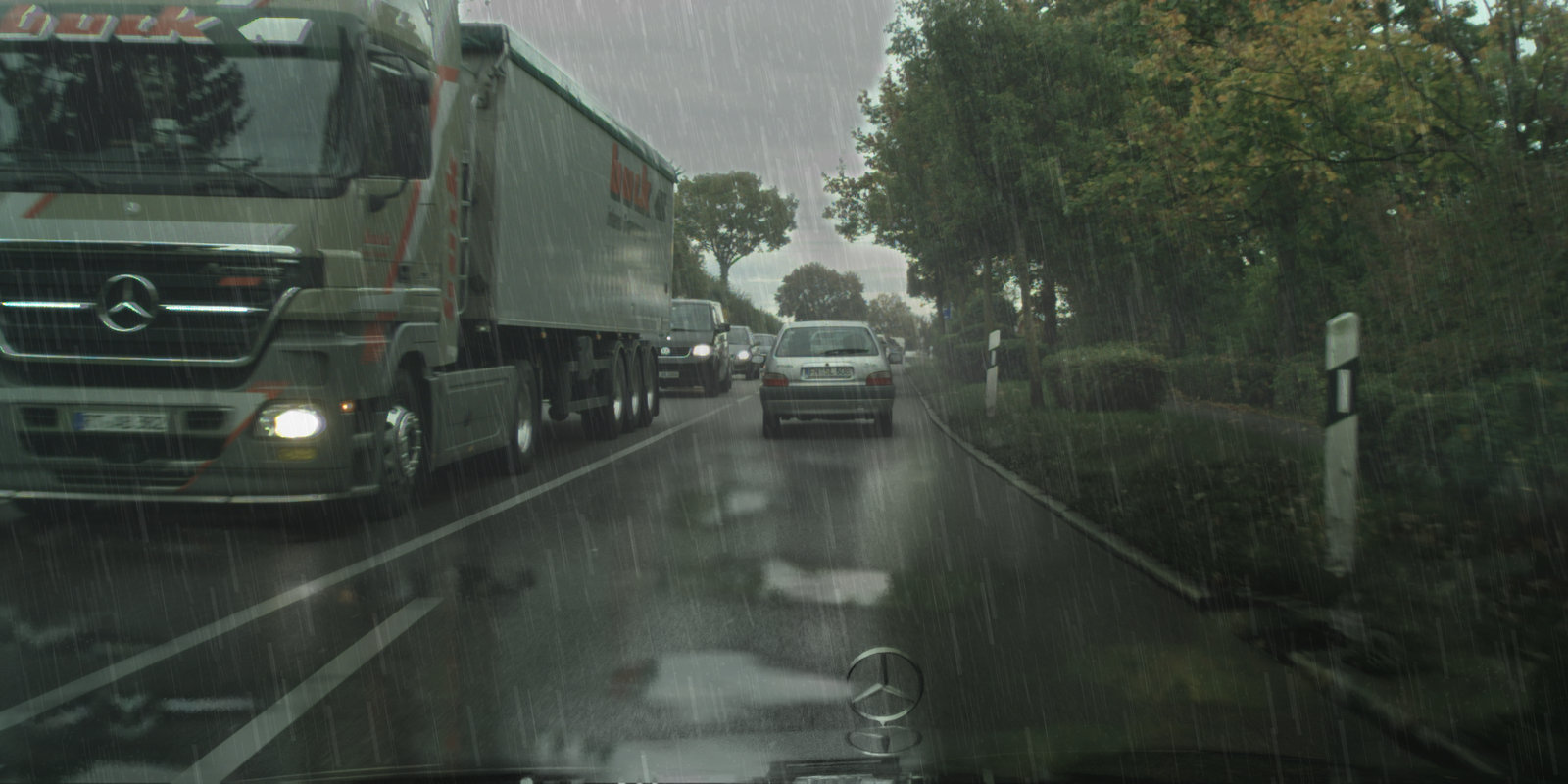} &
\includegraphics[width=\linewidth]{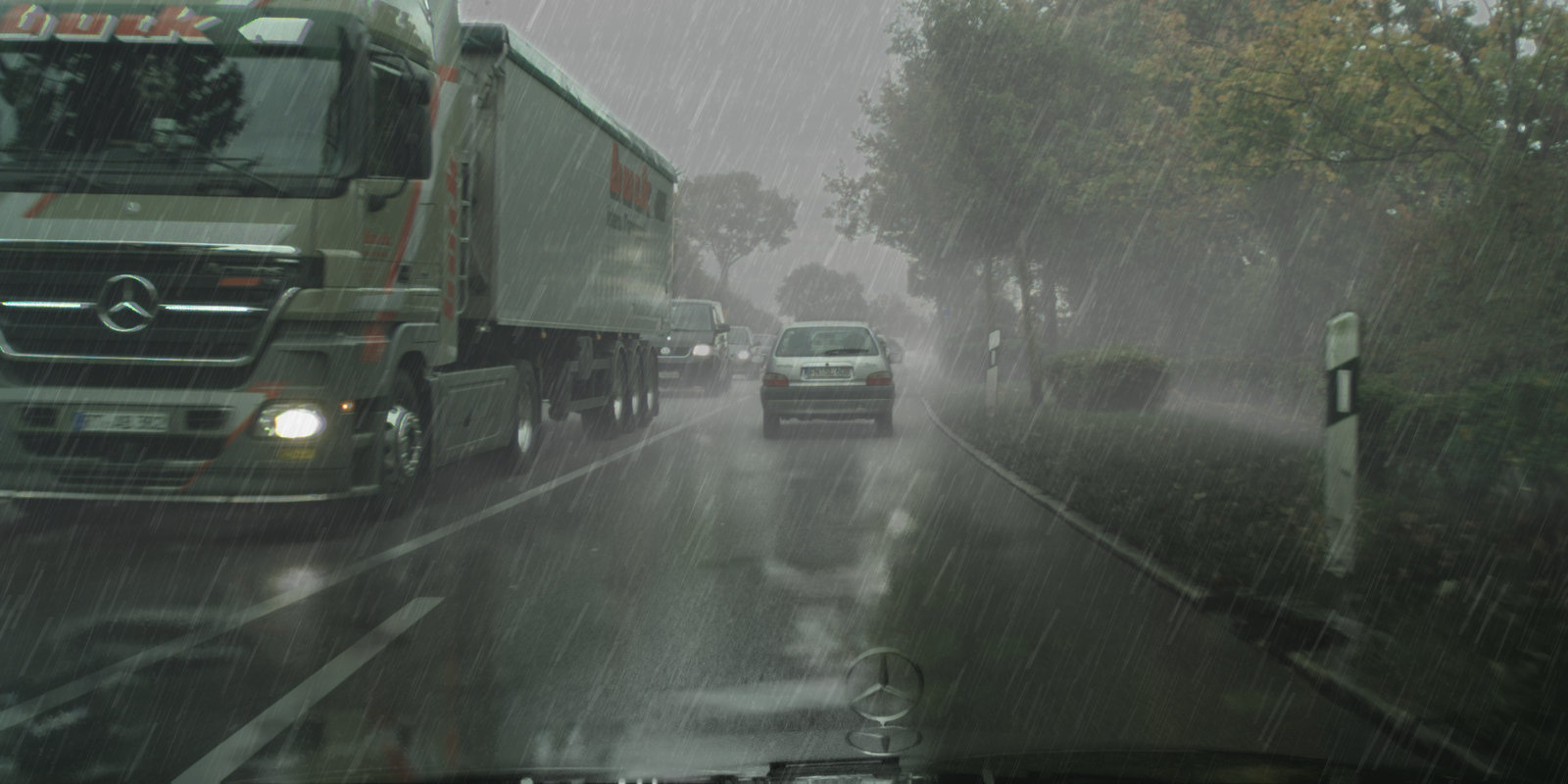} \\

\textbf{3} &
\includegraphics[width=\linewidth]{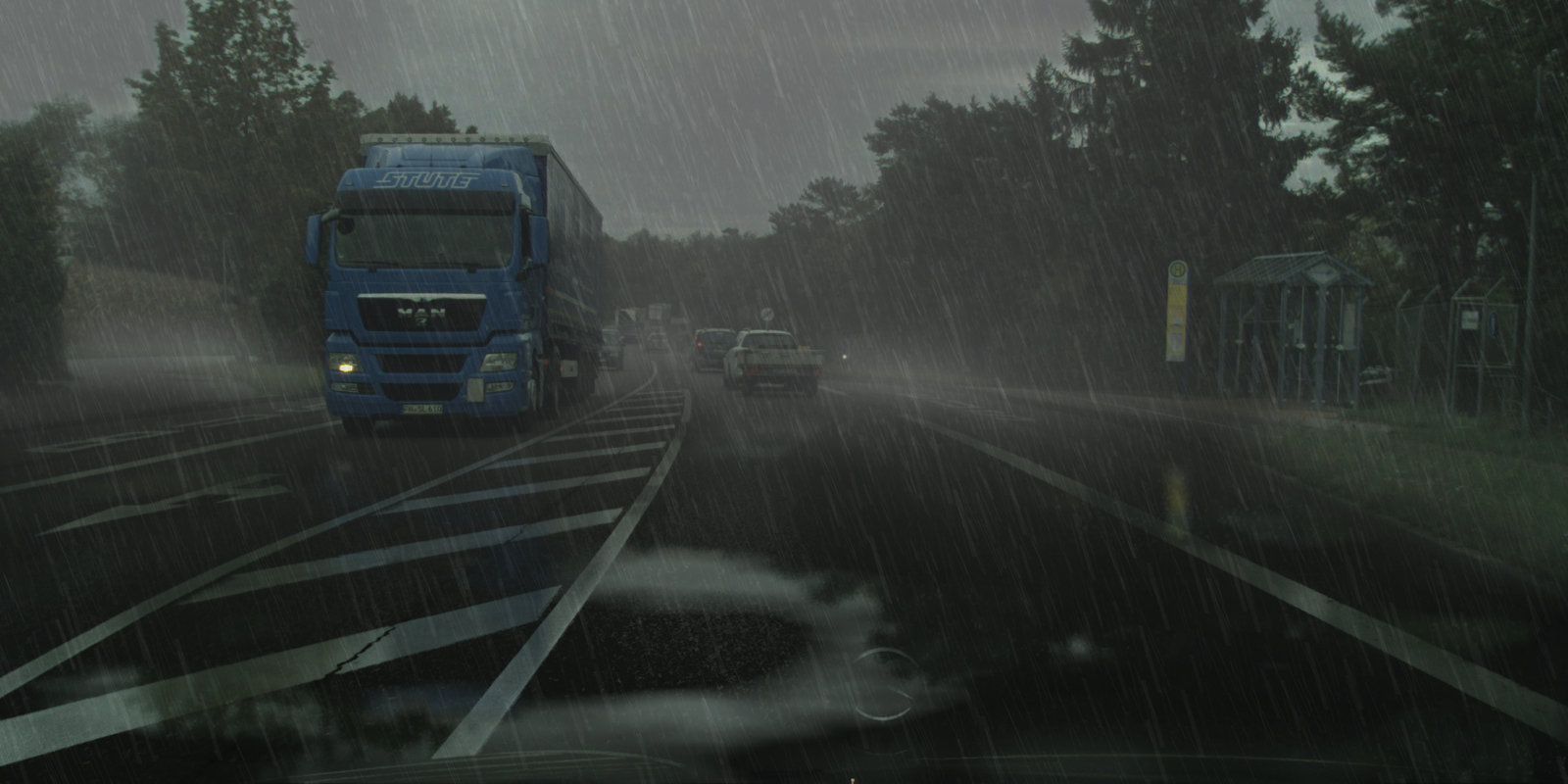} &
\includegraphics[width=\linewidth]{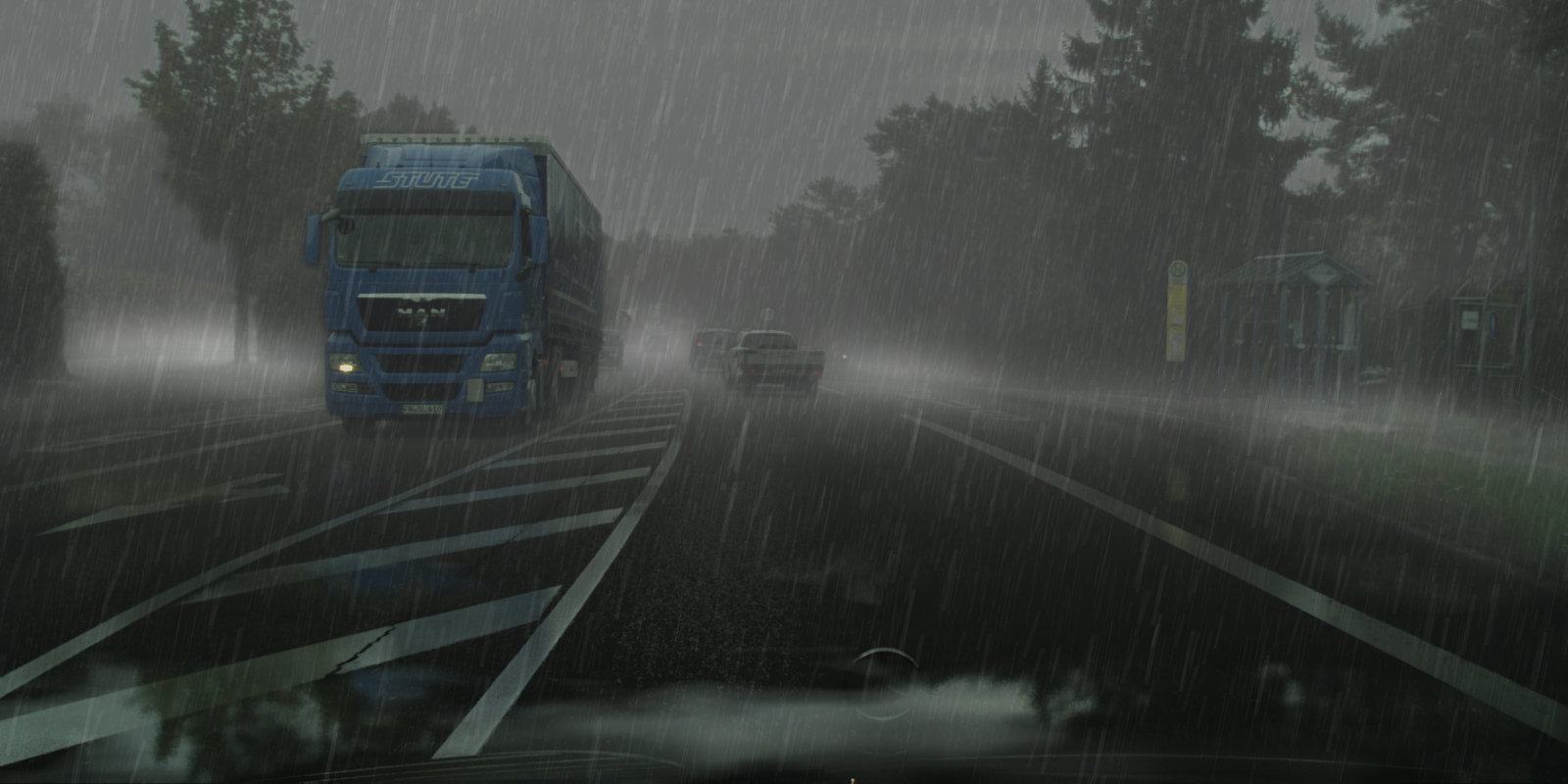} &
\includegraphics[width=\linewidth]{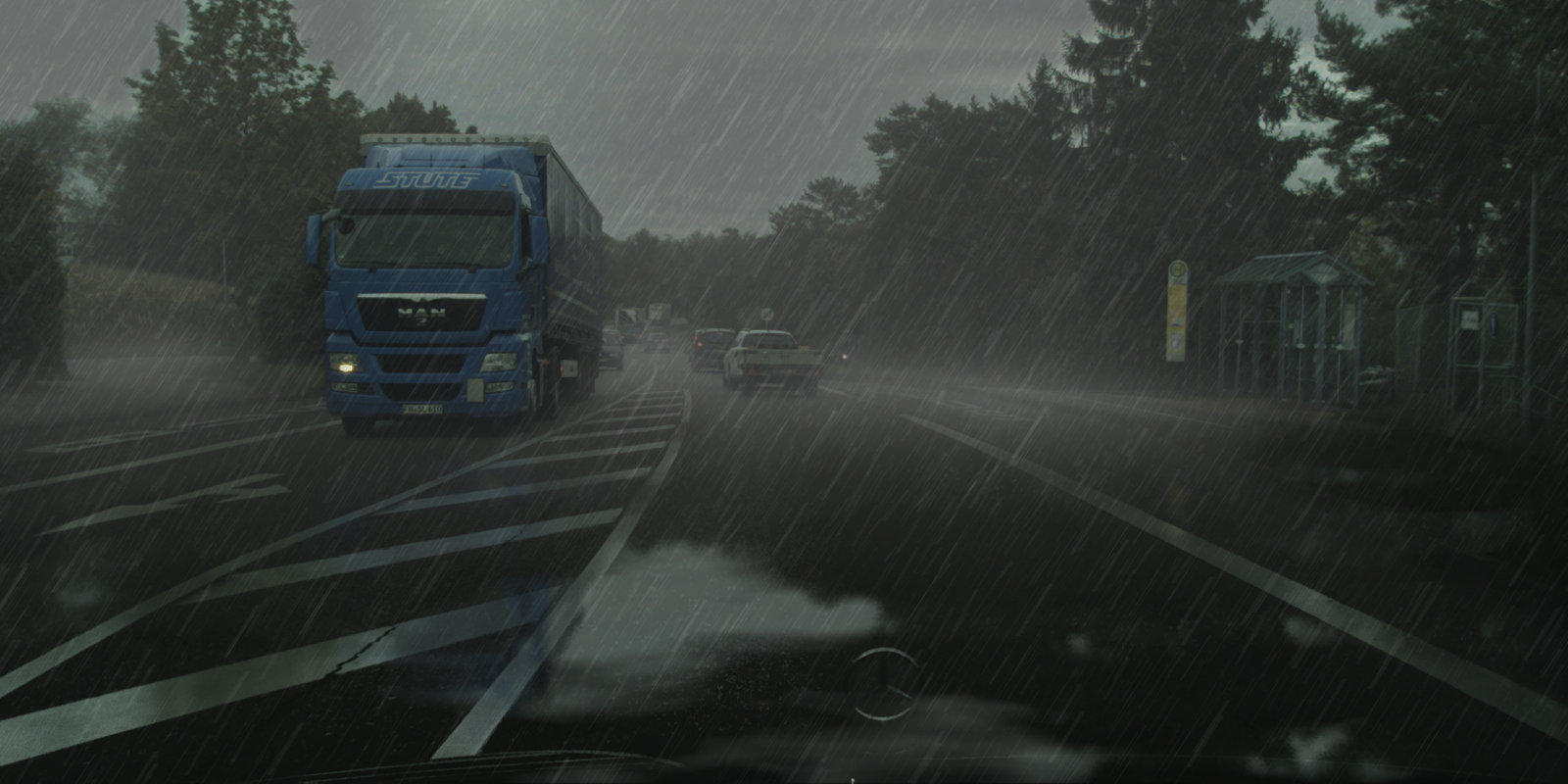} &
\includegraphics[width=\linewidth]{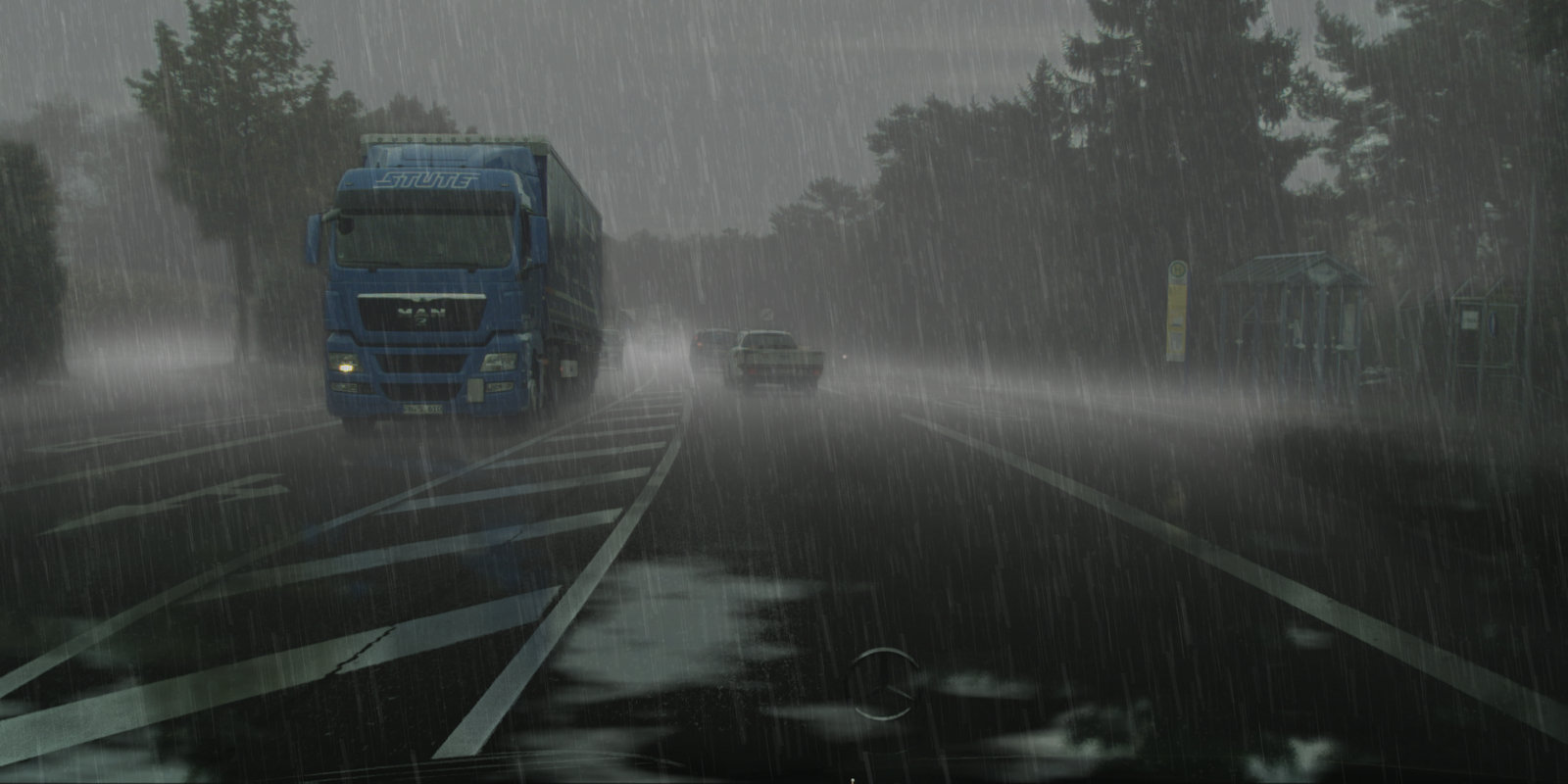} \\

\textbf{4} &
\includegraphics[width=\linewidth]{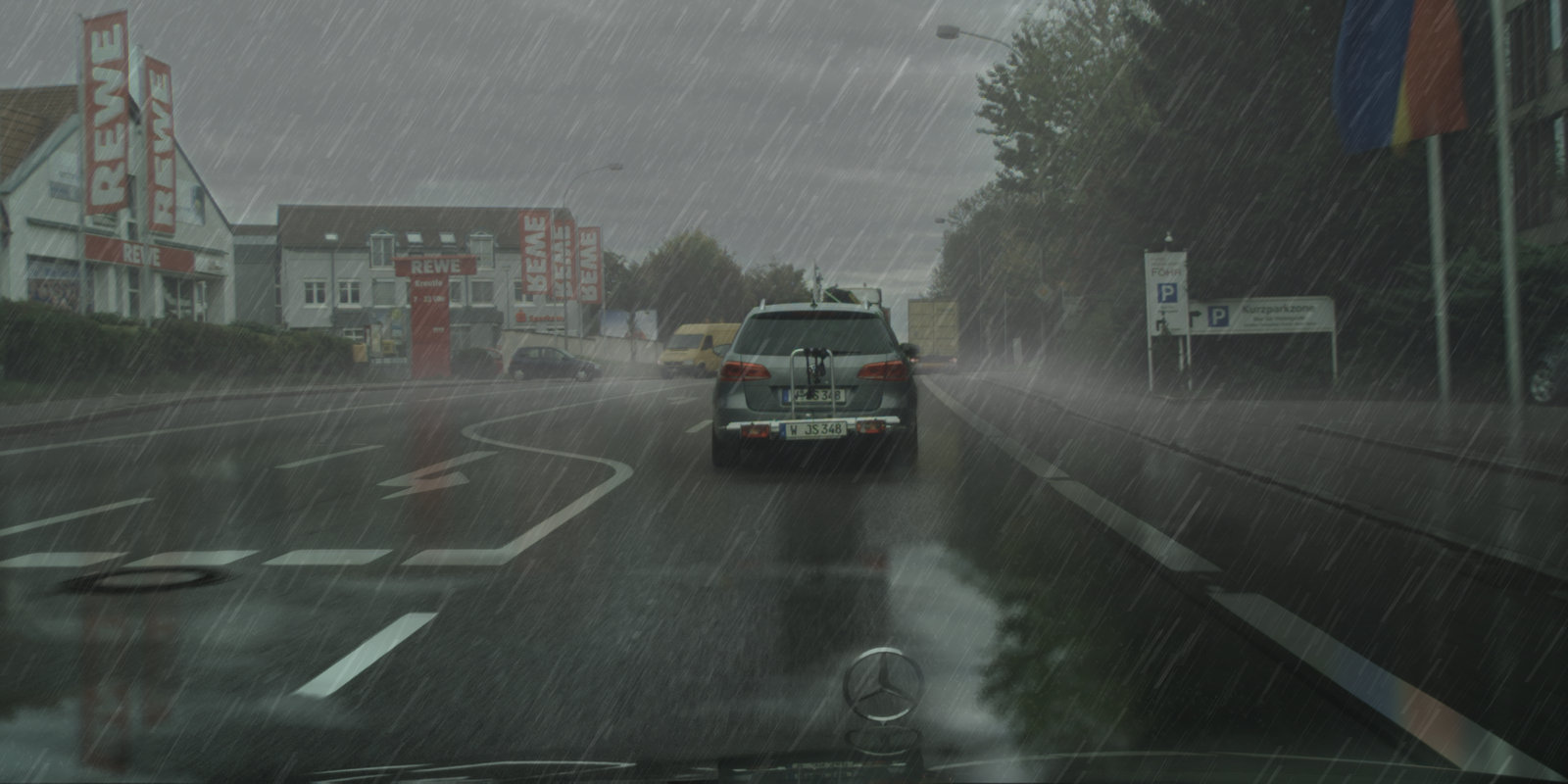} &
\includegraphics[width=\linewidth]{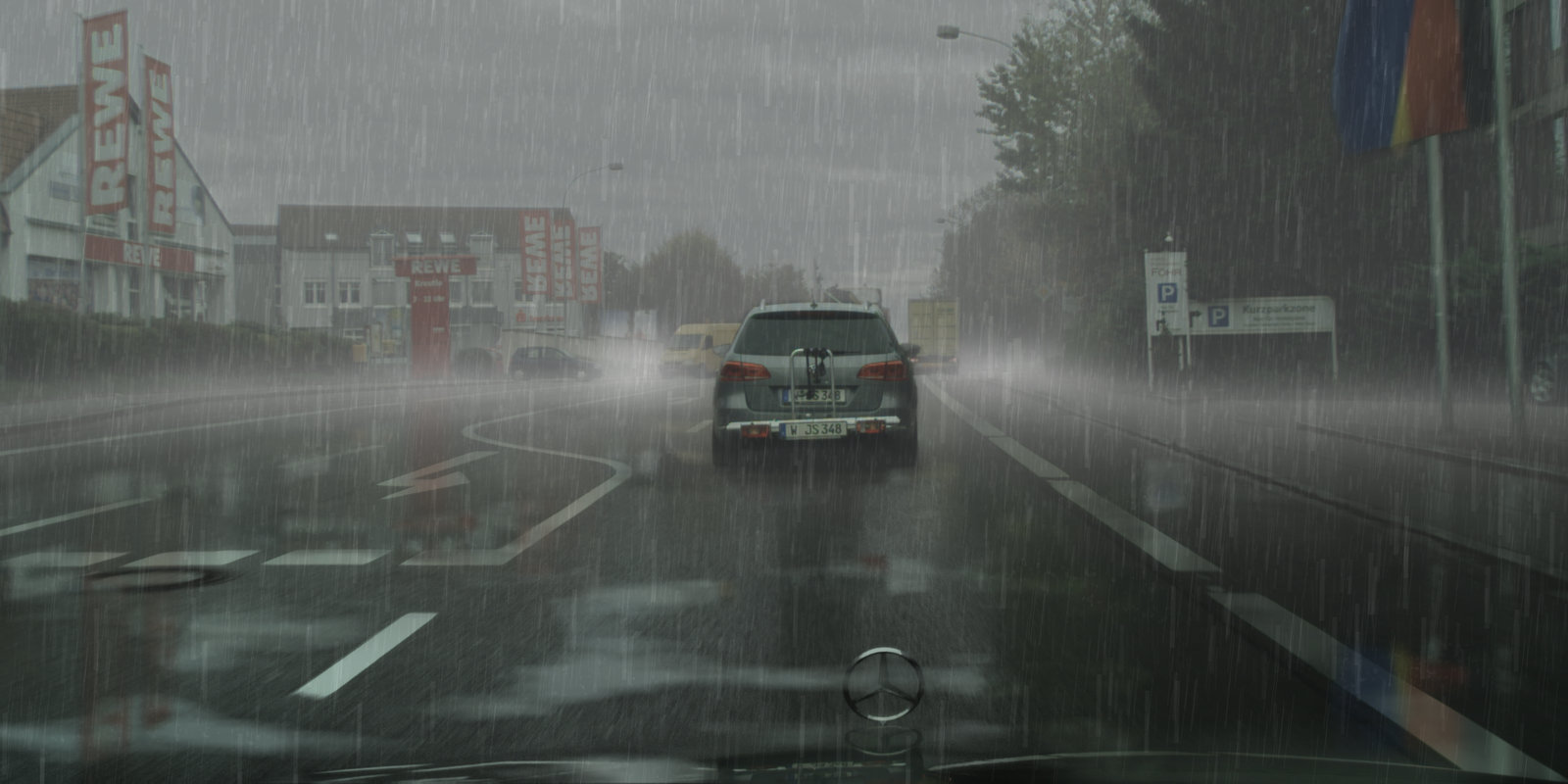} &
\includegraphics[width=\linewidth]{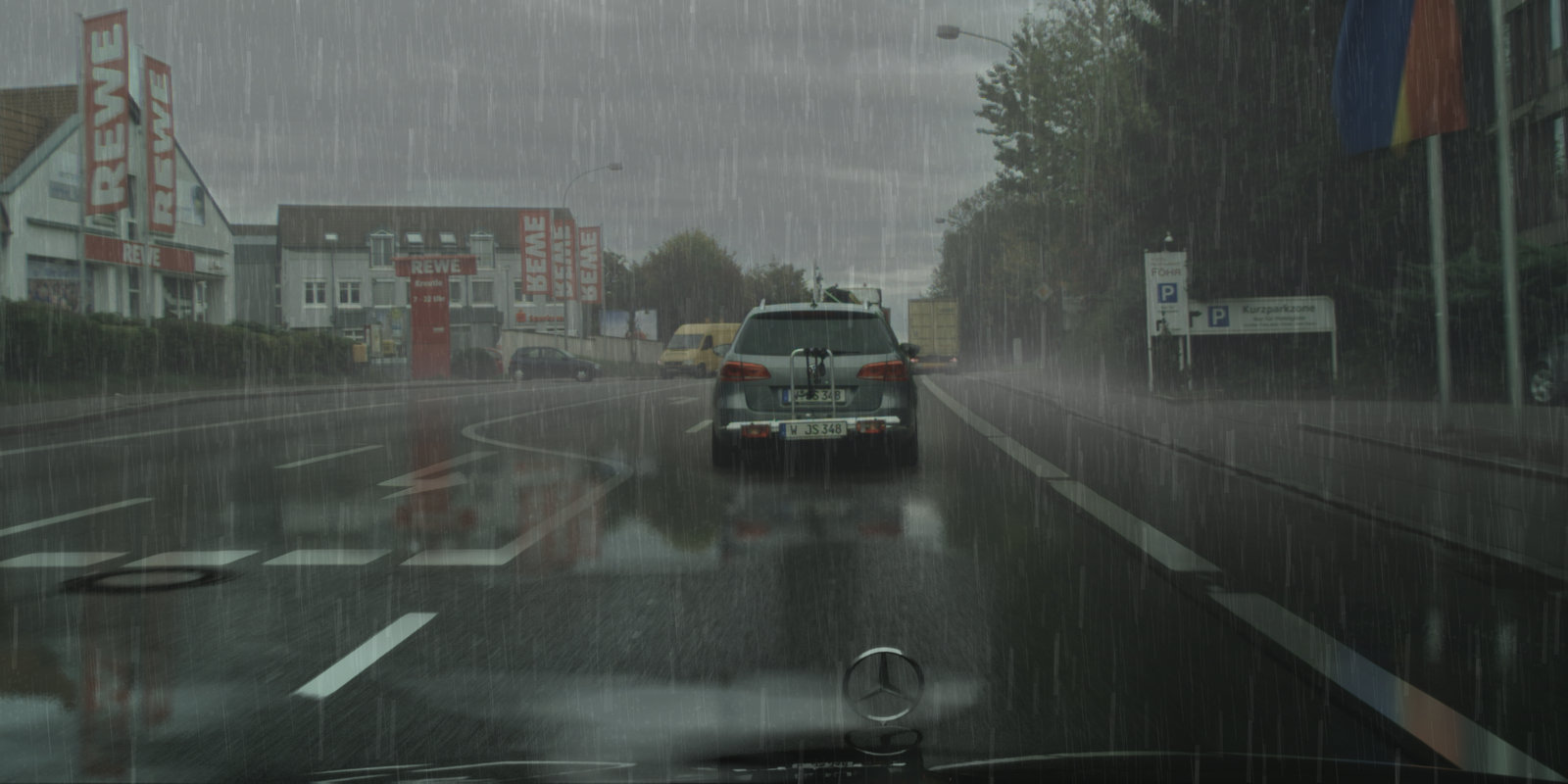} &
\includegraphics[width=\linewidth]{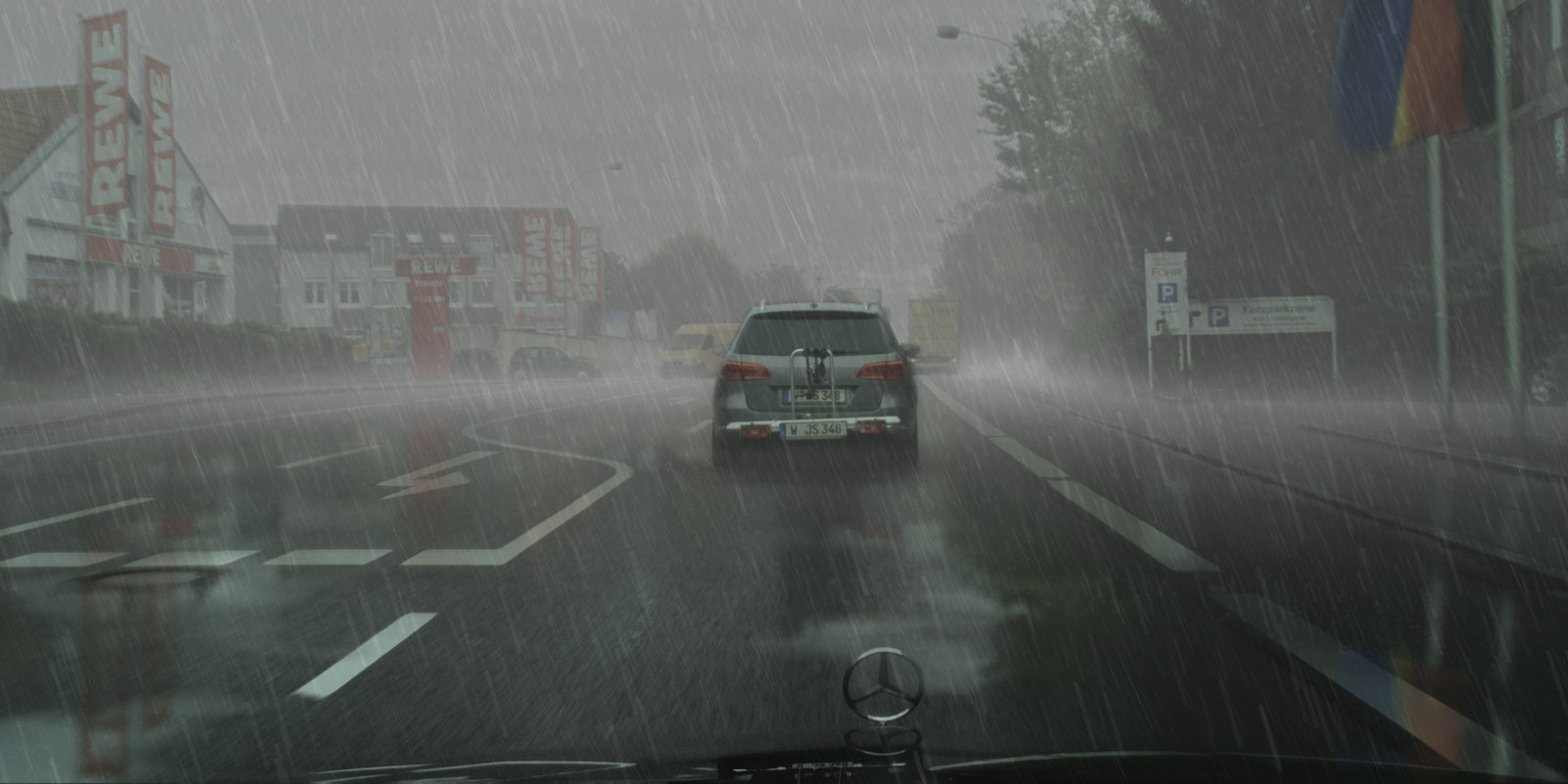} \\

\textbf{5} &
\includegraphics[width=\linewidth]{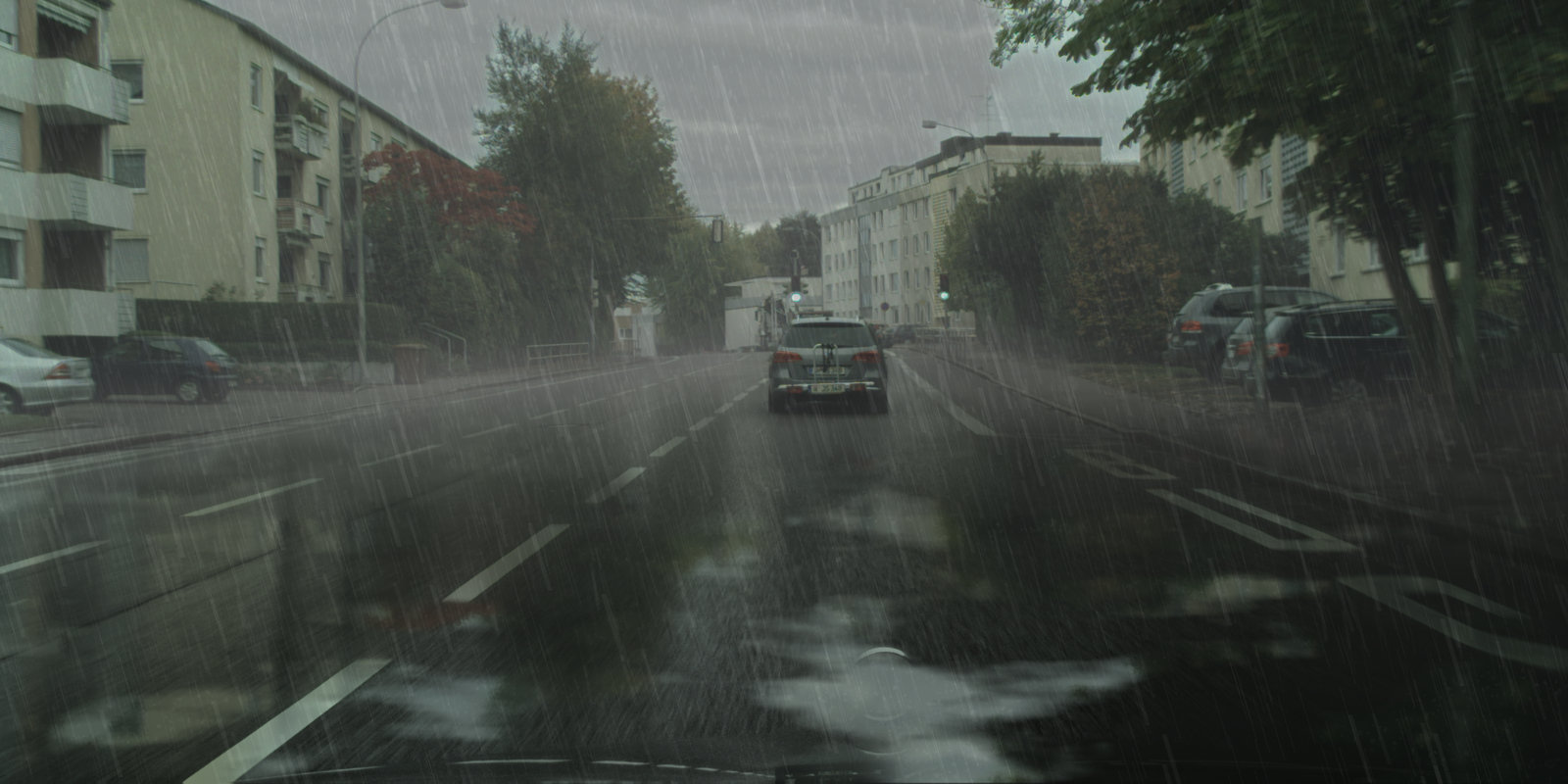} &
\includegraphics[width=\linewidth]{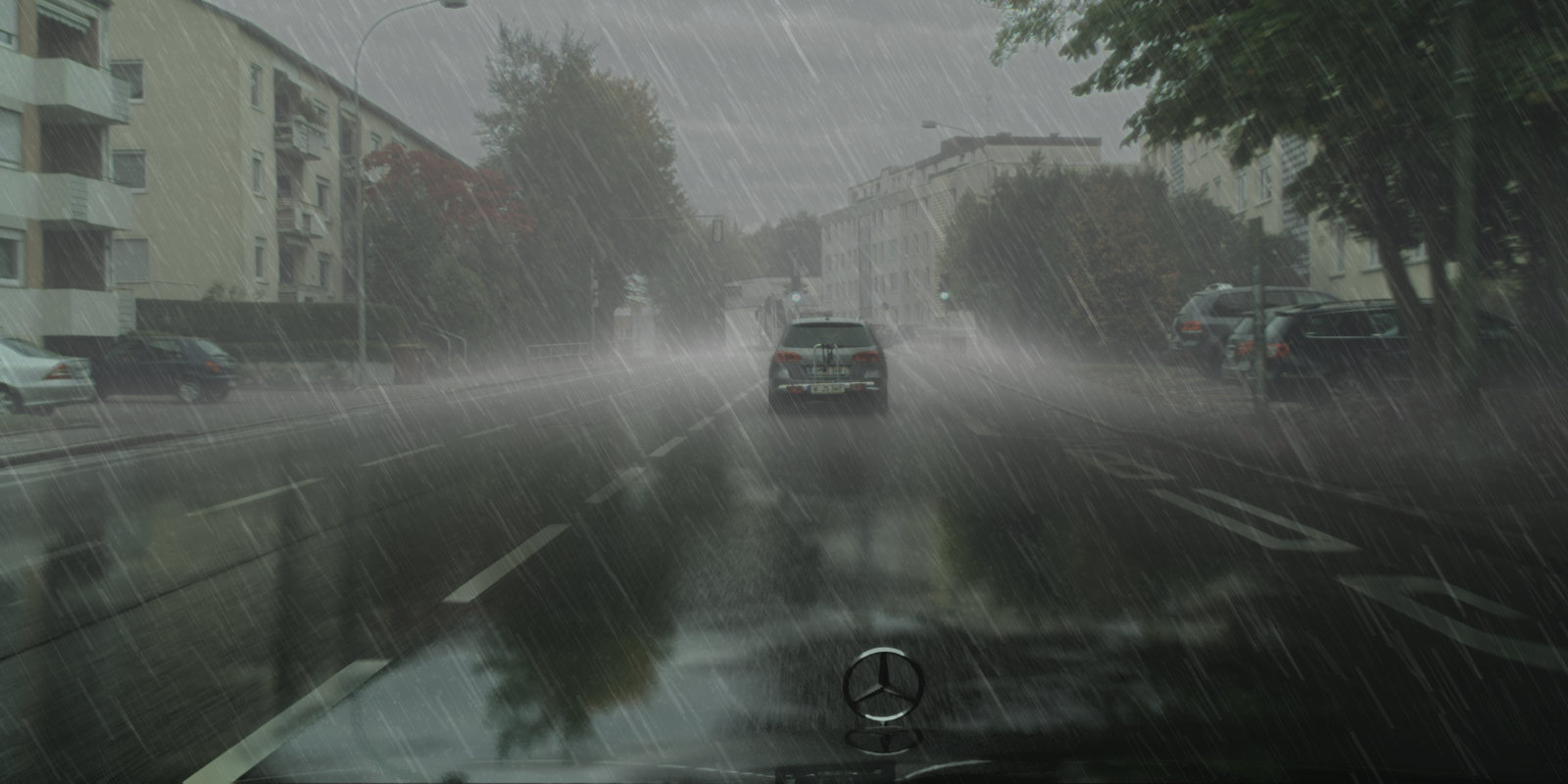} &
\includegraphics[width=\linewidth]{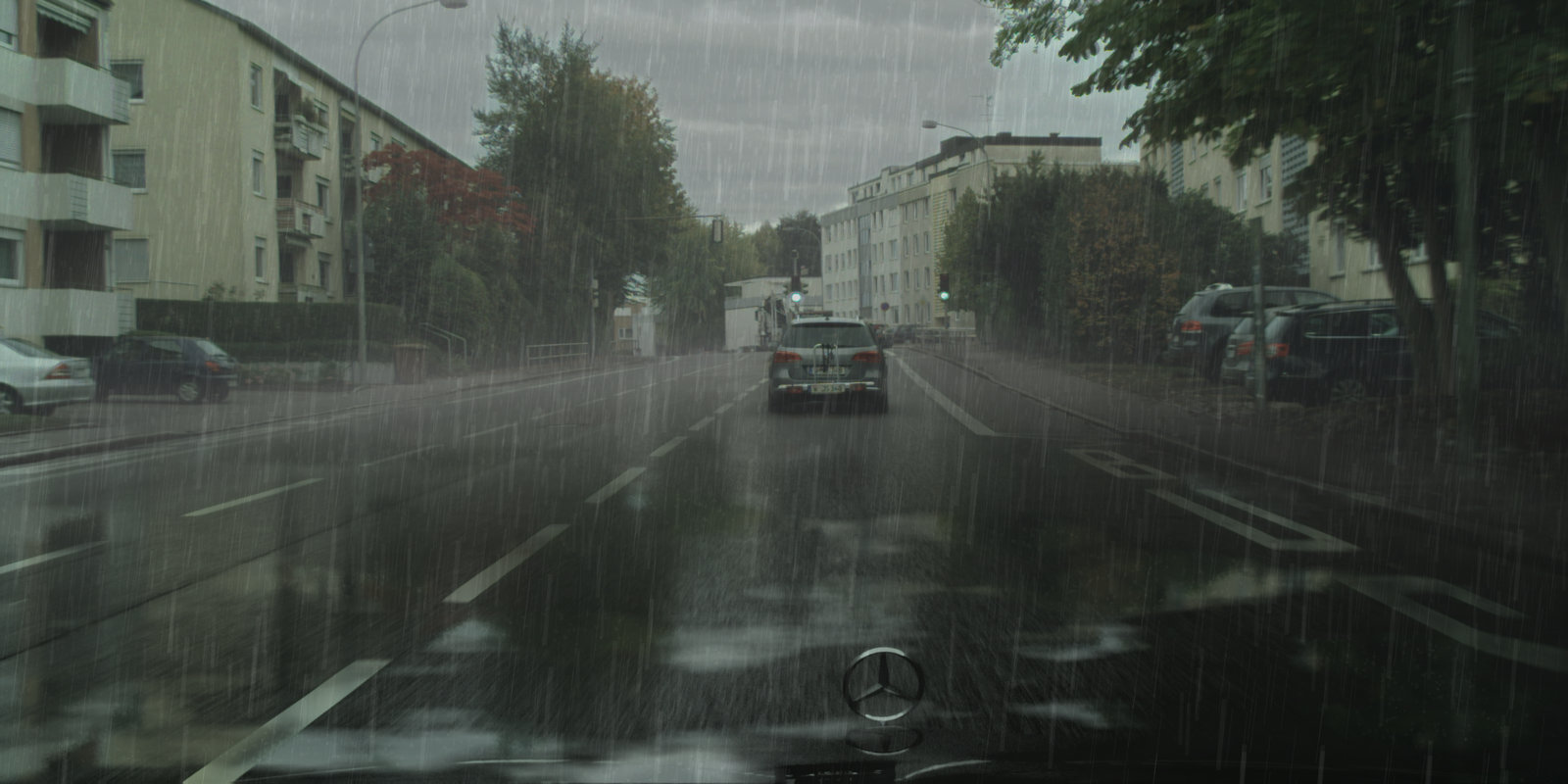} &
\includegraphics[width=\linewidth]{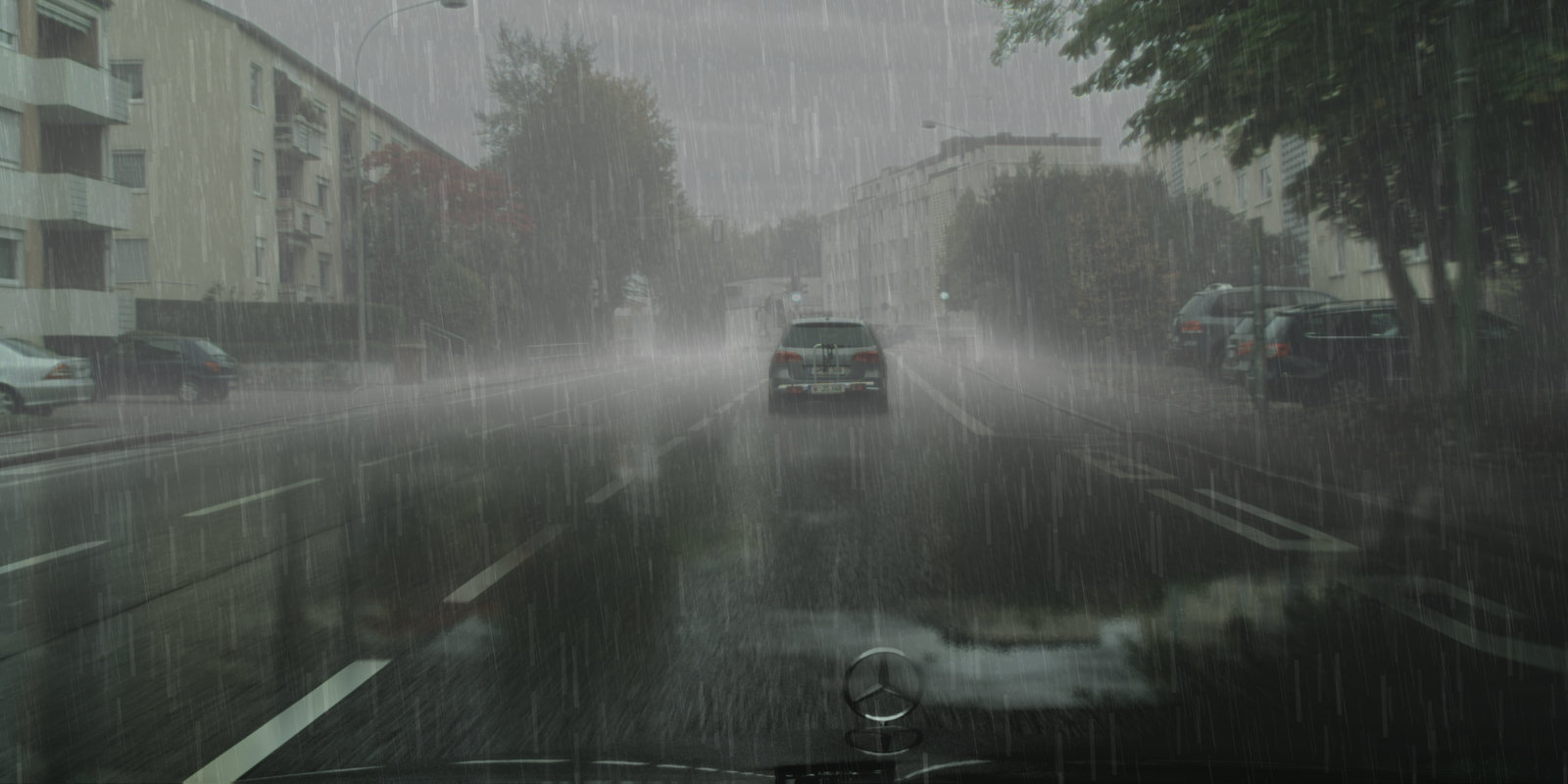} \\

\bottomrule
\end{tabular}
\caption{\textbf{Qualitative comparison of different pretrained depth and semantic segmentation model compositions.}
M = MiDaS, DA = DepthAnything, SegF = SegFormer, and FS = FastSeg. All images are rendered on rain intensity level 50.}
\label{fig:modularity_examples}
\vspace{-3mm}
\end{figure*}

\begin{figure}
    \centering
    \includegraphics[width=\linewidth]{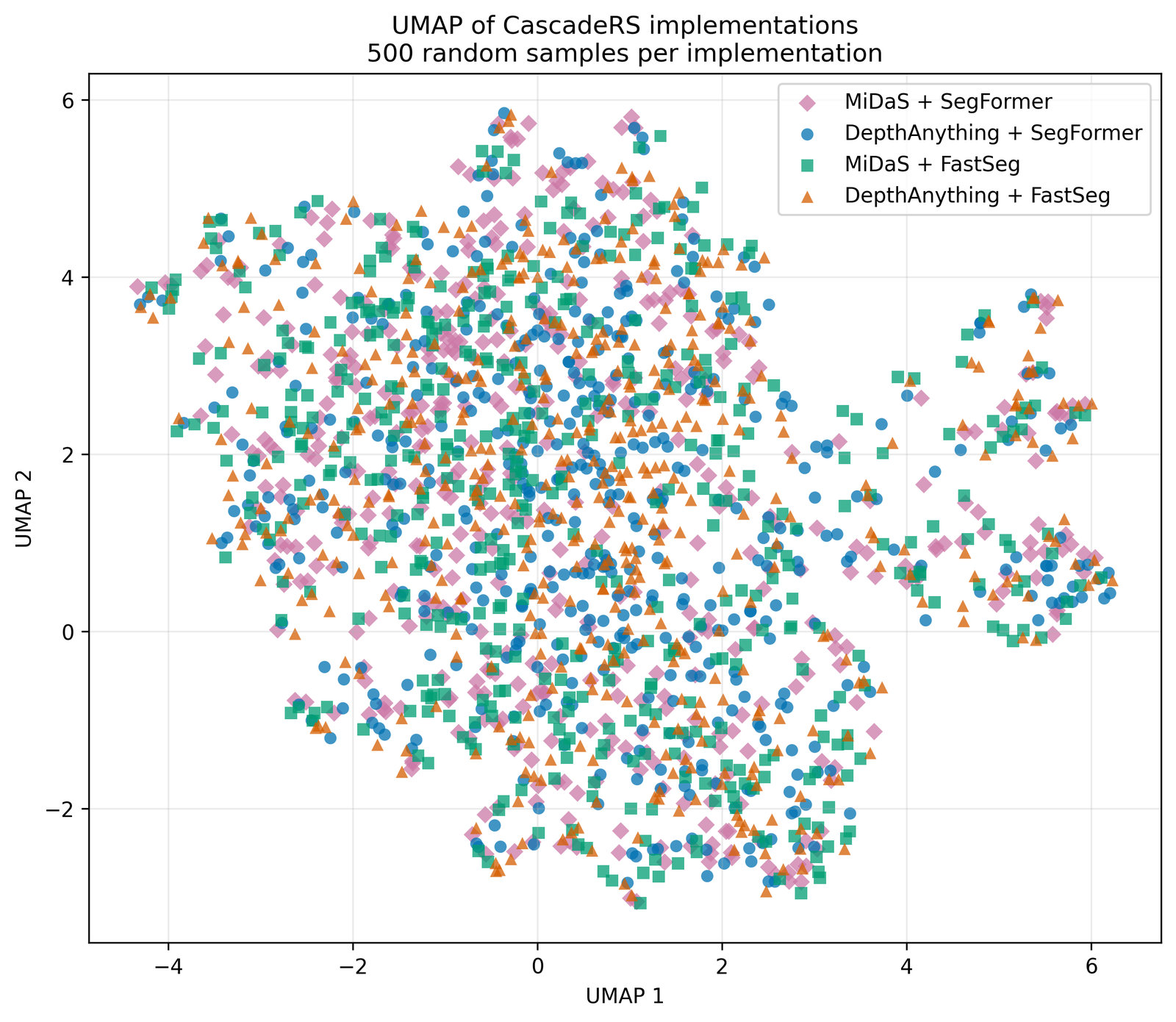}
    \caption{\textbf{UMAP visualization of DINOv3 embeddings} for four CascadeRS implementations using different pretrained depth and semantic segmentation models. Each implementation contains 500 randomly sampled images across all rain intensities. The substantial overlap between the embedding distributions indicates that replacing MiDaS with DepthAnything and/or SegFormer with FastSeg has little effect on the overall feature distribution of the synthesized images.}
    \label{fig:umap_embeddings}
\end{figure}

\section{Method implementation details \label{supmat: method implementation details}}
CascadeRS uses two global quantities that influence the remaining internal rendering parameters: the user-controlled \emph{rain intensity}~$\rho_r$ and the automatically computed \emph{scene brightness}~$\bar{L}$.
The rain-intensity parameter scales effects such as streak density and length, fog thickness, and the strength of wet-surface modules, while the global brightness modulates transparency, atmospheric light, and haze appearance.  
Both quantities reflect physically meaningful properties of the input scene, which helps the parameterization behave consistently across different environments and datasets.  
The following sections describe how $\rho_r$ and $\bar{L}$ influence the fog and rain-streak modules in more detail.

\paragraph{Fog generation (implementation details).}
This section provides the full parameterization of the fog-generation stage $\mathcal{R}_3$ referenced in the main paper. The effective transmittance is computed from the normalized depth map $D(x,y)$ as
\begin{equation}
    t_{\mathrm{eff}}(x,y) = \exp(-\beta\, D^{1.7}(x,y))^{\tau},
\end{equation}
where $\beta \in [4.7, 6.7]$ controls the fog density and visibility decay with distance. Before evaluation, the depth map is Gaussian-smoothed ($\sigma{=}5$) and slightly exponentiated ($D^{1.7}$) to enhance near--far contrast while reducing boundary artifacts.

The exponent $\tau$ determines the perceived haze thickness and depends on both global scene brightness $\bar{L}$ and rain intensity $\rho_r$:
\begin{equation}
    \tau = (0.9 - 2.3\bar{L} + 2.3\bar{L}^2)\left(\frac{\rho_r}{100}\right)^{1.2}.
\end{equation}
The brightness term $\bar{L}$ is derived from per-pixel luminance values,
\begin{equation}
    L(x, y) = 0.2126R + 0.7152G + 0.0722B,
\end{equation}
computed using perceptual channel weights~\cite{Luminance}. The global brightness $\bar{L}$ is defined as the \emph{proportion of pixels} with luminance above $0.7$; to ensure stable behavior across lighting conditions, $\bar{L}$ is clipped to a maximum of $0.5$.

The atmospheric-light color $A$ is determined from $\bar{L}$ through a quadratic mapping,
\begin{equation}
    A = a_{\min} + 2.1\bar{L} - 1.8\bar{L}^2 + \epsilon_A,
\end{equation}
with $a_{\min}{=}0.4$ and $\epsilon_A \sim \mathcal{U}(-0.05, 0.05)$ providing scene-level variation. Together, the parameters $(\beta, \tau, A)$ define the full fog behavior in our implementation, controlling haze density, opacity progression with depth, and the color of scattered atmospheric light.

\paragraph{Rain-streak generation (implementation details).}
This section provides the full parameterization used in the rain-streak synthesis stage $\mathcal{R}_4$. Rain streaks are modeled as elongated, semi-transparent structures whose appearance varies with depth: nearby streaks are thicker, brighter, and longer, while distant ones appear shorter, finer, and fainter. To reproduce this effect, we render a stack of $10$ depth-guided rain layers, each representing a distinct visibility range and streak profile.

\noindent Each layer $j$ is parameterized by 
\[
\theta_j=\{\rho_j,\;\mu_j,\;\gamma_j,\;\sigma_j,\;\delta_j,\;d_j^{\min},d_j^{\max},\;\phi_j\},
\]
where  
$\rho_j$ controls streak density (noise sparsity),  
$\mu_j$ the streak length,  
$\gamma_j$ the layer transparency,  
$\sigma_j$ the Gaussian smoothing width (softness),  
$\delta_j$ the horizontal dilation (streak thickness),  
$(d_j^{\min}, d_j^{\max}]$ the assigned depth interval,  
and $\phi_j$ the streak orientation.  

The base parameters $\rho_j$, $\sigma_j$, $\delta_j$, and depth intervals $(d_j^{\min}, d_j^{\max}]$ are predefined per layer to ensure a smooth progression from thick foreground streaks to subtle background streaks. In contrast, the streak length $\mu_j$ is dynamically adapted based on the 
global rain intensity. Each layer has a predefined minimum and maximum 
streak length, denoted $\mu_j^{\min}$ and $\mu_j^{\max}$.  
The implemented streak length is obtained through linear interpolation:
\begin{equation}
    \mu_j
    =
    \mu_j^{\min} 
    +
    (\mu_j^{\max} - \mu_j^{\min})
    \left( \frac{\rho_r - 10}{90} \right),
\end{equation}
optionally followed by a small stochastic offset
\protect{$\eta_j \sim \mathcal{U}(0, (\mu_j^{\max}-\mu_j^{\min})/5)$} 
to introduce per-image variability.

The transparency $\gamma_j$ of each streak layer is adapted to the global 
scene brightness~$\bar{L}$ to prevent overexposure in bright scenes.  
Each layer has a predefined minimum and maximum transparency 
$(\gamma_j^{\min}, \gamma_j^{\max})$, and the implemented value is obtained 
by linearly interpolating based on the brightness level:
\begin{equation}
\gamma_j = \gamma_j^{\min}
+ (\gamma_j^{\max} - \gamma_j^{\min})\!\left(\frac{0.35 - \bar{L}}{0.35}\right).
\end{equation}
Here, $\bar{L} \in [0, 0.35]$ denotes the clipped global brightness level.
This ensures that streaks appear more pronounced in dark scenes and are 
appropriately suppressed in bright, high-luminance conditions.

To avoid uniform patterns, the streak orientation $\phi_j$ is sampled around a near-vertical base angle of $80^\circ$ with small stochastic perturbations ($\pm5^\circ$ per layer). Sparse binary noise $\Psi_j$ with density $\rho_j$ is horizontally dilated by $\delta_j$, blurred using a motion-blur operator parameterized by $(\mu_j,\phi_j)$, and smoothed using a Gaussian kernel of width $\sigma_j$, producing the final streak texture. Together, these parameters determine the density, orientation, thickness, softness, depth placement, and intensity-dependent elongation of streaks, ensuring physically plausible and visually diverse rain patterns across the scene.




\section{Parameter robustness \label{supmat: parameter robustness}}
We evaluate the effect of varying five synthesis parameters in Figure~\ref{fig:parameter-sweep-full}: rain intensity, fog density $\beta$, puddle resolution $f_0$, puddle persistence $a$, and puddle reflection $\lambda_{p,0}$. Distribution similarity is assessed using Fr'echet Inception Distance (FID) and Kernel Inception Distance (KID) with ACDC \cite{sakaridis2024acdcadverseconditionsdataset} as the real-rain reference dataset, while image fidelity is evaluated using Learned Perceptual Image Patch Similarity (LPIPS), Structural Similarity Index Measure (SSIM), and Peak Signal-to-Noise Ratio (PSNR) relative to the corresponding original images. Downstream performance is evaluated through mean Intersection over Union (mIoU) using a pretrained SegFormer model \cite{xie2021segformersimpleefficientdesign}, with the reported standard deviation obtained through bootstrapping, and Average Precision (AP) and AP at an IoU threshold of 0.50 (AP@50) using a pretrained DINO-Teacher model~\cite{lavoie2025large}. Rain intensity is the primary user-controllable parameter and jointly influences the formation of rain streaks, fog, and puddles. Consequently, variations in rain intensity have the largest effect across distribution similarity, image fidelity, and downstream task performance. Fog density also affects these metrics, although to a lesser extent. In contrast, variations in the individual puddle parameters result in comparatively small changes. This behavior is consistent with the method's design: rain intensity affects multiple components of the synthesis pipeline and, therefore, the overall appearance of the image, whereas the puddle parameters control more localized properties of the generated puddles. The limited variation in the evaluated metrics across these puddle parameter settings further indicates that the method is relatively robust to their individual configuration.

\begin{figure*}
    \centering
    \includegraphics[width=\linewidth]{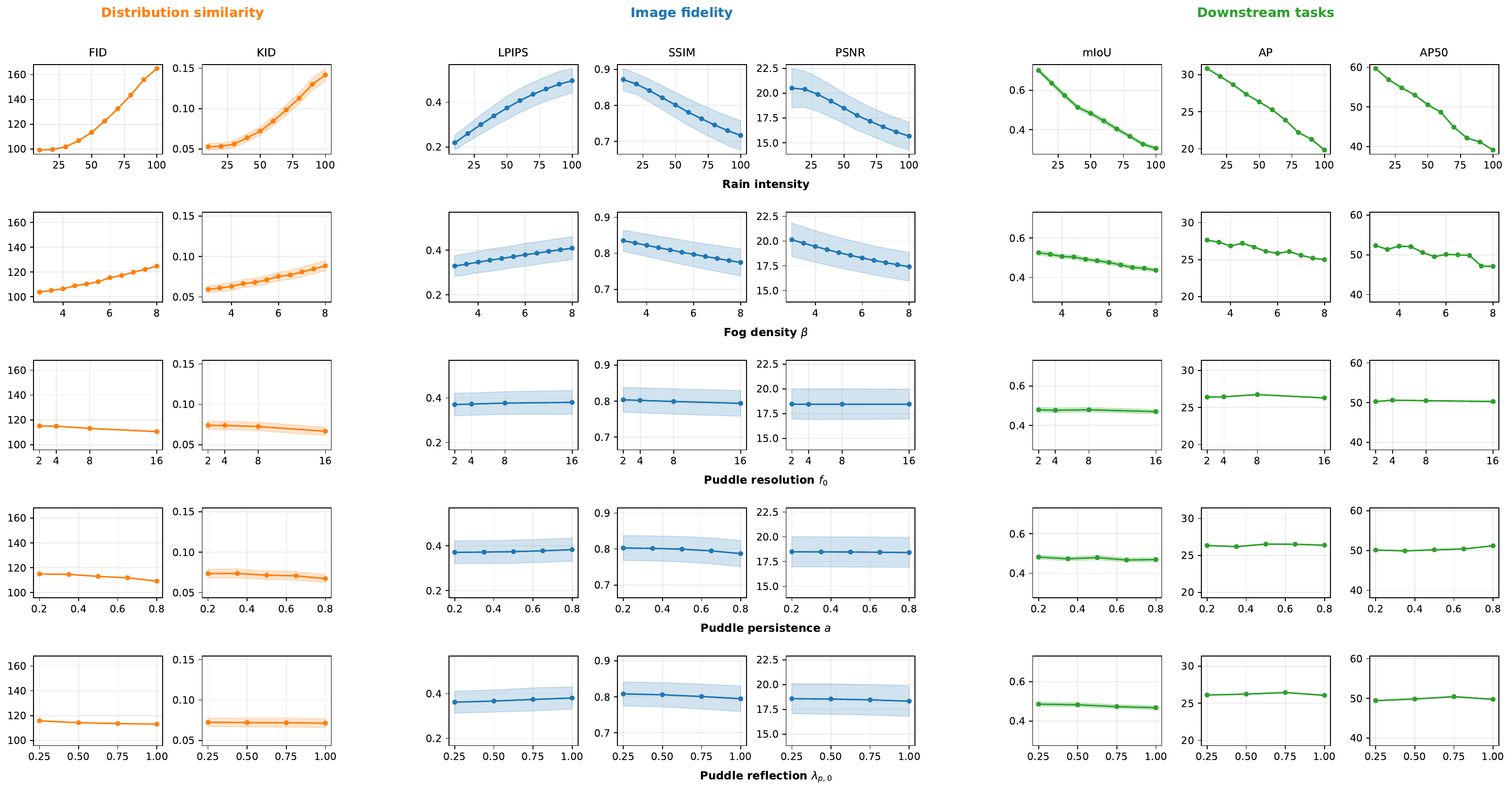}
    \caption{\textbf{Effect of synthesis parameters on image characteristics and downstream task performance.}
We vary rain intensity, fog density $\beta$, puddle resolution $f_0$, puddle persistence $a$, and puddle reflection $\lambda_{p,0}$ while keeping the remaining synthesis parameters at their default settings. Distribution similarity is evaluated using FID and KID with ACDC \cite{sakaridis2024acdcadverseconditionsdataset} as the reference distribution, while image fidelity with respect to the original Cityscapes images \cite{Cityscapes} is measured using LPIPS, SSIM, and PSNR. Downstream performance is evaluated using semantic segmentation mIoU with a pretrained SegFormer model \cite{xie2021segformersimpleefficientdesign} and object detection AP and AP@50 with a pretrained DINO-Teacher model \cite{lavoie2025large}. Shaded regions indicate standard deviation where applicable; for mIoU, the standard deviation is estimated by bootstrapping.}
    \label{fig:parameter-sweep-full}
\end{figure*}

\section{Effect Rain Intensity on Semantic Segmentation \label{supmat: effect rain intensity on segmentation}}

We analyze the effect of rain intensity on semantic segmentation in Figure~\ref{fig:miou_intensity} for SegFormer~\cite{xie2021segformersimpleefficientdesign}, FastSeg~\cite{zhang2020fastseg}, and GCNet~\cite{yang2025goldencudgel}. The figure reports the mIoU over the full dataset at each available rain intensity. Performance on real rainy images from ACDC~\cite{sakaridis2024acdcadverseconditionsdataset} is shown as a constant reference, as ACDC lacks metadata on rain intensity.
Across the synthetic datasets, increasing rain intensity generally leads to decreasing segmentation performance, although the magnitude of this effect depends on both the rain synthesis method and segmentation architecture. PBR~\cite{halder2019physicsbased} causes a degradation, with mIoU falling substantially below the ACDC reference at higher intensities. Cityscapes-Rain~\cite{DepthRainRemoval} produces a more gradual degradation, particularly for SegFormer, while TPSeNCE~\cite{zheng2023} has a comparatively limited effect on all three models.
CascadeRS provides a broad and controllable range of segmentation difficulty. For SegFormer, increasing rain intensity gradually reduces mIoU, spanning performance above and below the ACDC reference. FastSeg and GCNet are considerably more sensitive to increases in CascadeRS intensity, with performance decreasing rapidly at higher intensities. This architecture-dependent response explains the lower aggregated CascadeRS performance of FastSeg and GCNet observed in the main paper. Overall, the results demonstrate that CascadeRS can systematically vary the difficulty of rainy scenes through its rain-intensity parameter, while also revealing differences in robustness between segmentation architectures.
\begin{figure}[!t]
    \centering
    \includegraphics[width=\linewidth]{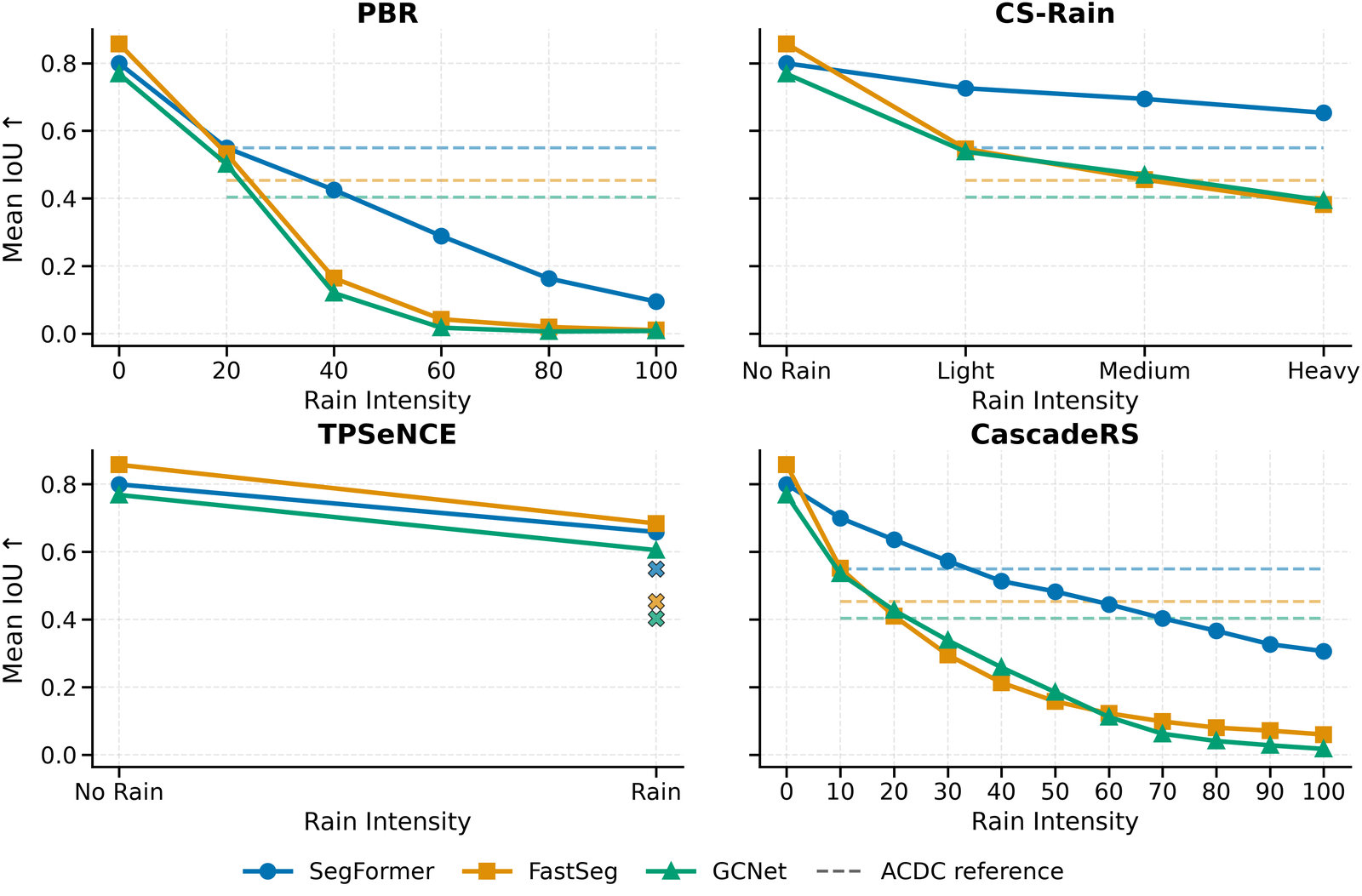}
    \caption{\textbf{Semantic segmentation performance across rain intensity levels.}
    Full-dataset mIoU for SegFormer~\cite{xie2021segformersimpleefficientdesign}, FastSeg~\cite{zhang2020fastseg}, and GCNet~\cite{yang2025goldencudgel} at each available rain intensity. Dashed lines denote the corresponding performance on real rainy ACDC images; for TPSeNCE, which does not provide multiple intensity levels, ACDC performance is shown as reference points.}
    \label{fig:miou_intensity}
\end{figure}

\section{Survey details \label{supmat: survey details}}
\noindent\textbf{Percentages Qualitative Evaluation}
Table \ref{tab:realism_score_distribution} reports the distribution of realism ratings from Part 1 of the qualitative evaluation. For each rain synthesis method and real rain, the table shows the percentage of ratings assigned to each score (1: not realistic, to 5: very realistic) for both the user survey and the Gemini evaluation.

\begin{table}[t]
    \centering
    \caption{\textbf{Distribution of realism ratings.}
    Percentage of ratings for each realism score (1: not realistic, to 5: very realistic), comparing the user survey and Gemini evaluation.}
    \label{tab:realism_score_distribution}

    \setlength{\tabcolsep}{3.5pt}
    \renewcommand{\arraystretch}{1.0}
    \scriptsize

    \begin{tabular}{@{}llccccc@{}}
        \toprule
        \textbf{Method} & \textbf{Evaluator} &
        \textbf{1} & \textbf{2} & \textbf{3} &
        \textbf{4} & \textbf{5} \\
        \midrule

        PBR
        & Survey & 28.4 & 16.7 & 21.3 & 21.6 & 12.0 \\
        & Gemini & 0.0 & 50.0 & 50.0 & 0.0 & 0.0 \\
        \addlinespace

        CS-Rain
        & Survey & 10.4 & 16.6 & 23.1 & 35.4 & 14.6 \\
        & Gemini & 0.0 & 12.5 & 25.0 & 62.5 & 0.0 \\
        \addlinespace

        TPSeNCE
        & Survey & 4.2 & 10.9 & 24.5 & 39.4 & 20.9 \\
        & Gemini & 0.0 & 12.5 & 0.0 & 75.0 & 12.5 \\
        \addlinespace

        CascadeRS
        & Survey & 5.0 & 10.9 & 23.2 & 41.6 & 19.3 \\
        & Gemini & 0.0 & 0.0 & 25.0 & 75.0 & 0.0 \\
        \addlinespace

        Real Rain
        & Survey & 5.4 & 8.8 & 15.4 & 31.7 & 38.8 \\
        & Gemini & 0.0 & 12.5 & 0.0 & 62.5 & 25.0 \\

        \bottomrule
    \end{tabular}
\end{table}

\noindent\textbf{Examples of Human Evaluation Questions.}
Figures~X–Z show representative questions from the three components of our perceptual study.

\begin{figure}[!t]
    \centering
    \includegraphics[width=\linewidth]{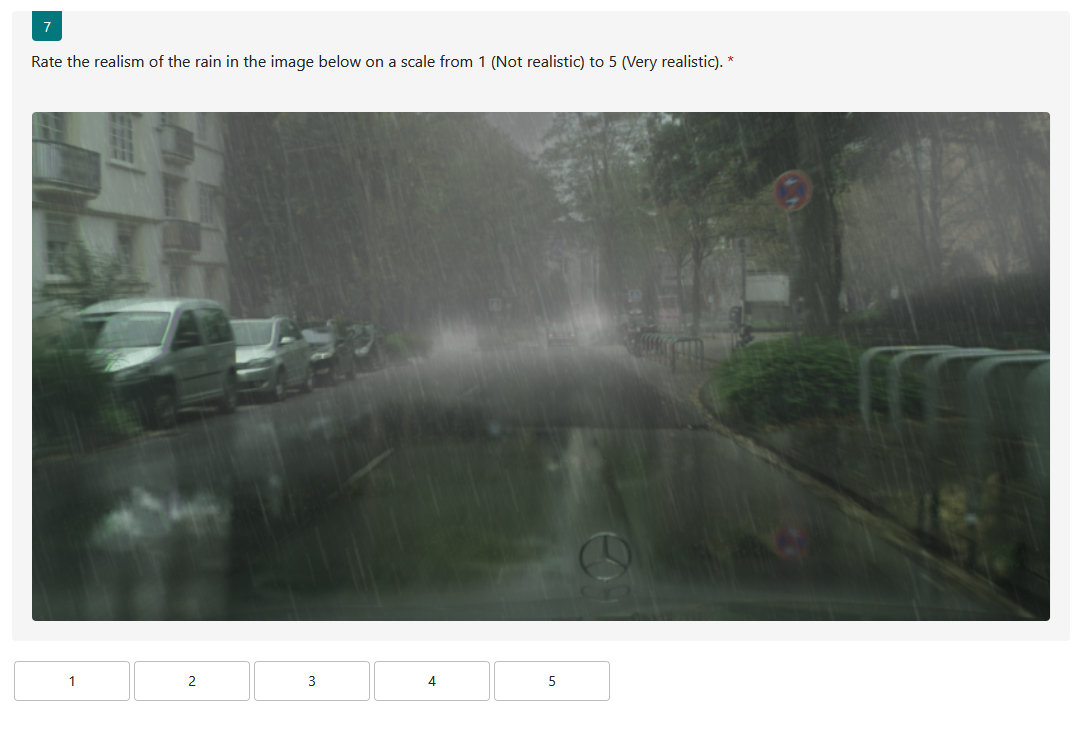}
    \caption{Example question of part 1 of the survey.}
    \label{fig:example_part1_question_survey}
\end{figure}

\begin{figure}[!t]
    \centering
    \includegraphics[width=\linewidth]{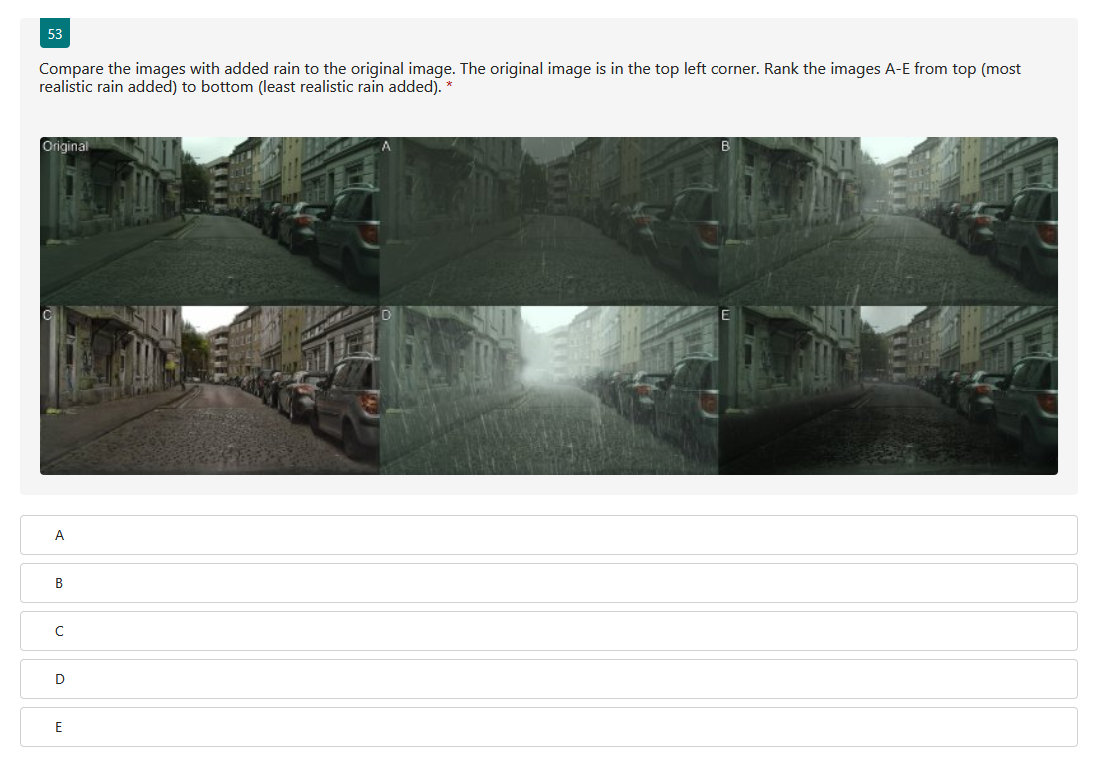}
    \caption{Example question of part 2 of the survey.}
    \label{fig:example_part2_question_survey}
\end{figure}

\begin{figure}[!t]
    \centering
    \includegraphics[width=\linewidth]{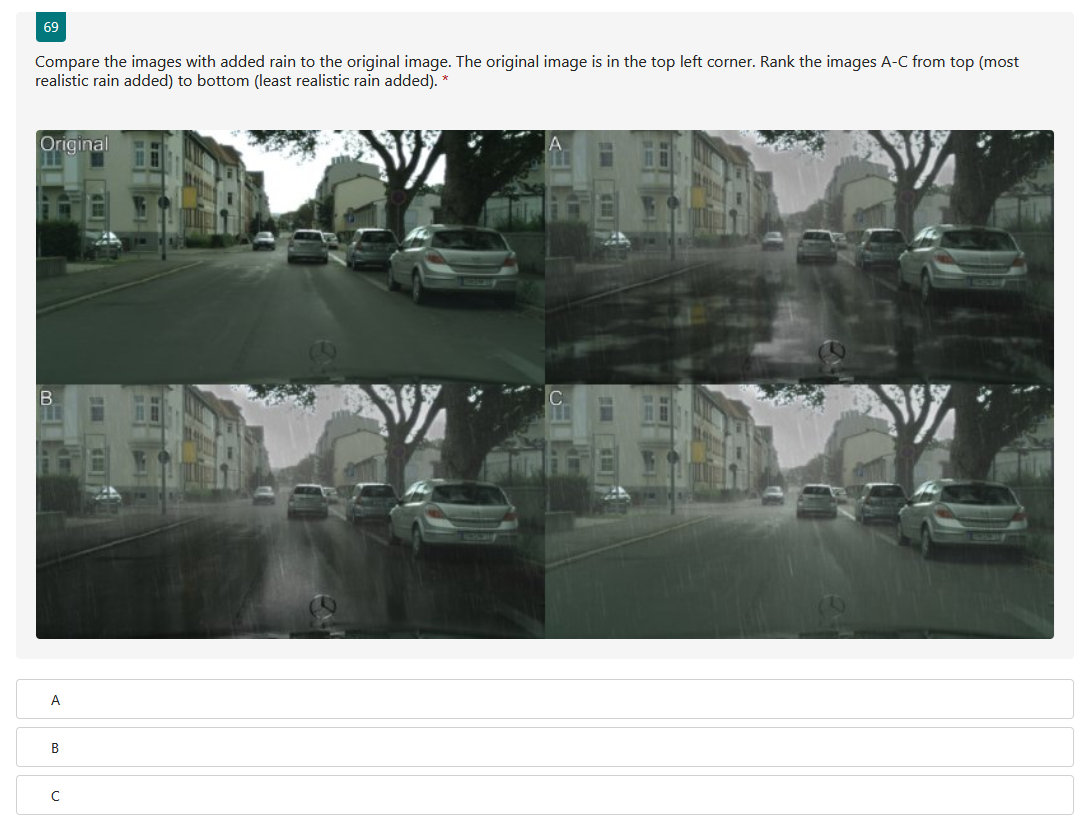}
    \caption{Example question of part 3 of the survey.}
    \label{fig:example_part3_question_survey}
\end{figure}

\textit{Part~1} (Fig.~\ref{fig:example_part1_question_survey}) presents the \emph{absolute realism} task. Participants are shown a single rain-augmented image and asked to rate its realism on a 1–5 Likert scale. This produces an absolute perceptual realism distribution for each method.

\textit{Part~2} (Fig.~\ref{fig:example_part2_question_survey}) illustrates the \emph{relative realism} task for full-method comparisons. Participants see five rain-augmented variants of the same scene alongside the original, and must rank the synthesized images from most to least realistic. This enables direct method-to-method comparison and yields the Borda-count rankings presented in the main paper.

\textit{Part~3} (Fig.~\ref{fig:example_part3_question_survey}) shows the \emph{relative realism ablation} task. Here, participants compare images generated using different subsets of our pipeline (rain only, rain+reflections, rain+puddles). As in Part~2, they rank the images from most to least realistic. This evaluates how individual components contribute to overall realism and supports the ablation results reported in the paper.

Together, these three tasks provide complementary measures of perceptual realism: an absolute quality score (Part~1), multi-method comparative judgments (Part~2), and a controlled ablation-based comparison (Part~3). These examples reflect the exact interface and question formats presented in the deployed human study.


\paragraph{Automated Weather Evaluation with Gemini.} 
To complement the human study and quantitative metrics reported in the main paper, we additionally perform an automated analysis of visual weather cues using the Gemini~2.5 multimodal API. The goal of this experiment is to obtain a structured, model-based assessment of the rendered rain, including visibility degradation, surface wetness, atmospheric appearance, and perceived realism.

\textit{Part~1} Each rendered image is sent to Gemini~2.5 Flash via a multimodal JSON API. For each image, the script transmits an instruction block (\emph{system prompt}) specifying the task, a detailed \emph{user query}, and a strict JSON schema that constrains all output fields to predefined enumerations. The returned analysis includes sky condition, precipitation type, lighting condition, visibility, ground-surface state, dominant weather effect, a confidence score representing the likelihood of rain, a brief natural-language justification, and a realism rating on a 1--5 scale. All responses are automatically saved as per-image JSON files for downstream use. The code will be available upon acceptance of the paper. Below we reproduce the exact prompts embedded in the code:
\newpage
\begin{samepage}
\begin{Verbatim}[breaklines,breakanywhere]
    
    SYSTEM_PROMPT = ( "You are an image analysis model specializing 
    in meteorological observations. Your task is to detect and 
    describe visible weather effects in the provided image. 
    Respond STRICTLY according to the JSON schema provided."
)

USER_QUERY = ( "Examine the image and provide a structured 
    analysis of the weather conditions. Ensure all categorical 
    fields use one of the specified allowed values. Provide a 
    brief reasoning, a confidence score that can be used to 
    rank the rainy effect of an image and rate the realism of 
    the rain in the image on a scale from 1 (not realistic) 
    to 5 (very realistic)."
)
\end{Verbatim}
\end{samepage}



\textit{Part~2}  We employ an automated realism-ranking procedure using the Gemini~2.5 Flash API. The script sends each rendered image to the model together with a targeted system prompt and a strict JSON schema, requesting a realism score from 1--10 and a brief justification focusing on physical plausibility, surface interactions, and light behavior. This enables efficient large-scale comparison of rain realism across thousands of images in a consistent manner. The code includes robust error handling, MIME-type inference, and exponential backoff for reliability. We will release the full implementation upon acceptance of the paper.


\begin{samepage}
\begin{Verbatim}[breaklines,breakanywhere]
system_prompt = ("You are an expert visual effects critic 
    specializing in photorealism and physics. Analyze the 
    provided image and assess the realism of the rain and 
    water effects (e.g., droplets, splashes, wet surfaces, 
    interaction with light/objects). Provide a score between 
    1 and 10 and a brief justification for the realism."
)

response_schema = {
    "type": "OBJECT",
    "properties": {
        "realism_score": {
            "type": "INTEGER",
            "description": "A score from 1 (low realism) to 10 
                        (perfect realism) for the rainy effects."
        },
        "justification": {
            "type": "STRING",
            "description": "A brief explanation of why this score 
            was assigned, focusing on elements like water movement,
            light interaction, and texture."
        }
    },
    "required": ["realism_score", "justification"]
}
\end{Verbatim}
\end{samepage}




\textit{Part~3)} To evaluate the perceptual contribution of individual components in CascadeRS (rain streaks, wet surfaces, and puddles), we apply a variant of the same Gemini-based realism evaluation used for the realism-ranking across different methods. In this case, the script processes, for each scene, the three ablation variants (Rain streaks, Reflections, and Puddles) and requests from Gemini a realism score on a 1--10 scale together with a brief justification. The system prompt is identical to the one used in the method ranking evaluation: both instruct Gemini to act as a visual-effects critic specializing in physical plausibility, surface interaction, and light behavior. This automated evaluation complements our human ablation study by offering an additional, reproducible perspective on which components most strongly influence perceived realism. The complete code will be released upon acceptance of the paper.  


\end{document}